%% file: neurips_2023_arxiv_camera_ready.tex
\documentclass{article}

\PassOptionsToPackage{numbers, compress}{natbib}

\usepackage[final]{neurips_2023}

\usepackage{natbib}
\usepackage[utf8]{inputenc} 
\usepackage[T1]{fontenc}    
\usepackage{url}            
\usepackage{booktabs}       
\usepackage{multirow}
\usepackage{array} 
\newcolumntype{H}{>{\setbox0=\hbox\bgroup}c<{\egroup}@{}}
\usepackage{amsfonts}       
\usepackage{amsmath}
\usepackage{amssymb}
\usepackage{mathtools}
\usepackage{flexisym}
\usepackage{breqn}
\usepackage{nicefrac}       
\usepackage[activate={true,nocompatibility},final,tracking=true,kerning=true,factor=1100,stretch=10,shrink=10]{microtype}
\microtypecontext{spacing=nonfrench}
\usepackage{xspace}
\usepackage[inline]{enumitem} 
\usepackage{caption}
\usepackage{subcaption}
\usepackage{csquotes}
\usepackage[dvipsnames]{xcolor}         
\usepackage{epsfig}
\usepackage{graphicx}
\usepackage{wrapfig}
\usepackage{rotating}
\usepackage{url}            

\usepackage[accsupp]{axessibility}  

\usepackage{subfiles} 

\usepackage{hyperref}       
\usepackage{bookmark}

\usepackage[capitalize]{cleveref}
\crefname{section}{Sec.}{Secs.}
\Crefname{section}{Section}{Sections}
\Crefname{table}{Table}{Tables}
\crefname{table}{Tab.}{Tabs.}

\newcommand{\persona}{}

\newcommand{\suppref}[1]{{#1}}

\newcommand*{\affaddr}[1]{#1}
\newcommand*{\affmark}[1][*]{\hspace{0.085em}\textsuperscript{#1}}

\DeclareMathOperator*{\argmax}{arg\,max}

\makeatletter
\newcommand{\maketitlepage}{%
    \let\orignaltitle\@title
    \gdef\@title{Supplementary Materials: \orignaltitle}%
    \let\thanks\@gobble
    \let\footnote\@gobble
    \if@twocolumn
      \ifnum \col@number=\@ne
        \@maketitle
      \else
        \twocolumn[\@maketitle]%
      \fi
    \else
      \@maketitle
    \fi
}
\makeatother

\title{In-Context Impersonation Reveals Large Language Models' Strengths and Biases}

\author{%
  Leonard Salewski\affmark[1,2]\\
  \And
  Stephan Alaniz\affmark[1,2]\\
  \And
  Isabel Rio-Torto\affmark[3,4]\thanks{Work done during a research visit at the University of Tübingen}\\
  \AND
  Eric Schulz\affmark[2,5]\\
  \And
  Zeynep Akata\affmark[1,2]\\
  \AND
  \vspace{-5mm}\\
  \affaddr{\affmark[1] University of Tübingen}\hspace{12pt}
  \affaddr{\affmark[2] Tübingen AI Center}\hspace{12pt}
  \affaddr{\affmark[3] University of Porto}\\
  \affaddr{\affmark[4] INESC TEC}\hspace{12pt}
  \affaddr{\affmark[5] Max Planck Institute for Biological Cybernetics}\hspace{5pt}
}

\hypersetup{
pdftitle={In-Context Impersonation Reveals Large Language Models' Strengths and Biases},
pdfauthor={Leonard Salewski, Stephan Alaniz, Isabel Rio-Torto, Eric Schulz, Zeynep Akata},
pdfkeywords={In-Context Impersonation, Large language models}
}

\begin{document}

\maketitle

\begin{abstract}
In everyday conversations, humans can take on different roles and adapt their vocabulary to their chosen roles. We explore whether LLMs can take on, that is impersonate, different roles when they generate text in-context. We ask LLMs to assume different personas before solving vision and language tasks. We do this by prefixing the prompt with a persona that is associated either with a social identity or domain expertise. In a multi-armed bandit task, we find that LLMs pretending to be children of different ages recover human-like developmental stages of exploration. In a language-based reasoning task, we find that LLMs impersonating domain experts perform better than LLMs impersonating non-domain experts. Finally, we test whether LLMs' impersonations are complementary to visual information when describing different categories. We find that impersonation can improve performance: an LLM prompted to be a bird expert describes birds better than one prompted to be a car expert. However, impersonation can also uncover LLMs' biases: an LLM prompted to be a man describes cars better than one prompted to be a woman. 
These findings demonstrate that LLMs are capable of taking on diverse roles and that this in-context impersonation can be used to uncover their strengths and hidden biases. 
Our code is available at \url{https://github.com/ExplainableML/in-context-impersonation}.
\end{abstract}

\section{Introduction}
Large Language Models (LLMs) can not only summarize documents and converse on a large range of topics~\cite{brown2020language}, but they have also shown other emergent abilities~\cite{webb2022emergent,wei2022emergent}. Because of their impressive abilities, LLMs are permeating into many applications~\cite{kasneci2023chatgpt,bommasani2021opportunities}. This means that there is a societal need to understand how these models ``tick''~\cite{tamkin2021understanding,bender2021dangers}. Traditionally, LLMs are provided with a context as a textual prompt and are asked to provide answers via text completion, thereby solving a variety of choice-based~\cite{binz2023using}, description-based~\cite{pilault2020extractive}, and reasoning tasks~\cite{wei2022chain}. Yet how in-context learning works is not fully understood. 
When Min et al.~\cite{min2022rethinking} prompted LLMs with random labels, they found that this did not drastically degrade performance, suggesting that the role of in-context demonstrations is to prime the model for a particular task. This is in line with other results suggesting that LLMs internally infer latent variables to make better predictions~\cite{xie2022bayesianexplanation}. It has been suggested that LLMs, and other large models, can change their behavior when asked to respond as a particular persona. When Deshpande et al.~\cite{deshpande2023toxicity} asked LLMs to respond as a hateful person, their toxicity score increased. When Wang and colleagues~\cite{wang2023can} asked LLMs to imagine being expert systematic reviewers, the quality of their literature search queries increased. That LLMs can impersonate specific people is also known; they can, for example, pretend to be Oscar Wilde, Carrie Bradshaw from Sex and the City, or Donald Trump~\cite{elkins2020can}. But how does in-context impersonation affect LLMs' behavior in language-based and other downstream tasks?

In the current work, we let LLMs impersonate, that is taking on different roles, in context. We do this by prefixing the prompt with \textit{``If you were a \{persona\}''} where \textit{persona} is replaced with the persona that the LLM is asked to impersonate. These personas are associated either with a social identity or a domain of expertise. In a first simulation using a multi-armed bandit task~\cite{binz2022modeling}, we find that LLMs impersonating children of different ages can recover the developmental stages of human-like exploration strategies. In language-based reasoning tasks, we find that LLMs impersonating domain experts perform better than LLMs impersonating non-domain experts. Finally, we ask LLMs to describe different classes of either birds or cars and then use their descriptions in a downstream, visual classification task. The results of this experiment corroborate our earlier results: LLMs become better as they pretend to be older, and they are also better when they pretend to be domain experts.
However, we also see how impersonating LLMs reproduce biases: LLMs impersonating a black person or a male describe cars better, while LLMs impersonating a white person or a female describe birds better. These results expand our understanding of in-context learning in LLMs and open up new research directions investigating role-taking and pretense in LLMs and beyond.

\section{Related Work}
In-context learning refers to an LLM's ability to improve at a given task after being provided with a number of task-relevant demonstrations~\cite{brown2020language}. This ability sets LLMs apart from traditional models and has led to a totally new paradigm -- one which does not require fine-tuning of weights on task-specific data but instead relies entirely on contextual information~\cite{lampinen2022can,wei2022chain,arora2022ask}.

This contextual information is normally delivered as textual prompts~\cite{zhou2022large}, where a task or scenario is described and a model is asked to solve the task or reason about the scenario by generating the next words of the provided text. Due to its flexibility, prompting has been widely used as a generic method for natural language tasks~\cite{schick2020exploiting,sanh2021multitask}. Importantly, the resulting in-context learning does not only work after LLMs have seen some examples, i.e.\ in the few-shot regime~\cite{wang2020generalizing}, but also without any examples, i.e.\ in the zero-shot regime~\cite{xian2018zero}. LLMs are reasonably proficient at solving arithmetic~\cite{yuan2023well} or reasoning tasks~\cite{kiciman2023causal} without having been prompted with example solutions but only after being asked to provide an answer to a given problem. 
LLMs can require careful engineering of the provided prompts, either manually~\cite{reynolds2021prompt} or automatically~\cite{sshin2020autoprompt}. Indeed, whole books have been written to provide guidelines on how to best perform prompt engineering~\cite{hunter2023art}, especially because engineering prompts can require a great amount of expertise~\cite{oppenlaender2023prompting}. 

One method known to influence LLMs behavior is to ask them to respond as a particular person~\cite{han2022meet,keskar2019ctrl}, an effect which is also described as role-taking~\cite{Shanahan2023RolePlayWL}. LLMs can take in the text of one famous author, e.g.\ Oscar Wilde, and rewrite it in the style of another famous author, e.g.\ James Joyce~\cite{yang2018unsupervised}. This is not only true for LLMs but for any large model that provides results based on prompts, such as text-to-image models~\cite{Crowson2022VQGANCLIPOD, Nichol2021GLIDETP,saharia2022photorealistic}. For example, using the artist's name for generative art prompting is known to boost the quality~\cite{oppenlaender2023prompting} or to substantially affect the style~\cite{Dehouche2023WhatsIA,Brack2022TheSA,Witteveen2022InvestigatingPE} of the generated images. To make LLMs respond more truthfully, Lin and colleagues introduced scenarios from the perspective of a fictional persona called ``Professor Smith''~\cite{lin2021truthfulqa}. Conversely, to make LLMs act maliciously, Wolf et al.~\cite{wolf2023fundamental} prompt LLMs adversarially to overcome alignment techniques. LLMs can also be used to simulate multiple humans which changes how they cooperate in economic games~\cite{aher2022using}.

LLMs can also have their own ``personalities'' which can be evoked in-context~\cite{pellert2023ai}. Although LLMs frequently behave like the average person~\cite{park2023correct}, their personality profiles can be tinkered with~\cite{karra2022ai}, e.g.\ by changing the context to be more or less emotional~\cite{coda2023inducing}. This has led researchers to use LLMs to simulate survey responses~\cite{DominguezOlmedo2023QuestioningTS} of subpopulations by conditioning them on socio-demographic descriptions~\cite{argyle2022out} or to ask them to respond in persona when writing about fictitious childhood events~\cite{jiang2023personallm}. 
Additionally, non-deterministic tasks such as open-ended questions have also been explored~\cite{Aher2022UsingLL}.

Semantics derived automatically from language corpora can contain human-like biases~\cite{caliskan2017semantics}. Thus, LLMs do not only reproduce human-like text but also replicate biases present in the training data~\cite{bender2021dangers,abid2021persistent}. Importantly, these biases can get exacerbated if LLMs are asked to provide answers in persona~\cite{coda2023inducing,deshpande2023toxicity,kang2023exploiting}.

LLMs are naturally combined with large vision-language models (VLMs)~\cite{jia2021align,singh2022flava} such as CLIP~\cite{radford2021clip} due to their versatility in a wide range of visual recognition tasks. Menon et al.~\cite{menon2023visual} used GPT-3~\cite{brown2020language} to generate a diverse set of short descriptions of a class that improve zero-shot classification when their CLIP scores are combined. Similarly, Yang et al.~\cite{Yang2022LanguageIA} used GPT-3 descriptions of classes as concept bottlenecks for interpretable image classification. LLMs can also be used as a knowledge base for visual question-answering (VQA) tasks~\cite{yang2022vqa}.

\section{In-context Impersonation Methodology}
Our methodology is composed of two steps. First, we prompt and query the LLM\@. Second, we evaluate the resulting text queries in three tasks, i.e.\ two-armed bandit, reasoning, and visual classification.

\subsection{Prompting and Querying the Large Language Model with Personas}
LLMs are trained to predict the most probable next token $t_k$ given previous tokens $t_1 \dots t_{k-1}$ by maximizing the likelihood function $p_{\text{LLM}}(t_k|t_1, \dots, t_{k-1})$. In this work, we use pre-trained LLMs without further finetuning them. Depending on the task, we generate one or more tokens given a task-specific context $\boldsymbol{c}$ that describes the task to the language model and prompts it for an answer. The context includes the instruction to impersonate using the phrase \emph{``If you were a \{persona\}''} where persona $p$ is replaced by the persona name.
Thus, we obtain generated tokens $\boldsymbol{t}$ by sampling from 
\begin{equation}
p_{\text{LLM}}(\boldsymbol{t}|\boldsymbol{c}^{(p)}) = \prod_{k=1}^K p_{\text{LLM}}(t_k|c_1^{(p)}, \dots, c_n^{(p)}, t_1, \ldots ,t_{k-1})    
\end{equation}
We refer to this type of contextualization as \emph{in-context impersonation}.

{\bf Personas Considered.}
The first interesting question to look at was if LLMs could impersonate the behavior of differently aged people. For this, we ask the LLM to imagine it is either a 2, 4, 7, 13, or 20-year-old.
We also evaluate whether the LLM is able to impersonate different fields of expertise. Depending on the task considered, the expertise profiles differ (more details below).
Finally, we evaluate whether LLMs have biases regarding gender and skin color. For this, we asked LLMs to imagine that they were either  a \persona{man} or a \persona{woman} or a \persona{black person} or a \persona{white person}. 

{\bf Large Language Models Considered.}
In this work, we evaluate two LLMs. For all of our tasks, we used the Vicuna-13B language model~\cite{chiang2023vicuna} which has 13 billion parameters and was trained to follow natural language instructions. Vicuna is a fine-tuned version of the LLAMA language model~\cite{touvron2023llama} using ShareGPT~\cite{shareGPT2023}
conversational data. We use an instruction fine-tuned model because it was optimized to follow user prompts.
Its weights are publicly available, allowing us to run the model locally. Vicuna is competitive with proprietary services such as ChatGPT in some domains~\cite{Zheng2023JudgingLW}\footnote{\url{https://chat.lmsys.org/?leaderboard}}.
In addition to Vicuna, we use the OpenAI API of ChatGPT~\cite{Ouyang2022TrainingLM} with the \texttt{gpt-3.5-turbo} model for the reasoning and vision tasks.
For the bandit task, however, running 12k games with 10 trials each is infeasible.

We do not further train the models, nor do we provide sample solutions in-context; thus, all experiments are conducted in a zero-shot fashion. By providing minimal guidance to perform the task, we avoid pre-conditioning the model such that answers can better reflect the internalized language of the LLM instead of relying on few-shot examples.
When sampling full sentences, we use a temperature of 0.7; to obtain the answer as a single symbol (token), we set it to 1 unless otherwise stated.
These different temperatures were chosen based on the recommended default values of each LLM\@.

\begin{figure}
    \centering
    \includegraphics[width=\linewidth]{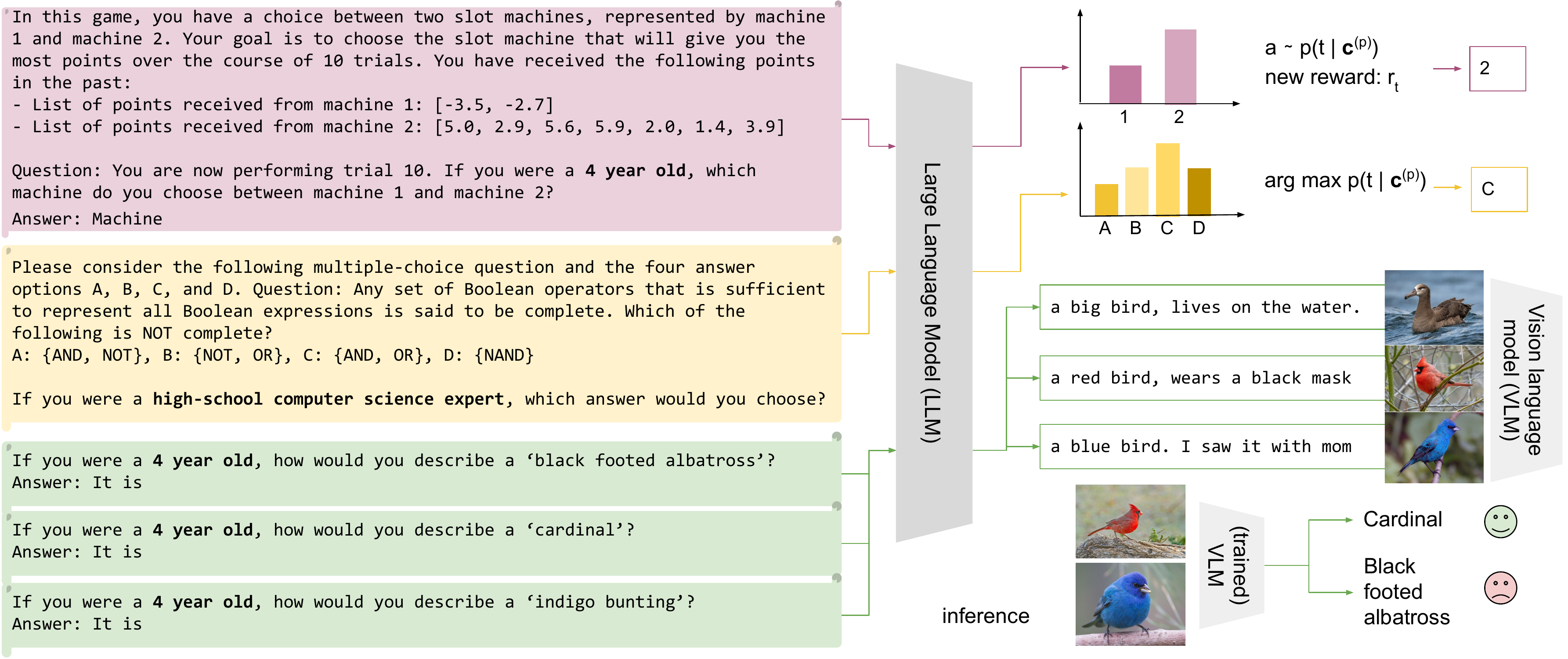}
    \caption{Our three tasks are designed to analyze the effect of \emph{in-context impersonation}. First, we investigate bandit tasks (pink) where the LLM must maximize the reward while impersonating different age groups. Second, we evaluate the effect of domain expert impersonation on natural language reasoning tasks (yellow). Third, we study the usefulness of descriptions generated with impersonation w.r.t.\ age, expertise, ethnicity, and gender for visual classification (green). 
    }%
    \label{fig:model}
\end{figure}

\subsection{Bandit Task Design}%
\label{subsec:bandit}
We asked LLMs to imagine being in different personalities while participating in a multi-armed bandit task~\cite{gershman2018deconstructing} taken from the psychology literature~\cite{schulz2019algorithmic} and already applied to LLMs~\cite{binz2023using}. 

An agent gets to interact with a two-armed bandit problem for $10$ trials. The mean reward for each arm $a$ is drawn from $p(\theta_a) = \mathcal{N}(0, 10)$ at the beginning of a task, and the reward for each trial is drawn from $p(r_t | a_t, \theta_{a_t}) = \mathcal{N}(\theta_{a_t}, 1)$. Feedback of past trials is provided via prompt-chaining, i.e.\ concatenating previous choices and their outcomes to the current prompt submitted to the LLM\@. We analyze the set of emerging exploration strategies, assuming that an agent uses Bayes’ rule to update its beliefs over unobserved parameters. If prior and rewards are normally distributed, then the posterior will be normally distributed and the corresponding updating rule is given by the Kalman filtering equations. Let $p(\theta_{a} | h_t) = \mathcal{N}(\mu_{a, t}, \sigma_{a, t})$ be the posterior distribution at time-step $t$. Based on the parameters of this posterior distribution, one can define a probit-regression model:
\begin{align} 
\label{eq:full}
    p(A_t = 1 | \mathbf{w}) = \pmb{\Phi}\left(\beta_1 \text{V}_t + \beta_2 \text{RU}_t\right)  
\end{align}
with $\pmb{\Phi}$ denoting the cumulative distribution function of a standard normal distribution. Here, $\text{V}_t = \mu_{1, t} - \mu_{2, t}$ represents the estimated difference in value and $\text{RU}_t = \sigma_{1, t} - \sigma_{2, t}$ the relative uncertainty. One can use Equation~\ref{eq:full} to analyze how much an agent engages in exploitation behavior by inspecting $\beta_1$ and how much the agent uses uncertainty to explore in a directed fashion by inspecting $\beta_2$~\cite{binz2022modeling}. 

For this bandit task, we consider personas of different ages. Specifically, we study ages 2, 4, 7, 13, and 20 to cover key developmental stages of early childhood, childhood, adolescence, and adulthood where the learning progress is most pronounced in humans.
The language model is prompted (see Figure~\ref{fig:model}, the pink path) to only answer ``1'' or ``2'' depending on which arm $a$ it would like to choose. The LLM receives rewards and the associated actions from previous trials inside the context in the form of a list.

With $\log d_{a_t} = \log p_{\text{LLM}}(t_1=a_t|\boldsymbol{c}^{(p)}, a_1, \dots, a_{t-1}, r_1, \dots, r_{t-1})$ being the unnormalized logits from the LLM for the token of arm $a$, for each trial we sample an action $\hat{a} \sim \sigma({\{\log d_{a_t}\}}_{a_t=1}^A)$ where we have two arms $A=2$. We do not apply temperature scaling in this case as we are only sampling a single token and want it to reflect the LLM decision-making as faithfully as possible.

\subsection{Reasoning Task Design}%
\label{subsec:reasoning}
In our reasoning task, the LLM has to answer a multiple-choice question regarding a given topic from the Multitask Language Understanding (MMLU) dataset~\cite{hendrycks2021measuring}, commonly used to benchmark LLMs~\cite{touvron2023llama}.
The MMLU dataset consists of 57 tasks from Science, Technology, Engineering, and Mathematics (STEM), Humanities, Social Sciences, and Other, ranging from elementary, high school, college, and professional levels of complexity. 
We start by prompting the LLM with the context:

{\begin{center}
    \texttt{
    Please consider the following multiple-choice question and the four answer options A, B, C, and D. 
    Question: \{task\} \\
    If you were a \{persona\}, which answer would you choose?}
\end{center}}

The \emph{task} is replaced by the question and the 4 possible answers, while the \emph{persona} is replaced by an expert (see Figure~\ref{fig:model}, the yellow path).
We consider three types of experts as personas. The task expert, e.g.\ for the high school computer science task, is ``\persona{high school computer science expert}''.
The domain expert is an aggregation of all the remaining experts in the same field as the task expert (but not the task expert himself), e.g.\ for high school computer science it would be any other STEM expert.
The non-domain expert is an aggregation of the task experts from the other domains, e.g.\ for high school computer science it would be all Humanities, Social Sciences and Other experts.

After feeding the prompt to the LLM, the LLM prediction of the first token following the context is $d = p_{\text{LLM}}(t_1|\boldsymbol{c}^{(p)})$ and the $N$ tokens for the possible answers of the multiple choice question are $o = {\{o_i\}}_{i=1}^N$ which in this case are A, B, C, and D. The predicted option is then given by 
\begin{equation}
    \hat{o} = \arg \max(\hat{c}_i), \text{ with } \hat{c}_i = d[c_i], i=1 \ldots N
\end{equation} 
which are the predicted probabilities of the language model.
With this approach, we are able to obtain the option with the highest probability according to the LLM and, thus, compare it with the ground truth label to measure the accuracy resulting from different in-context impersonations.

\subsection{Vision and Language Task Design}\label{subsec:vl-tasks}
Lastly, we want to evaluate the usefulness of descriptions generated by \emph{in-context impersonation} for downstream vision and language tasks.
We focus on challenging fine-grained classification tasks, as the generated descriptions need to be domain specific for these tasks to succeed. 
We ask the LLMs to generate a description of a class, from the perspective of a persona.
Our prompt is: 
{\begin{center}
    \texttt{
    If you were a \{persona\}, how would you answer the following question \\ in 45 words?
    Q: What is a/an \{class_name\}?
    A: It is}
\end{center}}

To avoid trivial solutions, i.e.\ the class name being mentioned in the description, we post-process the generated descriptions with a two-step approach: first, we replace class names used in noun phrases with an appropriate pronoun whilst respecting the given numerous. Second, if the class name is still not removed, we re-use the same language model to process the descriptions sentence by sentence. For this, we use 4 in-context examples, that demonstrate how to remove the class name information.
The full process is documented in suppl.\ Section \suppref{D.1}.

{\bf Vision-Language Models (VLMs).}
We use CLIP (or variants thereof)~\cite{radford2021clip,Cherti2022ReproducibleSL} to perform fine-grained visual classification as a means to evaluate the usefulness of the generated descriptions.
CLIP models are trained with contrastive image-text matching losses to rank matching image and text inputs highly and non-matching inputs lowly.
\cite{radford2021clip,Cherti2022ReproducibleSL} show that CLIP variants generalize well to match unseen texts, e.g.\ class names, an ability commonly referred to as zero-shot classification.

First, the image to classify is converted into a normalized feature representation $I$ using CLIP's pre-trained vision backbone. Then, the class names are embedded into normalized feature vectors $T_N$ using the pre-trained text backbone.
Next, all pairwise cosine similarities $I \cdot T_N$ of the respective feature representations are computed.
Finally, the $n^* = \argmax_N (I \cdot T_N)$ over these similarities reveals the most similar class $n^*$.

{\bf Inference.}
We generate a description $D_n^{(p)}$ with the above prompt for each class $n$ for each persona $p$ where we use a generative approach, i.e.\ we auto-regressively sample a random token from the predicted logits (see \Cref{fig:model}, the green path). For Vicuna-13B we use the default temperature of 0.7 and the default top-k value of $k=50$.
For ChatGPT we use the default temperature of 1.0.
This continues until the model emits an \textit{<end of sequence>} or the maximum number of tokens (96) is reached.
We did not tune these values.

For visual classification, we use the zero-shot classification capabilities of CLIP models, but instead of using the embedded class name itself ($T_n$), we use the embedding of the generated descriptions $D_n^{(p)}$ for each class $n$ and for each persona $p$.
The predicted class for each persona ${i^{(p)}}^*$ is:
\begin{align}
    {n^{(p)}}^* = \argmax (I \cdot D_n^{(p)})
\end{align}
Performance is measured by computing the classification accuracy of the test splits on both datasets.
As the descriptions are sampled from the LLM output, the results of the experiments are stochastic and we repeat them five times.
We report the mean performance as well as 95\% confidence intervals.

\section{Experiments}
Using Vicuna-13B, we evaluate the two-armed bandit and MMLU language reasoning tasks. For the zero-shot image classification task using a VLM we generate descriptions with both Vicuna-13B and ChatGPT\@. We focus on highlighting how the chosen persona changes the task performance of the LLM\@.
As LLMs seem to be sensitive to prompts~\cite{Arora2022AskMA}, we follow the meta-prompting approach from~\cite{reynolds2021prompt} to vary our impersonation prompts. We run all Vicuna-13B experiments with each of the six prompt variations, which are shown in the suppl.\ Section \suppref{A.1}.
All experiments are performed on the test splits using a single A100-40GB GPU and we mention inference times in suppl.\ Section \suppref{A.2}. 

\subsection{Age-based impersonation changes exploration strategies}
In the bandit task, for every age group that the LLM impersonates, we perform 2k two-armed bandit games of 10 trials each for each prompt variation. We evaluate the task performance in three ways. 

\begin{wrapfigure}[22]{r}{0.4\textwidth}
    \vspace{-2.5ex}
    \centering
    \includegraphics[width=0.35\textwidth,trim={7 0 8 2},clip]{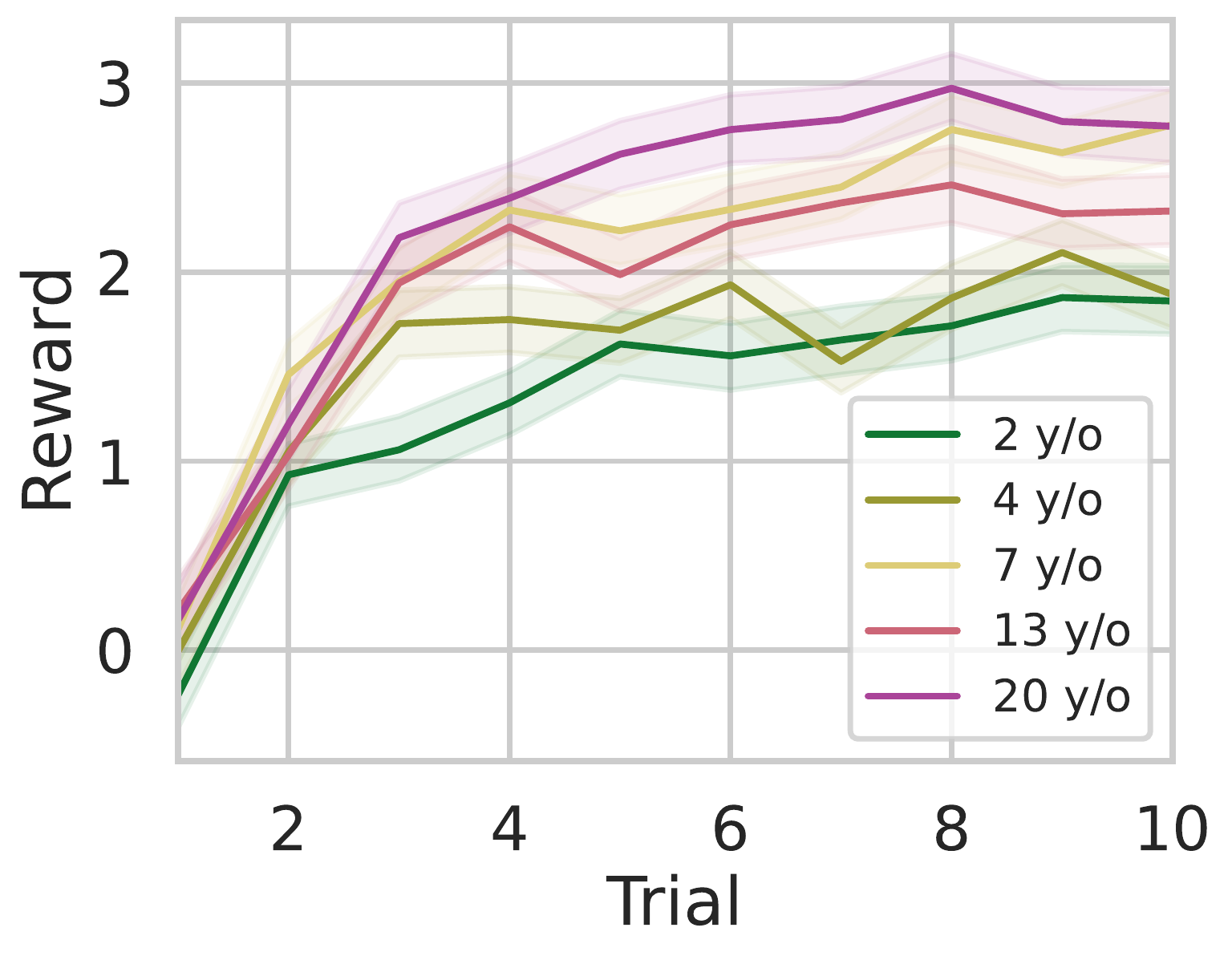}\\
    \begin{subfigure}{0.18\textwidth}
         \centering
         \includegraphics[width=\textwidth]{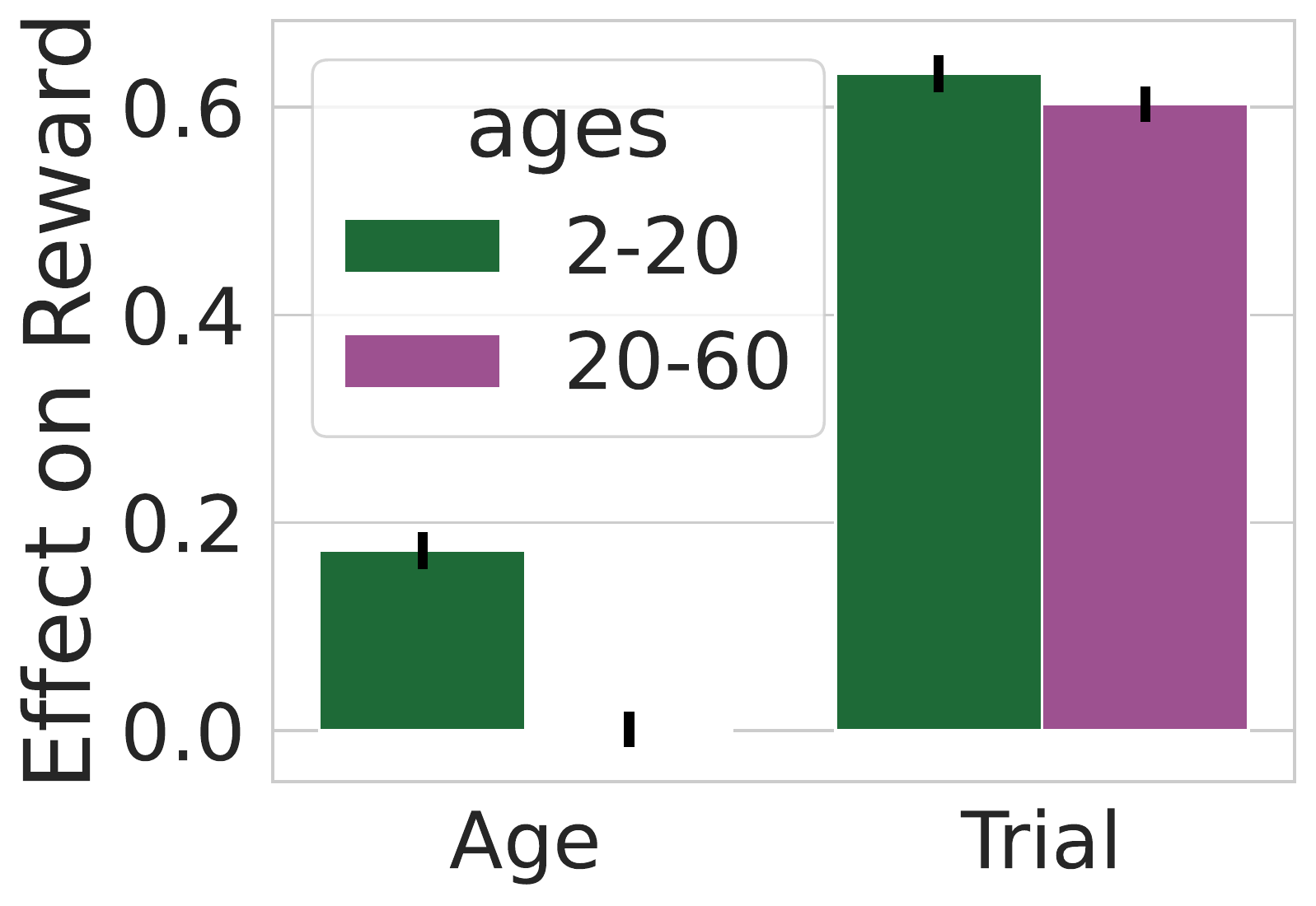}
     \end{subfigure} 
     \hfill
     \begin{subfigure}{0.2\textwidth}
         \centering
         \includegraphics[width=\textwidth]{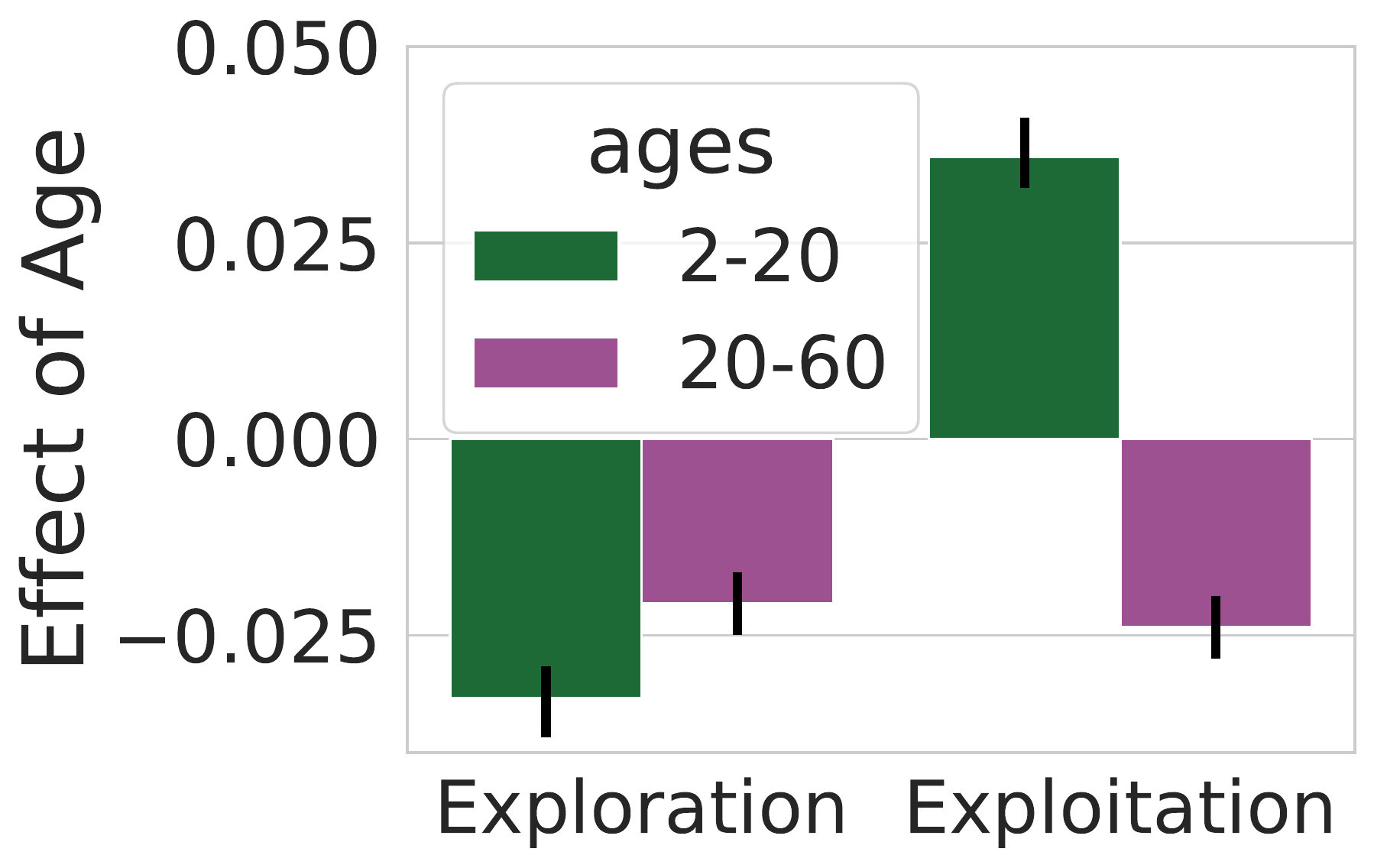}
     \end{subfigure} 
    \caption{Two-armed bandit task. Top: Average reward per persona (10k games of 10 trials), left: Age and \# of trials have a positive effect on the expected reward, right: With age, exploration decreases, and exploitation increases.}%
    \label{fig:bandits}
\end{wrapfigure}
First, we show the average reward per trial the LLM obtained with personas of increasing age in Figure~\ref{fig:bandits} (top). With an increasing number of trials, the LLM obtains a higher average reward, corroborating that Vicuna-13B is able to learn from past trials to improve its policy similarly to GPT-3 in~\cite{binz2023using}. Moreover, as the LLM takes on a persona of different ages, we observe a divergence of obtained rewards as the number of trials increases. Younger personas, i.e., \persona{2-} and \persona{4-year-old} personas, obtain a smaller reward than older ones, i.e., \persona{13-} and \persona{20-year-old} personas.

Secondly, we analyze the resulting rewards by using a regression, entering the trial number and age as independent variables. To extend the analysis, we evaluate two age groups, from 2 to 20 and from 20 to 60, where we evaluate ages in steps of 2 between 2 and 30 and steps of 5 from 30 to 60. We report these results in Figure~\ref{fig:bandits} (bottom left). We find that the impersonating LLMs generally improved over trials, i.e.\ they increase their rewards as they progressed over trials of a game ($\beta=0.63$, $p<.001$ for ages 2--20 and $\beta=0.60$, $p<.001$ for ages 20--60). Importantly, LLMs impersonating older participants generate higher average rewards until age 20 ($\beta=0.17$, $p<.001$), thereby replicating a general pattern found in the developmental literature~\cite{nussenbaum2019reinforcement}. We find no significant effect from ages 20--60, which also mirrors {observations of stagnating mental performance of adults.}

Lastly, we analyze how regression weights of the probit-regression were influenced by the age group the LLM is impersonating, again analyzing ages 2--20 and 20--60. Figure~\ref{fig:bandits} (bottom right) reveals that LLMs pretending to be older explored their environment less ($\beta=-0.03$, $p<.001$) and exploited more ($\beta=0.04$, $p<.001$) in the ages between 2--20.
This pattern is in line with several results from the psychological literature which also found that children show higher levels of directed exploration~\cite{Liquin2020ChildrenAM} than adults~\cite{schulz2019searching}.
These results suggest that impersonating LLMs can recover human-like developmental stages of exploration in a two-armed bandit task. If life is seen as an exploration-exploitation problem, then younger agents should show higher amounts of directed exploration~\cite{Giron2022DevelopmentalCI,Blanco2019SystematicEA}.
To the best of our knowledge we are the first to show that LLMs replicate similar trends when using in-context impersonation.

\subsection{Expertise-based impersonation changes reasoning abilities}
\begin{figure}[ht!]
     \centering
     \vspace{-3mm}
     \begin{subfigure}[t]{0.244\textwidth}
        \centering
         \begin{subfigure}[t]{\textwidth}
             \centering
             \includegraphics[width=\textwidth]{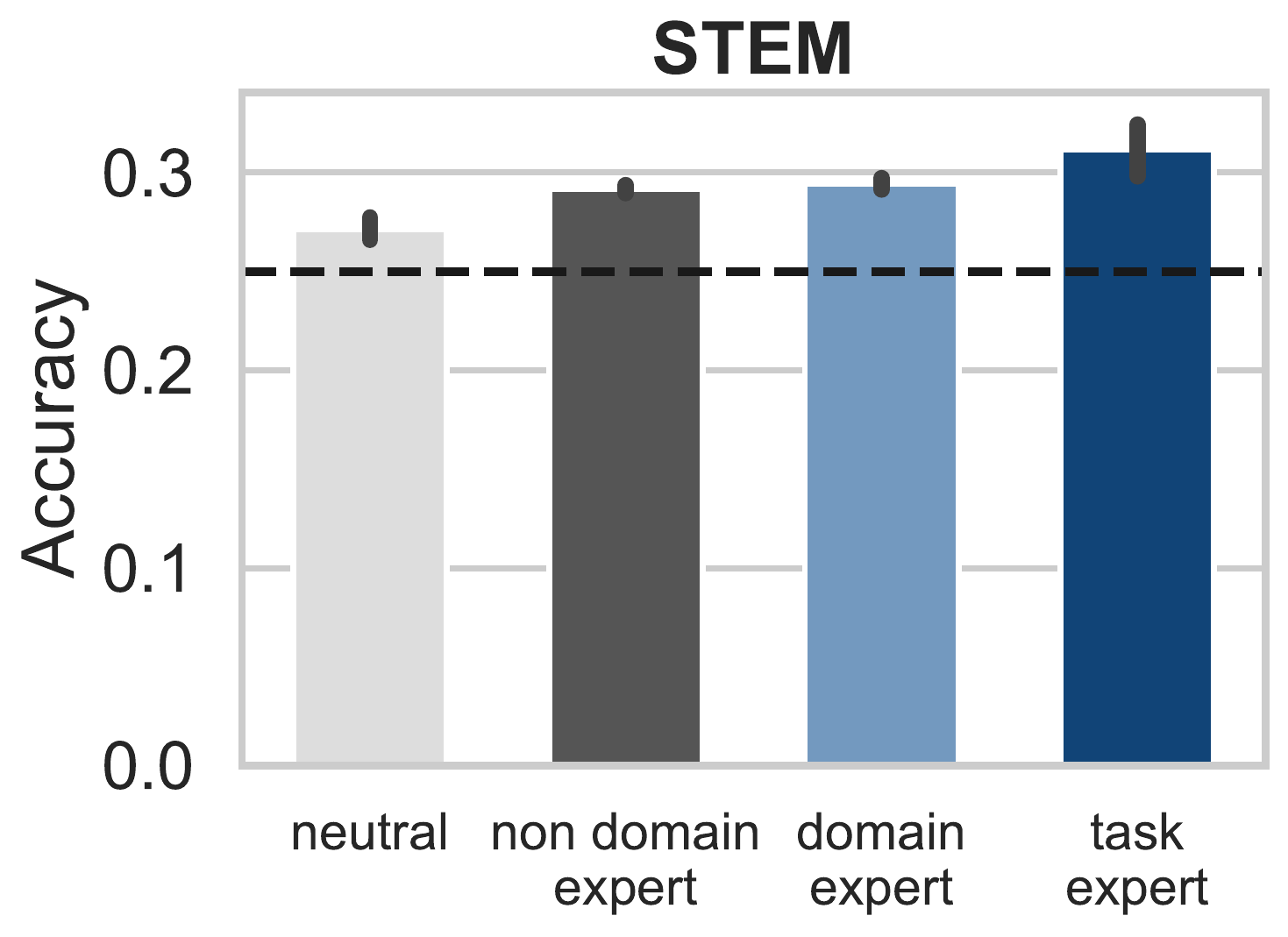}
         \end{subfigure}
         \\
         \begin{subfigure}[t]{0.85\textwidth}
             \centering
             \includegraphics[width=\textwidth]{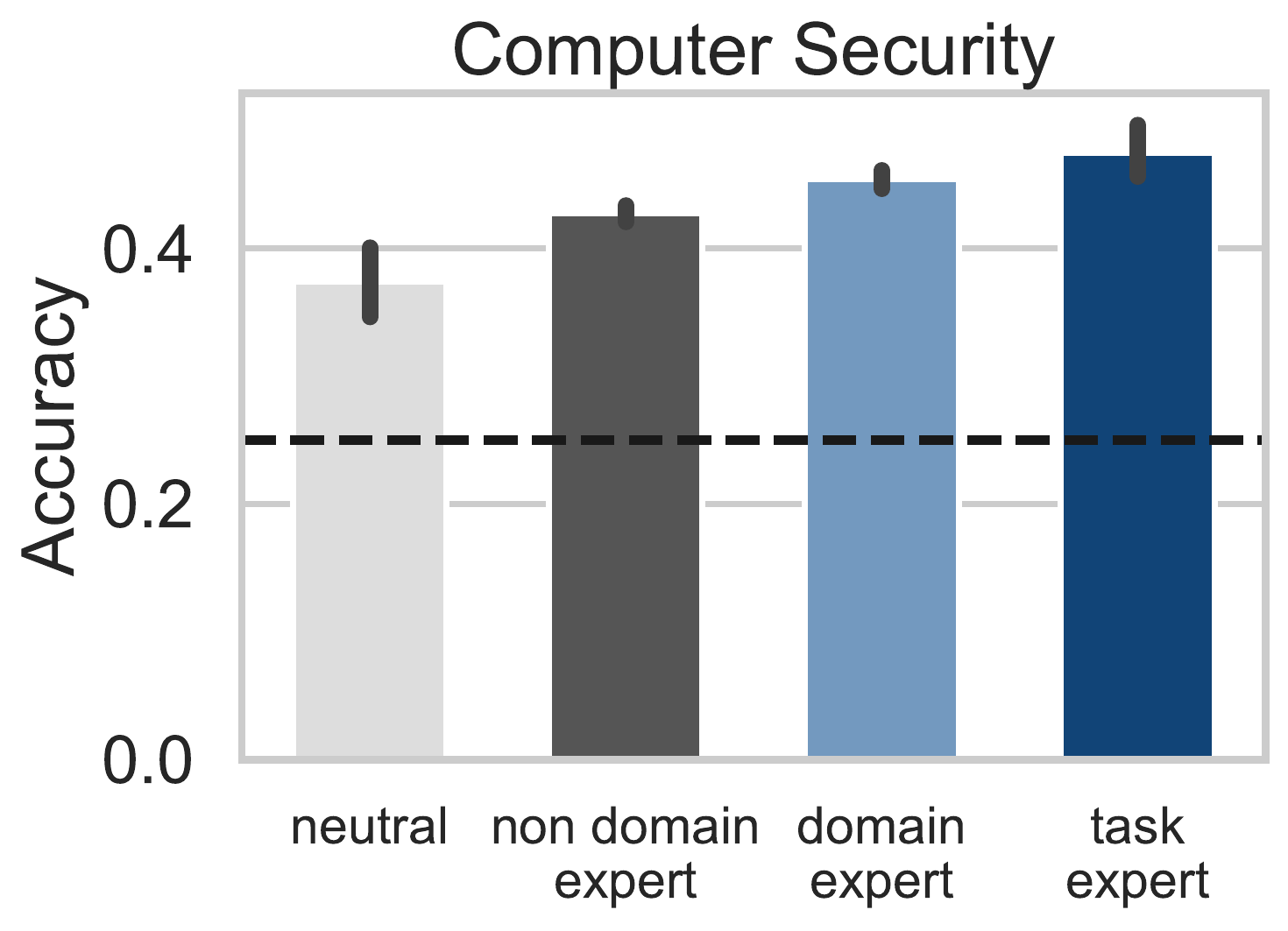}
         \end{subfigure}
    \end{subfigure} 
    \hfill
    \begin{subfigure}[t]{0.244\textwidth}
        \centering
         \begin{subfigure}[t]{\textwidth}
             \centering
             \includegraphics[width=\textwidth]{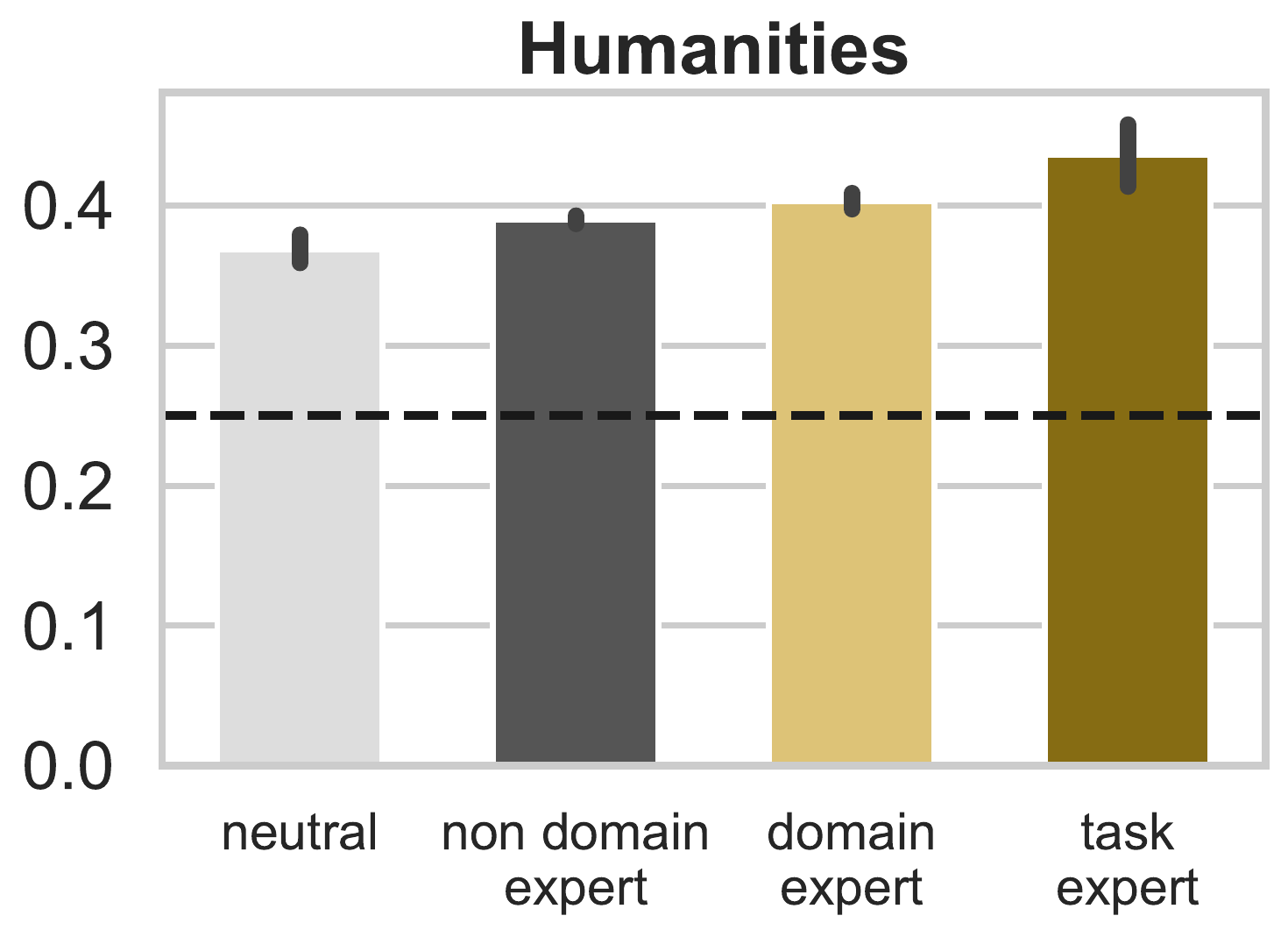}
        \end{subfigure}
        \\
        \begin{subfigure}[t]{0.85\textwidth}
             \centering
             \includegraphics[width=\textwidth]{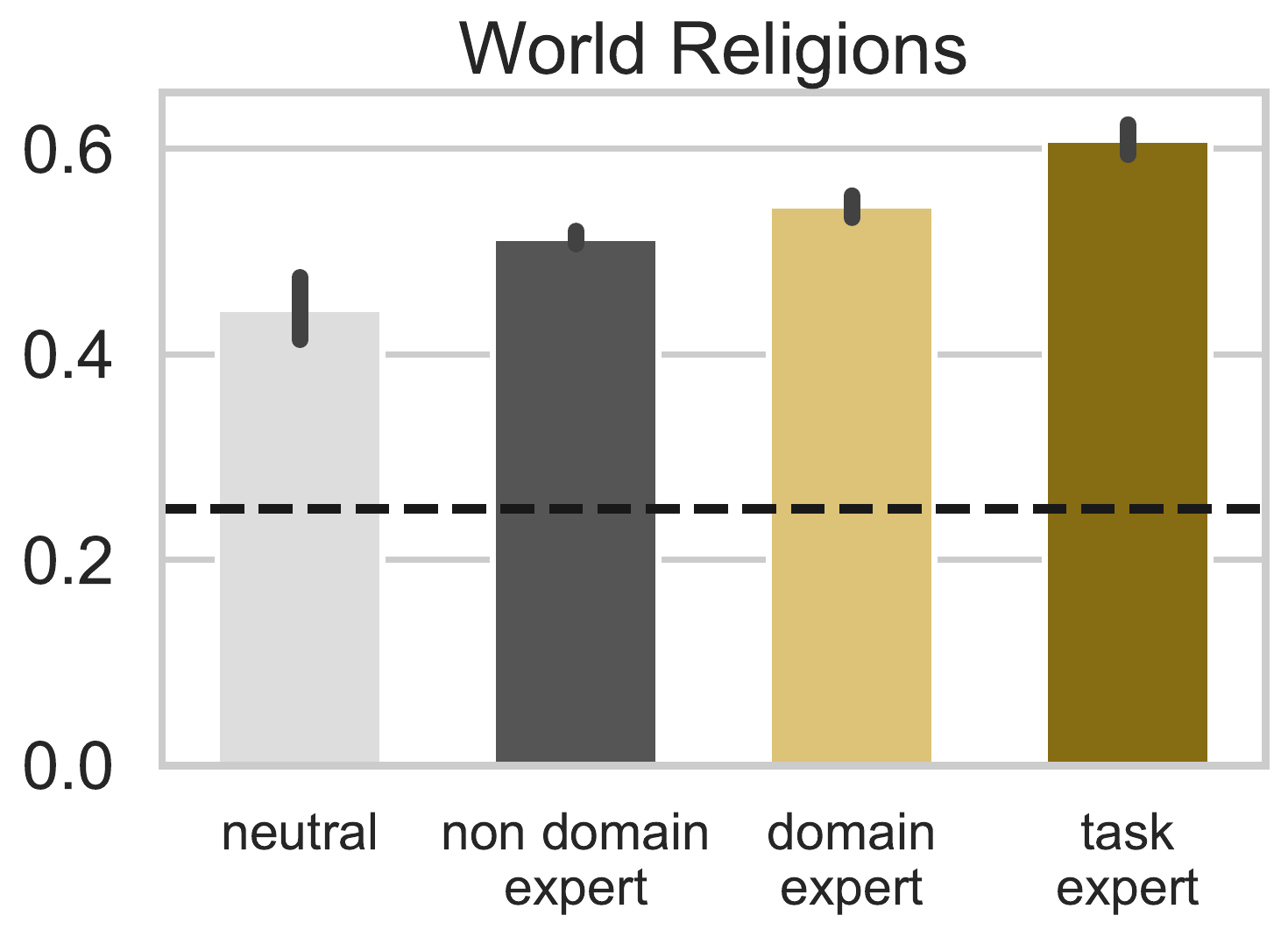}
         \end{subfigure}
    \end{subfigure}
    \hfill
    \begin{subfigure}[t]{0.244\textwidth}
        \centering
        \begin{subfigure}[t]{\textwidth}
             \centering
             \includegraphics[width=\textwidth]{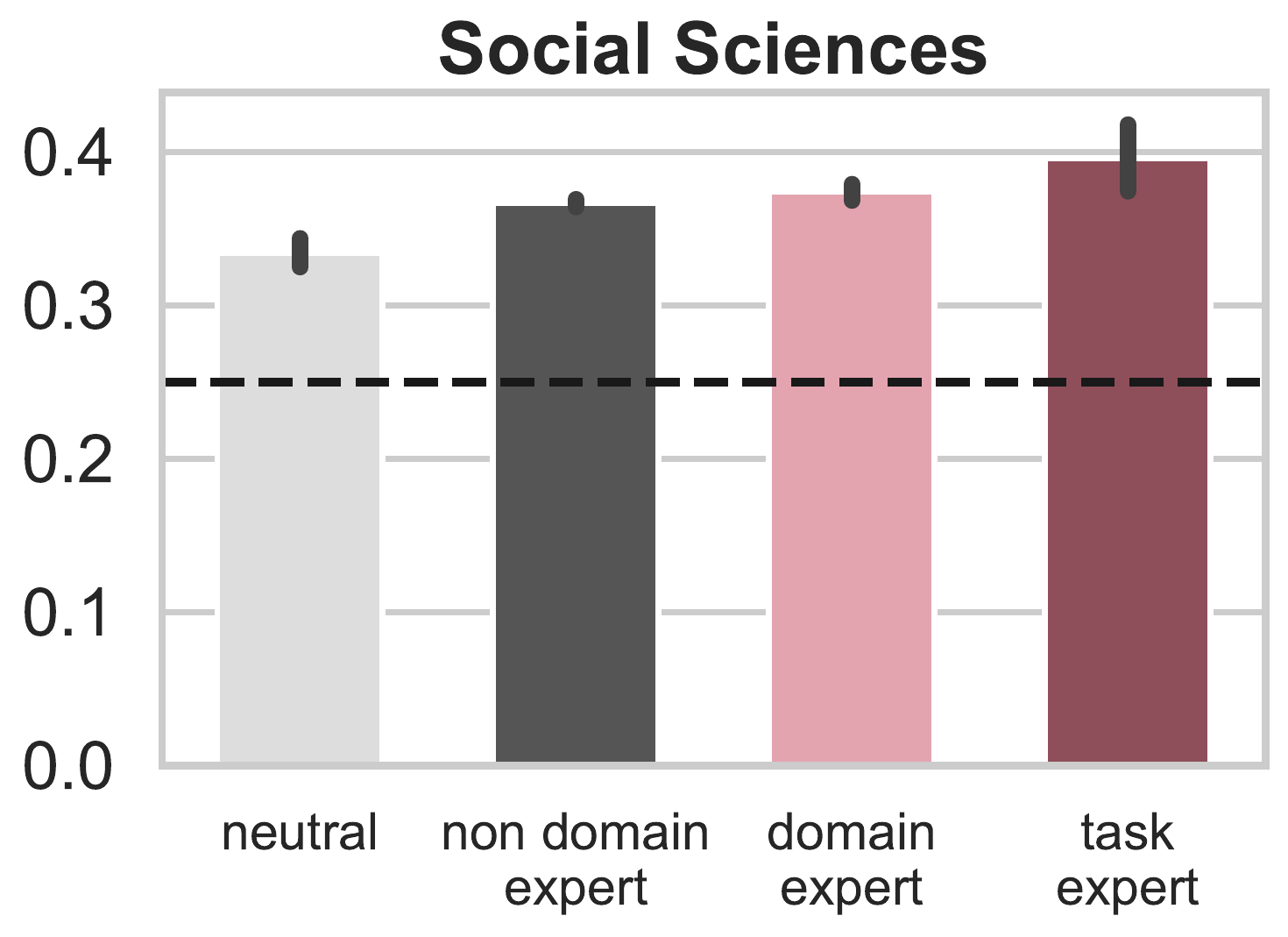}
        \end{subfigure}
        \\
        \begin{subfigure}[t]{0.85\textwidth}
             \centering
             \includegraphics[width=\textwidth]{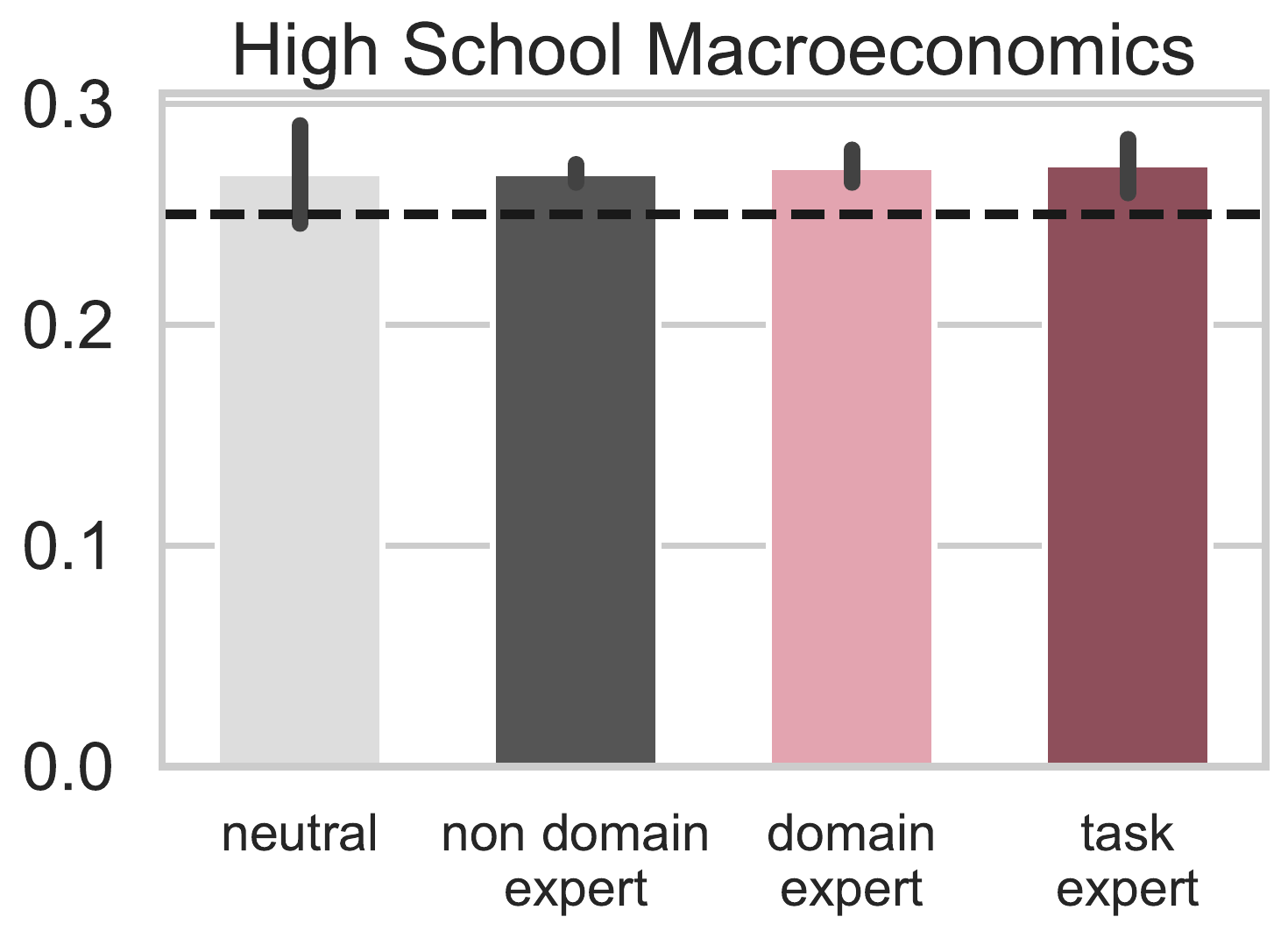}
         \end{subfigure}
    \end{subfigure}
    \hfill
    \begin{subfigure}[t]{0.244\textwidth}
        \centering
        \begin{subfigure}[t]{\textwidth}
             \centering
             \includegraphics[width=\textwidth]{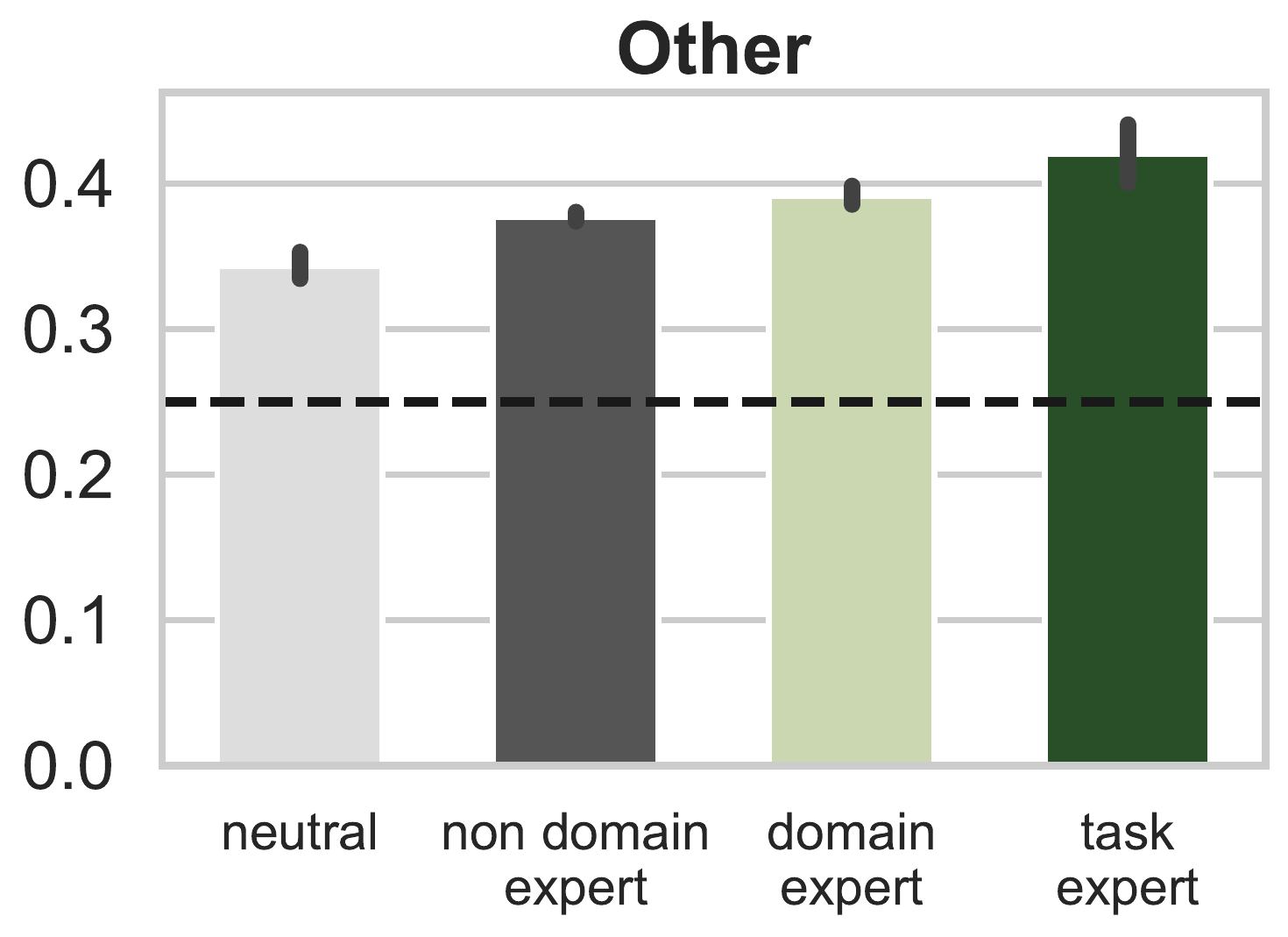}
        \end{subfigure}
        \\
         \begin{subfigure}[t]{0.85\textwidth}
             \centering
             \includegraphics[width=\textwidth]{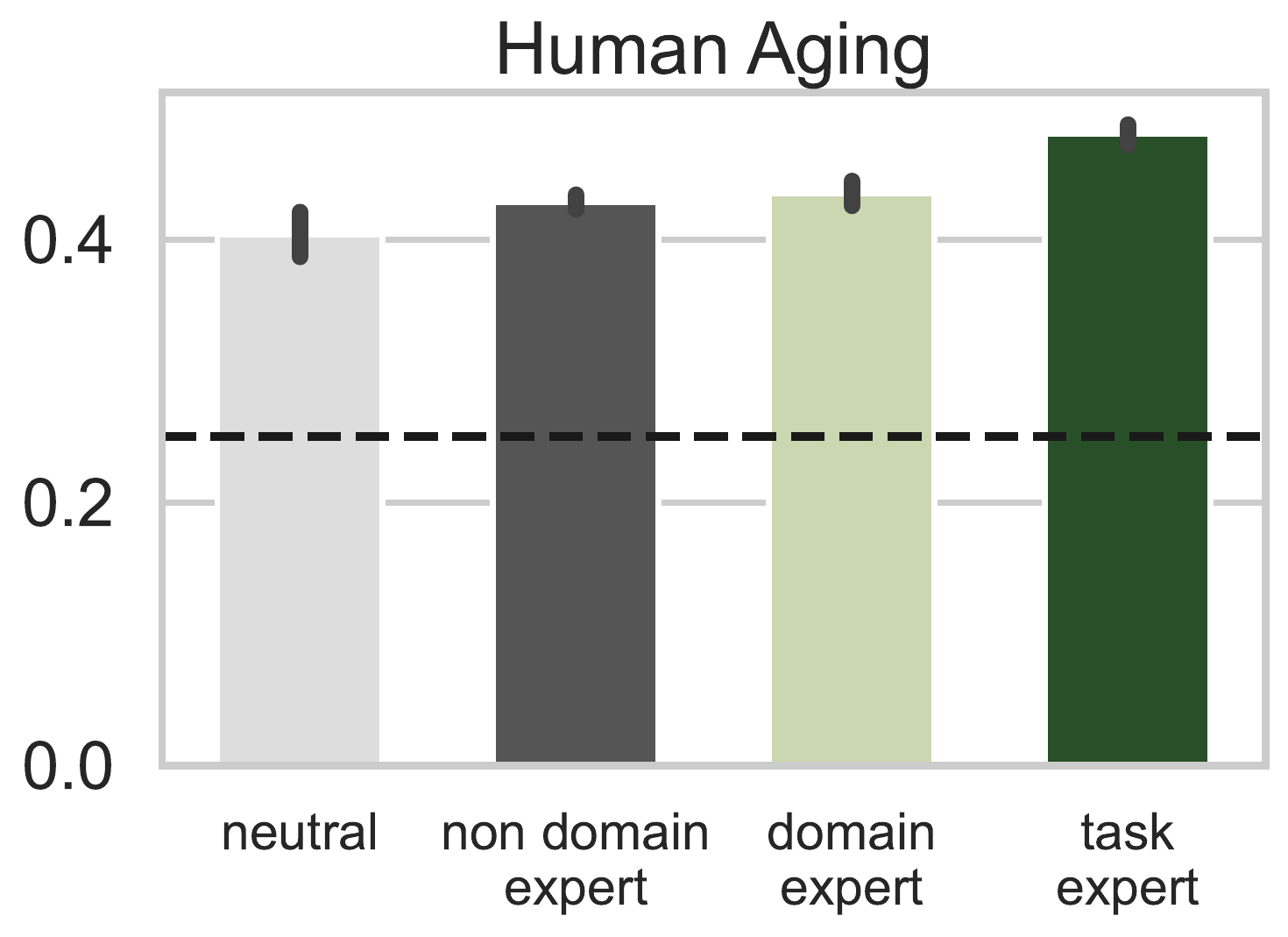}
         \end{subfigure}
    \end{subfigure}
     \caption{Expertise-based impersonation on all domains of the MMLU reasoning benchmark (top) and on exemplary individual tasks (bottom). For each task, we consider four personas: the neutral, the task expert, the domain experts (all experts from the same domain except the task expert) and the non-domain experts (all experts from all remaining domains). \mbox{The dashed line is the random baseline.}}%
    \label{fig:mmlu_experts}
    \vspace{-1ex}
\end{figure}
Our experiments on expertise-based impersonation (details in Section~\ref{subsec:reasoning}) are conducted on the MMLU dataset~\cite{hendrycks2021measuring}, for which we ask Vicuna-13B to impersonate experts from three different categories (task, domain, and non-domain). For each task we compute the task accuracy averaged over all task questions (95\% confidence intervals are computed over the average task accuracy). We compare the task expert results with the average of all {domain expert personas}, the average of all {non-domain expert personas}, the average of all neutral personas, and the random baseline (horizontal line). We consider four neutral personas, namely student, average student, person, and average person, and the six aforementioned prompt variations.

In Figure~\ref{fig:mmlu_experts} (top row), as expected, when the LLM is asked to impersonate the task expert, the performance is the highest. This shows that the LLM can indeed impersonate task experts with accuracy higher than random. Similarly, the domain expert personas perform better than the non-domain expert personas.
This trend holds for all four MMLU domains and thus for MMLU in its entirety. In general, we observe that the performance in the Humanities tasks is higher than the accuracy in the other domain tasks, which is in line with results reported in the literature~\cite{touvron2023llama, chowdhery2022palm, hoffmann2022chinchilla, hendrycks2021measuring}. Overall, these results suggest that LLMs can increase their performance when asked to impersonate task experts compared to non-task experts.

To provide more details on the individual behaviors of these personas, in the plots on the bottom row of Figure~\ref{fig:mmlu_experts}, we sample various expert personas, e.g.\ three positive and one negative case.  The first, second and last plots indicate that the task expert persona performs better than the domain expert persona, which, in turn, outperforms the non-domain expert persona. In those cases, all experts outperform the neutral persona. For the High School Macroeconomics task, the task expert persona performs close to random and to the non-domain expert persona. This may be because, as Hendrycks et al.~\cite{hendrycks2021measuring} observed, LLMs tend to perform worse on procedural problems that are calculation-heavy compared to purely verbal tasks. Furthermore, when the LLM performs close to or below the random baseline, i.e.\ the task is more difficult to solve for all types of experts, the impersonation trends are not as clear, since the model does not know how to solve the task well, irrespective of the persona.
Thus, while in the Social Sciences field, the High School Macroeconomics task has worse performance, we see that for World Religions, the exam result is higher than 60\%, i.e.\ a passing grade.
Especially for World Religions and Human Aging, we observe that the task expert performs much better than the corresponding domain expert personas. We show results for all tasks in Section \suppref{C.1} of the suppl.

Finally, since several MMLU evaluations~\cite{hendrycks2021measuring,Liang2023HolisticEO}, can lead to small variations when comparing different models’, we include results with the MMLU official prompt in suppl.\ Section \suppref{C.2}, where we verify that our findings on impersonation are not dependent on the formulation of the task.
Lastly, we also show MMLU results for social groups in \suppref{C.3}.

\subsection{Impersonation as categorical descriptions is complementary for visual categorization}
In this section, we provide experimental results on two state-of-the-art fine-grained visual categorization datasets, i.e.\ Caltech UCSD Birds (CUB)~\cite{Wah2011TheCB} and Stanford Cars~\cite{Krause20133DOR}, with 200 and 196 classes of birds and cars, respectively.
Additional results for FGVC Aircraft~\cite{maji13fine-grained} and Oxford Flowers~\cite{Nilsback08flower} can be found in Section \suppref{D.2} of the supplementary.
We first compare how different VLMs make use of the generated descriptions, then compare different LLMs in our in-context impersonation tasks and finally provide some qualitative results. 

\begin{figure}[t]
     \centering
      \begin{minipage}[b]{0.03\textwidth}
        \centering
        \begin{turn}{90}
          \begin{minipage}{0.12\textheight}
            \centering
            CUB
          \end{minipage}          
        \end{turn}
      \end{minipage}
     \hfill
     \begin{subfigure}[b]{0.22\textwidth}
         \centering
         \includegraphics[width=\textwidth]{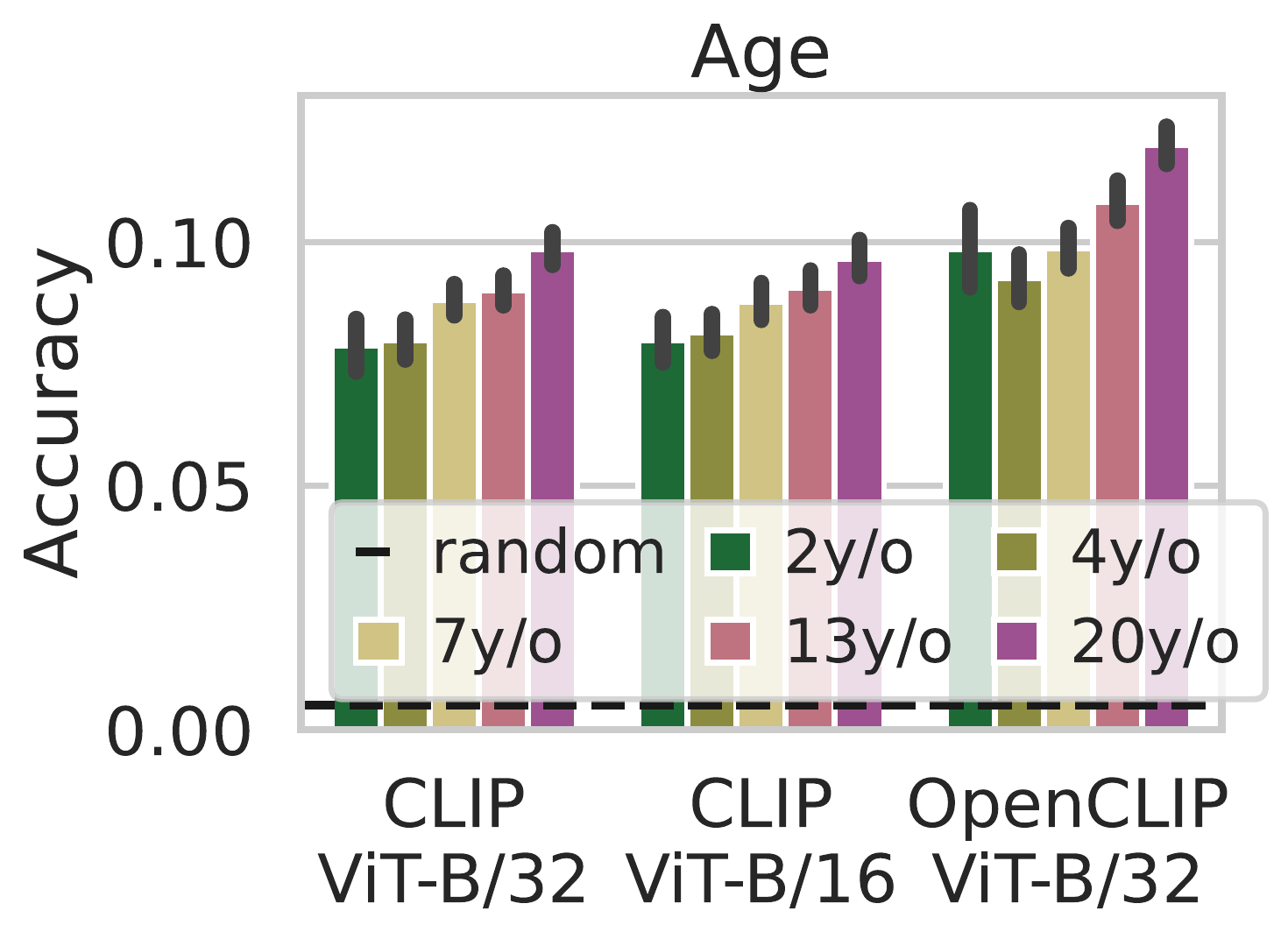}
     \end{subfigure}
     \hfill
     \begin{subfigure}[b]{0.22\textwidth}
         \centering
         \includegraphics[width=\textwidth]{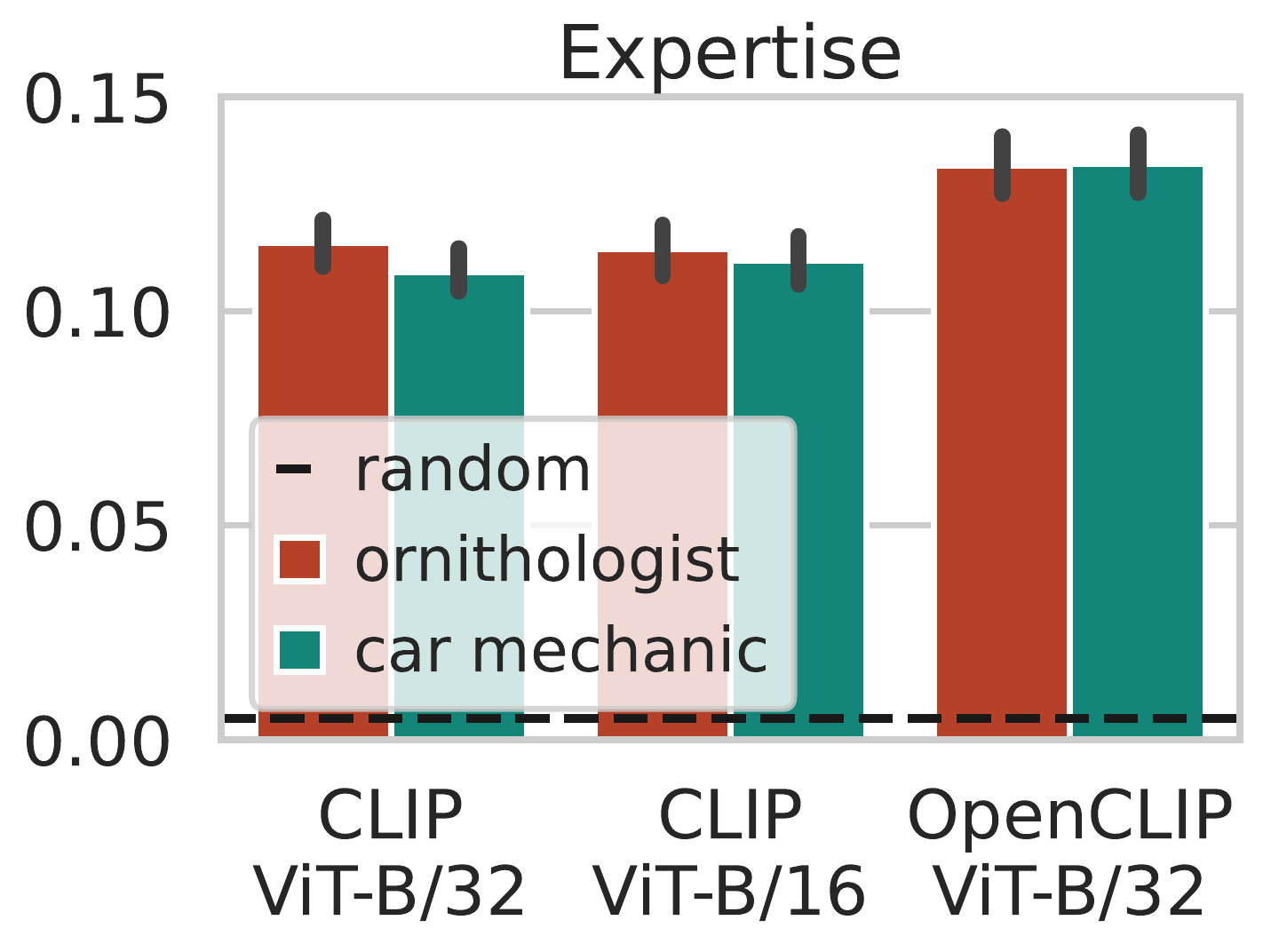}
     \end{subfigure}
     \hfill
     \begin{subfigure}[b]{0.22\textwidth}
         \centering
         \includegraphics[width=\textwidth]{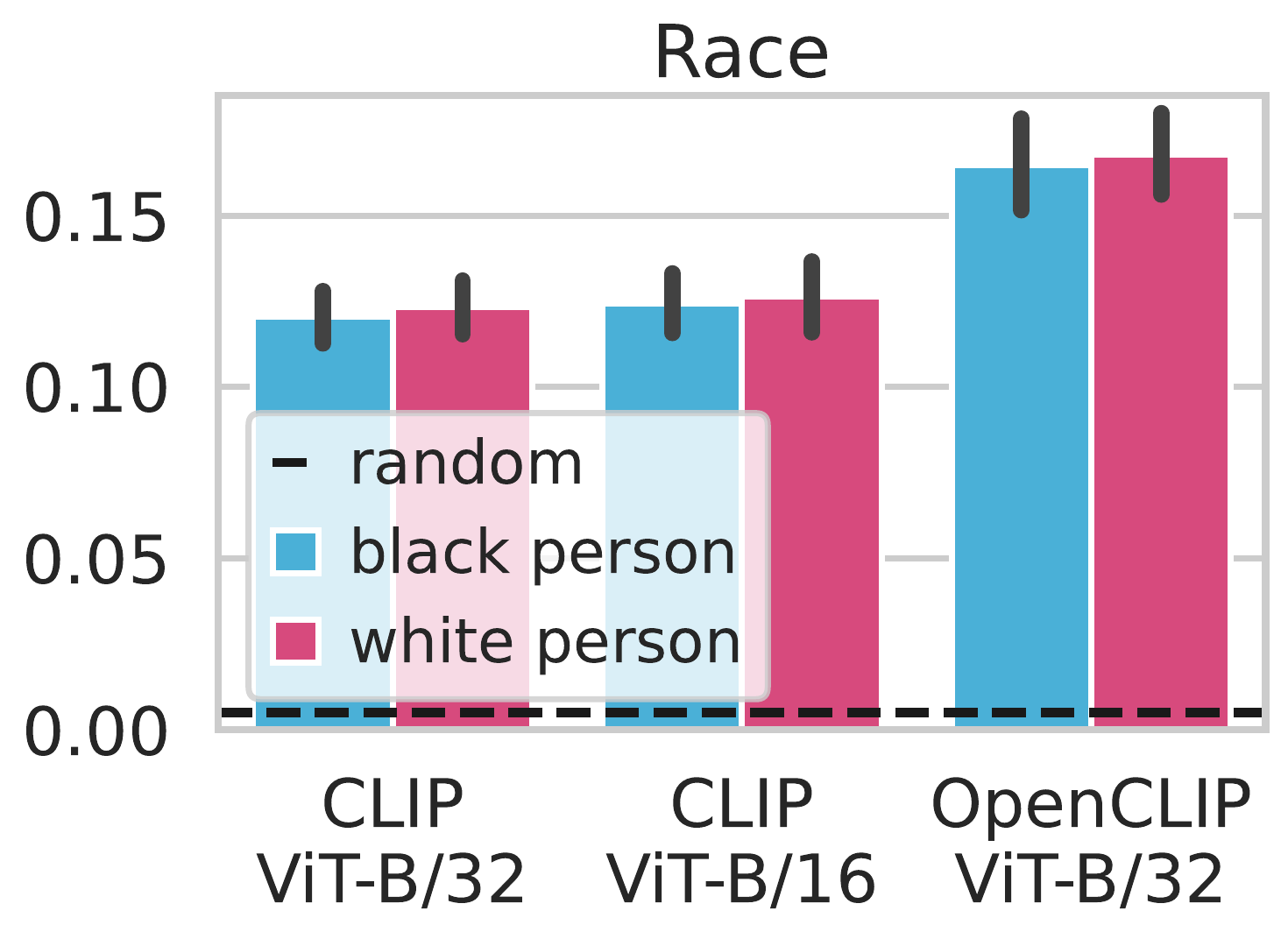}
     \end{subfigure}
      \hfill
     \begin{subfigure}[b]{0.22\textwidth}
         \centering
         \includegraphics[width=\textwidth]{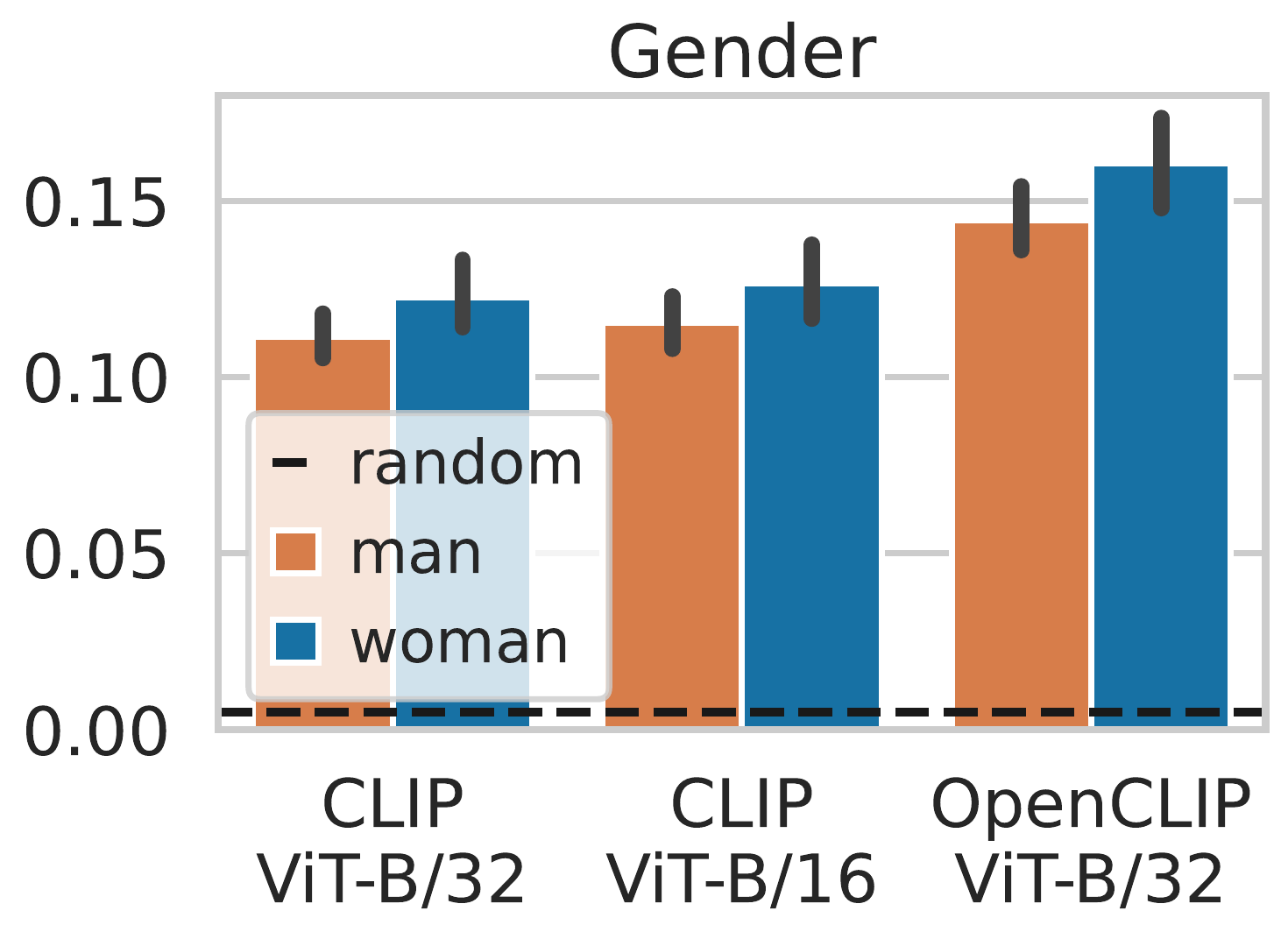}
     \end{subfigure}
     \\
      \begin{minipage}[b]{0.03\textwidth}
        \centering
        \begin{turn}{90}
          \begin{minipage}{0.12\textheight}
            \centering
            Stanford Cars
          \end{minipage}          
        \end{turn}
      \end{minipage}
     \hfill
     \begin{subfigure}[b]{0.22\textwidth}
         \centering
         \includegraphics[width=\textwidth]{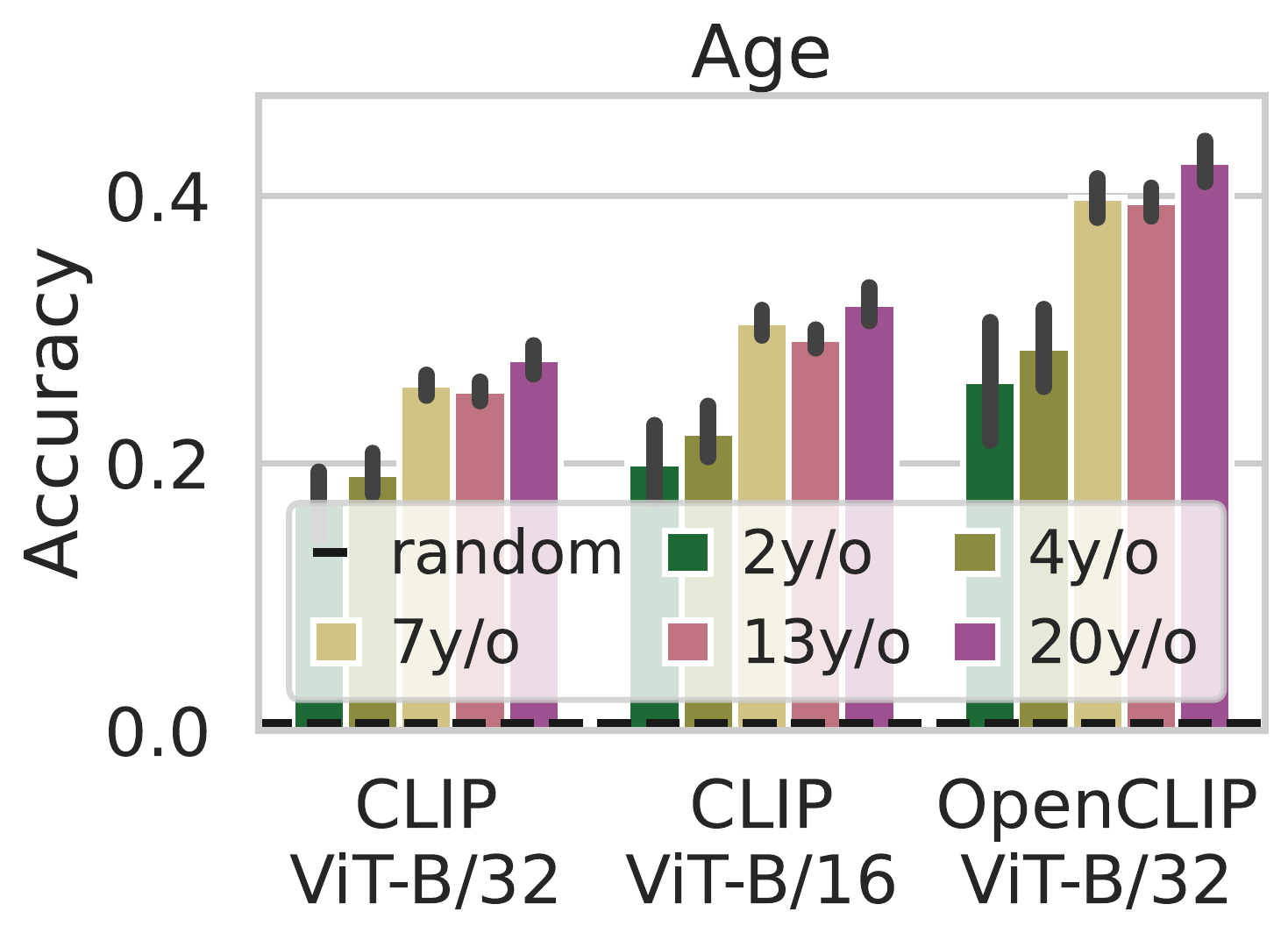}
     \end{subfigure}
     \hfill
     \begin{subfigure}[b]{0.22\textwidth}
         \centering
         \includegraphics[width=\textwidth]{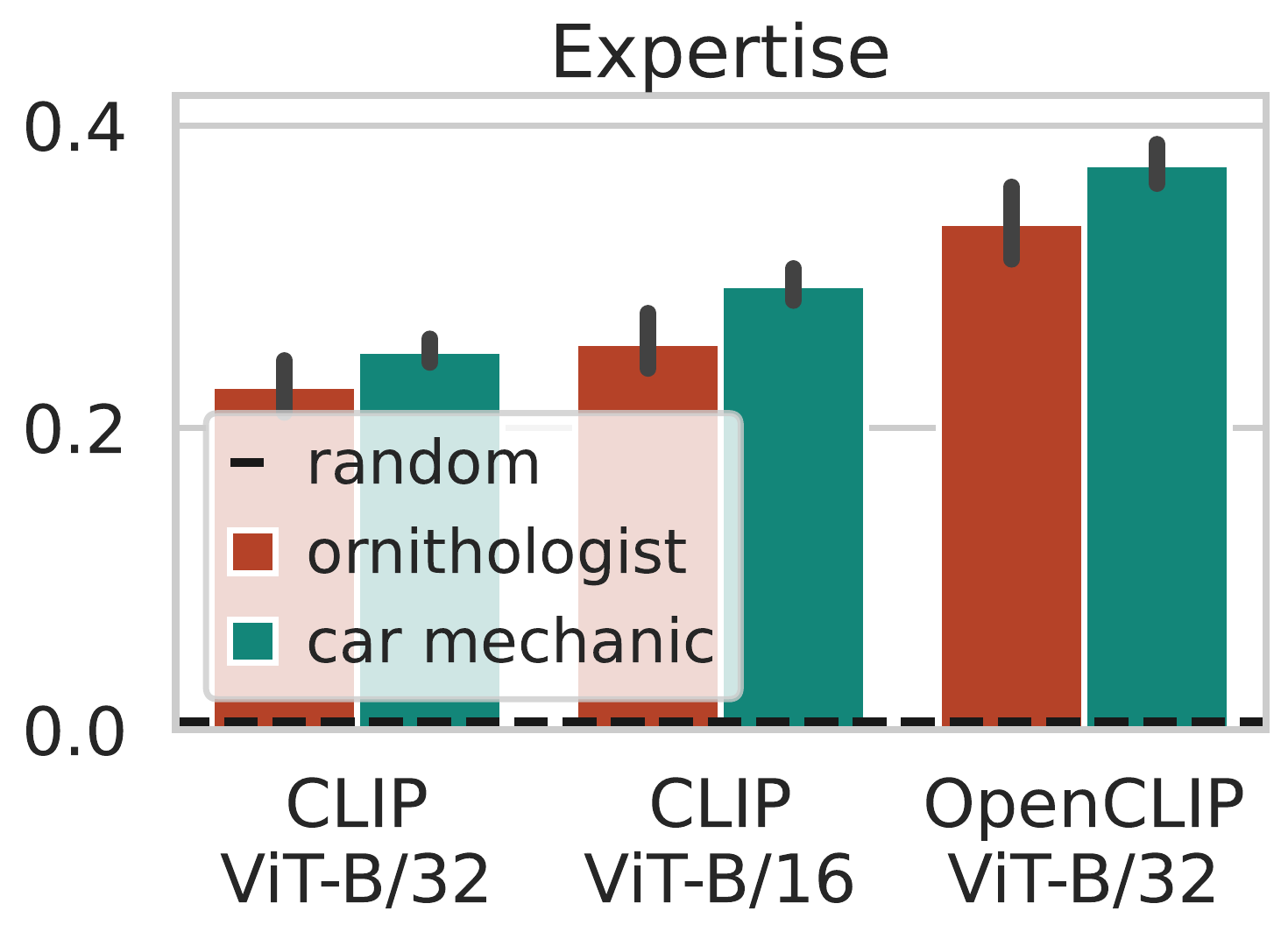}
     \end{subfigure}
     \hfill
     \begin{subfigure}[b]{0.22\textwidth}
         \centering
         \includegraphics[width=\textwidth]{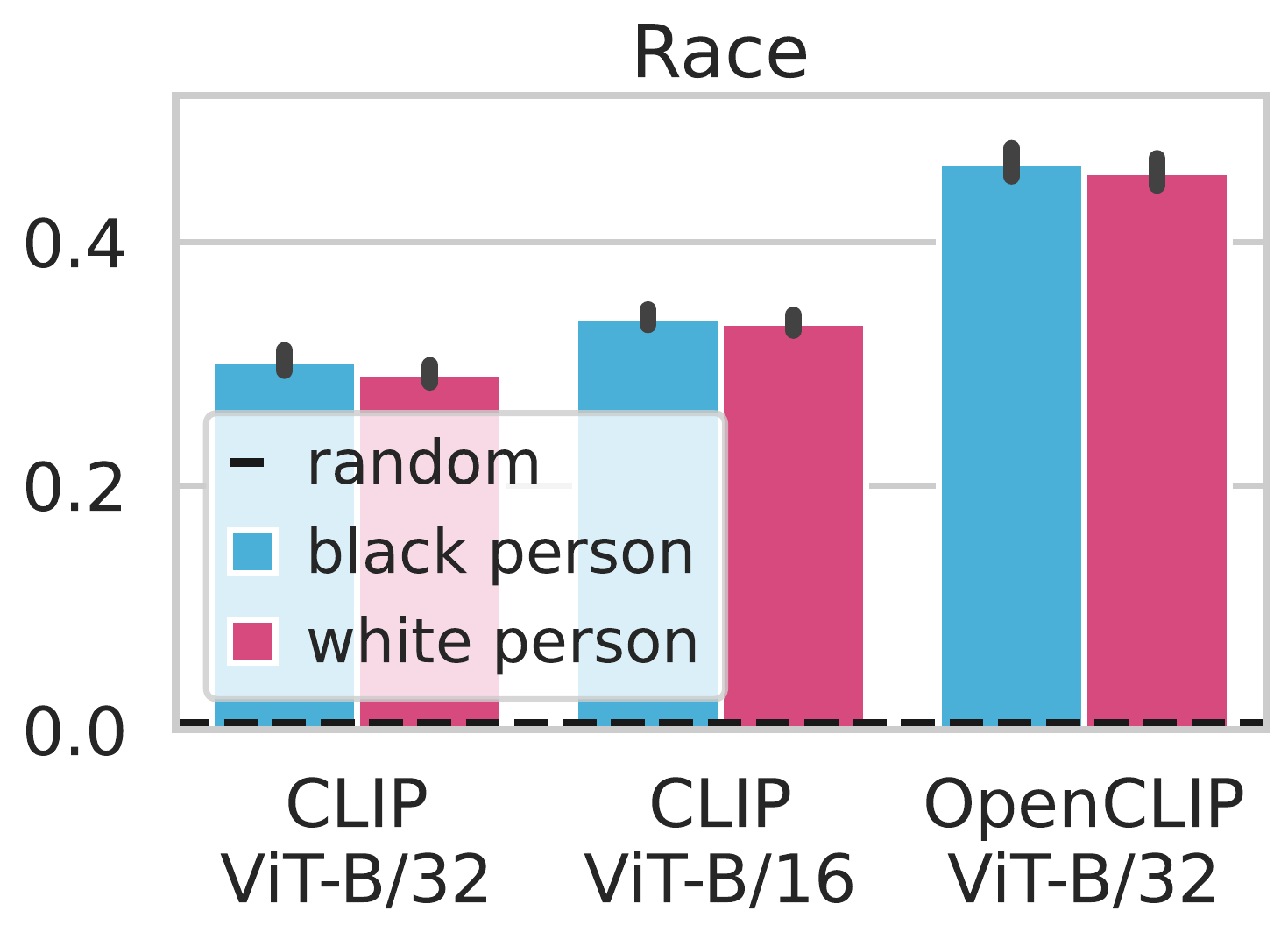}
     \end{subfigure}
      \hfill
     \begin{subfigure}[b]{0.22\textwidth}
         \centering
         \includegraphics[width=\textwidth]{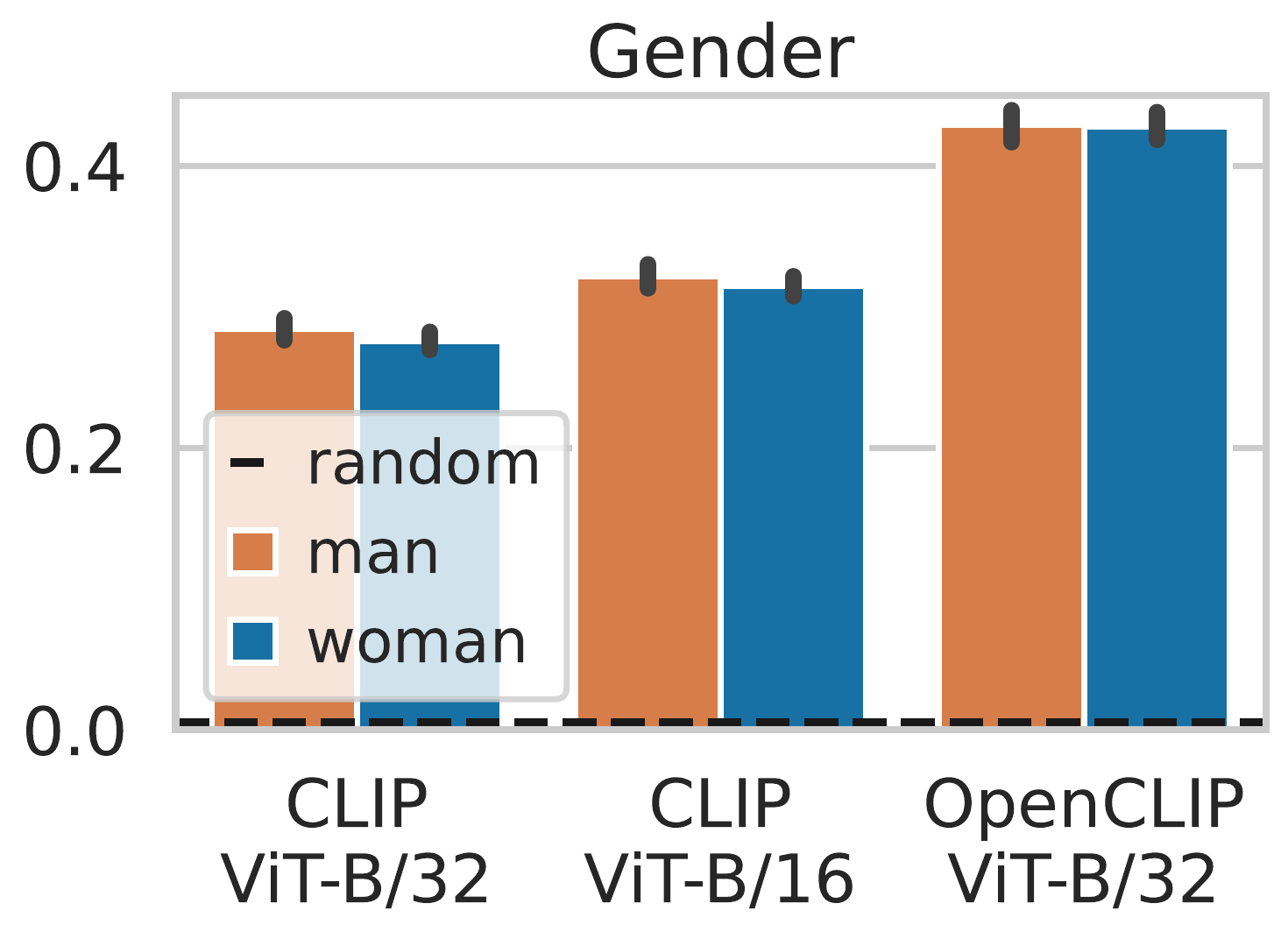}
     \end{subfigure}
    \caption{Comparing CLIP-32, CLIP-16 and OpenCLIP as VLMs (the language input comes from Vicuna-13B) on CUB (top) and Stanford Cars (bottom) datasets. We observe the effects of age, expertise, ethnicity and gender independent of the VLM used for fine-grained visual classification. The dashed line represents the random baseline.
    }%
    \label{fig:vlm-comparison}
    \vspace{-1.63ex}
\end{figure}
\textbf{Comparing VLM variants.}
We first compare the classification accuracy of different VLMs when the Vicuna-13B generated descriptions of classes are fed to the language encoder of the VLM\@.
For the vision encoders we consider the Vision Transformer (ViT)~\cite{Dosovitskiy2020AnII} based B/32 and B/16 variants of the official CLIP implementation~\cite{radford2021clip} as well as the OpenCLIP B/32 ViT variant~\cite{Cherti2022ReproducibleSL}. The latter is a replication of the original CLIP trained on a larger dataset (Laion 5B~\cite{Schuhmann2022LAION5BAO}). For each CLIP variant, we use the corresponding causal transformer text encoders, which might not encode text as well as Vicuna but are able to embed the text into a shared multi-modal space.

Our results in Figure~\ref{fig:vlm-comparison} show that across all three CLIP variants increased age in the impersonated persona increases performance for both bird and car classification. Interestingly, there is a significant increase in performance at 7 years of age when recognizing cars.
Our expertise evaluation shows that the car mechanic persona's descriptions performs better than ornithologist's when recognizing cars. 
Interestingly, racial (column 3) and gender (column 4) personas, reveal consistent biases. While the \persona{black} performs better in car classification, the \persona{white} performs better in bird classification. This may indicate that there are stereotypical biases in the training data. Similarly, while the \persona{woman} performs clearly better than \persona{man} for bird classification, the trend is not as strong for car classification although \persona{man} performs slightly better than \persona{woman}.
The language encoder of VLMs potentially being weaker than Vicuna, we expect these results to improve overall with a stronger language encoder in the VLM but this is an orthogonal direction to explore.
To confirm the significance of our results, we run $\text{Chi}^2$ tests for expertise, race and gender. We consider the three CLIP models, five different seeds and the six different impersonation prompt variations. We find that for all experiments considered, \{CUB, Stanford Cars\} x \{man/woman, black/white, ornithologist/car mechanic\}, p<0.001. Thus, we conclude that our results are significant.

\begin{figure}[t]
    \centering
      \begin{minipage}[b]{0.03\textwidth}
        \centering
        \begin{turn}{90}
          \begin{minipage}{0.12\textheight}
            \centering
            CUB
          \end{minipage}          
        \end{turn}
      \end{minipage}
     \hfill
     \begin{subfigure}[b]{0.236\textwidth}
         \centering
         \includegraphics[width=\textwidth]{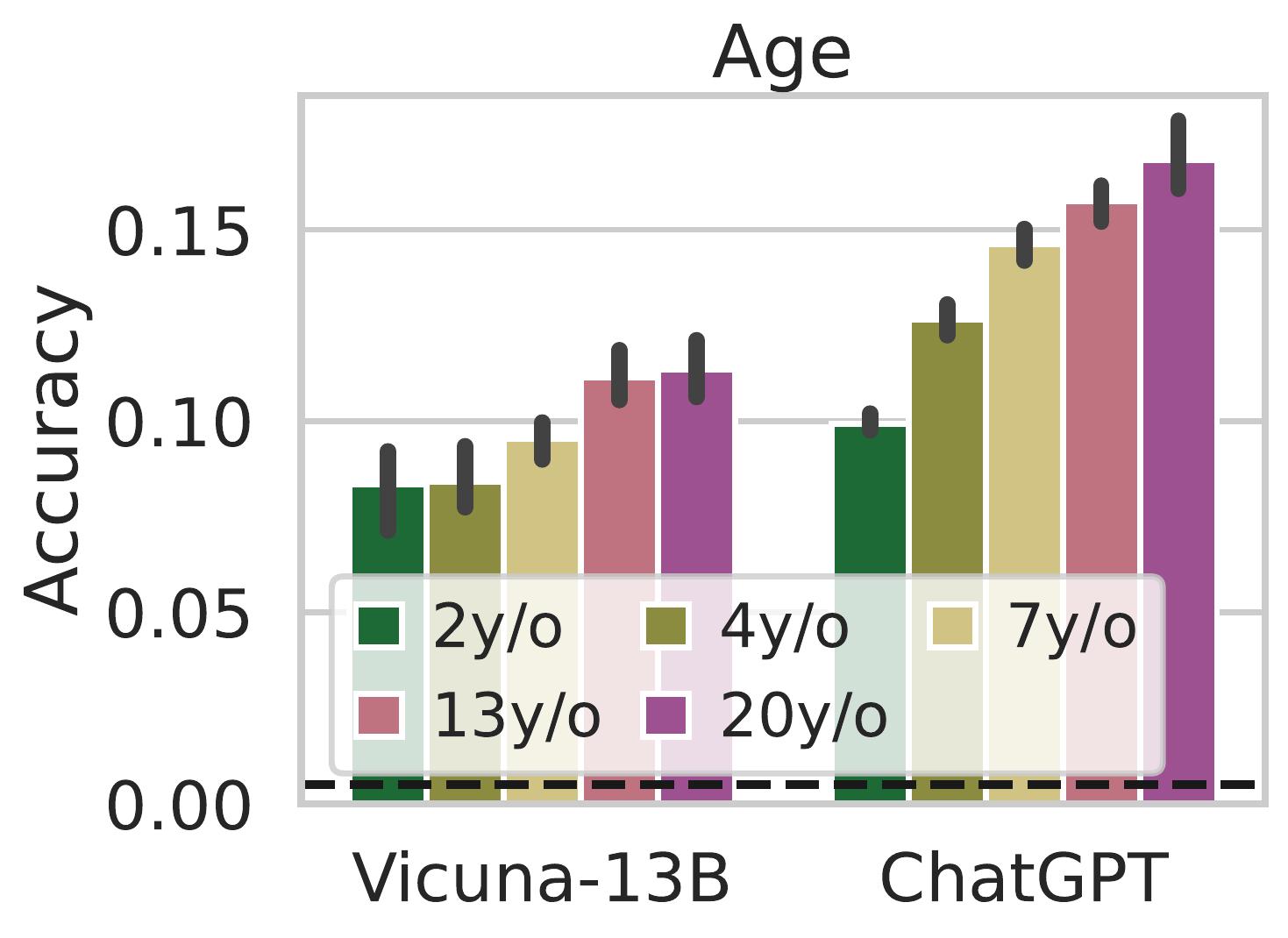}
     \end{subfigure}
     \hfill
     \begin{subfigure}[b]{0.236\textwidth}
         \centering
         \includegraphics[width=\textwidth]{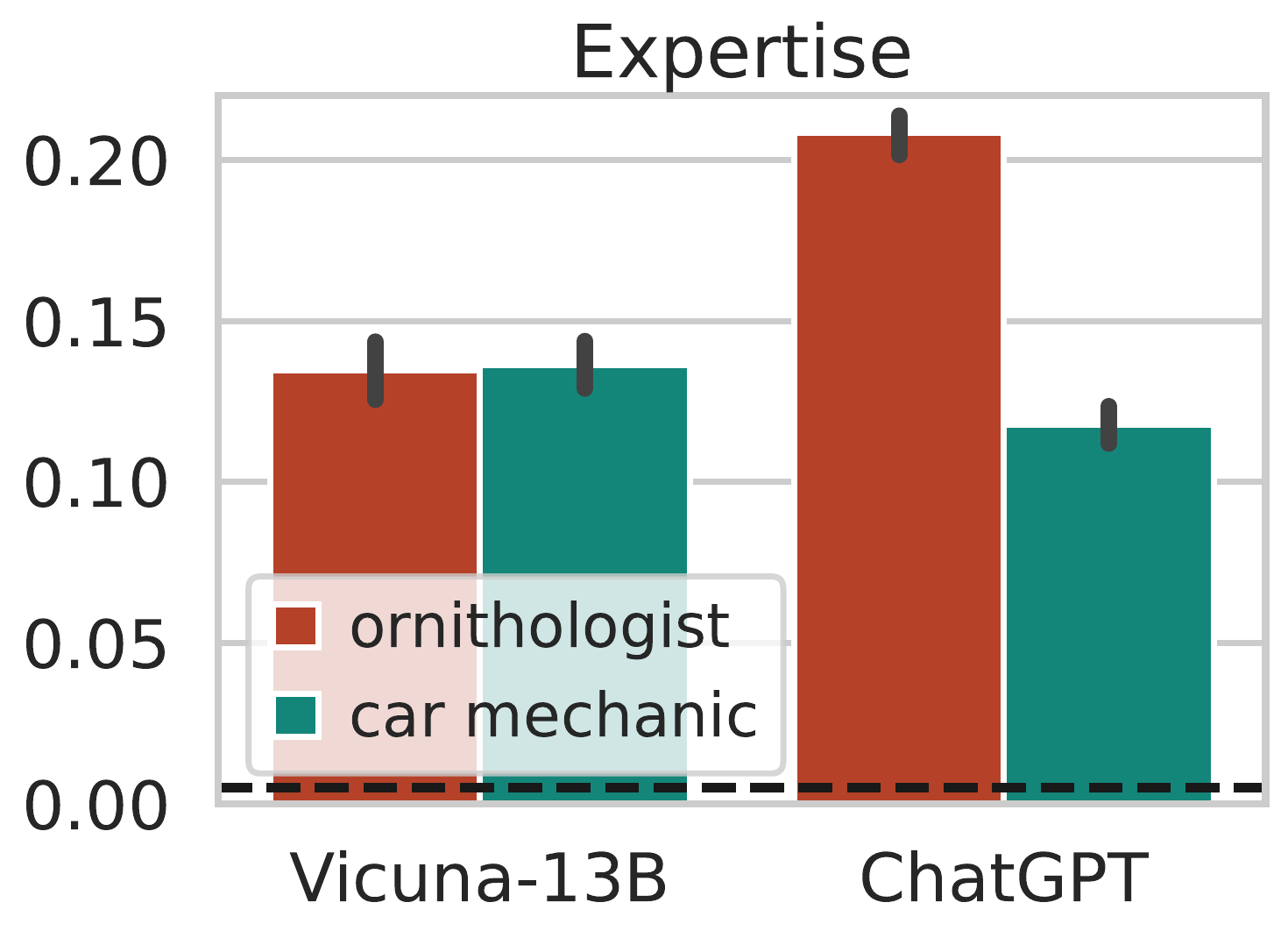}
     \end{subfigure}
     \hfill
     \begin{subfigure}[b]{0.236\textwidth}
         \centering
         \includegraphics[width=\textwidth]{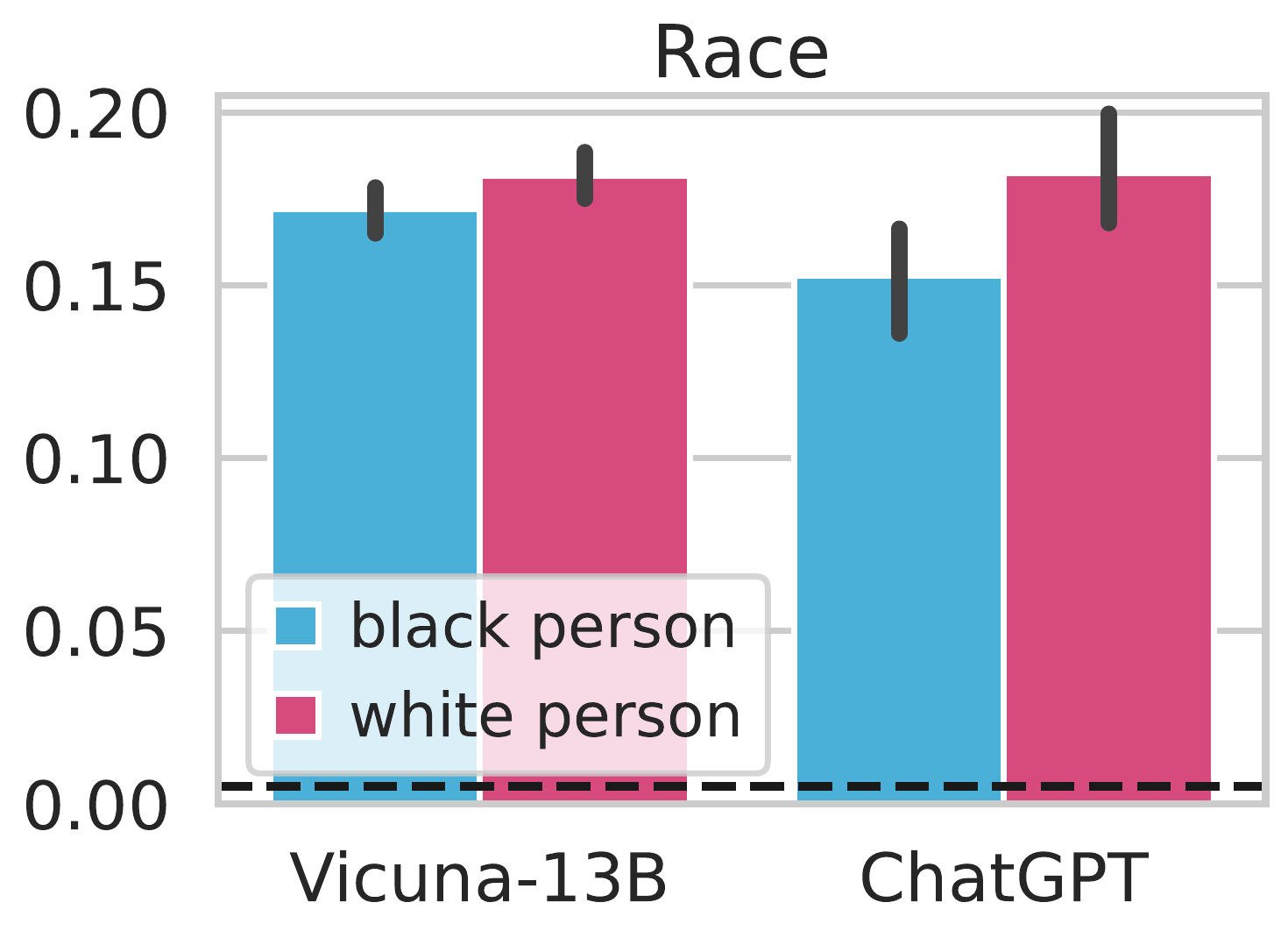}
     \end{subfigure}
      \hfill
     \begin{subfigure}[b]{0.236\textwidth}
         \centering
         \includegraphics[width=\textwidth]{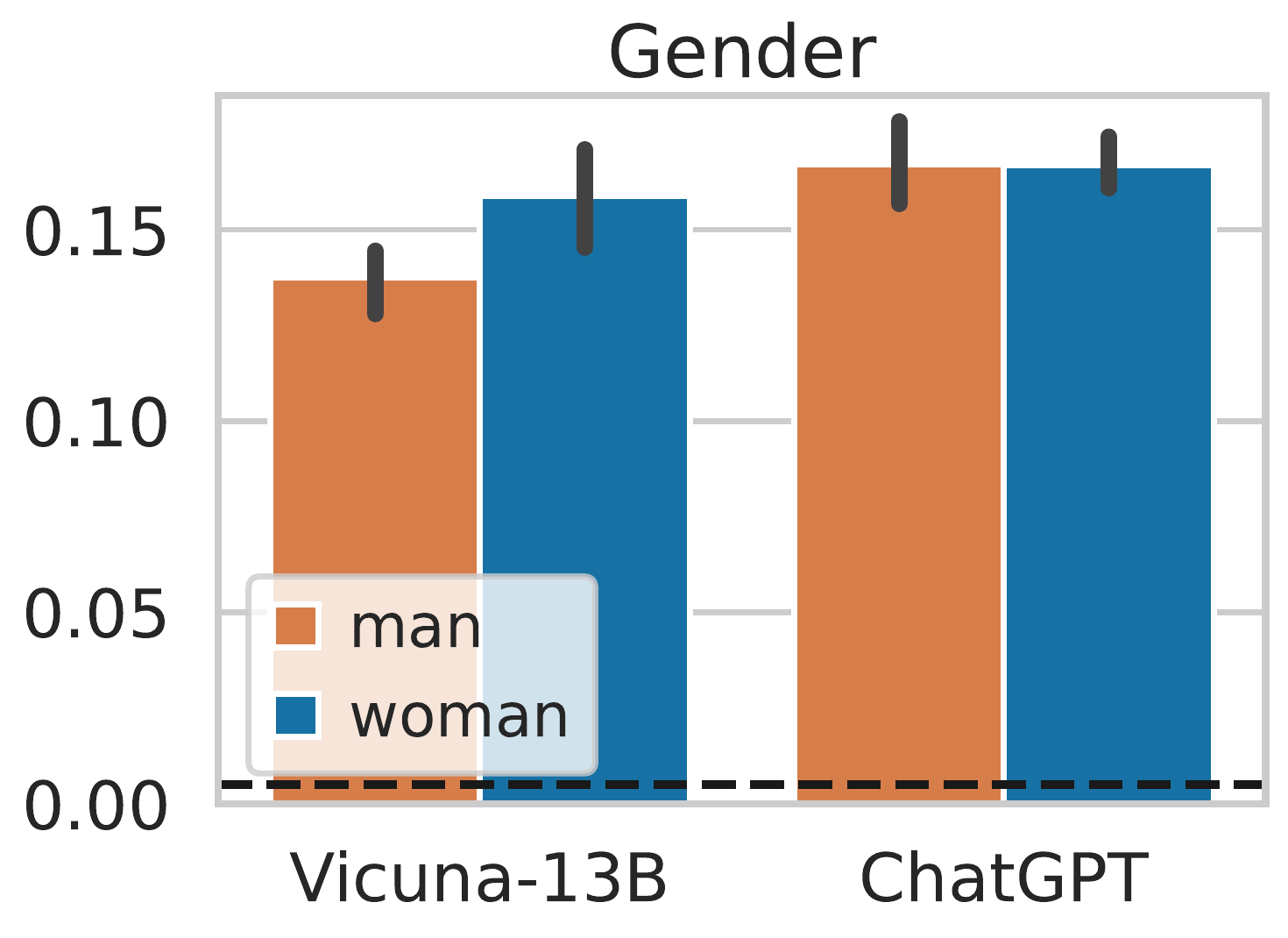}
     \end{subfigure}
     \centering
      \begin{minipage}[b]{0.03\textwidth}
        \centering
        \begin{turn}{90}
          \begin{minipage}{0.12\textheight}
            \centering
            Stanford Cars
          \end{minipage}          
        \end{turn}
      \end{minipage}
     \hfill
     \begin{subfigure}[b]{0.236\textwidth}
         \centering
         \includegraphics[width=\textwidth]{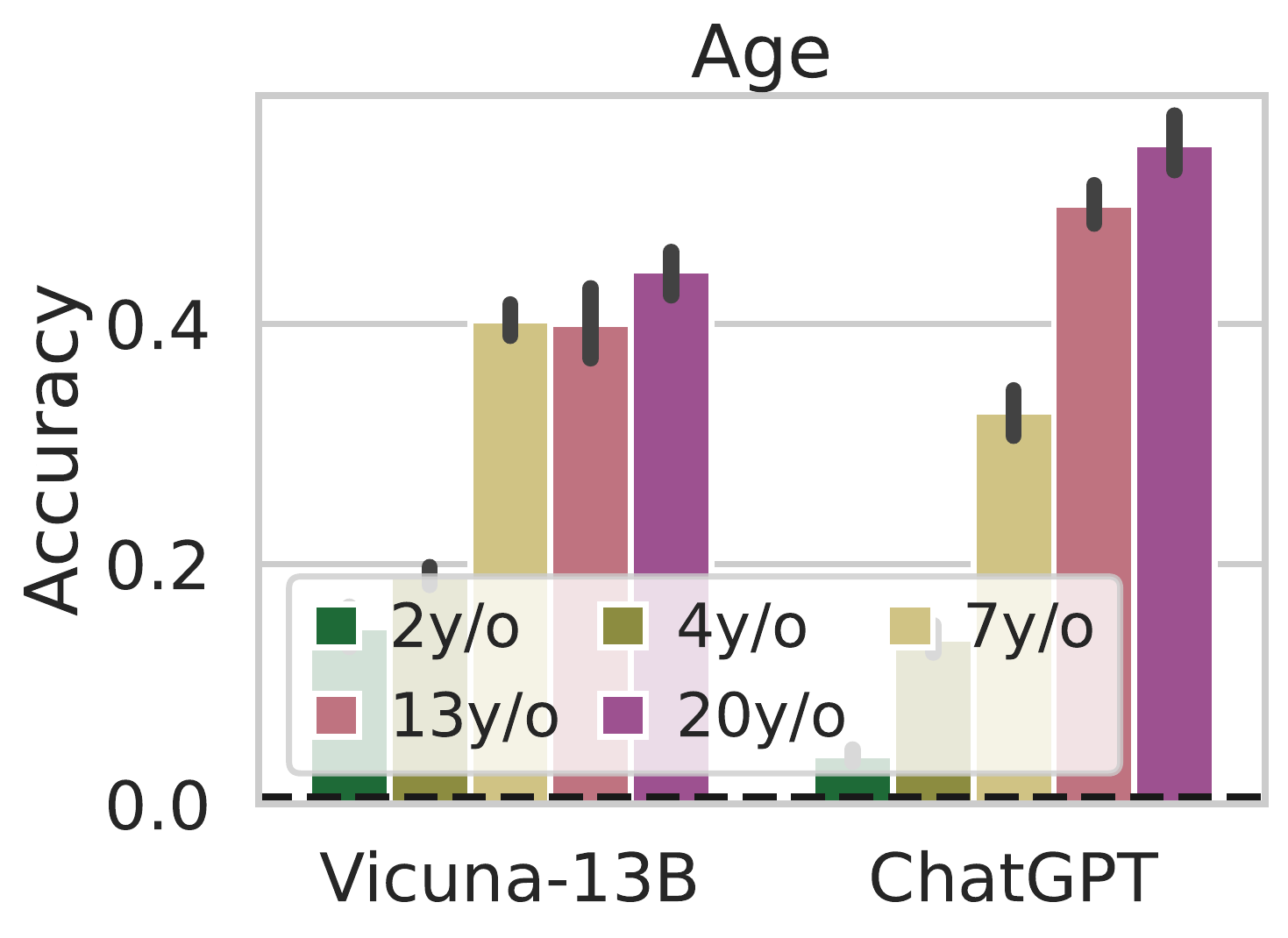}
     \end{subfigure}
     \hfill
     \begin{subfigure}[b]{0.236\textwidth}
         \centering
         \includegraphics[width=\textwidth]{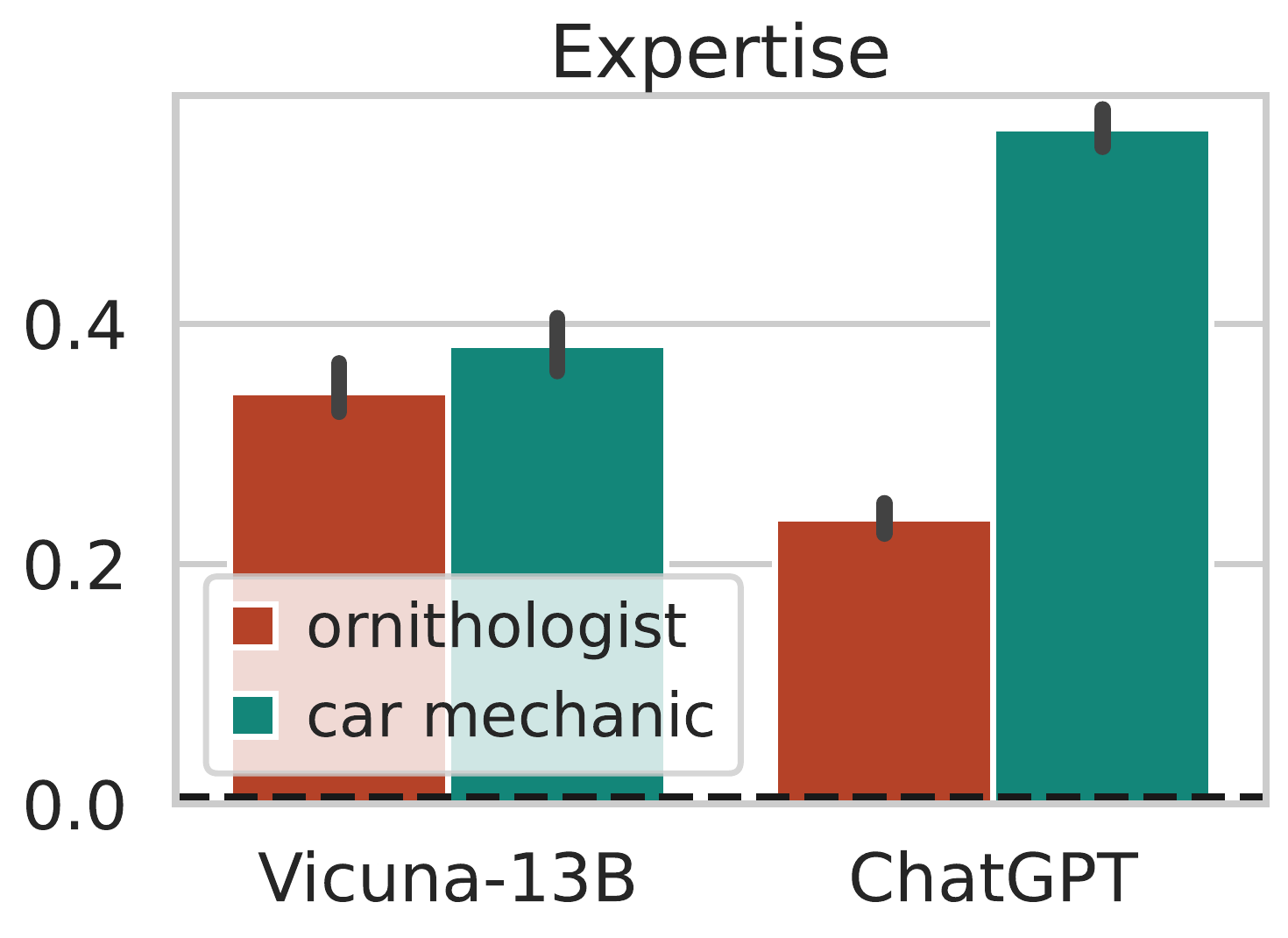}
     \end{subfigure}
     \hfill
     \begin{subfigure}[b]{0.236\textwidth}
         \centering
         \includegraphics[width=\textwidth]{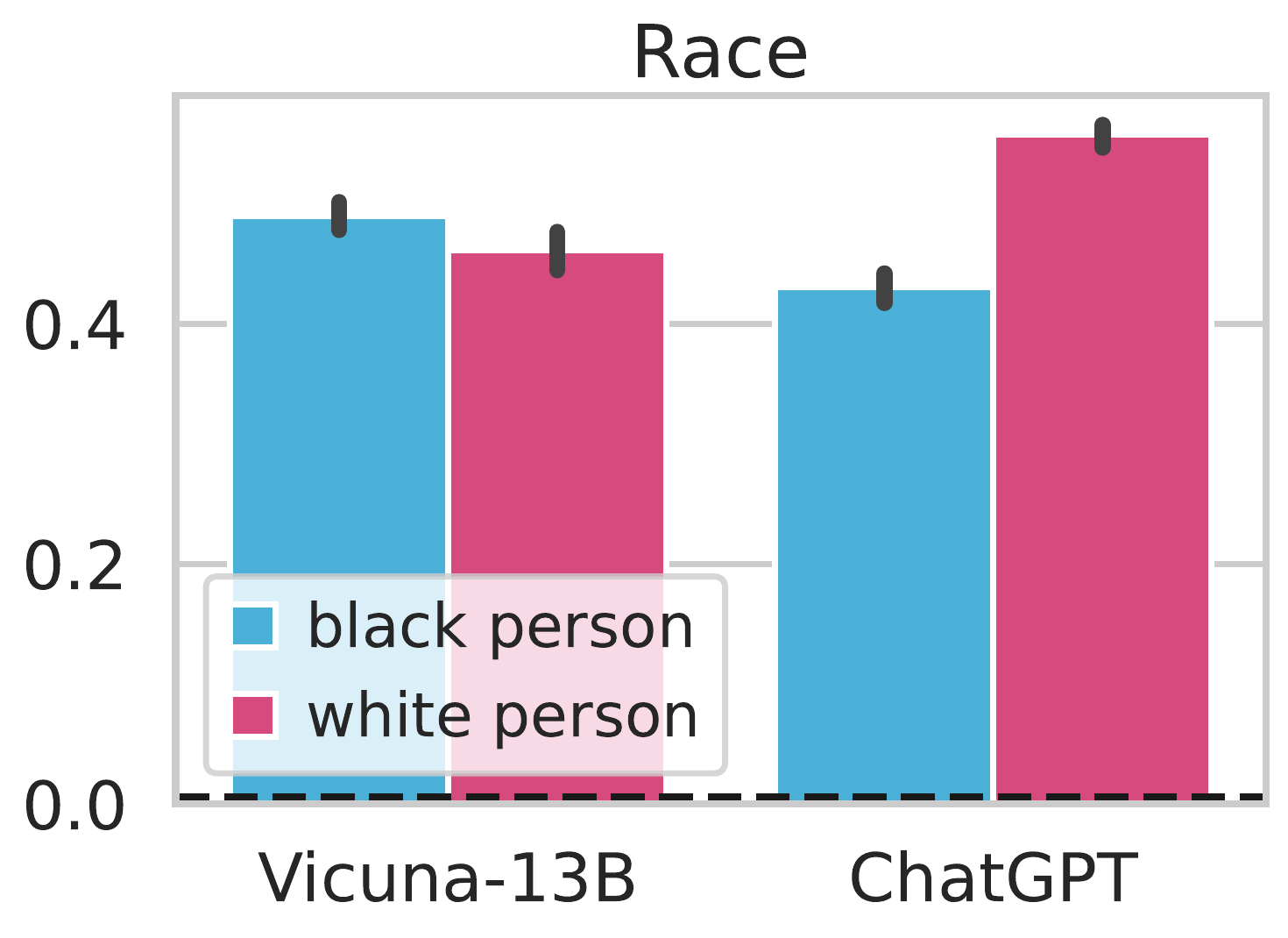}
     \end{subfigure}
      \hfill
     \begin{subfigure}[b]{0.236\textwidth}
         \centering
         \includegraphics[width=\textwidth]{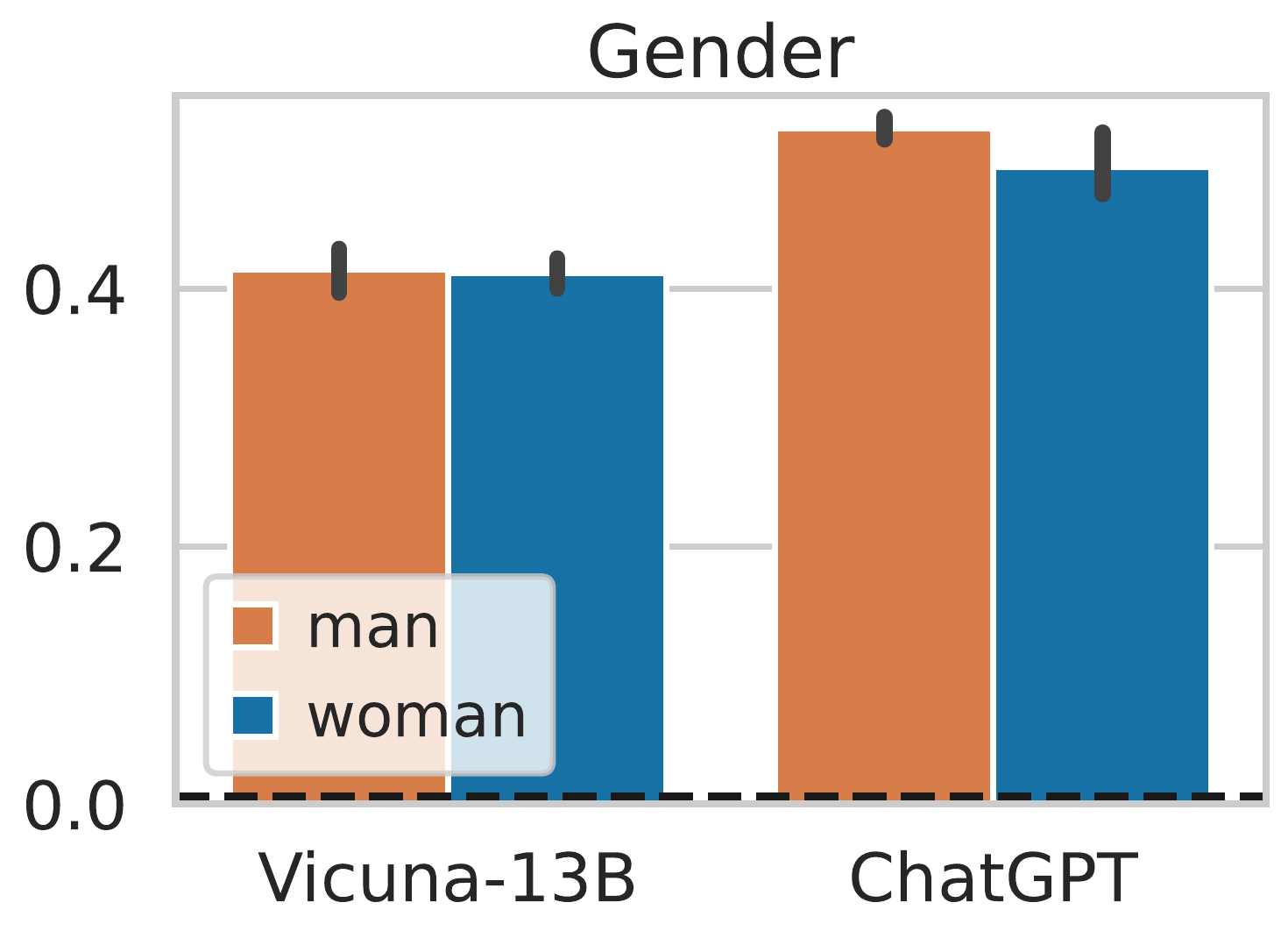}
     \end{subfigure}
     \vspace{-5mm}
    \caption{Comparing Vicuna-13B and ChatGPT as LLM variants (OpenCLIP is the VLM) on CUB and Stanford Cars.
    For both LLMs, the accuracy increases with increasing age, the expert persona on the respective dataset performs better and both LLMs are not free of biases, and impersonation of different genders or race affects their performance. The dashed line represents the random baseline.
    }%
    \label{fig:llm-comparison}
    \vspace{-3ex}
\end{figure}

\begin{wrapfigure}[11]{r}{0.27835\textwidth}
    \vspace{-3ex}
    \begin{center}
        \includegraphics[width=\linewidth,trim={0 0 0 30},clip]{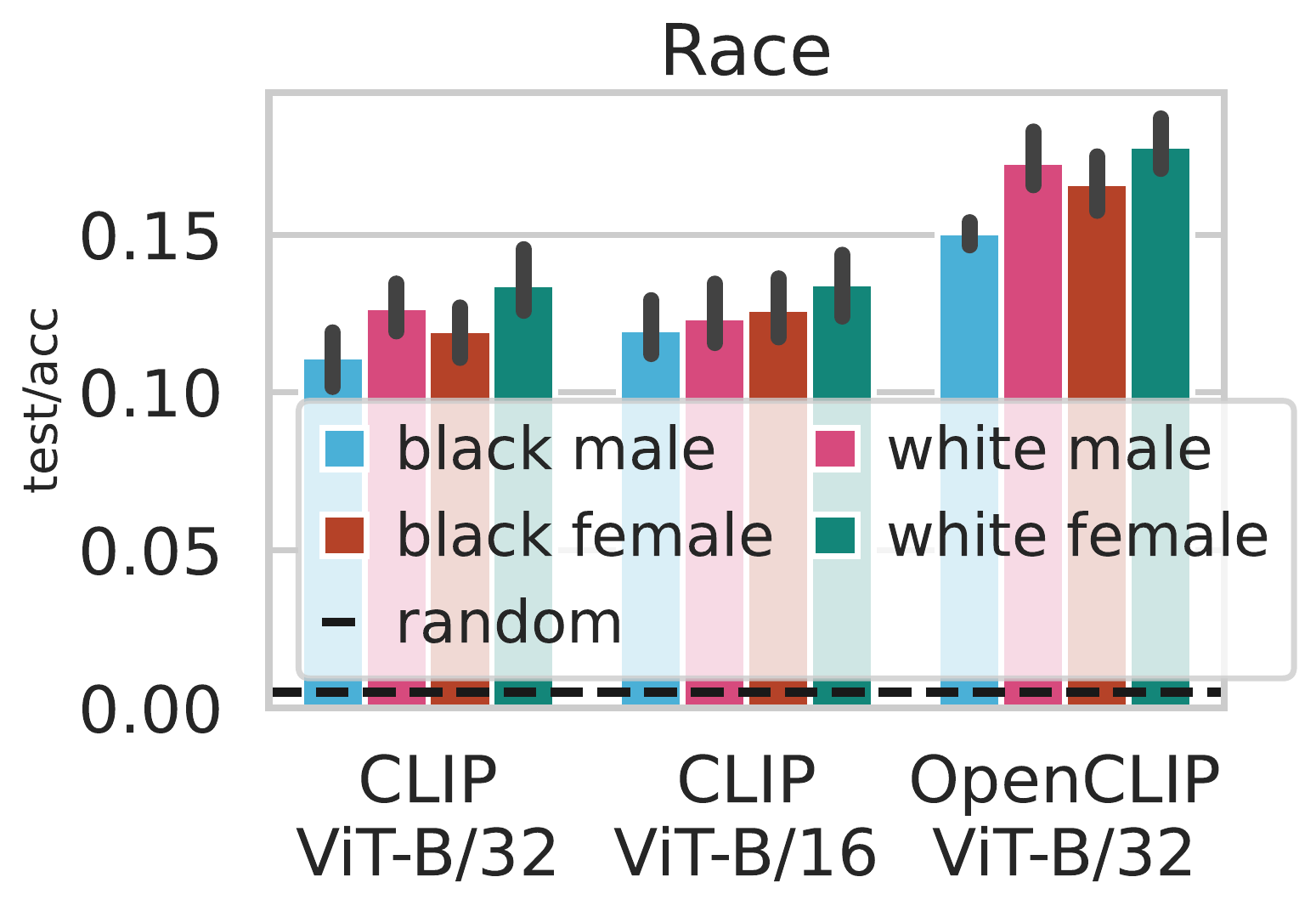}
    \end{center}
    \vspace{-1ex}
    \caption{Composition of personas on CUB for Vicuna-13B.}%
    \label{fig:bias-combination}
\end{wrapfigure}
We also investigate the effects of composing personas for a computationally feasible subset of persons. More specifically, we study all possible combinations of \{Black, White\} × \{Female, Male\} for the CUB dataset for 5 different seeds (\Cref{fig:bias-combination}).
With Vicuna-13B we see weak evidence that the biases co-construct: Individually the white persona outperforms the black persona and the same applies to the female persona outperforming the male persona. Combined, the white female persona outperforms both the black female persona (change in race) and the white male persona (change in gender).
Furthermore, we also study performance of additional genders (agender and non-binary) and races (indian, asian and hispanic) in the suppl.\ in Section \suppref{D.5}.

\textbf{Comparing LLM variants}
We evaluate how different LLMs, namely Vicuna-13B and ChatGPT, generate descriptions of the classes of interest. In these experiments, we keep the VLM fixed to OpenCLIP, as it is the best of the CLIP variants tested above. For computational reasons, we only evaluate on our original impersonation prompt.
Figure~\ref{fig:llm-comparison} shows the effect of LLM impersonation on the generated descriptions evaluated on zero-shot image classification.

For the age personas, we observe a clear trend of increased performance for both LLMs as they impersonate older characters. The progression is particularly pronounced for ChatGPT, where on Stanford Cars the \persona{2-year-old} persona describes different cars with similar expressions leading to $\sim 4\%$ accuracy, but as ChatGPT's persona gets older, it becomes more accurate in describing cars, e.g.\ 54.9\% for persona of age \persona{20}. This indicates that LLMs can replicate human language at different development stages, varying their language both in terms of vocabulary and general knowledge for accurately describing these objects as discussed in~\cite{oates2004cognitive}.
Similarly to the reasoning task, LLMs exhibit higher expertise on the topic when we ask them to impersonate a bird expert (``\persona{ornithologist}'' persona) and a car expert (``\persona{car mechanic}'' persona). The respective domain expert persona performs approximately twice as well as the non-domain expert persona when using ChatGPT\@. Impersonating an expert, the LLM tends to describe a class in more detail and mention more discriminative features. 

We also observe that impersonation can reveal biases encoded in the LLMs. A race bias becomes apparent when we ask the LLMs to impersonate a ``\persona{black}'' or ``\persona{white}'' person. ChatGPT tends to describe both birds and cars better when posing as a \persona{white} person. Vicuna-13B, on the other hand, provides better descriptions of cars as a \persona{black} person. Gender biases are a bit less noticeable, but we still find Vicuna-13B giving better bird descriptions as a \persona{woman} persona and ChatGPT identifying cars better as a \persona{man} persona. While instruction-based fine-tuning~\cite{Ouyang2022TrainingLM} tries to remedy social biases encoded in LLMs to some extent, we can still expose them through in-context impersonation.

Overall, we find that ChatGPT shows larger effects, probably due to its access to more diverse (fine-tuning) data. The fact that the effects described above can be found with two very different language models suggests that they are a result of the overall language modeling and instruction following training on internet data instead of specific model artifacts.

\begin{figure}[t]
    \centering
    \includegraphics[width=\linewidth]{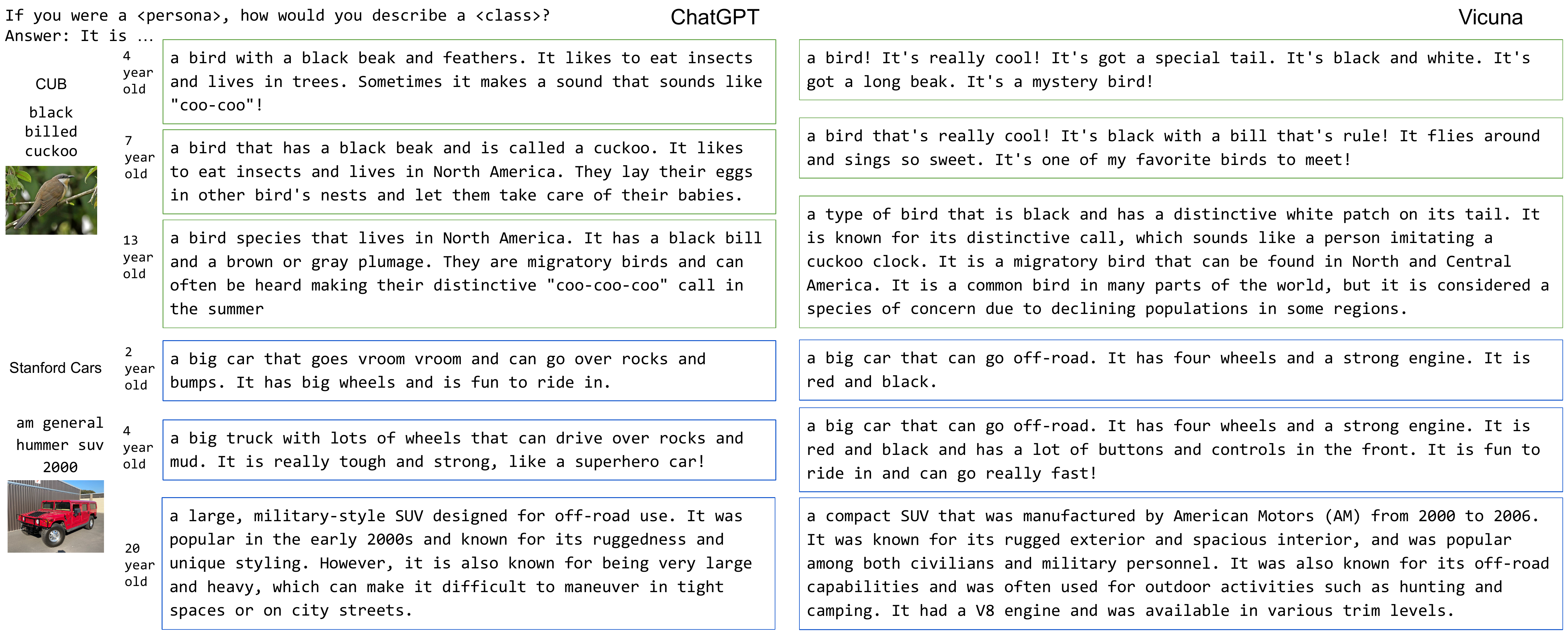}
    \caption{Qualitative results sampling all the age personas (\persona{2}, \persona{4}, \persona{7}, \persona{13} and \persona{20-year-old} personas) for two classes, i.e.\ Black Billed Cuckoo (CUB) and AM General Hummer SUV 2000 (Stanford Cars) classes. The results are obtained by querying ChatGPT and Vicuna.}%
    \label{fig:qualitative}
\end{figure}
\textbf{Qualitative results and limitations.} In Figure~\ref{fig:qualitative}, we provide the descriptions generated by ChatGPT and Vicuna for one class, i.e.\ black billed cuckoo, from the CUB dataset and one class, i.e.\ AM General Hummer SUV 2000, from the Stanford Cars dataset. As personas, we sample all the age personas we considered in our experiments, namely \persona{2}, \persona{4}, \persona{7}, \persona{13} and \persona{20-year-old} personas. 

For both LLMs, in both datasets, we observe that with increasing age, the complexity of the vocabulary and attributes of the mentioned objects increases. A \persona{2-year-old} persona talks about the sound the bird or the car makes, the shapes of the wings or wheels, and the emotions attached to seeing or riding it. A \persona{4-year-old} persona interestingly mentions experiences seeing the bird or the car more distinctly. A \persona{7-year-old} persona starts using more complicated adjective phrases, e.g.\ can drive on rough roads and outside places, whereas a \persona{13-year-old} persona takes it one step further, e.g.\ brownish-gray body with distinctive rusty colored markings. Finally, a \persona{20-year-old} persona makes a more complete description of the object including where the bird is found or what the car is mainly used for. This is in line with~\cite{durant2023sophistication} where the authors show that given the same length of text, smaller children use less diverse and non-academic vocabulary, and repeat a lot. Even though LLM's may not faithfully represent the language of children, we qualitatively observe similar patterns. We show more examples and quantize the properties of the generated descriptions in suppl.\ Section \suppref{D.3}. 

One obvious difference between these two LLMs to point out is that the descriptions obtained from Vicuna appear to be longer and more detailed. Further, at earlier ages, e.g.\ 2 or 4, especially on CUB, the descriptions of Vicuna seem poetic. The difference between the semantic content of the descriptions of the 13-year-old persona and the 20-year-old persona seems to be less distinct in Vicuna than in ChatGPT\@. One final interesting observation is that Vicuna descriptions talk about the color of the car whereas the color can not be a distinguishing property of a car. 

\section{Broader Impact}
We believe that a better understanding of in-context impersonation, as well as its resulting downstream effects, can not only help to mitigate the risk of fraud but also to understand how these newly-powerful agents behave more generally~\cite{burnell2023rethink}. We have already seen that in-context impersonation boosts performance and produces biases; these results could be followed up by investigating how these characteristics emerge during training, change with increasing model size~\cite{kaplan2020scaling}, or adapt with additional fine-tuning~\cite{ziegler2019fine}.
Additionally, LLM providers could quantitatively test for these biases before releasing new models.
We specifically discourage crafting (system) prompts for maximal performance by exploiting biases, as this may have unexpected side effects, reinforce societal biases and poison training data obtained with such prompts.
Other misuses may include amplification of stereotypical biases through generated content and using impersonation to invoke fake trust. However, we believe systematically studying these biases raises awareness in the ML community and general society and serves as a first step to research mitigation strategies.
Lastly, we discuss limitations of our work in suppl.\ Section \suppref{E}.

\section{Conclusion}
We presented evidence that \emph{in-context impersonation}, that is asking LLMs to take on different roles in context, can change their performance and reveal their biases. Asking LLMs to impersonate differently aged people in a two-armed bandit task, LLMs could reproduce human-like developmental stages of exploration behavior. Asking LLMs to impersonate domain experts, they performed better than LLMs that were asked to impersonate a non-domain expert. Finally, asking LLMs to impersonate various roles in a vision-language task revealed not only that impersonation can boost relative performance but also recovered societal biases about a person's age, gender, and race. 

We have demonstrated the effects of in-context impersonation on single agents performing relatively simple tasks across a limited range of personas. In future work, we want to scale up this approach to multiple LLMs impersonating a variety of personas across complex and interactive tasks~\cite{park2023generative}. Finally, we believe that in-context impersonation can also be applied to other modalities, for example to large models for video generation~\cite{wang2022internvideo}.

\section{Acknowledgements}
The authors thank IMPRS-IS for supporting Leonard Salewski. This work was partially funded by the Portuguese Foundation for Science and Technology (FCT) under PhD grant 2020.07034.BD, the Max Planck Society, the Volkswagen Foundation, the BMBF Tübingen AI Center (FKZ: 01IS18039A), DFG (EXC number 2064/1 – Project number 390727645) and ERC (853489-DEXIM).

{\small
\bibliography{bib}
}

\newpage
\setcounter{section}{0}
\renewcommand*{\theHsection}{chX.\the\value{section}}
\renewcommand\thesection{\Alph{section}}

\maketitlepage

In this supplementary materials we show additional results mentioned in the main paper.
First, we give experimental details in \Cref{sec:experimental_details}. Next, we show results for Llama 2 on the bandit task in \Cref{sec:bandit_llama2}.
Afterwards, we show in \Cref{sec:additional-mmlu-results} additional quantitative results for the expertise-based impersonation results.
\Cref{sec:vision_and_language_task} provides additional details about the vision and language tasks.
Finally, \Cref{sec:limitations} discusses some limitations of our work.

The code to reproduce our results is available here: \url{https://github.com/ExplainableML/in-context-impersonation}. For more details on the code please refer to the \texttt{README.md} file.

\vspace{5ex}

\section{Experimental Details}%
\label{sec:experimental_details}
This section describes the meta-prompt that we used to generate the prompt variations (Section \Cref{subsec:prompt_vars}) and the amount of compute required to reproduce our experiments (Section \Cref{subsec:compute-reproduction})

\subsection{Prompt variations generated by meta-prompting}%
\label{subsec:prompt_vars}
As LLMs have been found to be sensitive to specific prompts~\cite{Arora2022AskMA}, we follow the meta-prompting approach from~\cite{reynolds2021prompt} to vary our original impersonation prompt \texttt{If you were a \{persona\}}.
We generated five additional variations with the following meta-prompt:
\begin{center}
    \texttt{
Write 5 different grammatical and linguistic variations of the following instruction. You shall not fill in the curly brackets:\\
If you were a \{persona\}
    }
\end{center}

The following enumeration lists all generated prompts, which we used in addition to the original prompt:
\texttt{
\begin{itemize}
    \item Should you be transformed into a \{persona\}
    \item Imagine you are a \{persona\}
    \item Should you assume the role of a \{persona\}
    \item Were you to take on the persona of a \{persona\}
    \item In the case of you being a \{persona\}
\end{itemize}
}

\subsection{Compute and Reproduction}%
\label{subsec:compute-reproduction}
For all Vicuna-13B based experiments (bandit, reasoning and vision) we used a single Nvidia A100-40GB GPU\@.
The weights for this language model can be obtained from its open source documentation\footnote{\url{https://github.com/lm-sys/FastChat}}, making our Vicuna-13B based experiments fully reproducible.
For our ChatGPT vision experiments we used Nvidia 2080ti 11GB GPUs to run the CLIP models.

For the bandit task, we chose to run 12k games (2k per prompt variation) with Vicuna-13B to obtain a large sample size for our analysis. Trials and games were run sequentially for approximately 3.5 hours per persona. Processing games and personas in parallel through batching could reduce the time needed for this experiment significantly.

For the Vicuna-13B reasoning experiments, running (sequentially) all 57 tasks and personas considered takes about 12 hours for a single prompt variation.

For the Vicuna-13B vision and language experiments, generating the descriptions for a single persona and for 196 (Stanford Cars) or 200 (CUB) classes and running CLIP zero shot classification with them on the entire test splits takes approximately an hour for a single impersonation prompt. 

\section{Bandit Task --- Results for Llama 2}%
\label{sec:bandit_llama2}
Most open-source models such as Vicuna are fine-tuned from the same base model Llama~\cite{touvron2023llama}. Recently, a new foundational open-source model, Llama 2 (70B, Chat variant)~\cite{Touvron2023Llama2O} has been released which is significantly larger than Vicuna-13B and has been trained on more data. We rerun the bandit experiments using Llama 2 and come to the same conclusions. The effect of age in the range of 2--20 on the reward is $\beta=0.17$ ($p<.001$) and $\beta=0.26$ ($p<.001$) for Vicuna-13B and Llama 2, respectively.

\section{Reasoning Task}
This section describes additional results regarding the MMLU reasoning task. We start by complementing the results of the main paper by presenting all 57 individual task plots in Section~\ref{sec:additional-mmlu-results}. We then present a comparison between our prompt and the official MMLU prompt~\cite{hendrycks2021measuring} in Section~\ref{subsec:mmlu_prompt} and, lastly, present results on race and gender social categories in Section~\ref{subsec:social_mmlu}. All experiments are conducted on both Vicuna-13B and ChatGPT\@.

\subsection{Additional quantitative results for expertise-based impersonation}%
\label{sec:additional-mmlu-results}

In the main paper, only a part of the Vicuna~\cite{chiang2023vicuna} related results were included for the MMLU~\cite{hendrycks2021measuring} reasoning task, for which the LLM is prompted with a question and four answer options. Thus, in this section, we simultaneously provide the Vicuna-13B individual results for all 57 tasks considered, and a comparison with ChatGPT\@. These experiments are the result of the six prompt variations described in Section~\ref{subsec:prompt_vars}.

Contrary to Vicuna, which is an open source model, ChatGPT does not offer direct access to the token probabilities. Therefore, we add the following expression to the Vicuna prompt mentioned in the main paper \texttt{Answer:} \texttt{The answer is option}, in order to force ChatGPT to provide one of the 4 options as the first generated token. This generated token is then taken as the LLM prediction. When ChatGPT does not provide one of the options as the first token, we repeat the question until a valid option is generated or until a maximum of 10 tries. If none of these conditions are met, we discard the sample. For example, for the STEM and Humanities domains, in about 250k questions (7835 unique questions, each of which evaluated for the 32 personas of these domains), only 178 were discarded (0.07\%).

The aforementioned results are presented in Figures~\ref{fig:mmlu_stem},~\ref{fig:mmlu_humanities},~\ref{fig:mmlu_social_sciences}, and~\ref{fig:mmlu_other}, for the STEM, Humanities, Social Sciences, and Other domains, respectively. ChatGPT performs consistently better than Vicuna-13B, which is also in line with the expectation given that ChatGPT is a larger model trained on more and higher quality (human feedback) data. Furthermore, as discussed in the main paper for Vicuna and again observed for ChatGPT, the performance on Humanities tasks is consistently higher than on STEM tasks, which aligns with previous literature.
For Vicuna-13B, the tasks where the trend is not verified (i.e.\ where the task expert does not outperform the domain expert and/or where the domain expert does not surpass the non-domain expert), coincide with tasks that the model could not perform well in general, i.e.\ had accuracies close to or below the random baseline for all personas considered (see Formal Logic in Figure~\ref{fig:mmlu_humanities} or, for example, College Chemistry, College Computer Science, High School Statistics).
For ChatGPT, the tasks where the trend is not as clear coincide with tasks where Vicuna also had worse results.
Interestingly, the neutral persona performs on par with the domain expert. 
Additionally, for the Other domain, ChatGPTs' expertise trends are not as clear, which might be due to the fact that this domain includes tasks from a very wide range of domains, such as Nutrition and Business Ethics, for example.
Nevertheless, the non-domain expert is outperformed by the domain expert, who in turn is outperformed by the task expert for all four domains.

\include{figures_supp/mmlu_tasks_ChatGPTVicuna_figure_CR}

\subsection{MMLU Task Formulation}%
\begin{figure}[ht!]
     \centering
     \vspace{-3mm}
     \begin{subfigure}[t]{0.244\textwidth}
         \centering
         \includegraphics[width=\textwidth]{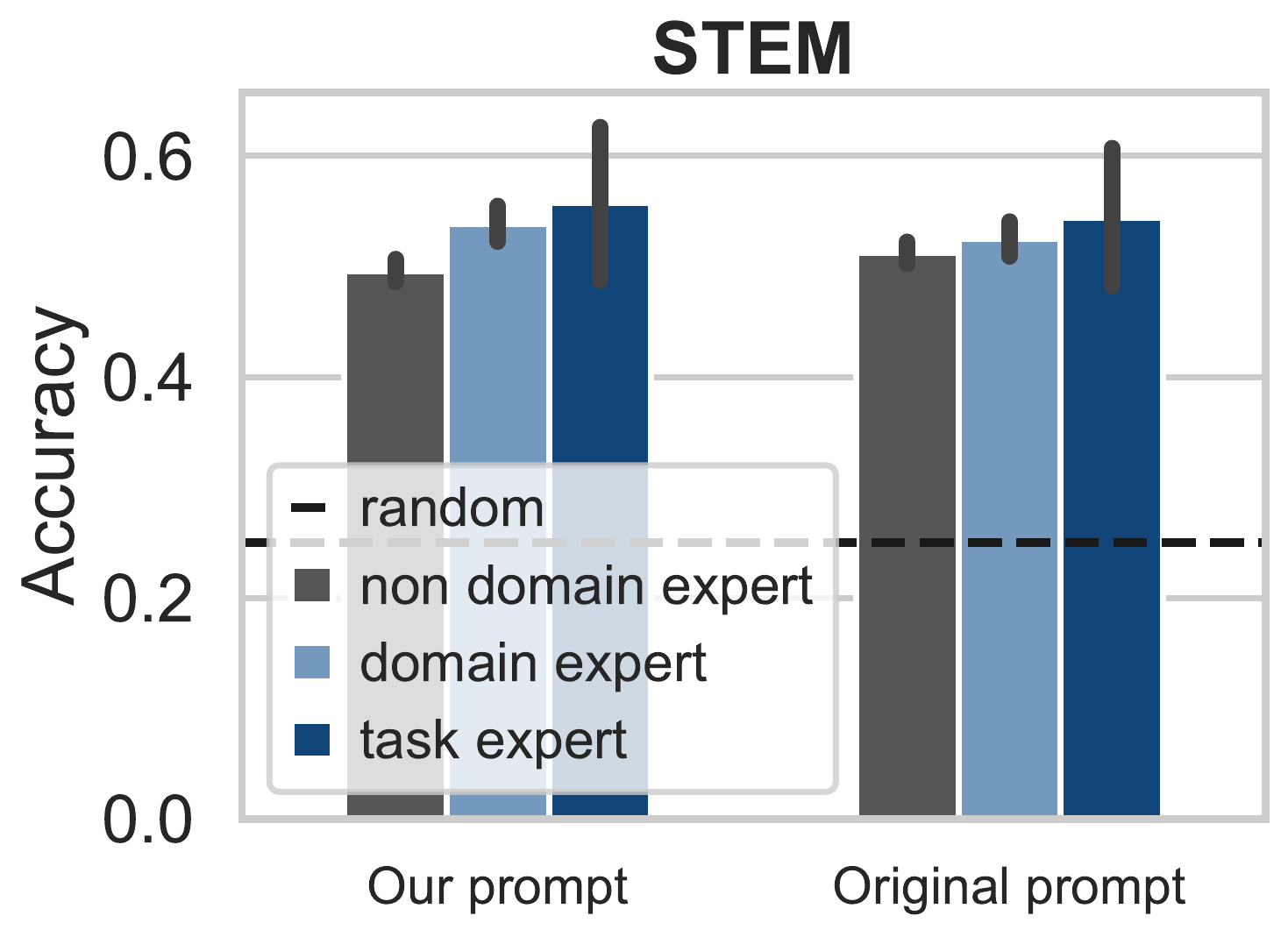}
     \end{subfigure}
     \hfill
     \begin{subfigure}[t]{0.244\textwidth}
         \centering
         \includegraphics[width=\textwidth]{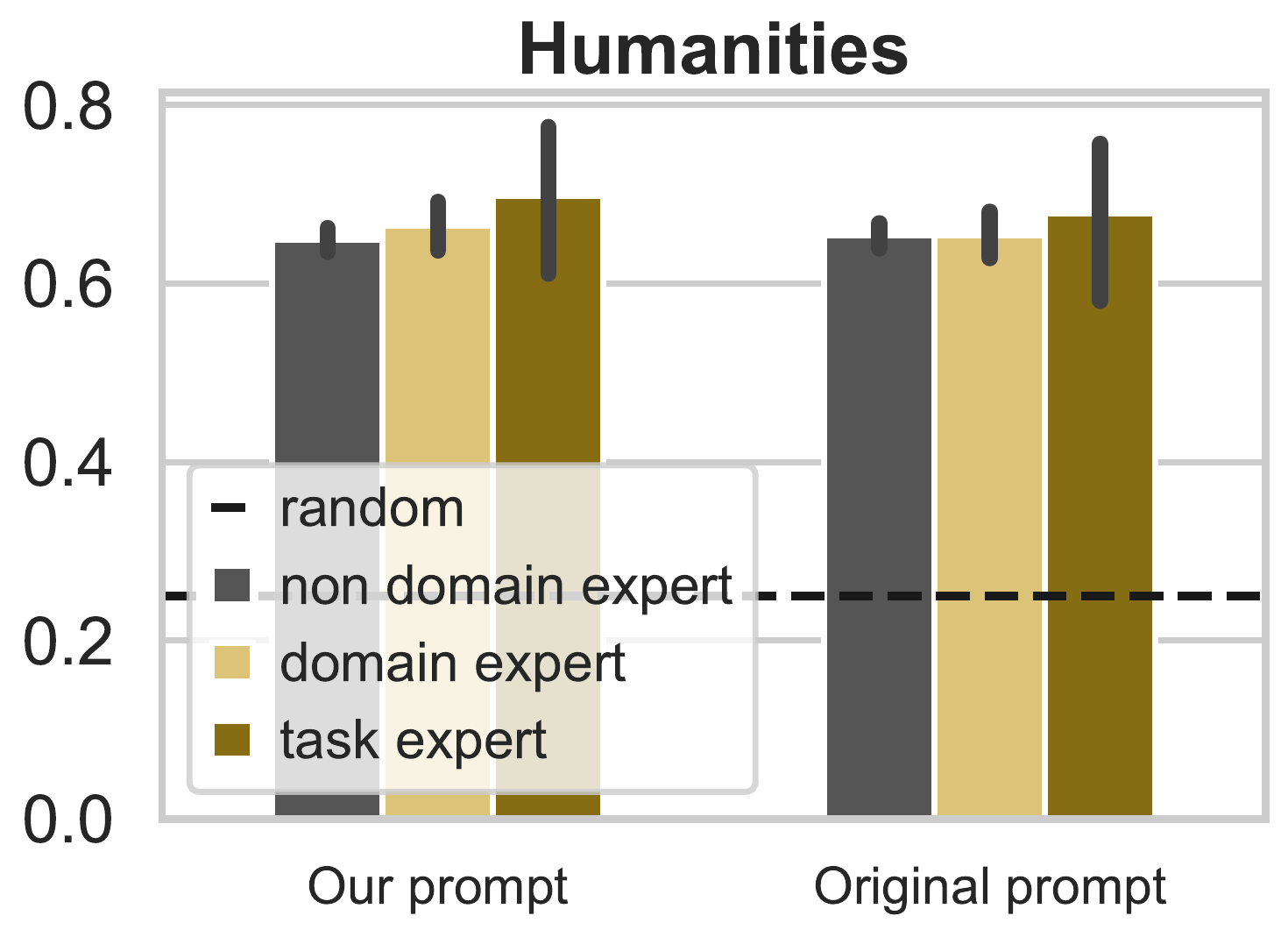}
     \end{subfigure}
    \hfill
    \begin{subfigure}[t]{0.244\textwidth}
     \centering
     \includegraphics[width=\textwidth]{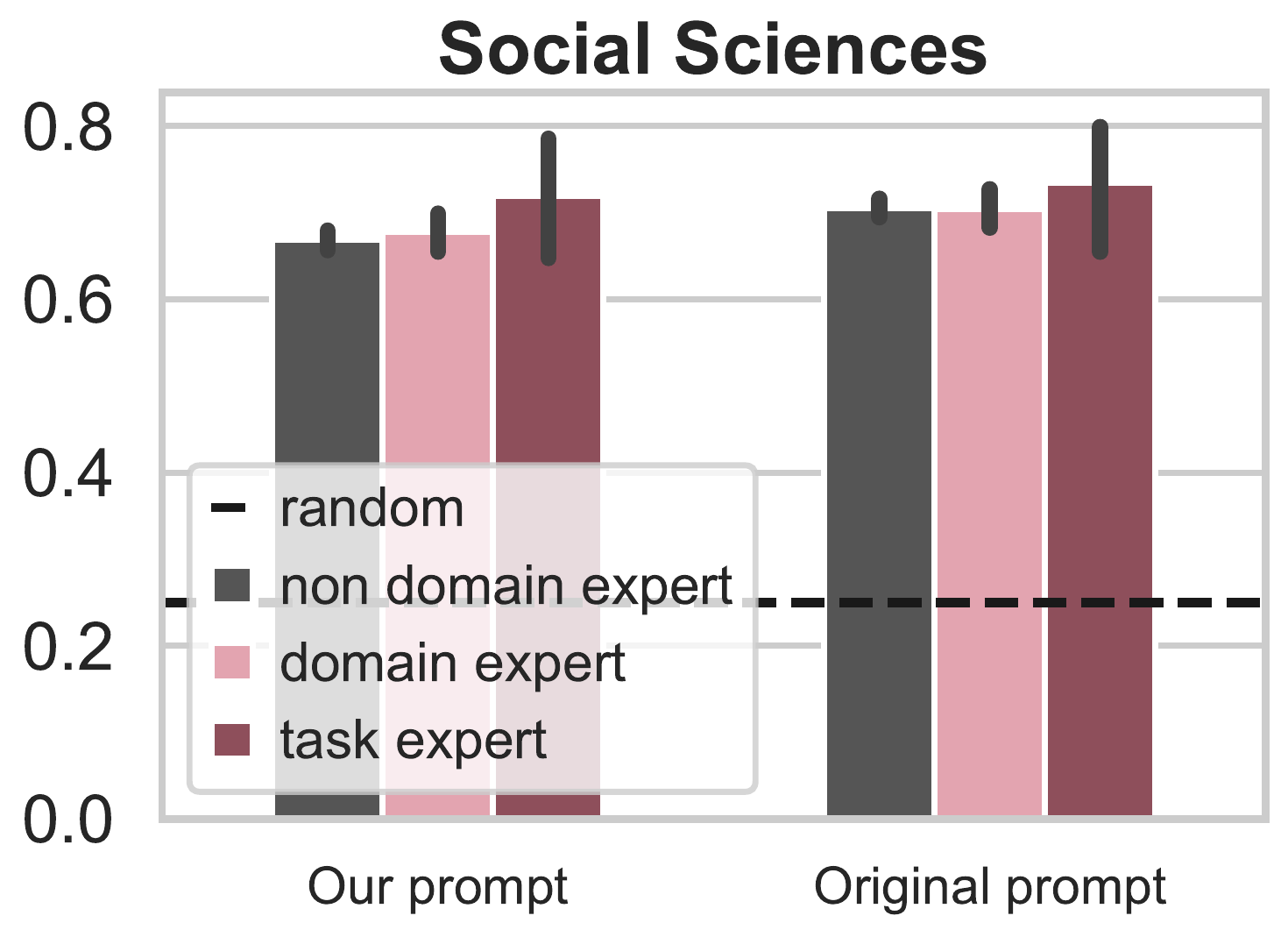}
    \end{subfigure}
    \hfill
    \begin{subfigure}[t]{0.244\textwidth}
         \centering
         \includegraphics[width=\textwidth]{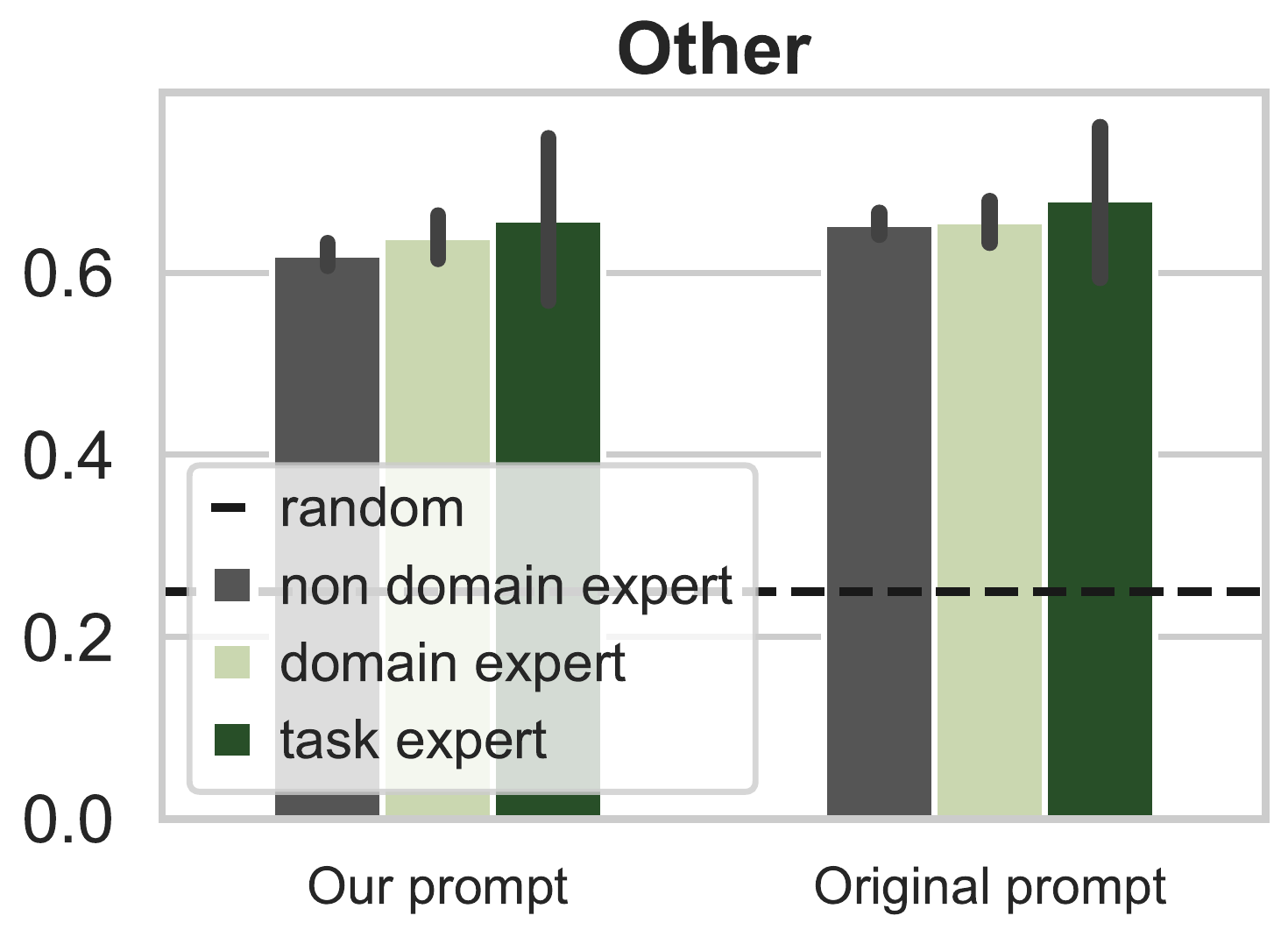}
     \end{subfigure}
    \\
    \begin{subfigure}[t]{0.244\textwidth}
        \centering
         \includegraphics[width=\textwidth]{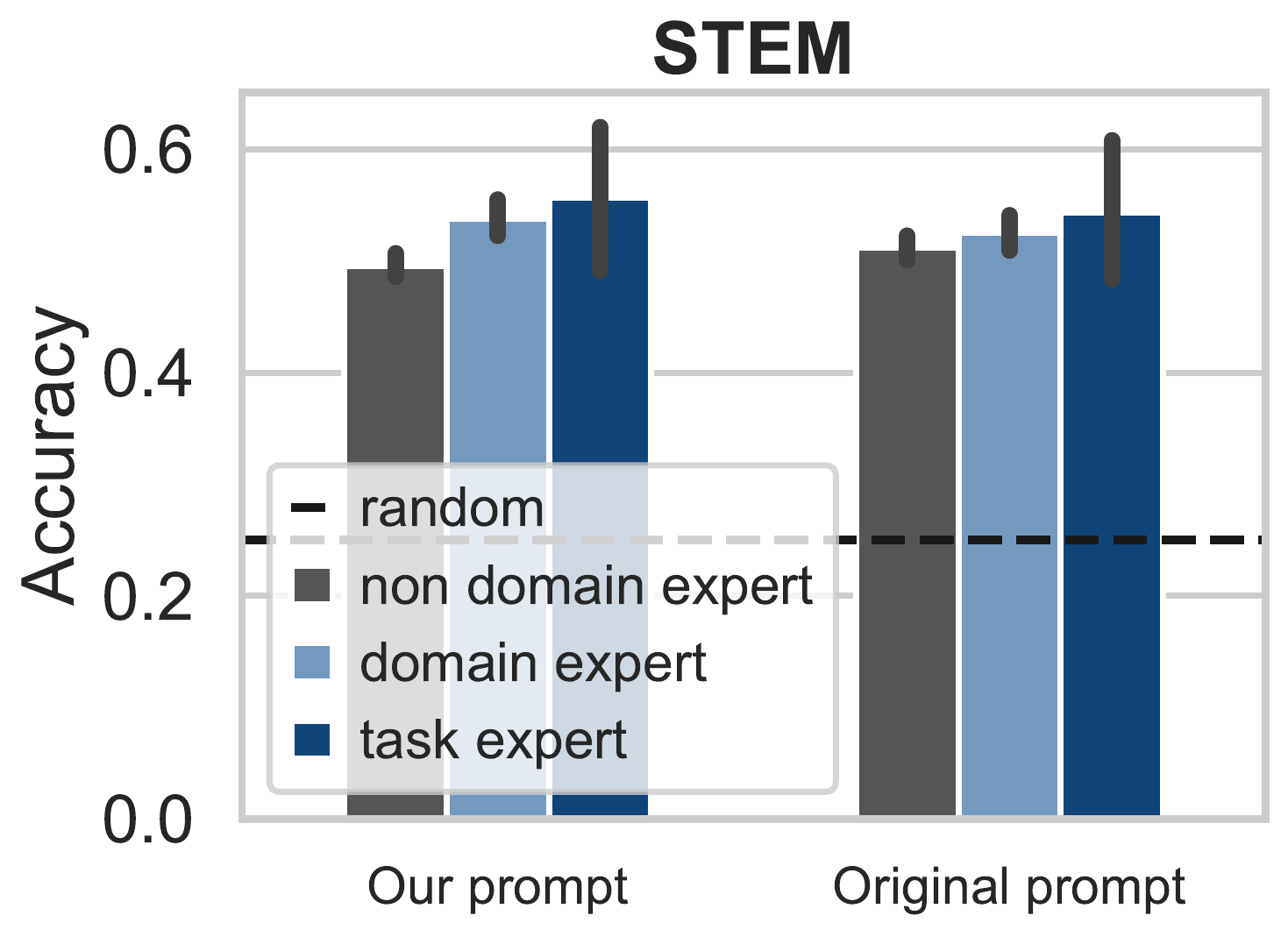}
    \end{subfigure}
    \hfill
    \begin{subfigure}[t]{0.244\textwidth}
         \centering
         \includegraphics[width=\textwidth]{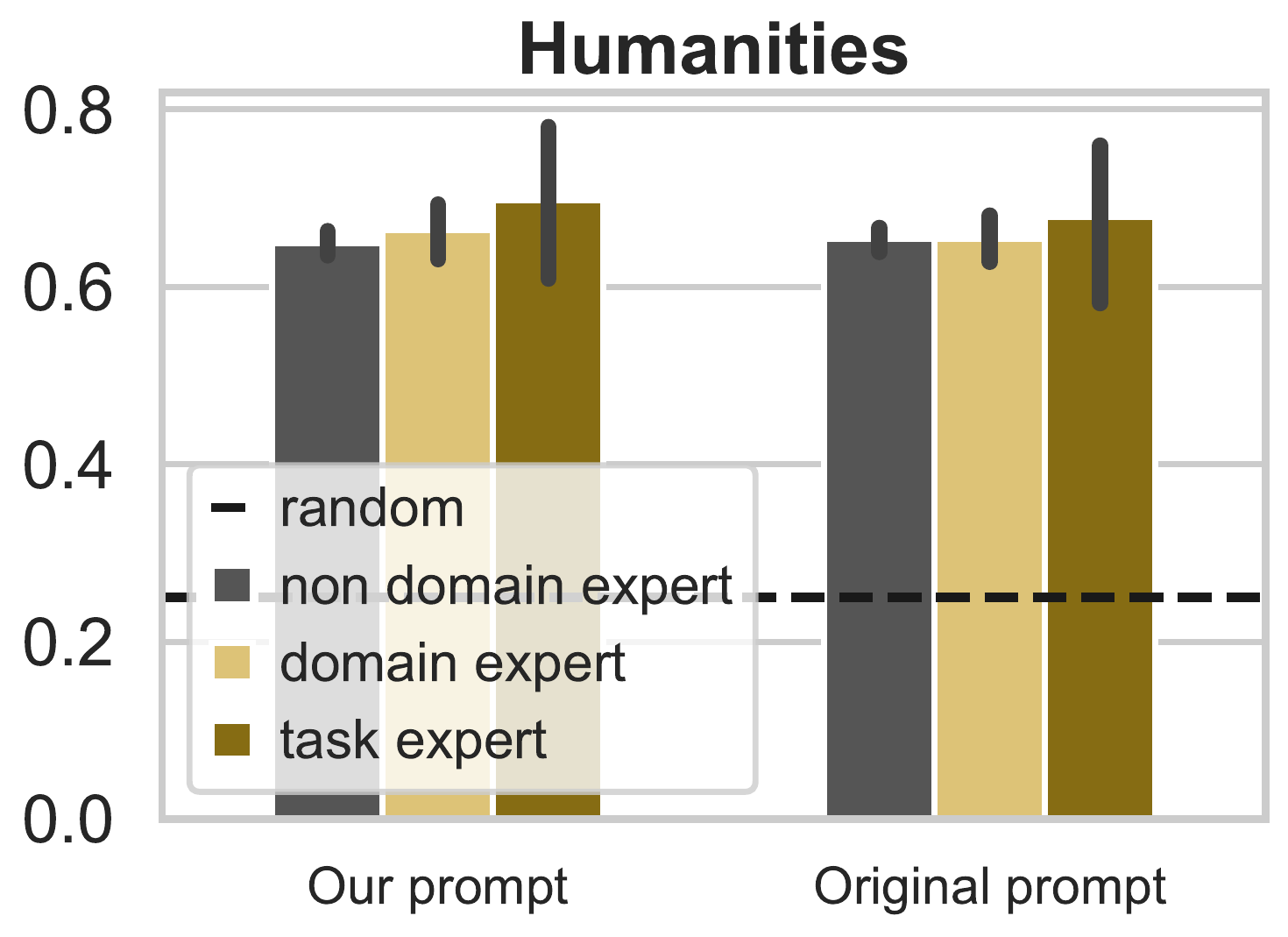}
     \end{subfigure}
    \hfill
    \begin{subfigure}[t]{0.244\textwidth}
        \centering
         \includegraphics[width=\textwidth]{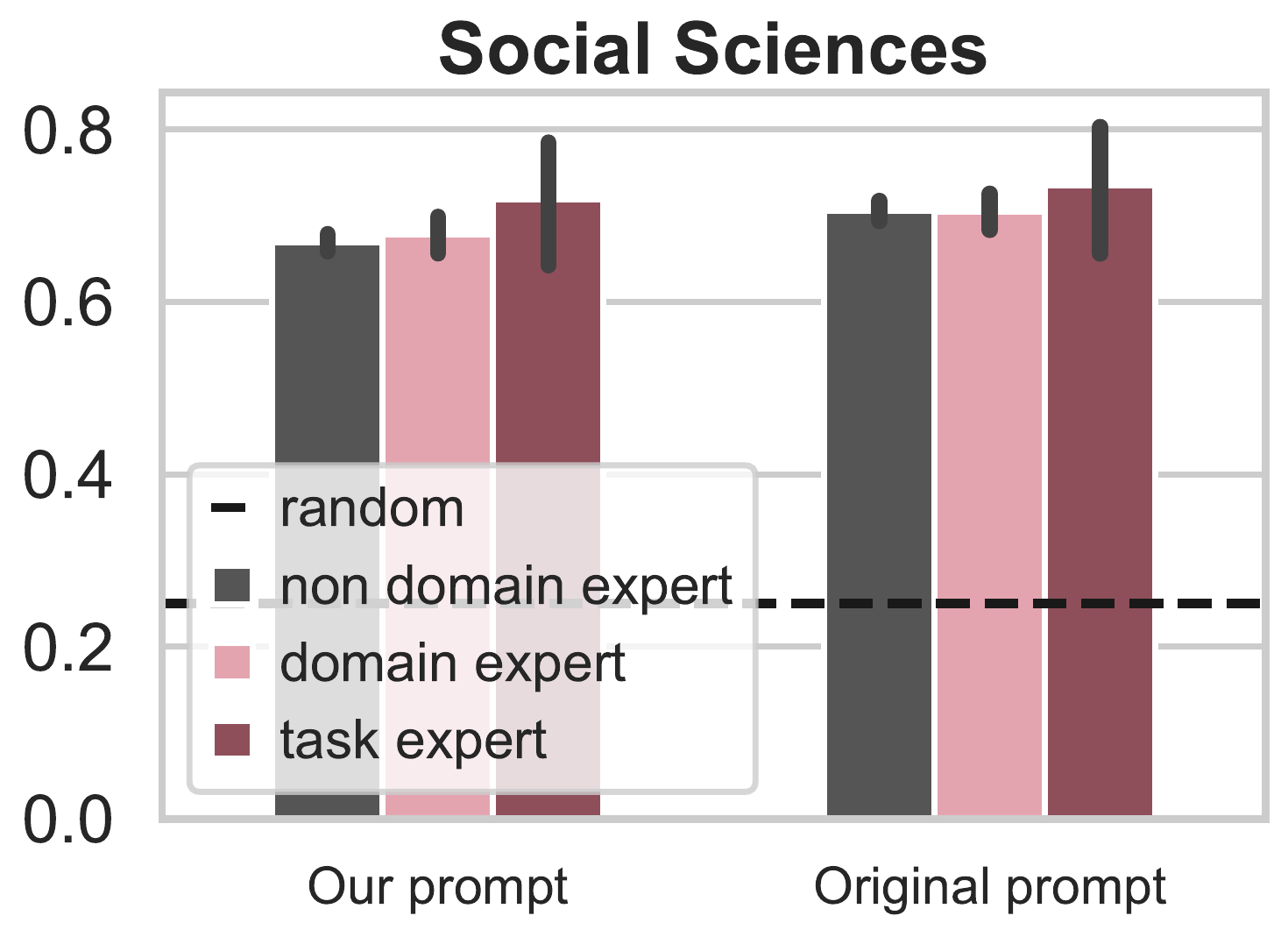}
    \end{subfigure}
    \hfill
    \begin{subfigure}[t]{0.244\textwidth}
         \centering
         \includegraphics[width=\textwidth]{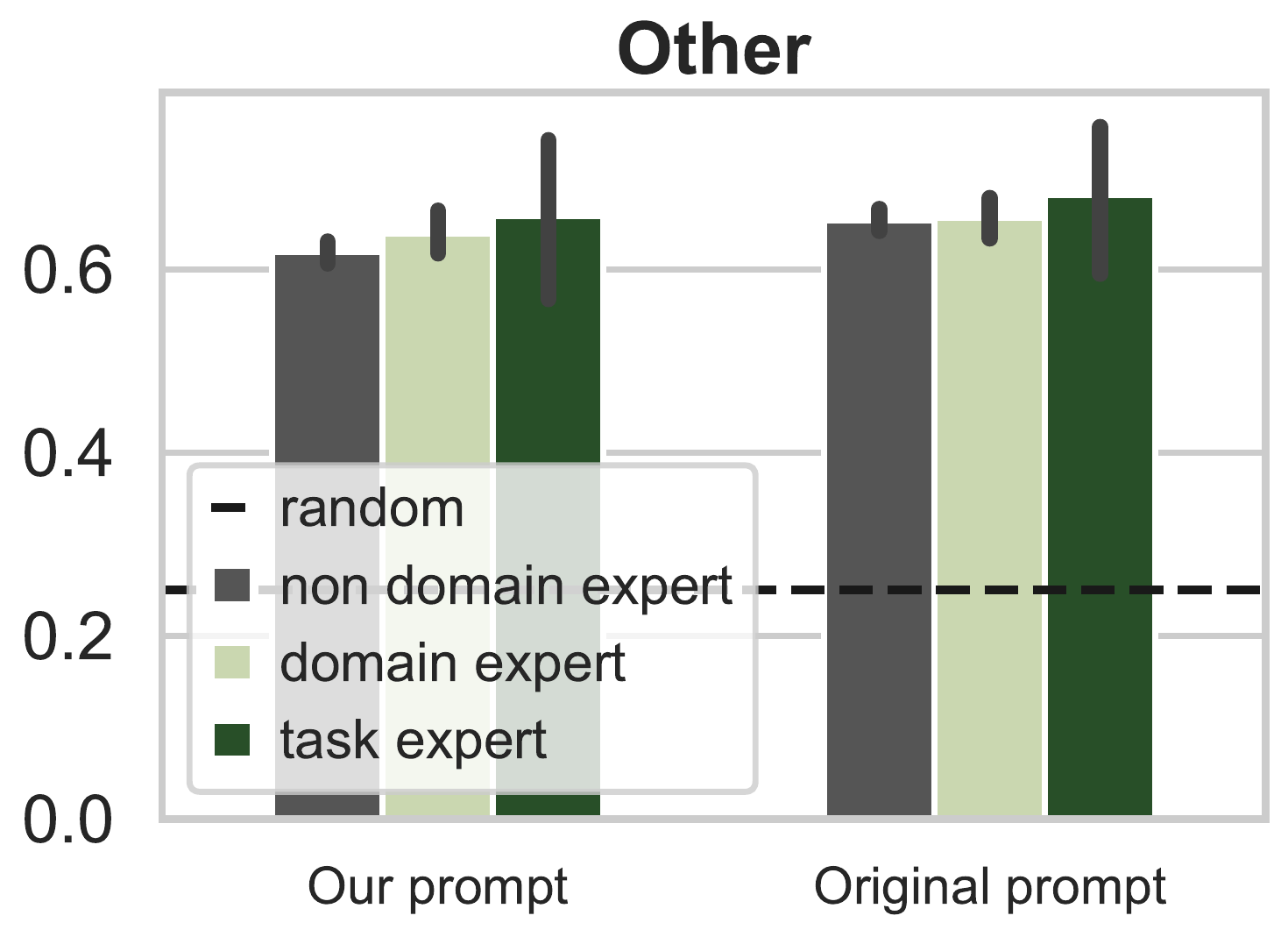}
     \end{subfigure}
     \caption{Comparison between our task formulation (Our prompt) and the official MMLU prompt~\cite{hendrycks2021measuring} (Original prompt), for Vicuna-13B (top) and ChatGPT (bottom).}%
    \label{fig:mmlu_prompt}
\end{figure}

\label{subsec:mmlu_prompt}
Since several MMLU evaluations~\cite{hendrycks2021measuring, Liang2023HolisticEO}, may lead to small variations when comparing different models’ ranks, we include results with the MMLU official prompt (see Figure~\ref{fig:mmlu_prompt}), i.e.\ by using the MMLU prompt at the start and keeping our impersonation strategy. Our expertise-based impersonation trends still hold, and the absolute accuracy values improve. This increase in accuracy might be explained by the fact that the official MMLU prompt includes the task name in the prompt, which might provide additional clues to the LLM\@. Thus, we conclude that our findings on impersonation are not dependent on the formulation of the task.

\subsection{Social Categories on MMLU}%
\label{subsec:social_mmlu}

\begin{figure}[ht!]
     \centering
     \vspace{-3mm}
     \begin{subfigure}[t]{0.244\textwidth}
         \centering
         \includegraphics[width=\textwidth]{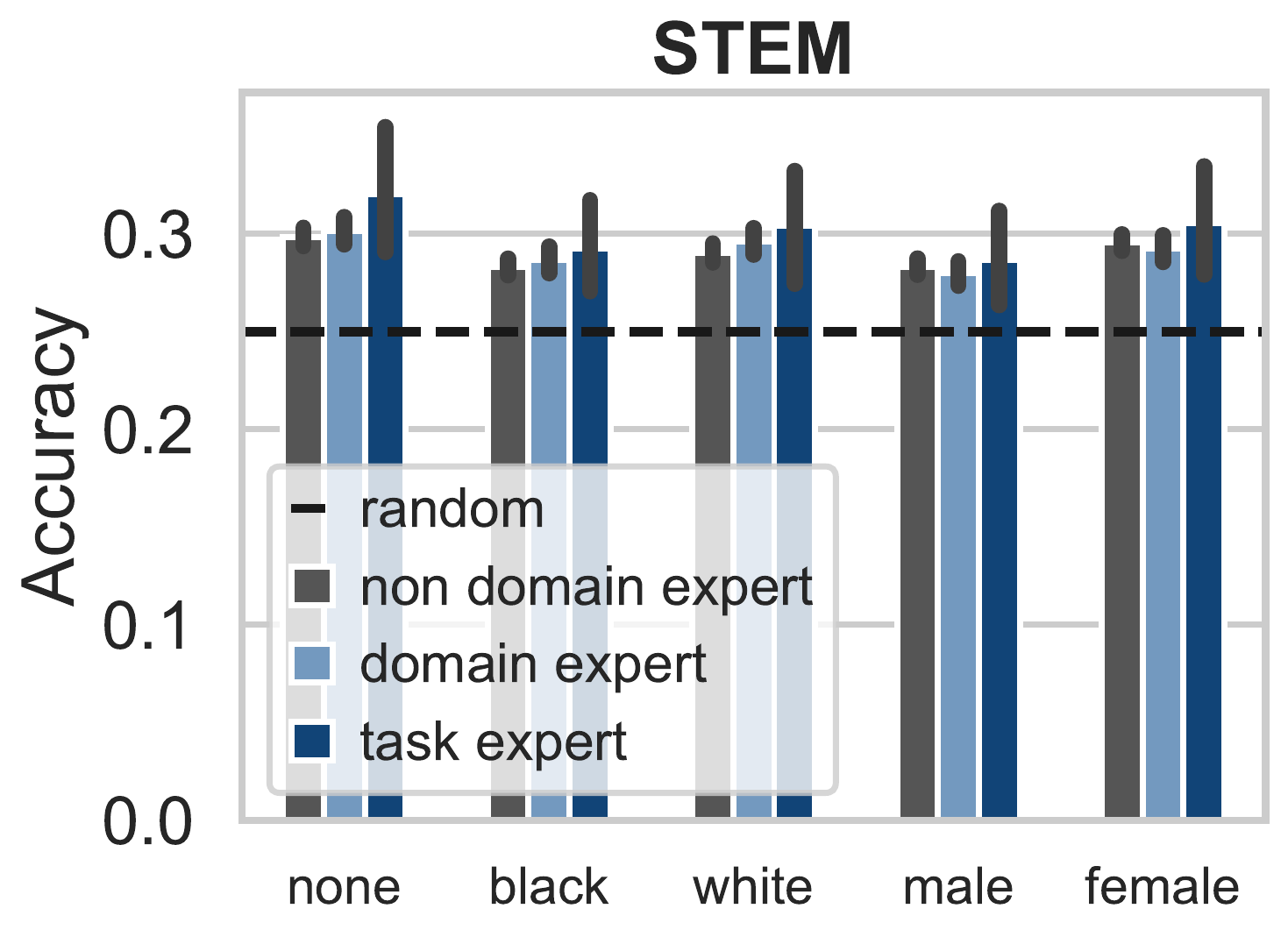}
     \end{subfigure}
     \hfill
     \begin{subfigure}[t]{0.244\textwidth}
         \centering
         \includegraphics[width=\textwidth]{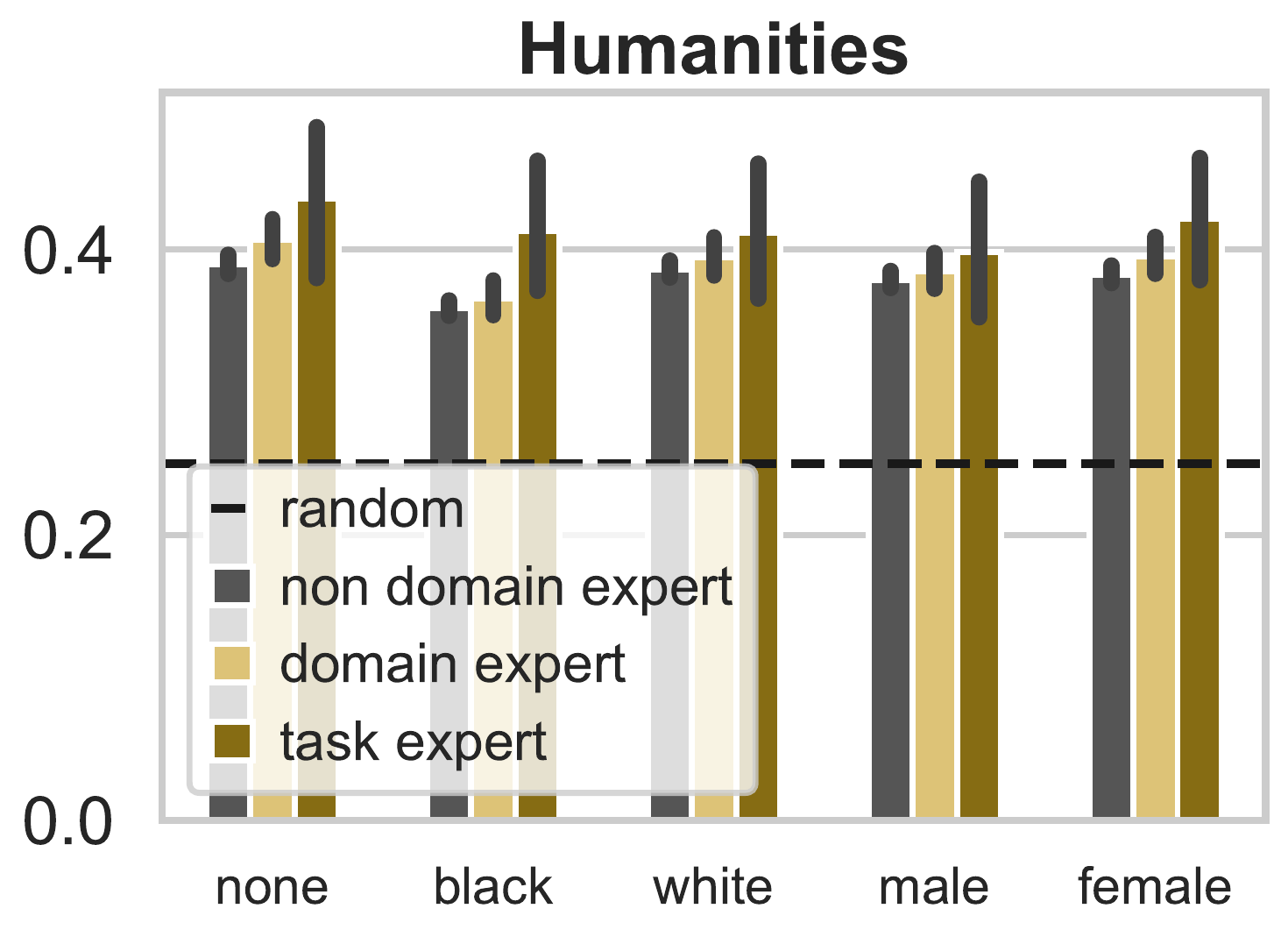}
     \end{subfigure}
    \hfill
    \begin{subfigure}[t]{0.244\textwidth}
     \centering
     \includegraphics[width=\textwidth]{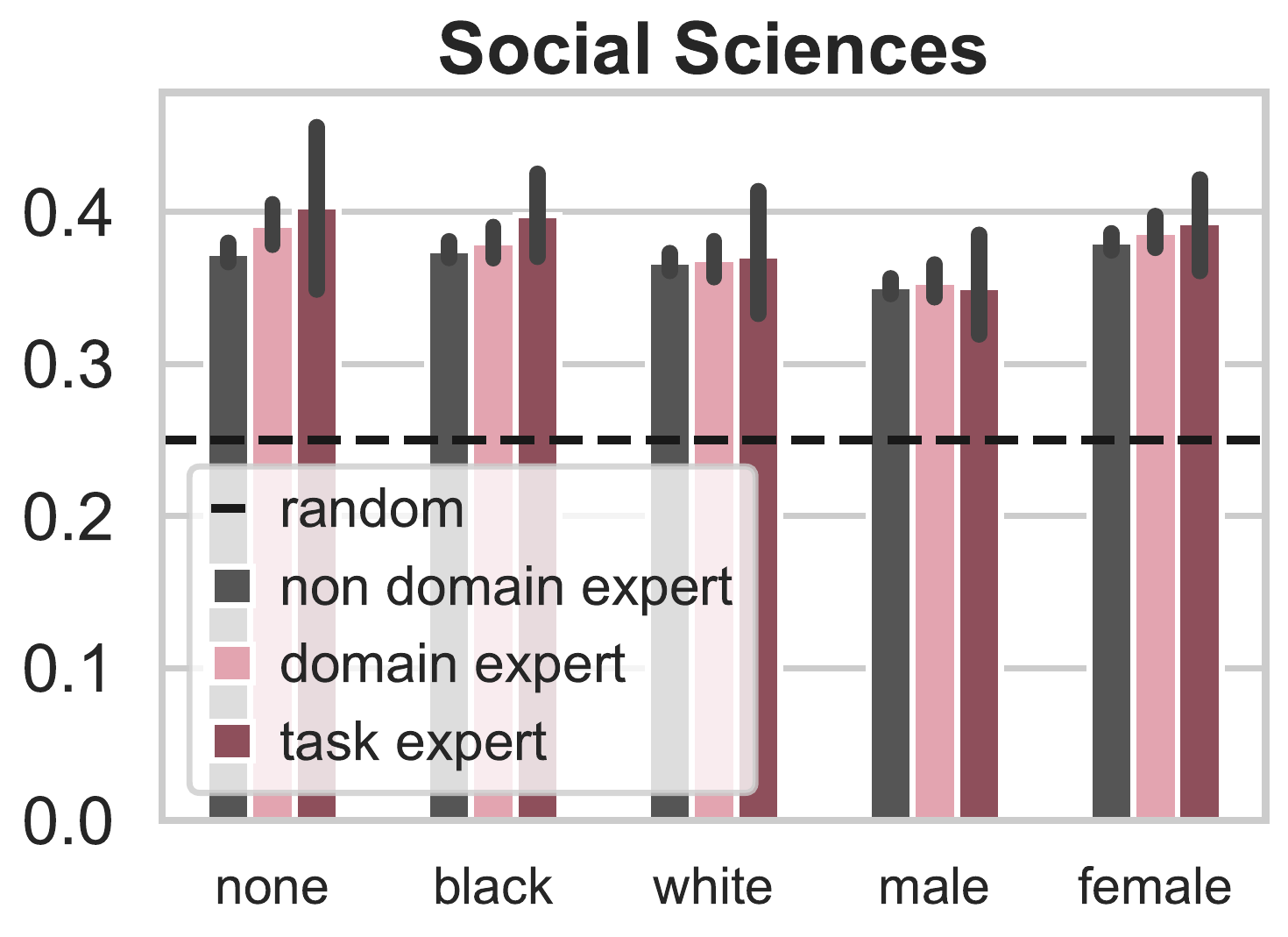}
    \end{subfigure}
    \hfill
    \begin{subfigure}[t]{0.244\textwidth}
         \centering
         \includegraphics[width=\textwidth]{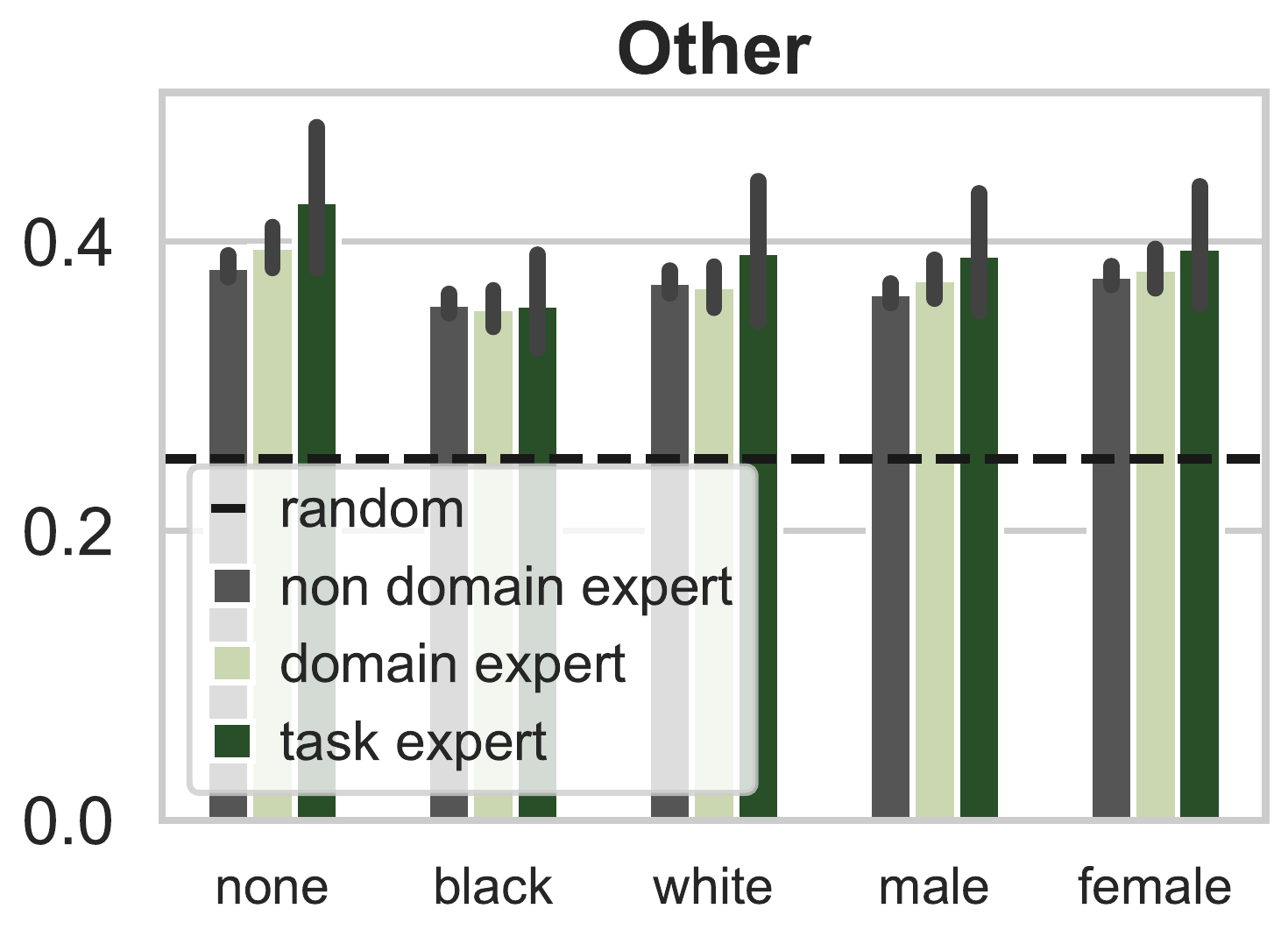}
     \end{subfigure}
    \\
    \begin{subfigure}[t]{0.244\textwidth}
        \centering
         \includegraphics[width=\textwidth]{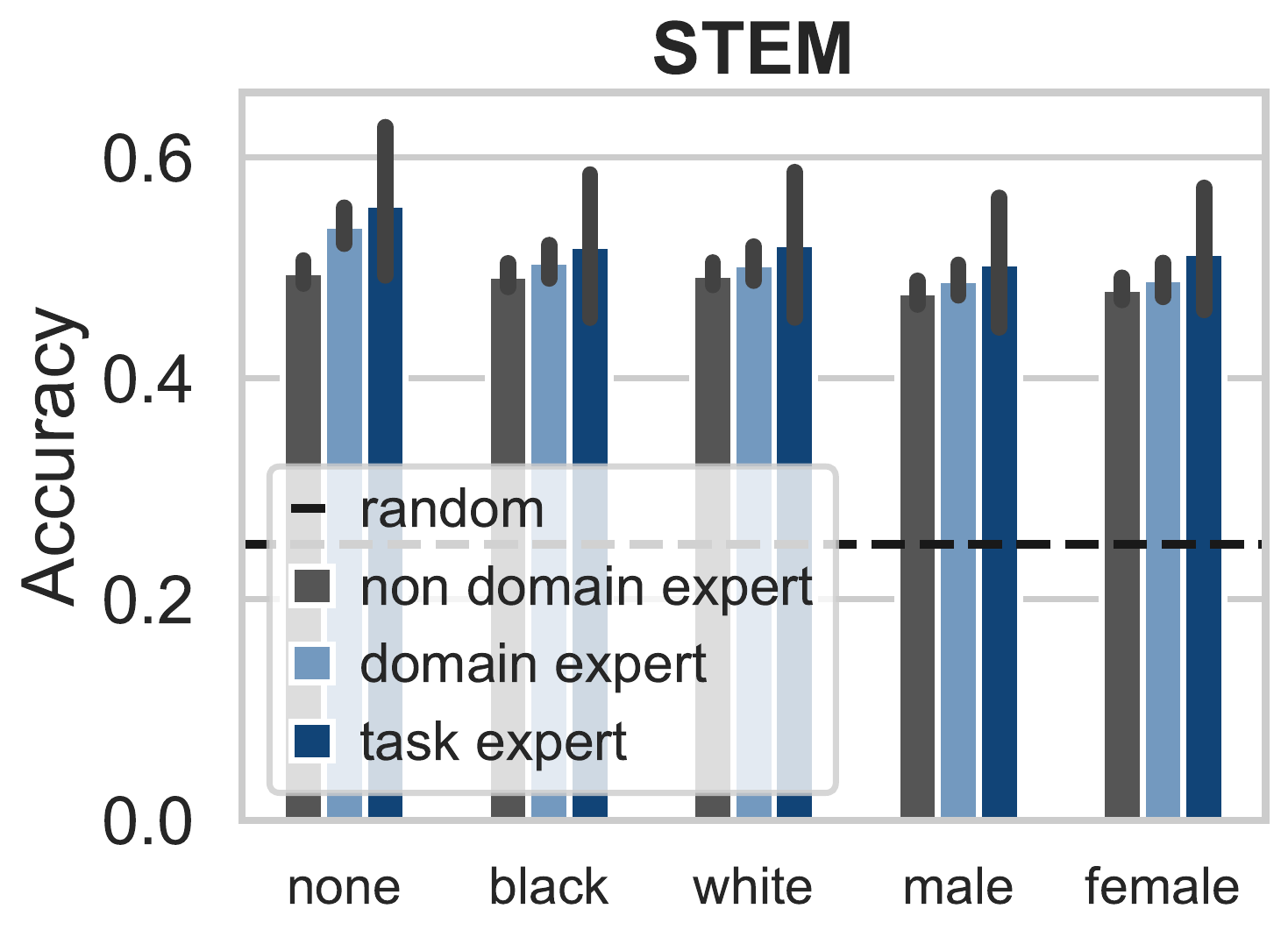}
    \end{subfigure}
    \hfill
    \begin{subfigure}[t]{0.244\textwidth}
         \centering
         \includegraphics[width=\textwidth]{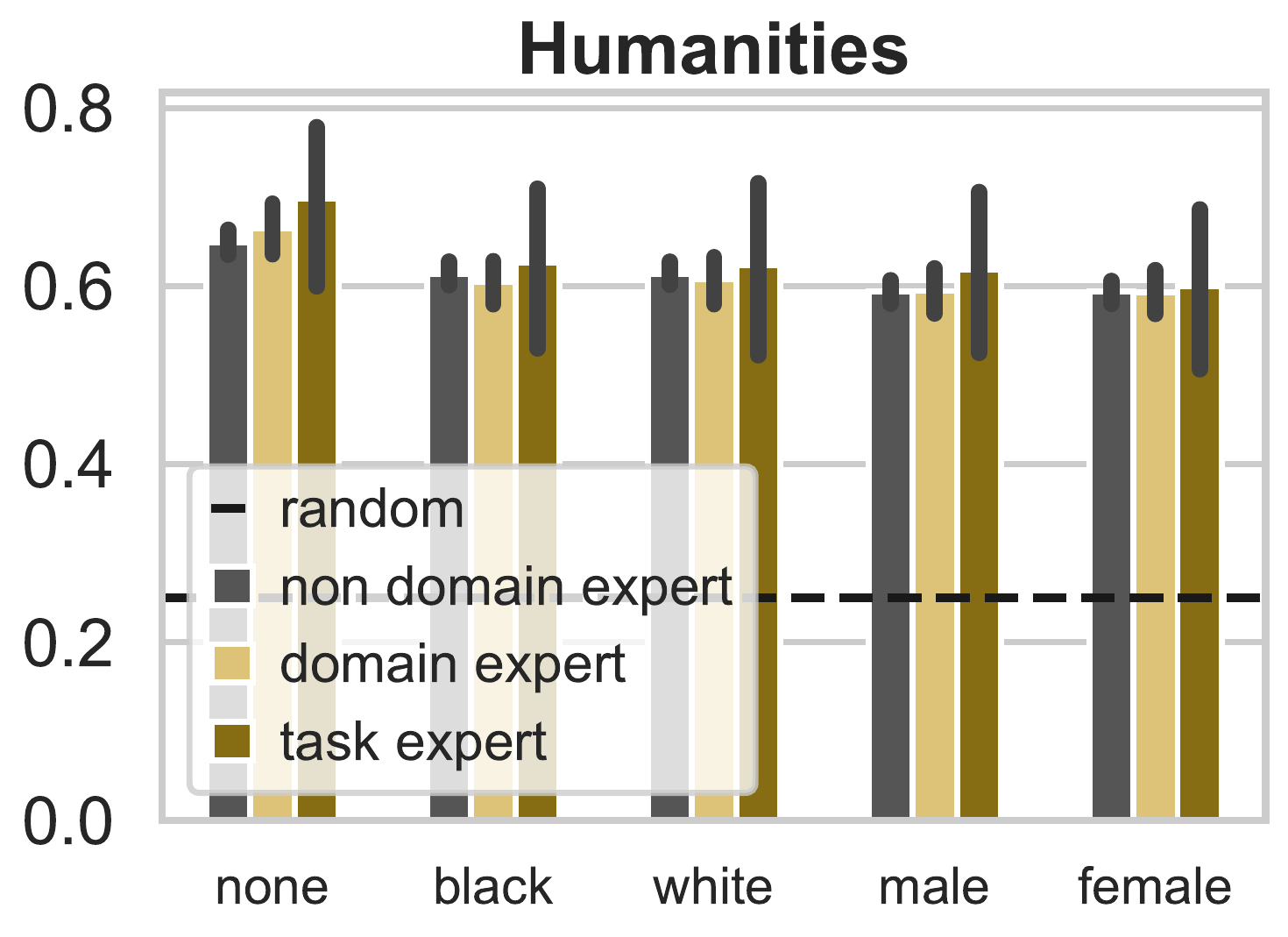}
     \end{subfigure}
    \hfill
    \begin{subfigure}[t]{0.244\textwidth}
        \centering
         \includegraphics[width=\textwidth]{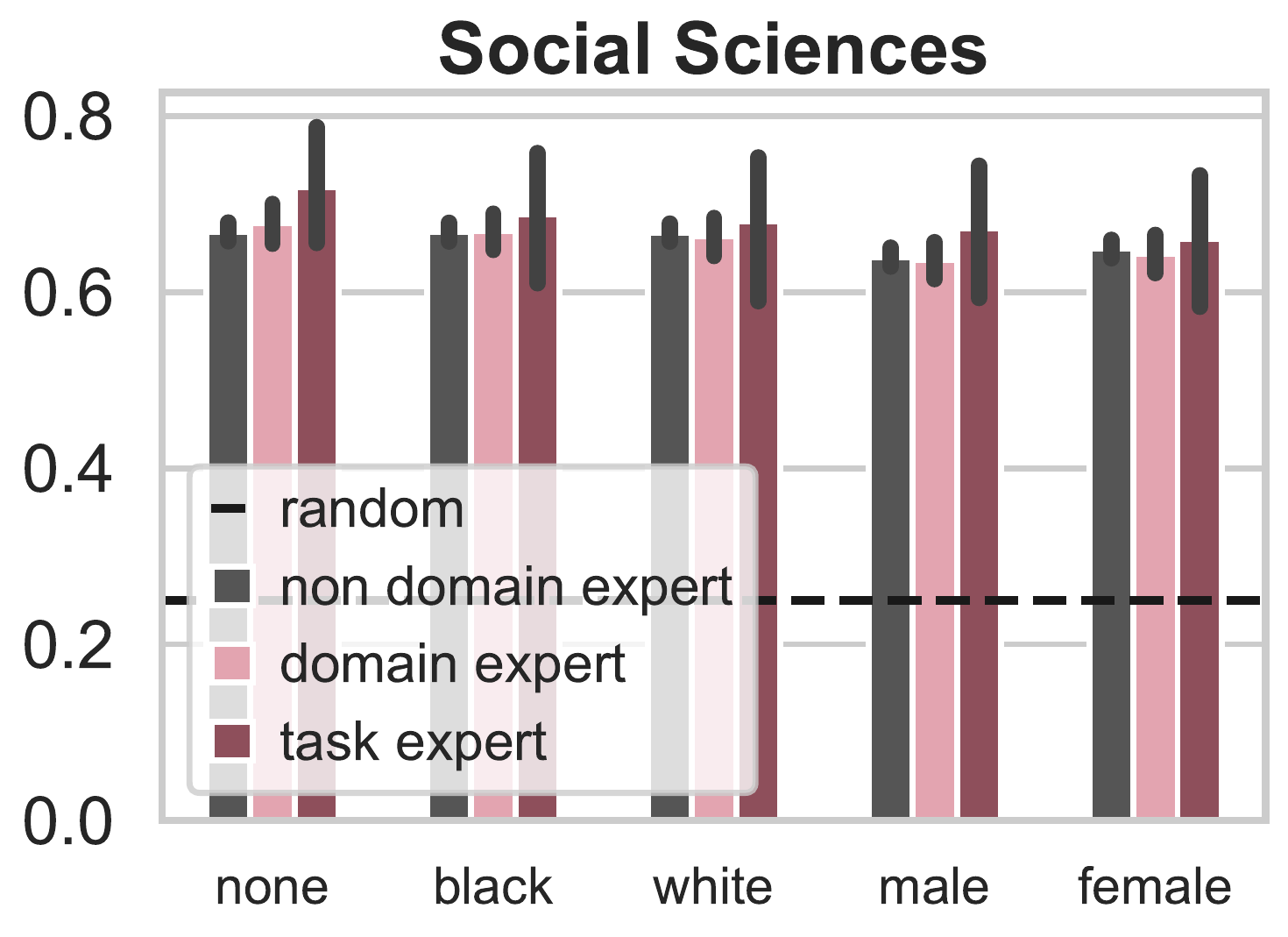}
    \end{subfigure}
    \hfill
    \begin{subfigure}[t]{0.244\textwidth}
         \centering
         \includegraphics[width=\textwidth]{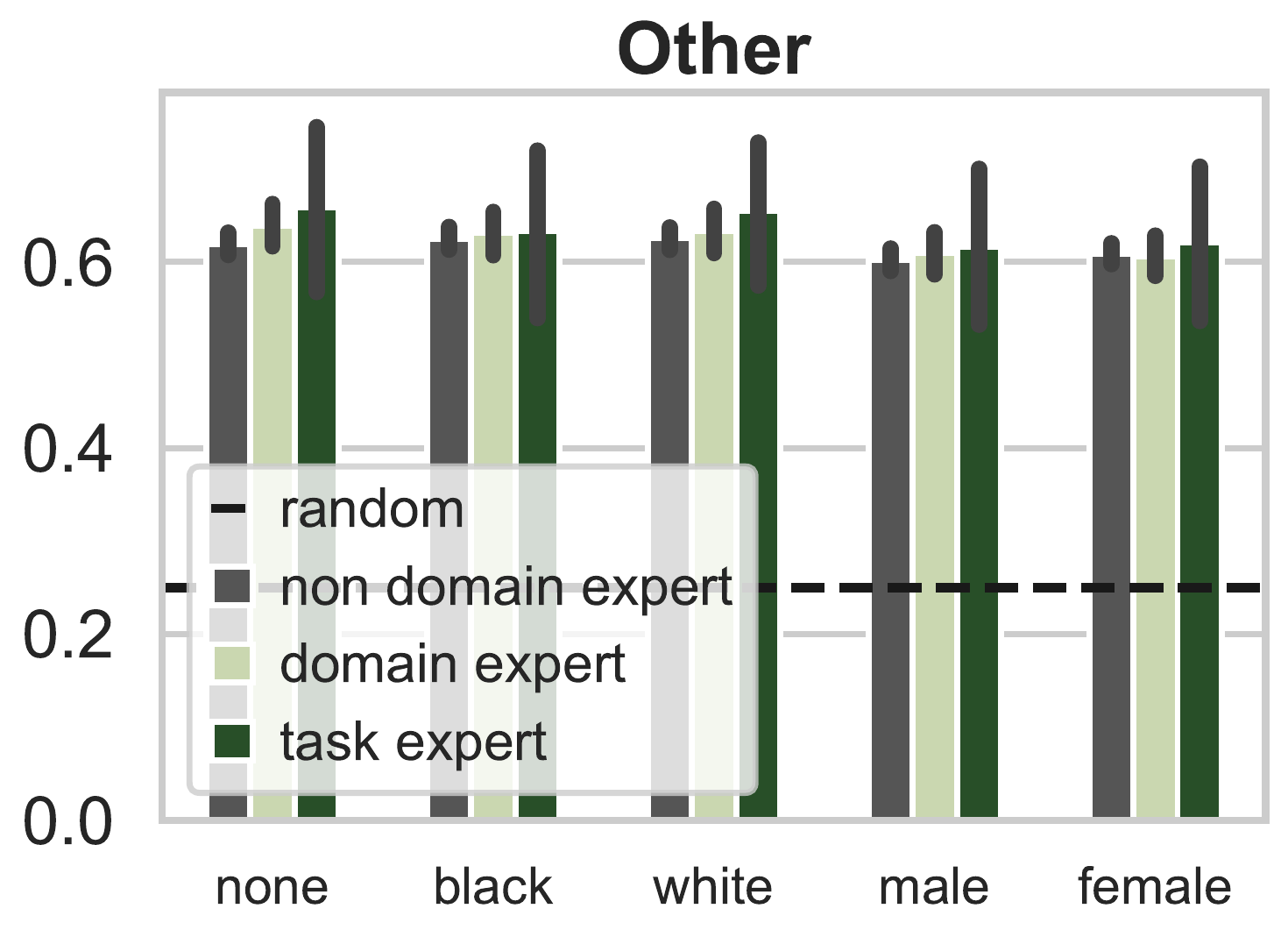}
     \end{subfigure}
     \caption{Expertise-based impersonation results with social category prefixes (black, white, male, and female) for Vicuna-13B (top) and ChatGPT (bottom).}%
     \label{fig:mmlu_social_categs}
\end{figure}

We present in Figure~\ref{fig:mmlu_social_categs} results for both Vicuna-13B (top) and ChatGPT (bottom) on MMLU when considering different social category prefixes (black, white, male, and female). We observe that, for both models, the performance when impersonating experts while adding these prefixes is consistently lower than when no prefix is added (i.e.\ the none columns). For Vicuna, the black persona obtains lower accuracies than the white persona, especially regarding the non-task experts, and a female expert outperforms a male expert. For ChatGPT, all prefixed personas' performance is similar.

\section{Vision and Language Task}%
\label{sec:vision_and_language_task}
In this section we give additional details for the vision and language task.
First, in \Cref{subsec:cleaning-class-names} we describe how class names were removed from the generated visual descriptions to avoid trivial solutions.
Then we show more results on two additional fine-grained visual classification datasets in \Cref{subsec:additional_visual_datasets}.
Next, we show more qualitative examples of the class descriptions generated by Vicuna-13B and ChatGPT in \Cref{subsec:example-descriptions}.
Afterwards, we show more quantitative results on more LLM / VLM pairs in \Cref{subsec:quantitative-results-pairs}.
Lastly, we show more results for additional races and genders (\Cref{subsec:additional_bias_groups}) and for Google PaLM (\Cref{subsec:google_palm}).

\subsection{Removing class names from visual descriptions}%
\label{subsec:cleaning-class-names}
\begin{wrapfigure}[21]{r}{0.4\textwidth}
    \vspace{-2.5ex}
    \centering
    \includegraphics[width=\linewidth]{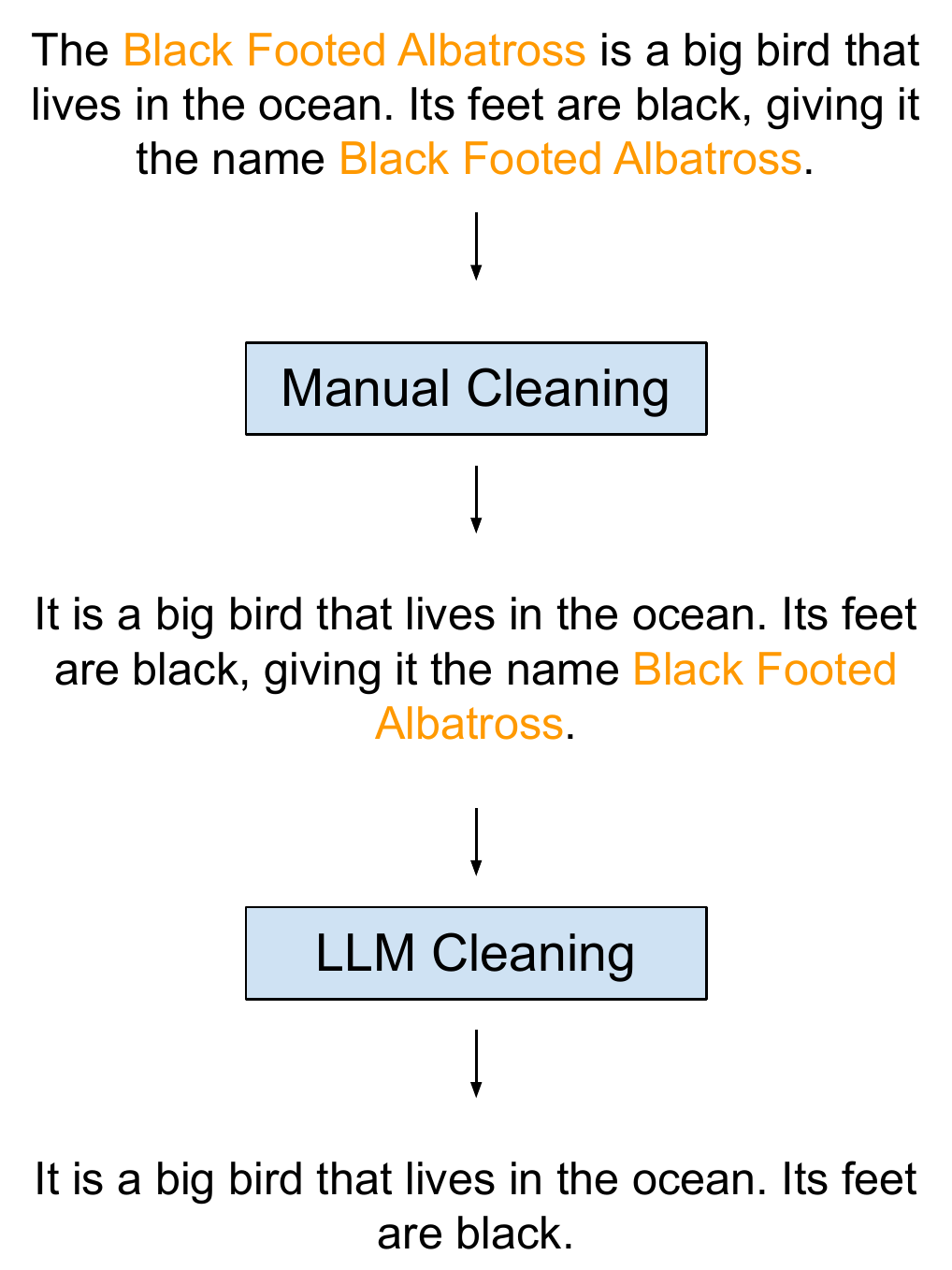} 
    \caption{Example of the two step class name removal process.}%
    \label{fig:cleaning-fig}
\end{wrapfigure}

When the class name is included inside the generated description, it has a significant effect on the downstream performance of the vision task. In such cases CLIP can classify images well without the need of additional descriptions. We find that both language models occasionally use the class name in their output. For example at the beginning of a new sentence. To actually measure how well a persona describes a class, we use a two step process to remove the class name from the descriptions.

\paragraph{Manual cleaning.}
We use a set of heuristics to remove the class name, e.g.\ replacing \texttt{A \{class name\} \{verb\}} with \texttt{It \{verb\}}. 
These heuristics account for the numerous (singular or plural) of the class name as well as for lower and upper casing variants.
Whilst this approach is very fast, it does not scale to all possible variants how the class name could be mentioned in the generated descriptions.

\paragraph{LLM based cleaning.}
For the LLM based cleaning we first split the descriptions into individual sentences with spacy~\cite{spacy2017Natural}.
This simplifies the task for the LLM\@.
To remove the class name in more complex settings we prompt the same LLM used for generating the descriptions with four in-context examples.
Empirically, we find this cleaning approach works well and can also handle more complex cases, e.g.\ removing parts of a sentence if needed. An example of this is shown in~\Cref{fig:cleaning-fig}.
Lastly, if the result still contains the class name we use the original sentence, to avoid introducing any malformed LLM output.

\subsection{Additional visual datasets}%
\begin{figure}[h]
     \centering
      \begin{minipage}[b]{0.03\textwidth}
        \centering
        \begin{turn}{90}
          \begin{minipage}{0.10\textheight}
            \centering
            FGVC Aircraft
          \end{minipage}          
        \end{turn}
      \end{minipage}
     \hfill
     \begin{subfigure}[b]{0.22\textwidth}
         \centering
         \includegraphics[width=\textwidth]{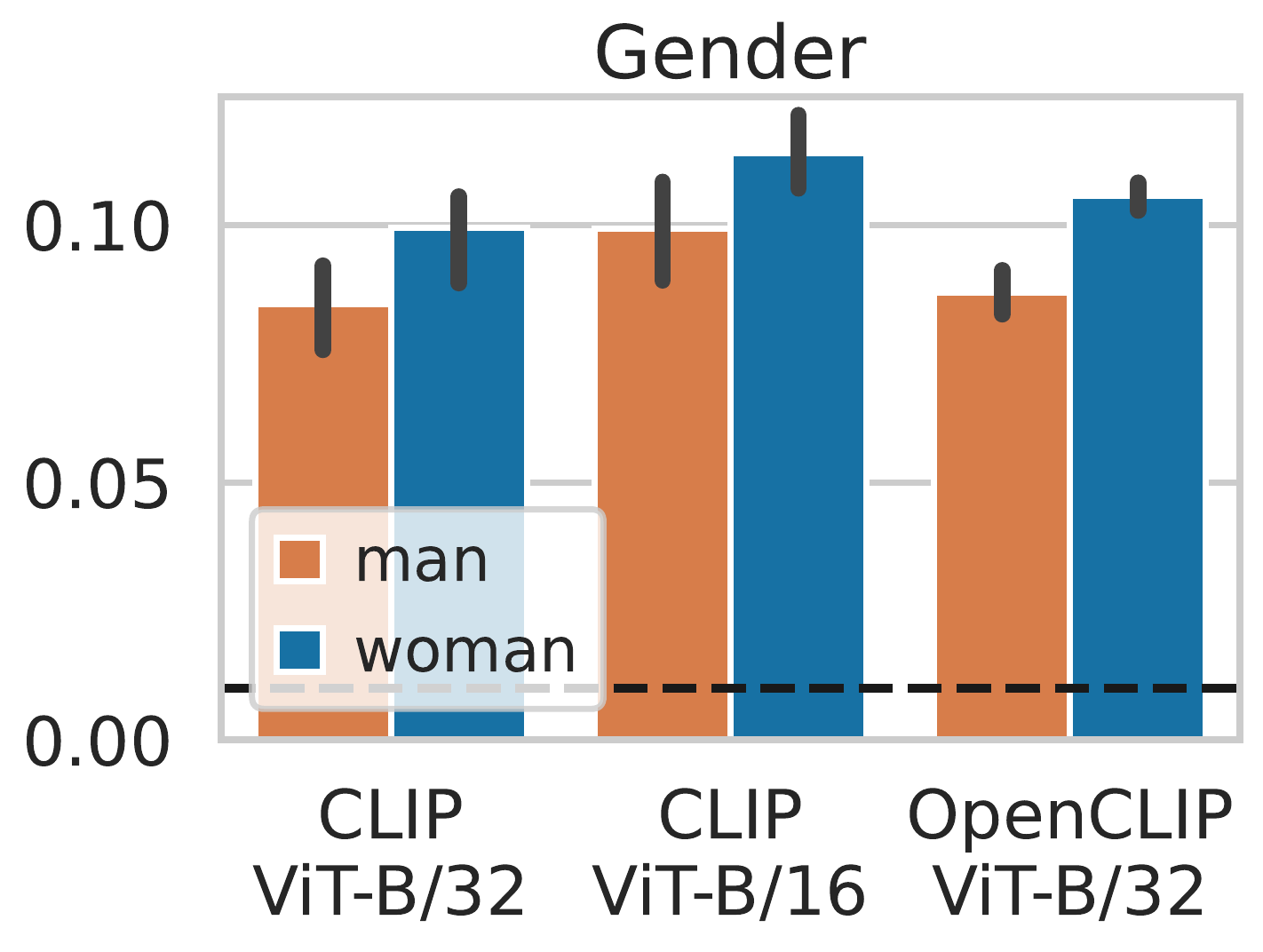}
     \end{subfigure}
      \hfill
     \begin{subfigure}[b]{0.22\textwidth}
         \centering
         \includegraphics[width=\textwidth]{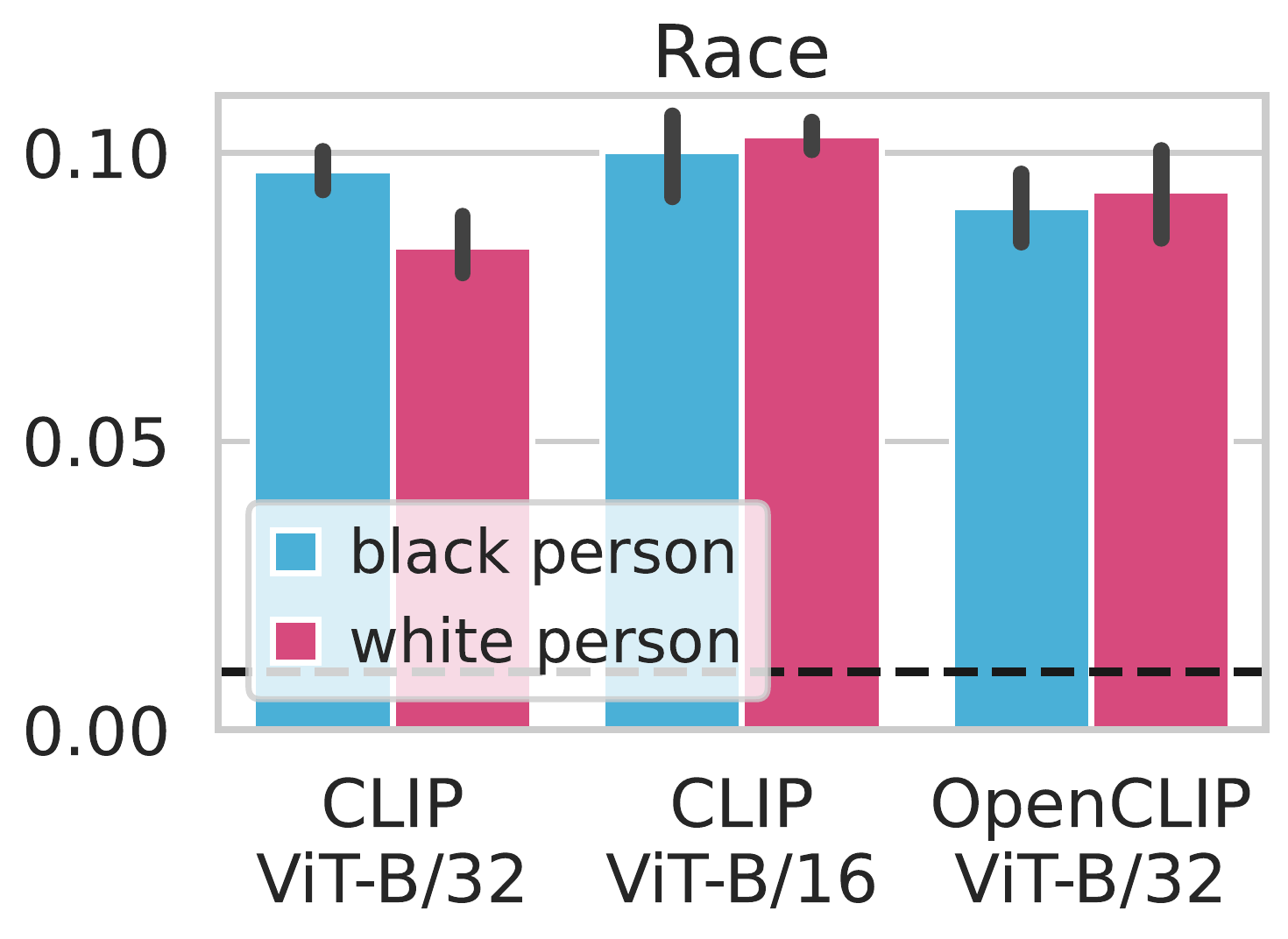}
     \end{subfigure}
      \begin{minipage}[b]{0.03\textwidth}
        \centering
        \begin{turn}{90}
          \begin{minipage}{0.10\textheight}
            \centering
            Oxford Flowers
          \end{minipage}          
        \end{turn}
      \end{minipage}
     \hfill
     \begin{subfigure}[b]{0.22\textwidth}
         \centering
         \includegraphics[width=\textwidth]{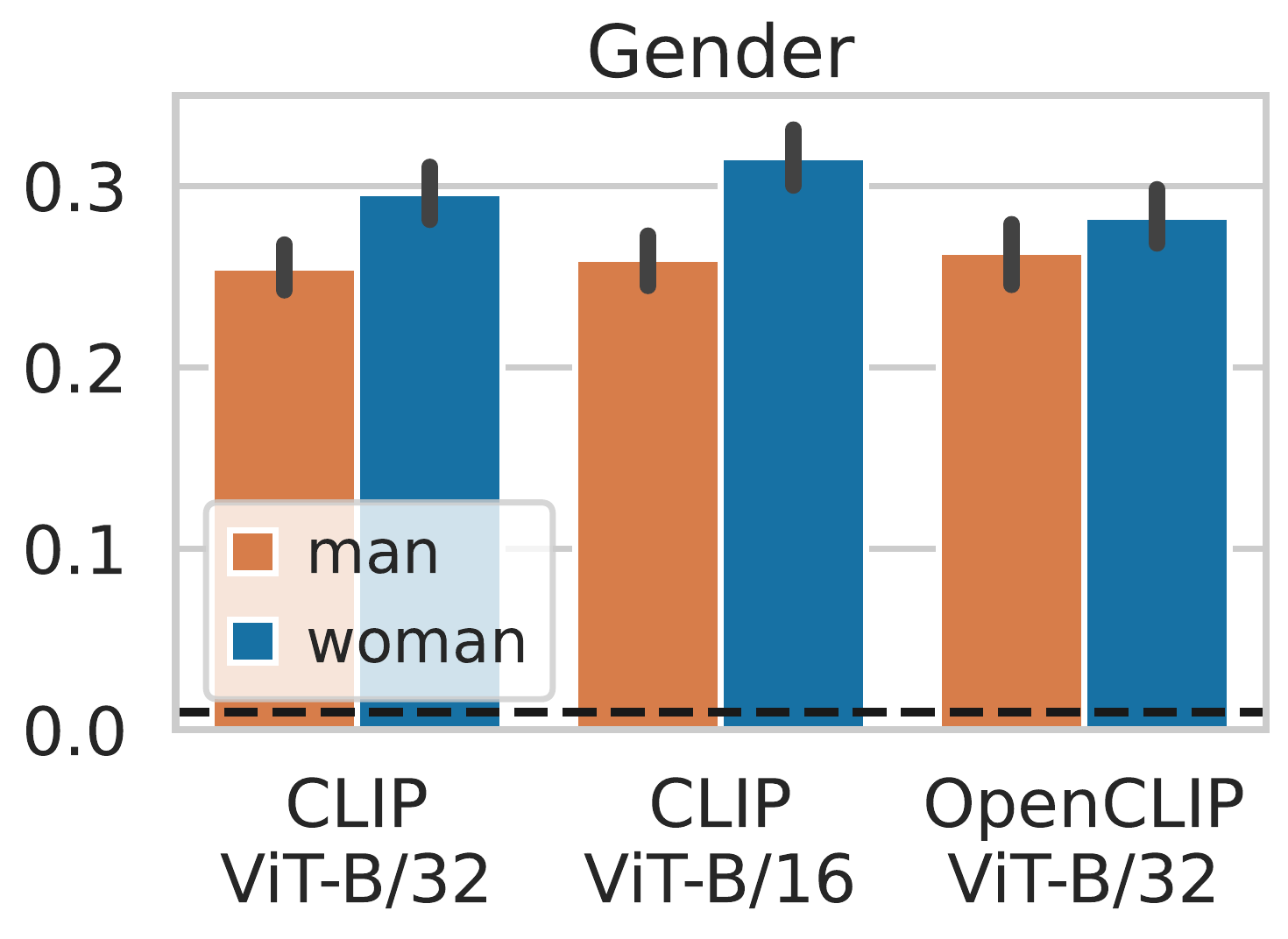}
     \end{subfigure}
      \hfill
     \begin{subfigure}[b]{0.22\textwidth}
         \centering
         \includegraphics[width=\textwidth]{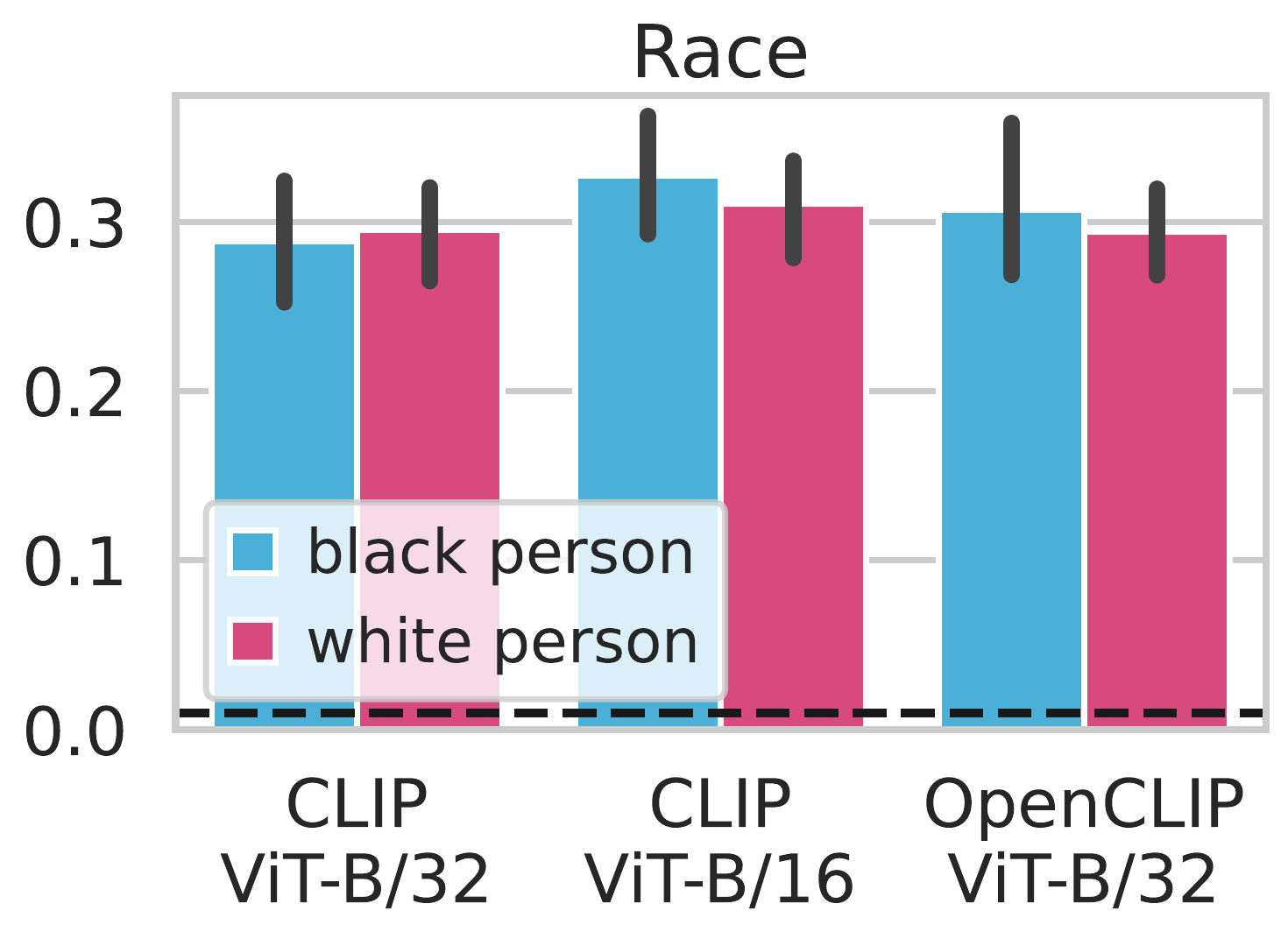}
     \end{subfigure}
    \caption{Evaluating bias of Vicuna-13B on more object categories (FGVC Aircraft~\cite{maji13fine-grained} and Oxford Flowers~\cite{Nilsback08flower}). The dashed line is the random baseline.}%
    \label{fig:vlm-additional-datasets}
\end{figure}

\label{subsec:additional_visual_datasets}
We extend our analysis to other datasets and more categories by using FGVC Aircraft~\cite{maji13fine-grained} (100 categories of aircraft from different manufacturers and eras) and Oxford Flowers~\cite{Nilsback08flower} (102 categories of flowers with large scale, pose and light variations).
For gender, we find significant performance differences when evaluating the descriptions generated by Vicuna-13B on the two additional datasets, strengthening our original argument that these LLMs exhibit biases (\Cref{fig:vlm-additional-datasets}). This means that descriptions generated by female personas outperform those generated by male personas across all three tested VLMs. For racial biases, we see only smaller differences across evaluation with different VLMs. The same trends hold for ChatGPT\@.

\subsection{Example descriptions for the visual classification task}%
\label{subsec:example-descriptions}
Recall that for the visual classification task we ask the impersonating language models to generate a description for each class of the dataset.
In the main paper we showed descriptions for two examples of a subset of all personas considered (4-, 7- and 13-year-old for CUB~\cite{Wah2011TheCB} and 2-, 4- and 20-year-old for Stanford Cars~\cite{Krause20133DOR}).
In this section we show and discuss example descriptions on both vision datasets (CUB and Stanford Cars) generated to for all age ranges included in our paper (2-, 4-, 7-, 13- and 20-year-old).
\begin{figure}[h!]
    \centering
    \includegraphics[width=\linewidth]{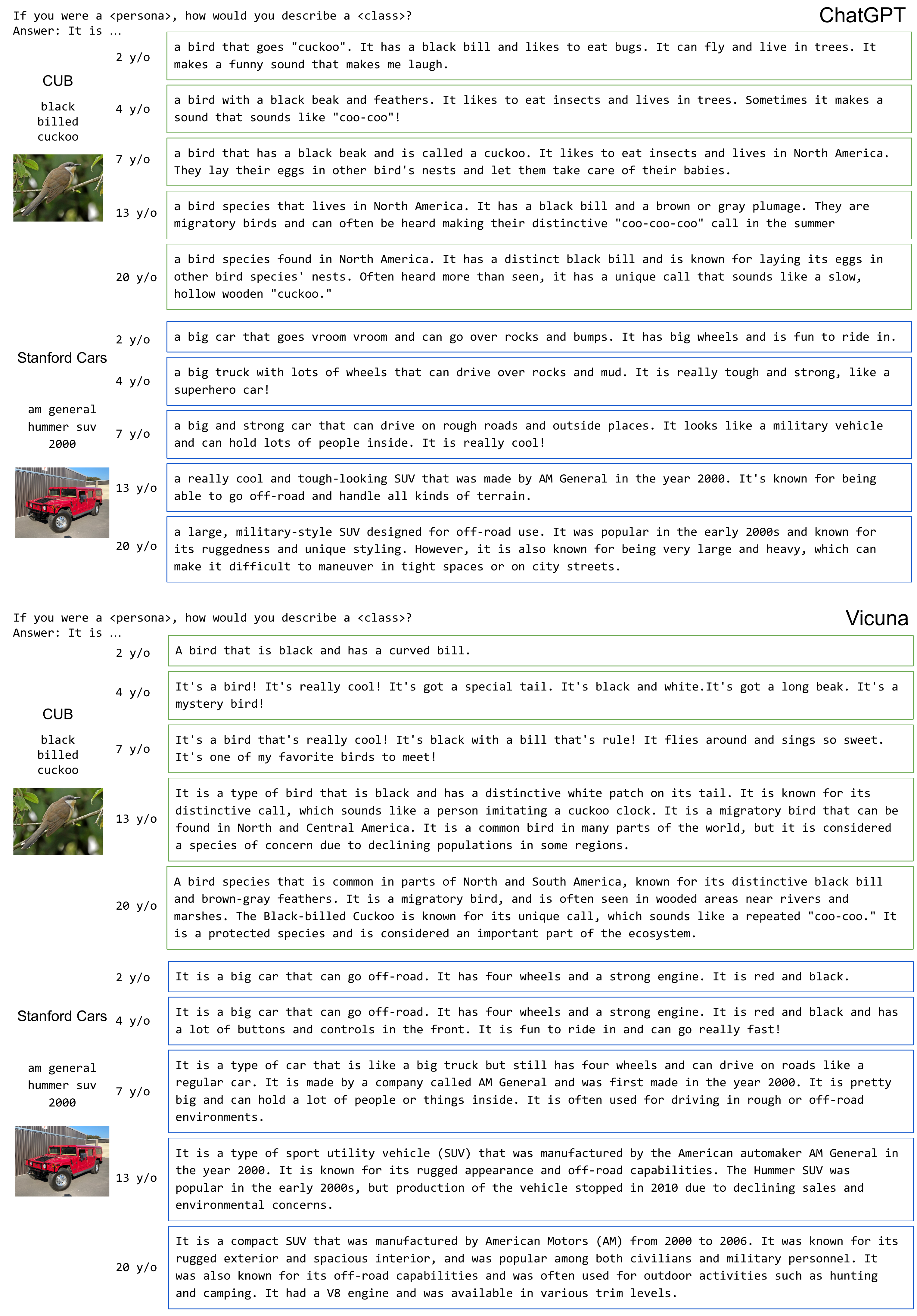}
    \caption{Qualitative results for all the age personas (\persona{2}, \persona{4}, \persona{7}, \persona{13} and \persona{20-year-old} personas) for two classes, i.e.\ Black Billed Cuckoo (CUB) and AM General Hummer SUV 2000 (Stanford Cars) classes. The results are obtained by querying ChatGPT and Vicuna.}%
    \label{fig:qualitative-full}
    \vspace{1em}
\end{figure}

The examples in \Cref{fig:qualitative-full} support our findings from the main paper, that with increasing age of the persona the complexity w.r.t.\ e.g.\ vocabulary increases.
For CUB we additionally show examples for the 2- and 20-year-old's and the differences in the wording are very apparent.
For both language models the descriptions generated for the 2-year-old are short and have simple grammatical structures.
In contrast, for the 20-year-old the descriptions exhaust much more of the 45 word instruction and use words that are not part of the vocabulary of a 2-year-old (e.g.\ \texttt{migratory bird} or \texttt{protected species}).\\
For Stanford Cars we additionally include the descriptions generated for the 7- and 13-year-old personas.
In contrast to the 4-year-old both descriptions are much longer, including many facts about the cars (e.g.\ the manufacturer of the car).

\subsubsection{Analysis of text complexity for different age groups}
In Figure 6 of the main paper as well as in \Cref{fig:qualitative-full} we qualitatively described how the text changes as we vary the age of the impersonated person.
To understand how the generated descriptions quantitatively change we also evaluate the complexity of the generated descriptions.

\begin{figure}[h!]
    \centering
     \begin{subfigure}[b]{0.49\textwidth}
         \centering
     \begin{subfigure}[b]{0.49\textwidth}
         \centering
         \includegraphics[width=\textwidth]{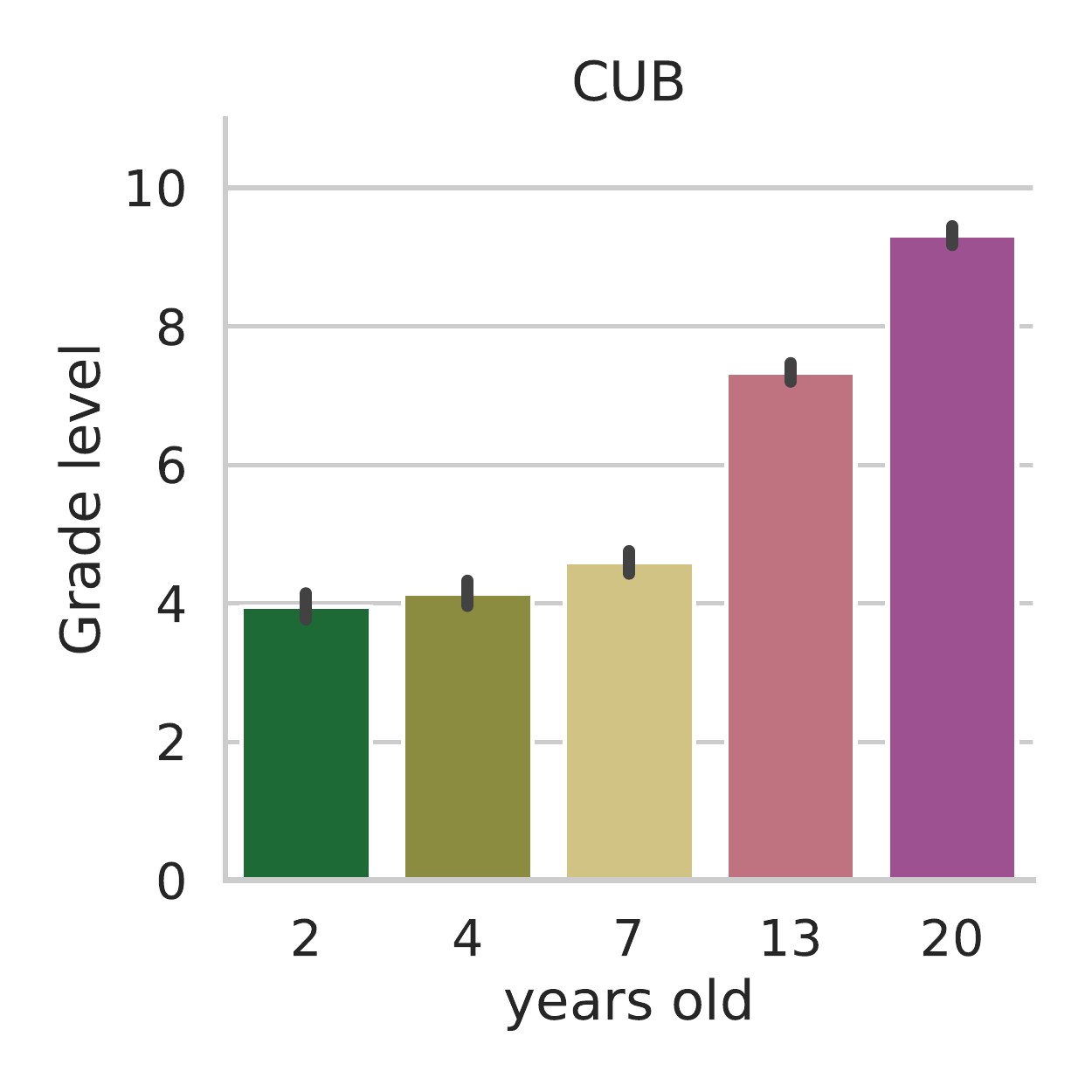}
     \end{subfigure}
     \hfill
     \begin{subfigure}[b]{0.49\textwidth}
         \centering
         \includegraphics[width=\textwidth]{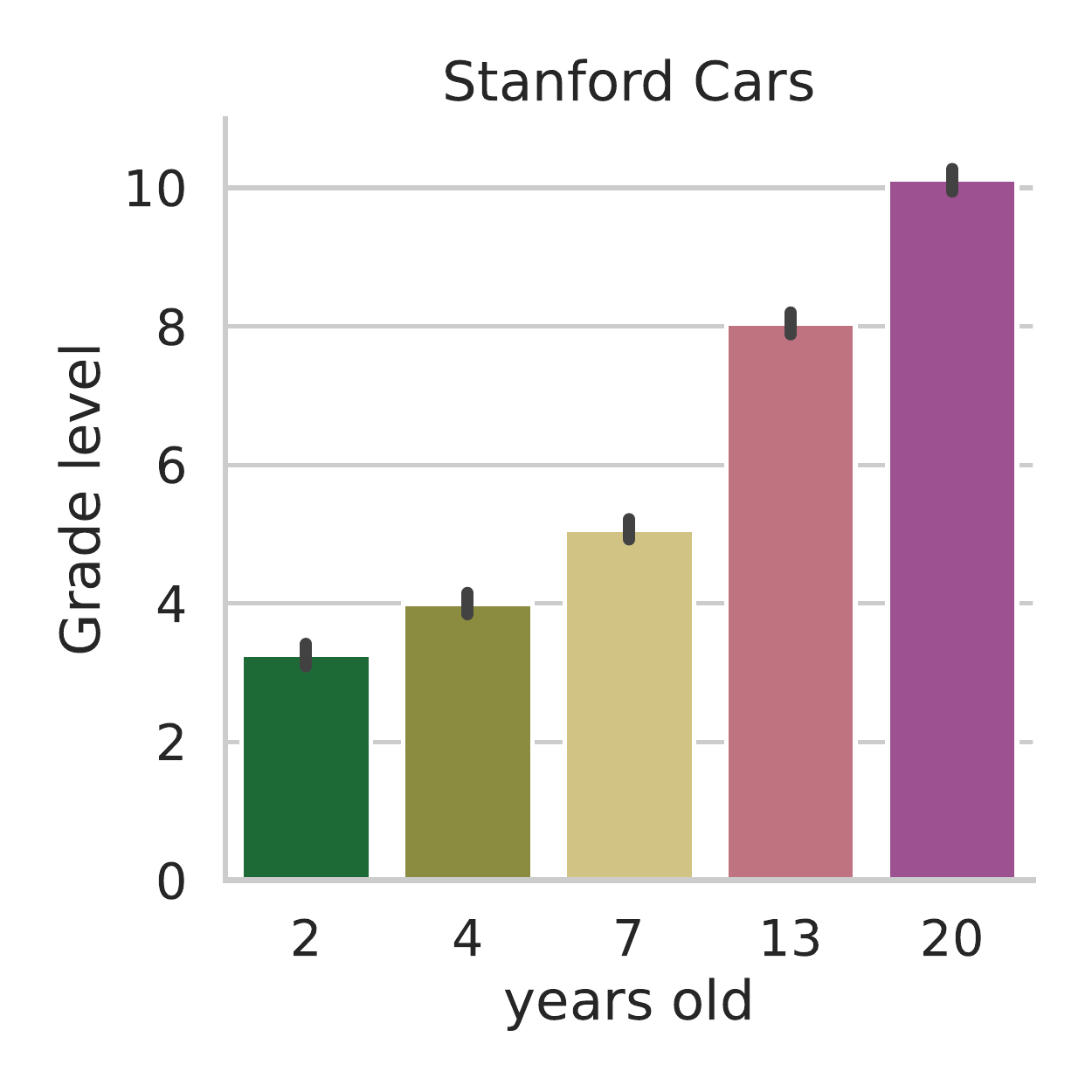}
     \end{subfigure}
     \caption{Vicuna-13B}
      \end{subfigure}
     \hfill
  \begin{subfigure}[b]{0.49\textwidth}
         \centering
     \begin{subfigure}[b]{0.49\textwidth}
         \centering
         \includegraphics[width=\textwidth]{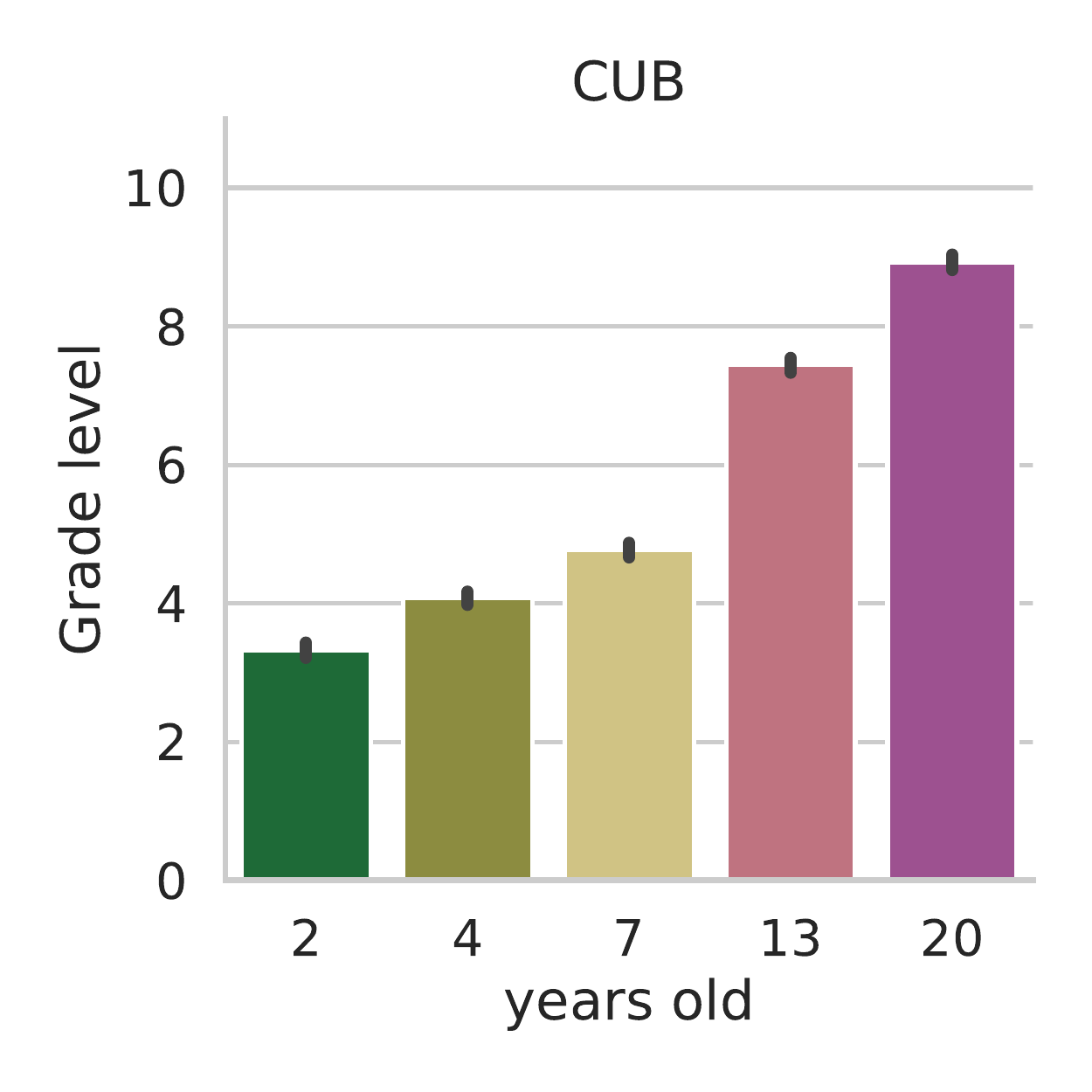}
     \end{subfigure}
     \hfill
     \begin{subfigure}[b]{0.49\textwidth}
         \centering
         \includegraphics[width=\textwidth]{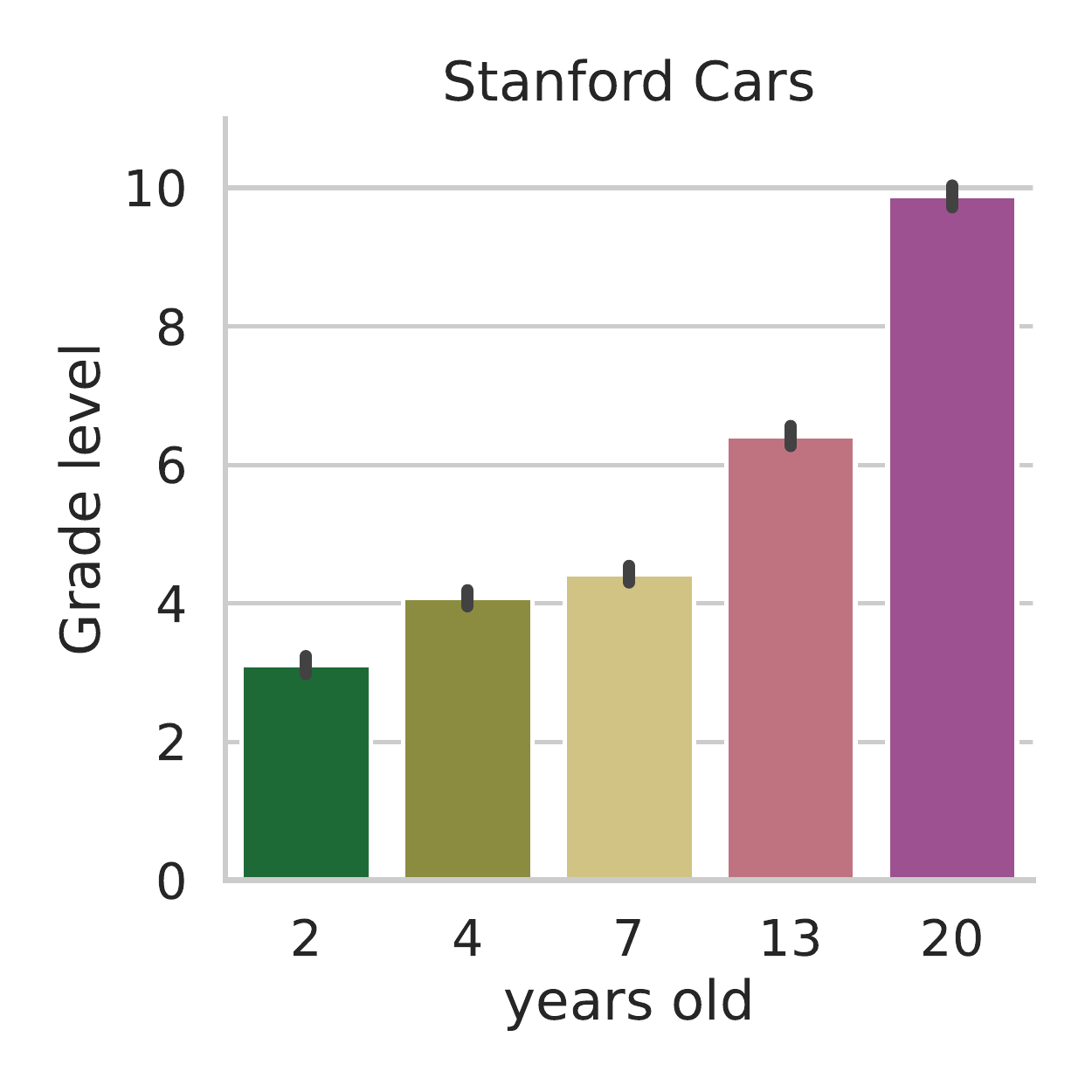}
     \end{subfigure}
          \caption{ChatGPT}
      \end{subfigure}
    \caption{Text complexity on Vicuna-13B (left) and ChatGPT (right) for CUB and Stanford Cars.}%
    \label{fig:descriptions-complexity}
\end{figure}

We use the \texttt{textstat} package\footnote{\url{https://github.com/textstat/textstat}}, which runs several different text complexity metrics~\cite{Kincaid1975DerivationON,Mclaughlin1969SMOGG,Coleman1975ACR,Dale1948AFF,Chall1995ReadabilityR,Klare1974Assessing,Gunning1968TheTO} and creates an aggregate consensus score that indicates which grade in school is at least required to read the texts.
In \Cref{fig:descriptions-complexity} we show the results for both, CUB and Stanford Cars.

We find, that across all language models and both datasets the impersonation of differently aged personas increases the required grade level to read the descriptions.
For CUB the grade level increases not as much (from 4th to 9th grade) than on Stanford Cars (from 3rd to approx.\ 10th grade). This might be due to the fact that more descriptions of the oldest personas mention complex terms like manufacturers for the Stanford Cars dataset.

\subsection{Quantitative results on LLM / VLM pairs}\label{subsec:quantitative-results-pairs}
\begin{figure}[t!]
     \centering
      \begin{minipage}[b]{0.03\textwidth}
        \centering
        \begin{turn}{90}
          \begin{minipage}{0.12\textheight}
            \centering
            CUB
          \end{minipage}          
        \end{turn}
      \end{minipage}
     \hfill
     \begin{subfigure}[b]{0.236\textwidth}
         \centering
         \includegraphics[width=\textwidth]{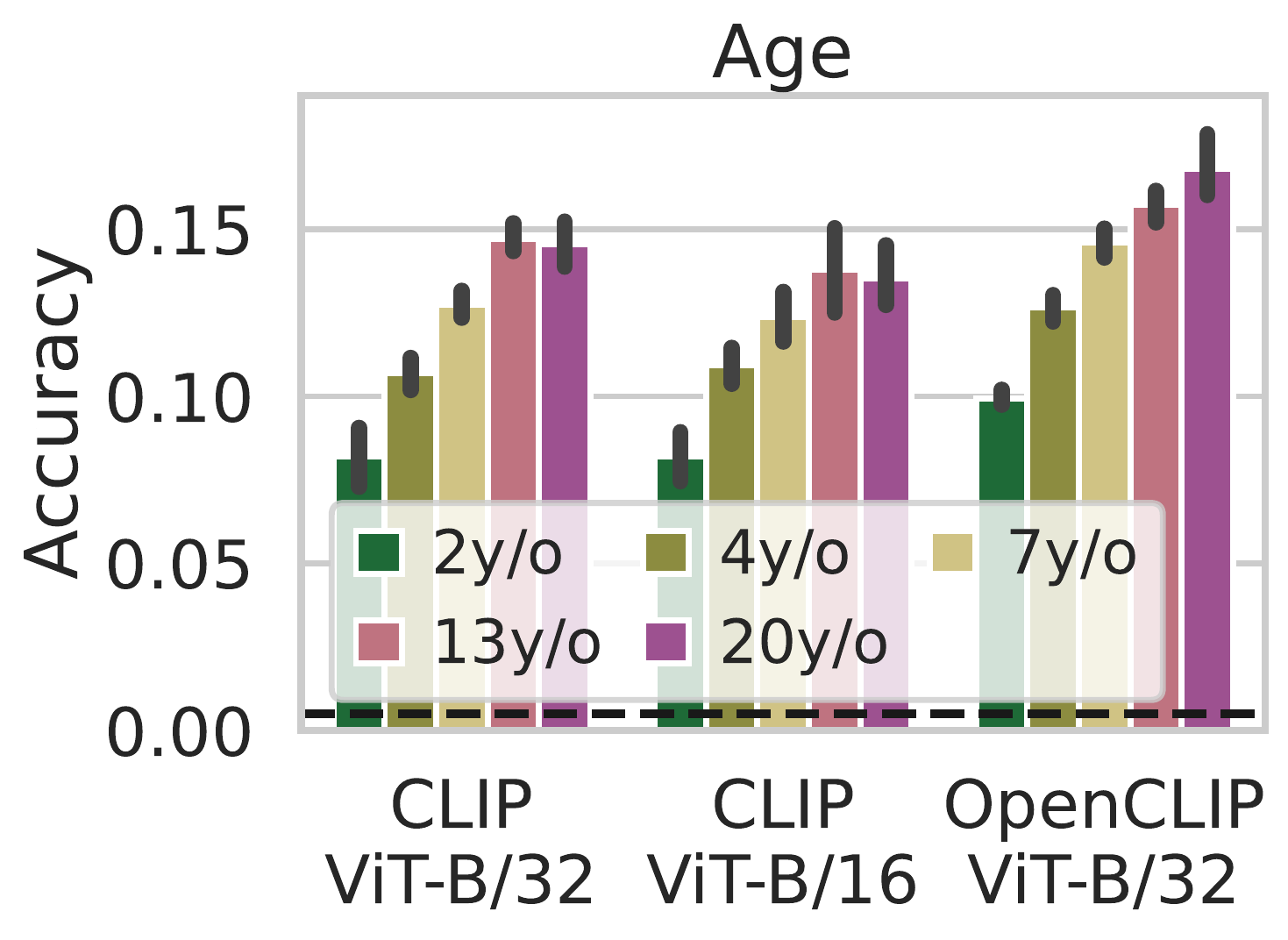}
     \end{subfigure}
     \hfill
     \begin{subfigure}[b]{0.236\textwidth}
         \centering
         \includegraphics[width=\textwidth]{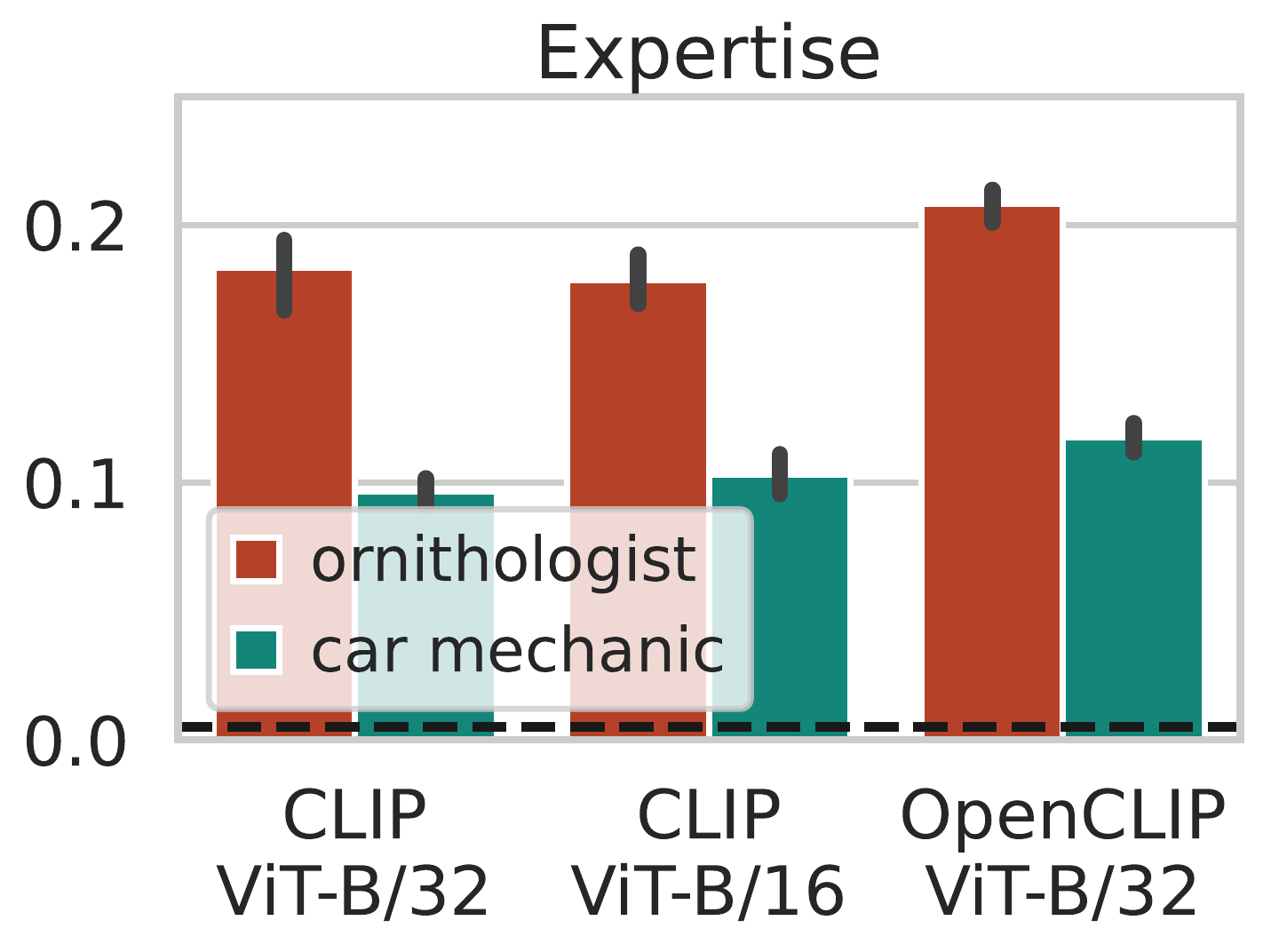}
     \end{subfigure}
     \hfill
     \begin{subfigure}[b]{0.236\textwidth}
         \centering
         \includegraphics[width=\textwidth]{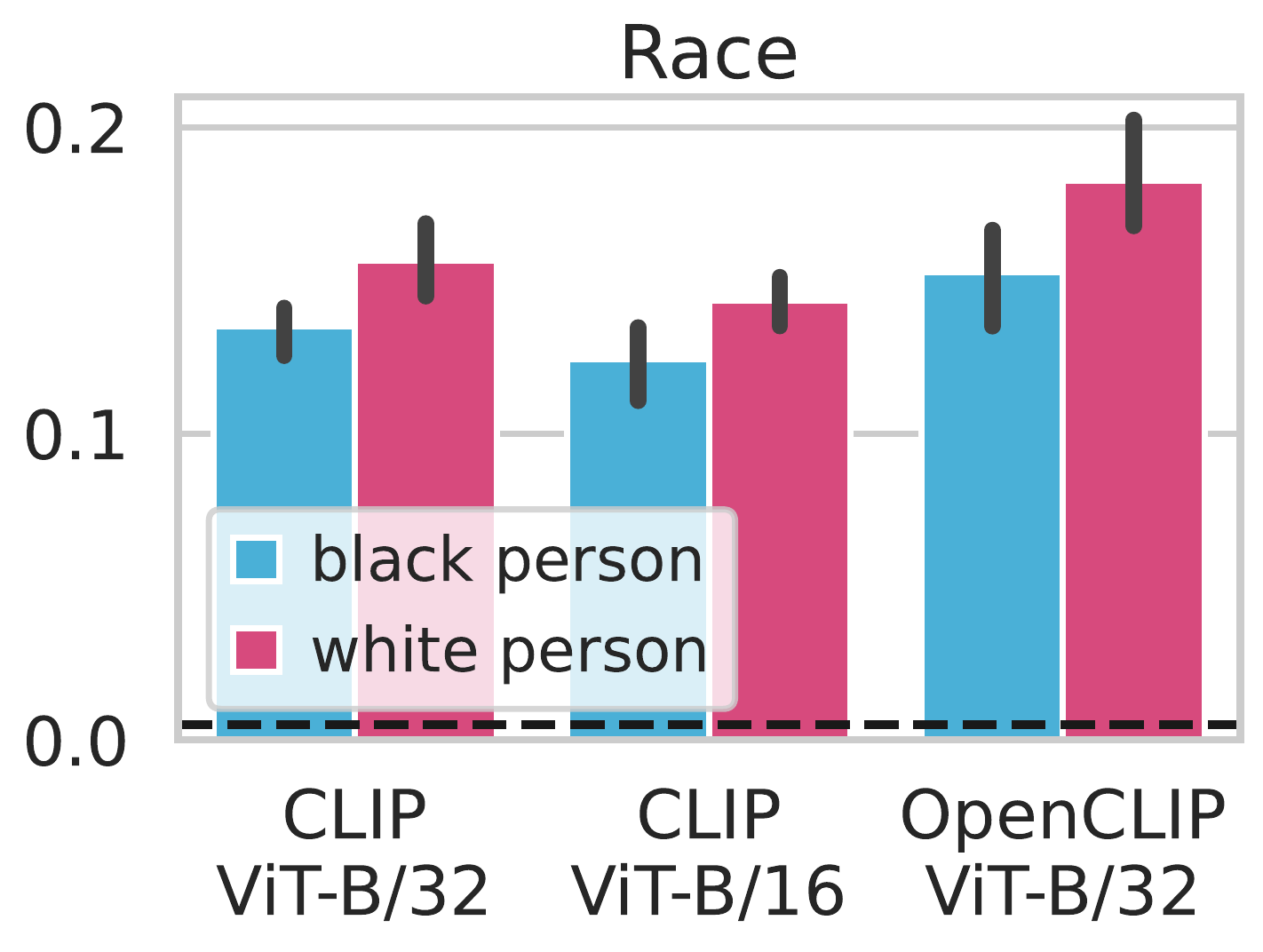}
     \end{subfigure}
      \hfill
     \begin{subfigure}[b]{0.236\textwidth}
         \centering
         \includegraphics[width=\textwidth]{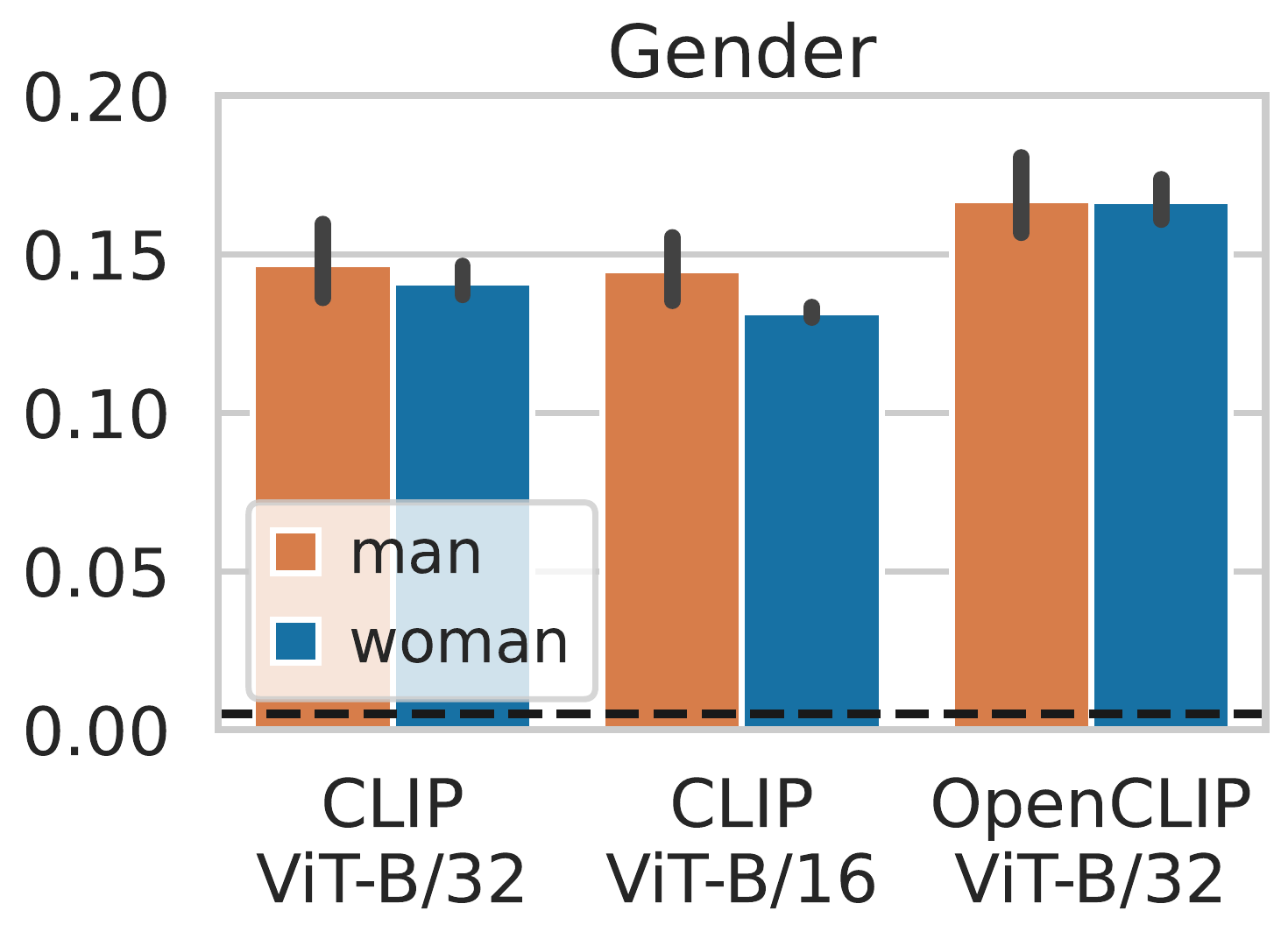}
     \end{subfigure}
      \begin{minipage}[b]{0.03\textwidth}
        \centering
        \begin{turn}{90}
          \begin{minipage}{0.12\textheight}
            \centering
            Stanford Cars
          \end{minipage}          
        \end{turn}
      \end{minipage}
     \hfill
     \begin{subfigure}[b]{0.236\textwidth}
         \centering
         \includegraphics[width=\textwidth]{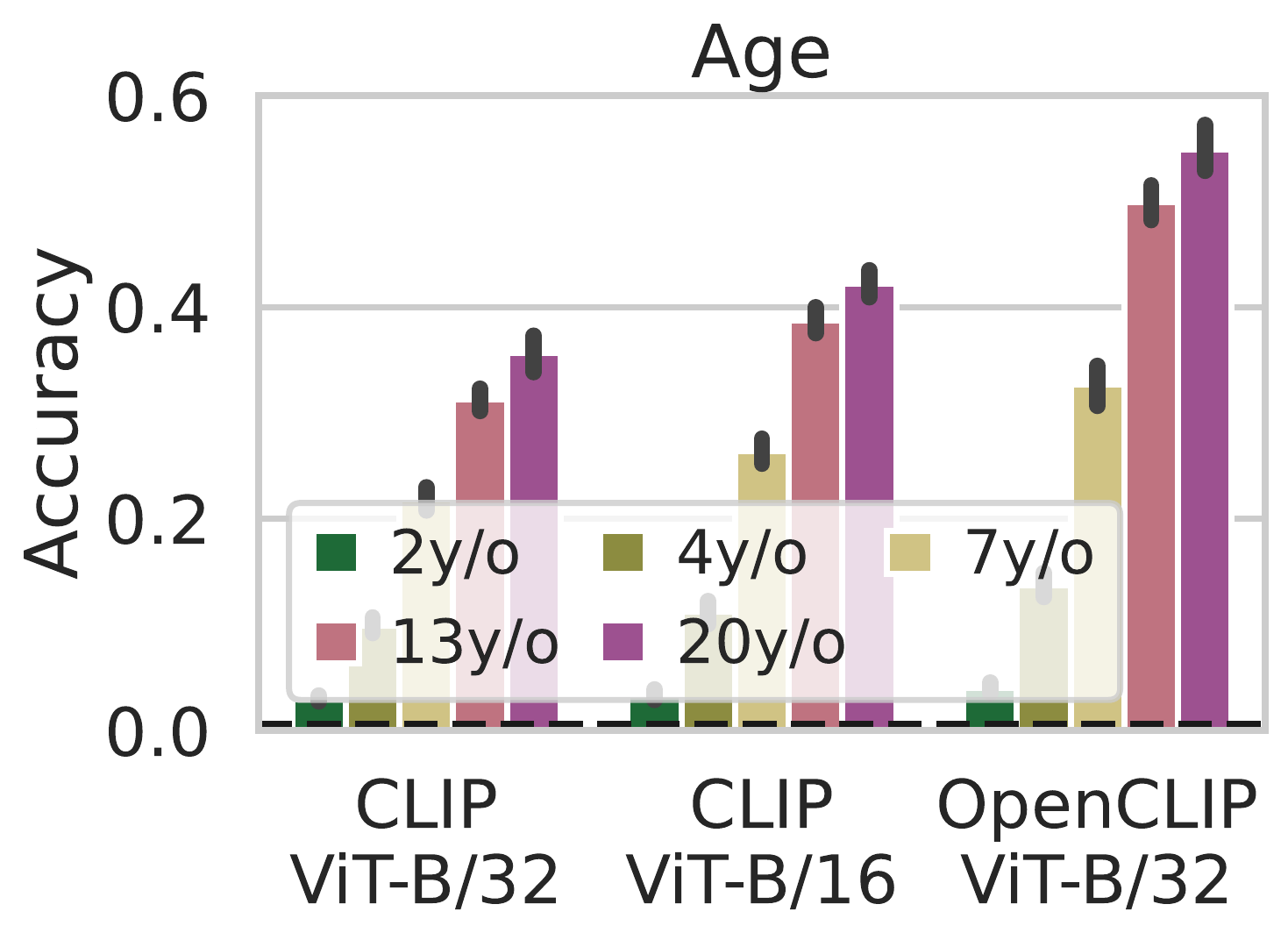}
     \end{subfigure}
     \hfill
     \begin{subfigure}[b]{0.236\textwidth}
         \centering
         \includegraphics[width=\textwidth]{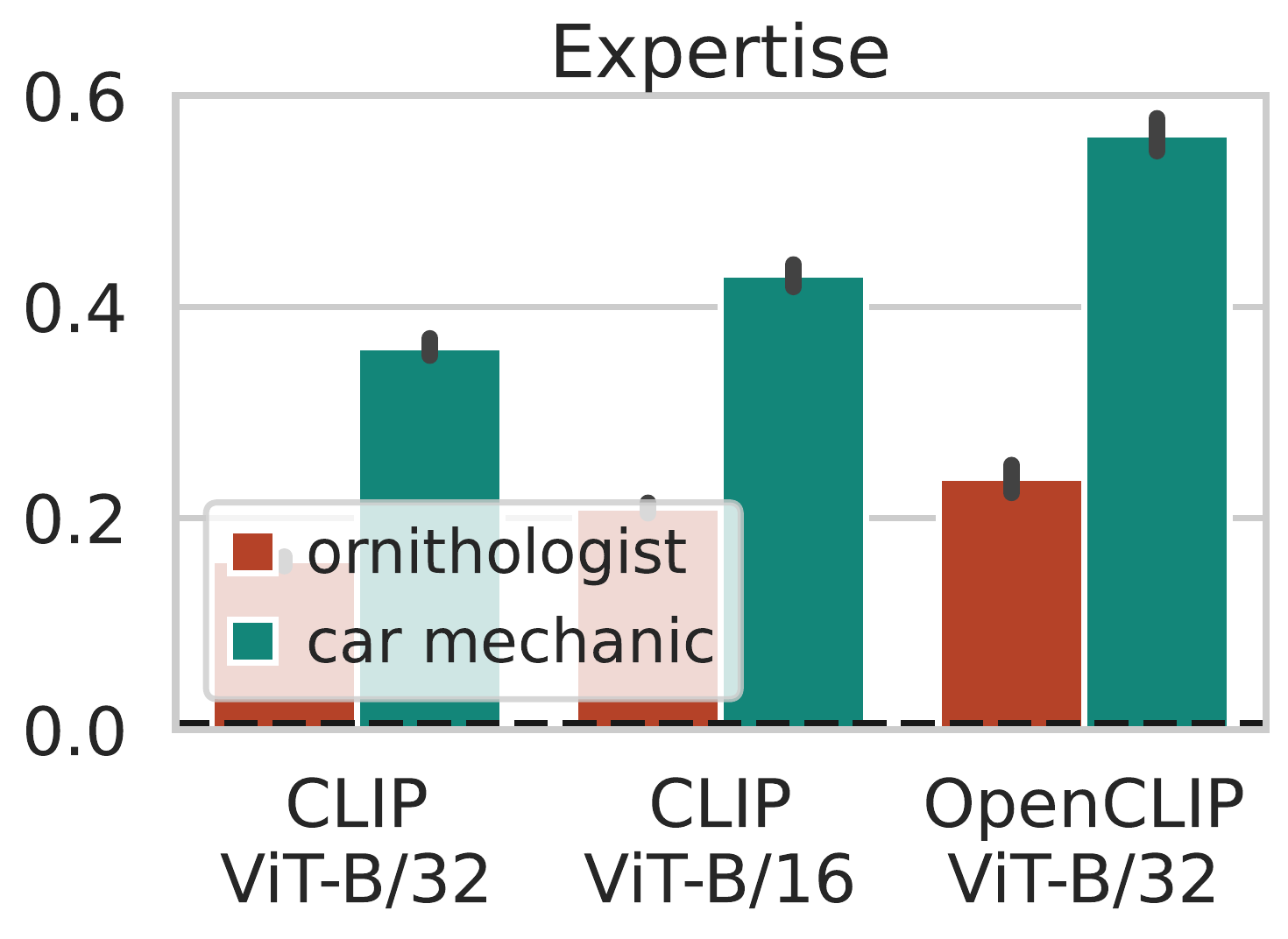}
     \end{subfigure}
     \hfill
     \begin{subfigure}[b]{0.236\textwidth}
         \centering
         \includegraphics[width=\textwidth]{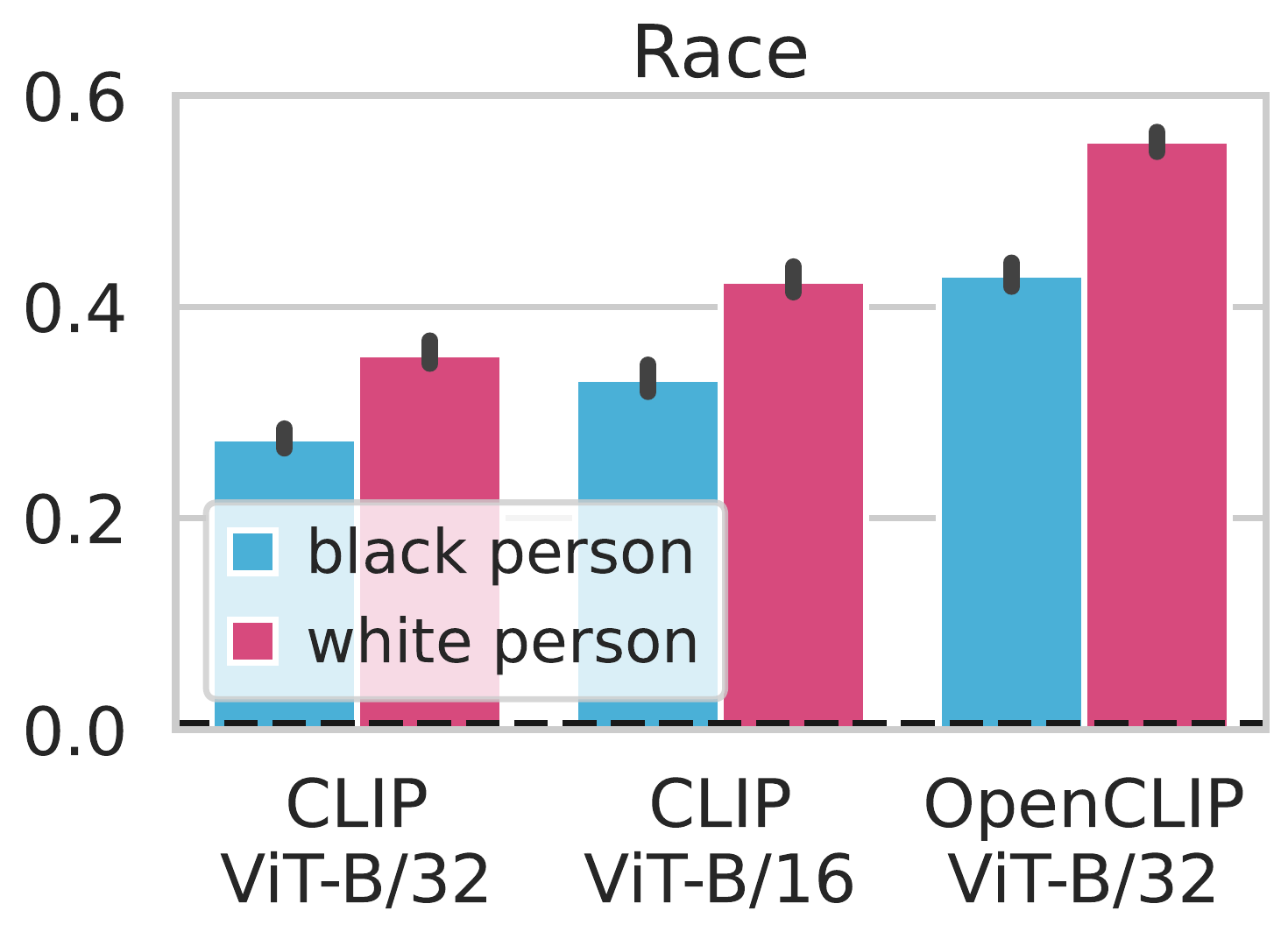}
     \end{subfigure}
      \hfill
     \begin{subfigure}[b]{0.236\textwidth}
         \centering
         \includegraphics[width=\textwidth]{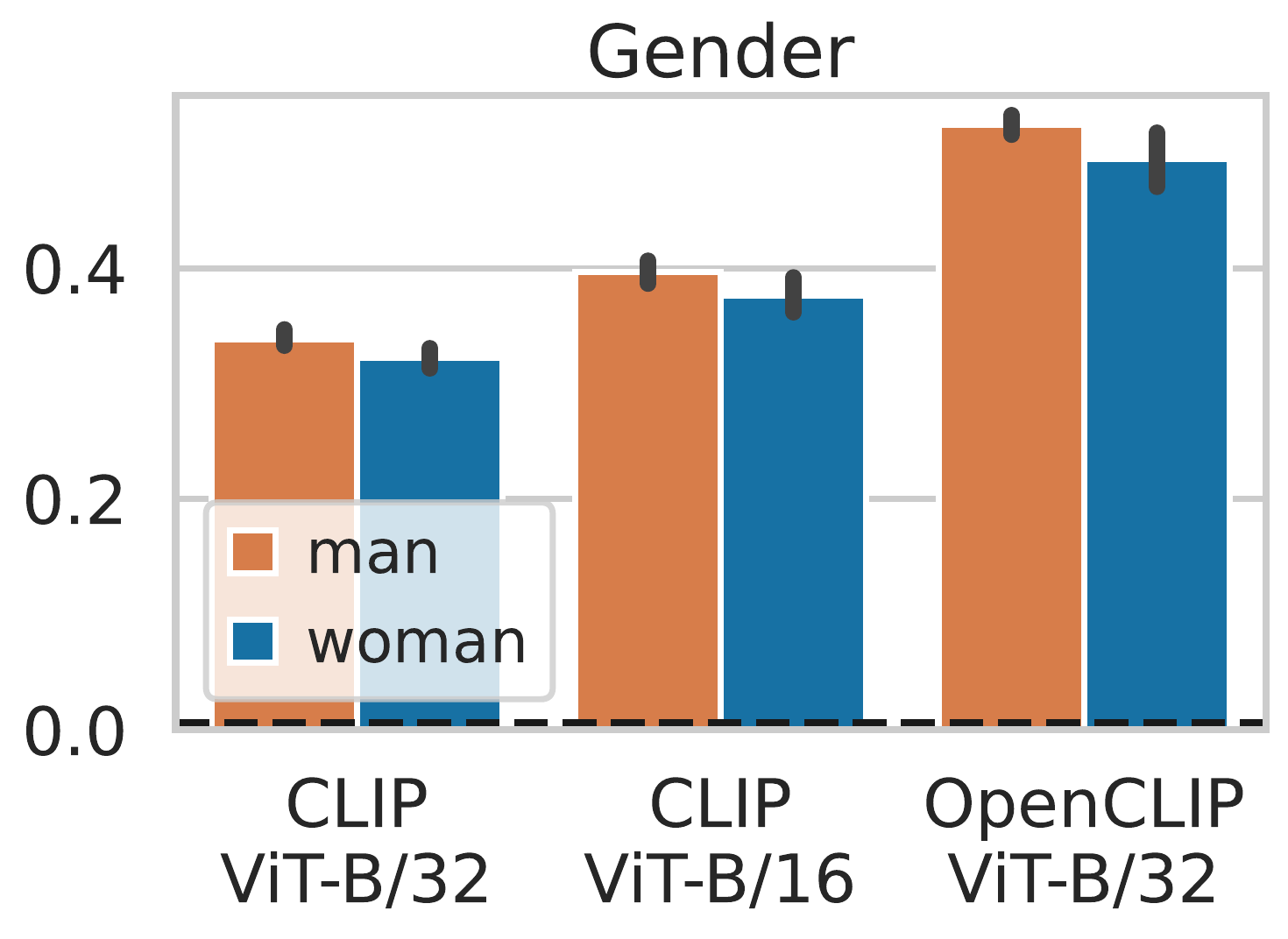}
     \end{subfigure}
    \caption{Comparing CLIP-32, CLIP-16 and OpenCLIP as VLMs. In contrast to Figure 4 in the paper (which shows Vicuna-13B results), the language input here comes from ChatGPT\@. We observe the effects of age, expertise, race and gender independent of the VLM used for fine-grained visual classification on the CUB (top) and Stanford Cars (bottom) datasets. The dashed line is the random baseline.}%
    \label{fig:vlm-comparison-chatgpt}
\end{figure}

In Section 4.3, Figure 4 of the main paper, we show results for the three different CLIP variants (CLIP with ViT B/32, ViT B/16 and OpenCLIP~\cite{Cherti2022ReproducibleSL}) based on the descriptions generated by the Vicuna-13B LLM\@.
Here we additionally show these results for descriptions generated with ChatGPT in \Cref{fig:vlm-comparison-chatgpt}.

Similar to the findings on Vicuna-13B the descriptions generated by ChatGPT exhibit an increase in fine-grained visual classification performance as the age of the impersonated person increases.
For ChatGPT this effect is more clear on Stanford Cars than on CUB\@.
Additionally, these results confirm our finding that expert impersonations perform better than non-experts. However, for ChatGPT the effect is even more clear; the expert performs roughly twice as well as the non-expert across all VLMs.
Regarding race, ChatGPT descriptions' seem to have more bias than those generated by Vicuna.
Lastly, for different genders, we find ChatGPTs' descriptions of female impersonation to perform consistently worse than those of male impersonation.

\subsection{Additional bias groups}%
\begin{figure}[h!]
    \centering
        \begin{minipage}[b]{0.03\textwidth}
            \centering
            \begin{turn}{90}
              \begin{minipage}{0.12\textheight}
                \centering
                CUB
              \end{minipage}          
            \end{turn}
          \end{minipage}
         \hfill
         \begin{subfigure}[b]{0.22\textwidth}
             \centering
             \includegraphics[width=\textwidth]{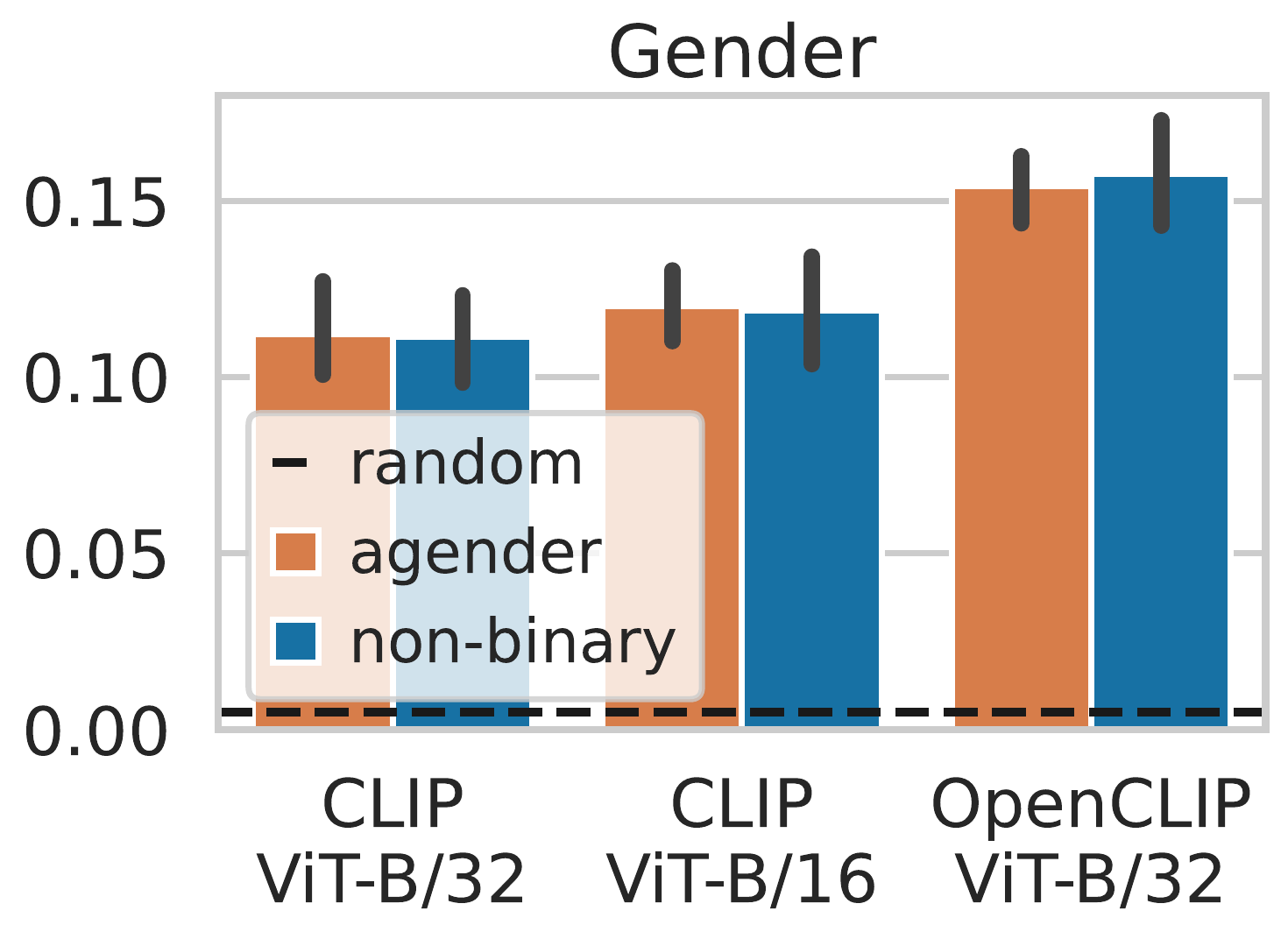}
         \end{subfigure}
          \hfill
         \begin{subfigure}[b]{0.22\textwidth}
             \centering
             \includegraphics[width=\textwidth]{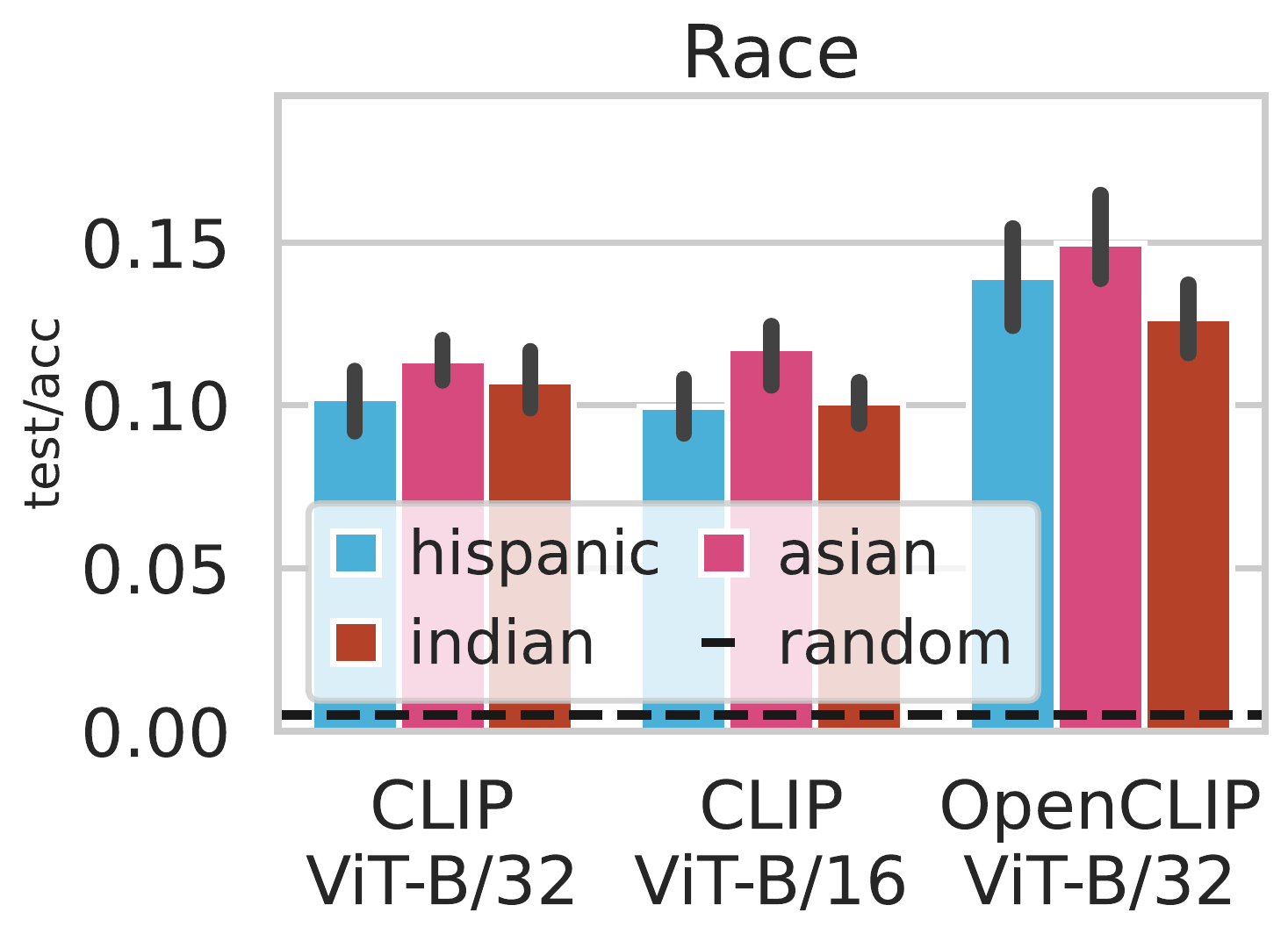}
         \end{subfigure}
         \hfill
          \begin{minipage}[b]{0.03\textwidth}
            \centering
            \begin{turn}{90}
              \begin{minipage}{0.09\textheight}
                \centering
                Stanford Cars
              \end{minipage}          
            \end{turn}
          \end{minipage}
         \hfill
         \begin{subfigure}[b]{0.22\textwidth}
             \centering
             \includegraphics[width=\textwidth]{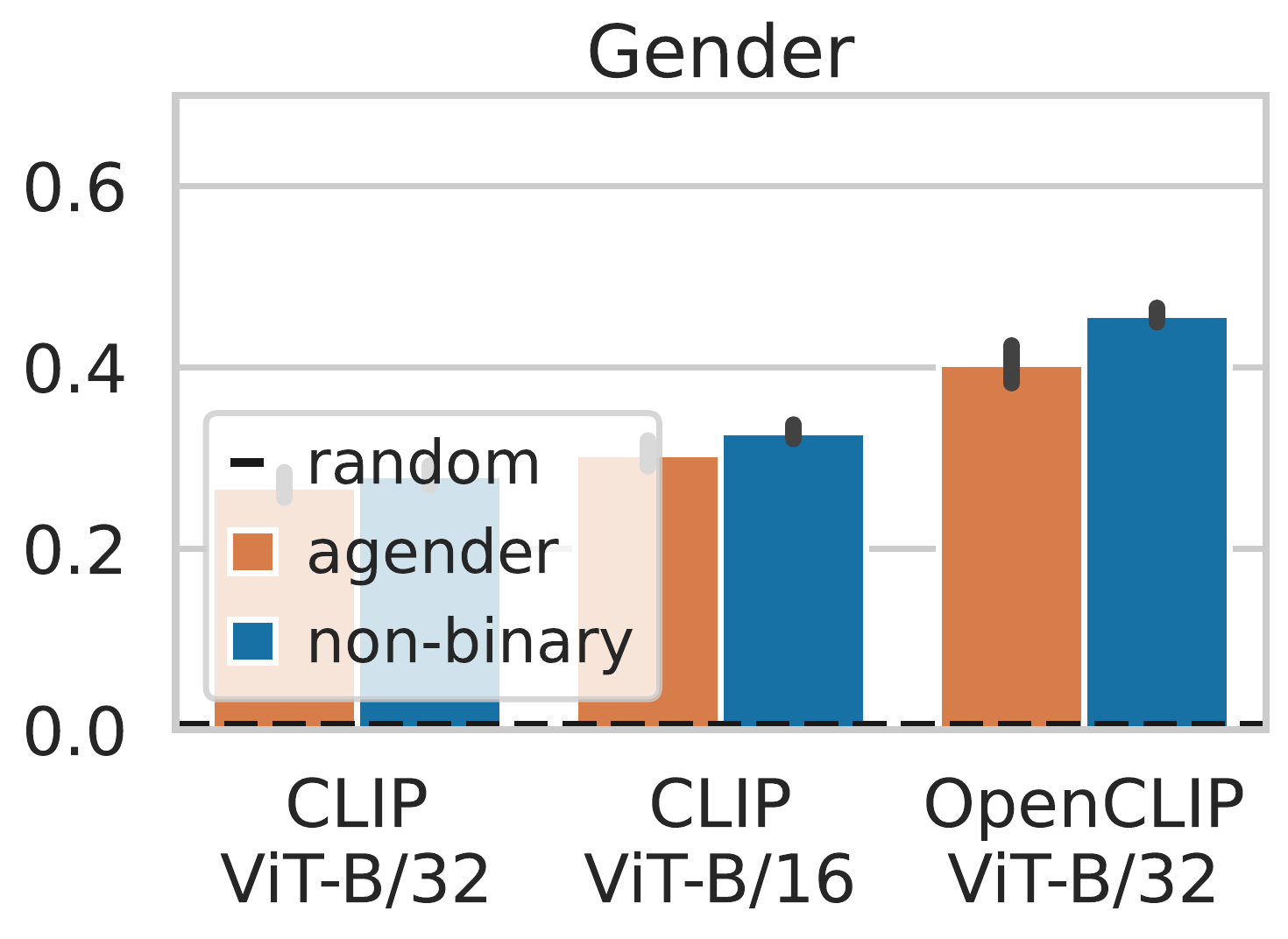}
         \end{subfigure}
          \hfill
         \begin{subfigure}[b]{0.22\textwidth}
             \centering
             \includegraphics[width=\textwidth]{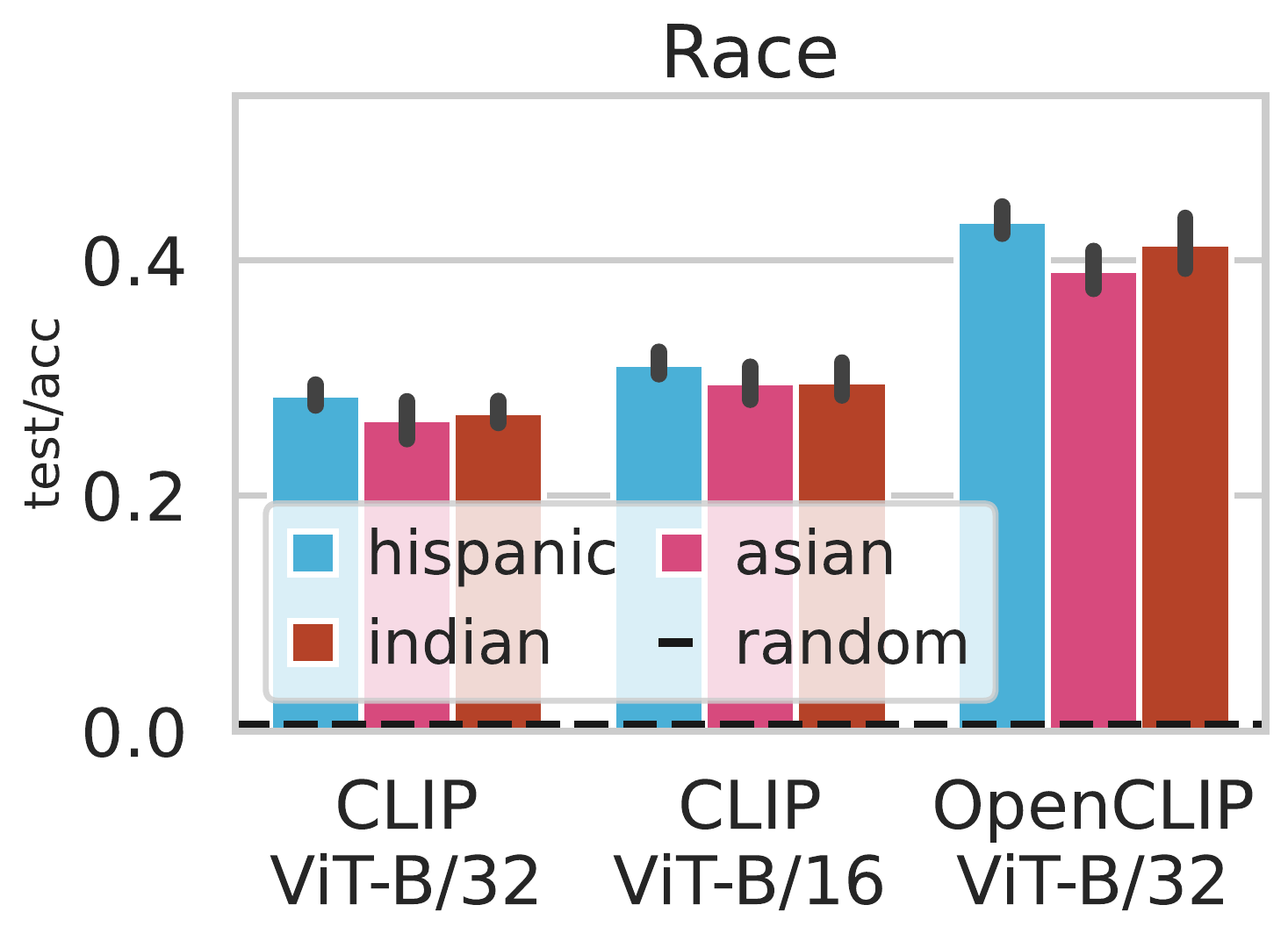}
         \end{subfigure}
        \caption{Evaluating more genders and races for Vicuna-13B on CUB and Stanford Cars.}%
        \label{fig:vlm-more-biases}
\end{figure}

\label{subsec:additional_bias_groups}
We also study additional bias groups for race and gender in \Cref{fig:vlm-more-biases}. For gender we study agender and non-binary and for race we study indian person, asian person and hispanic person. For CUB we find for gender that performance of the agender and non-binary personas is similar to the performance of the female persona. On Stanford Cars the non-binary persona outperforms the agender persona. For race we find overall worse performance compared to the white and black personas. Overall we conclude that biases can also be found in additional bias groups.

\subsection{Results for Google PaLM}%
\label{subsec:google_palm}
Additionally, we ran exploratory experiments using the proprietary Google PaLM model~\cite{chowdhery2022palm,Anil2023PaLM2T} via their API\@. When we apply in-context impersonation of racial biases, the API does not provide an answer because the LLM model output is flagged by a text classifier to be unsafe. Hence, there are already safeguards in place for some commercial services. These safeguards seem to be less sensitive for impersonation of age and gender. However, they prevent us from reliably evaluating the underlying LLM\@.

\section{Limitations}\label{sec:limitations}
Our vision based experiments are a two step process.
Thus, a limitation of our work is that the results on the vision datasets fundamentally depend on the performance and biases of the VLM models as well. We try to alleviate this fact by evaluating multiple different CLIP variants.
Additionally, the results obtained with proprietary models such as ChatGPT may be hard or costly to reproduce and the training regime and data as well as the systems prompts are unknown.

\end{document}

%% file: figures_supp/mmlu_tasks_ChatGPTVicuna_figure_CR.tex
\newcommand{\stemscale}{0.85}
\begin{figure}[h!]
     \centering
     \resizebox{\stemscale\textwidth}{!}{
     \begin{subfigure}[c]{0.35\textwidth}
         \centering
         \includegraphics[width=\textwidth]{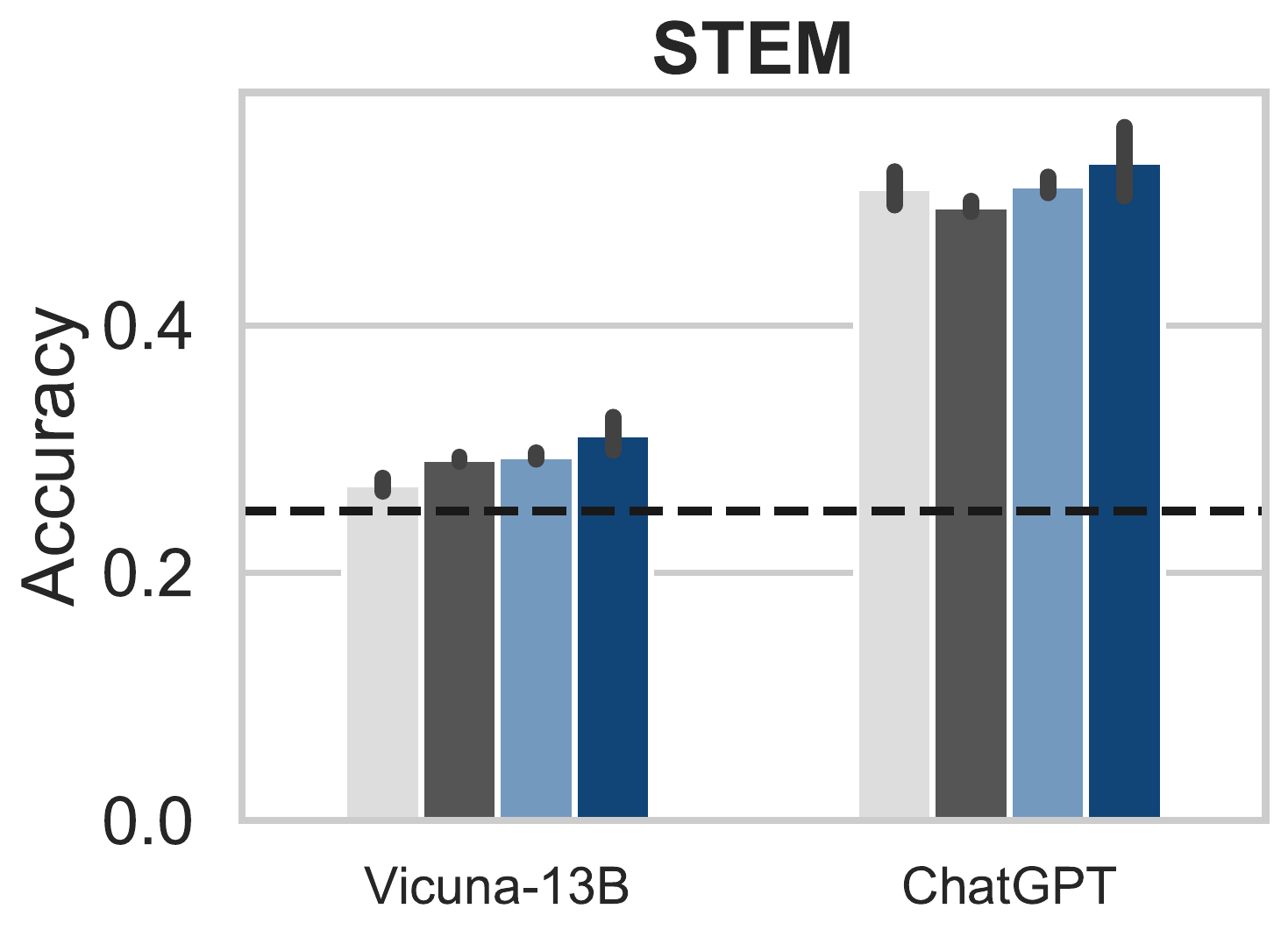}
     \end{subfigure}
     \hfill
     \begin{subfigure}[c]{0.31\textwidth}
         \centering
         \includegraphics[width=\textwidth]{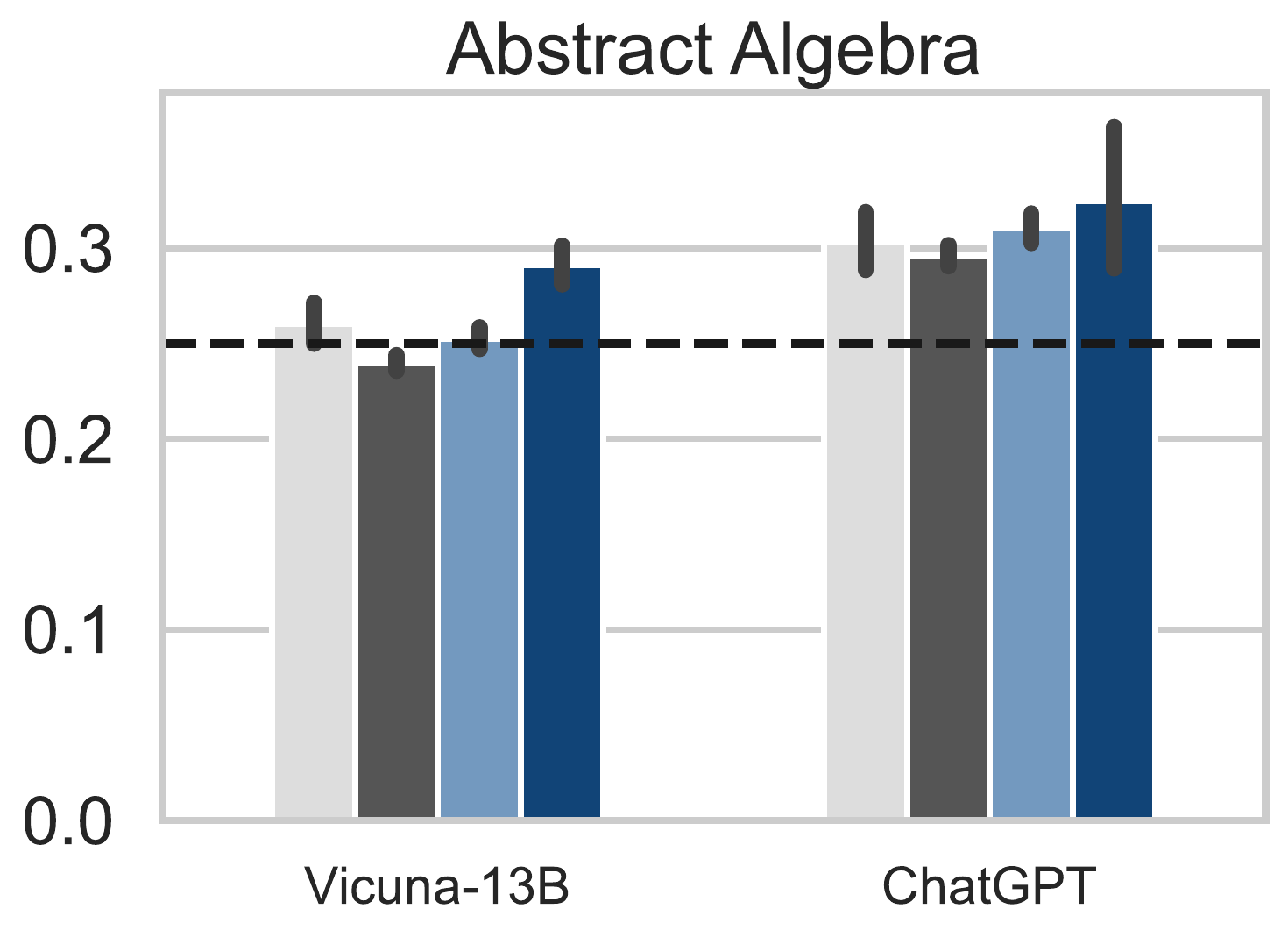}
     \end{subfigure}
     \hfill
     \begin{subfigure}[c]{0.31\textwidth}
         \centering
         \includegraphics[width=\textwidth]{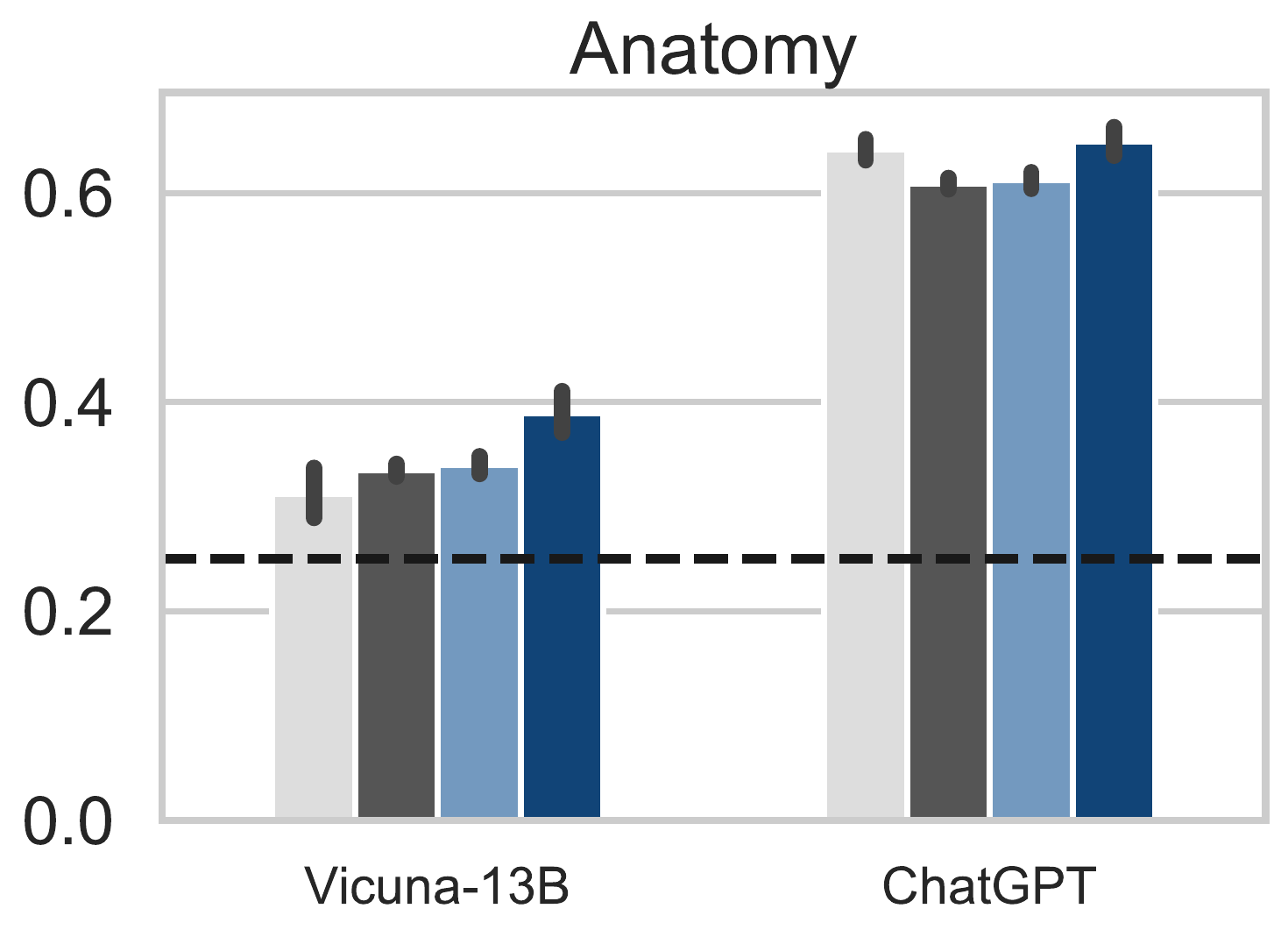}
     \end{subfigure}
     }\\
     \resizebox{\stemscale\textwidth}{!}{
     \begin{subfigure}[c]{0.31\textwidth}
         \centering
         \includegraphics[width=\textwidth]{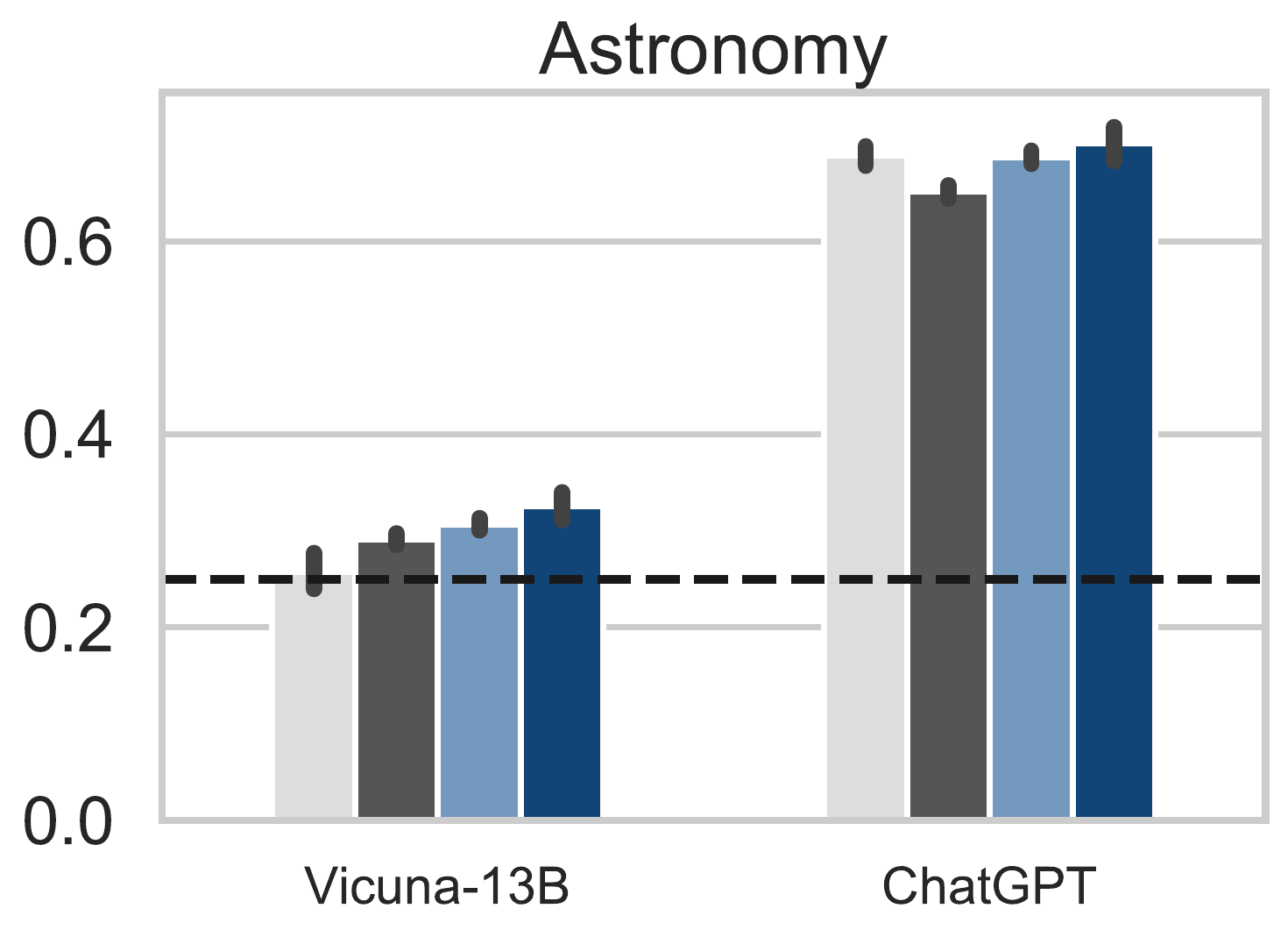}
     \end{subfigure}
     \hfill
     \begin{subfigure}[c]{0.31\textwidth}
         \centering
         \includegraphics[width=\textwidth]{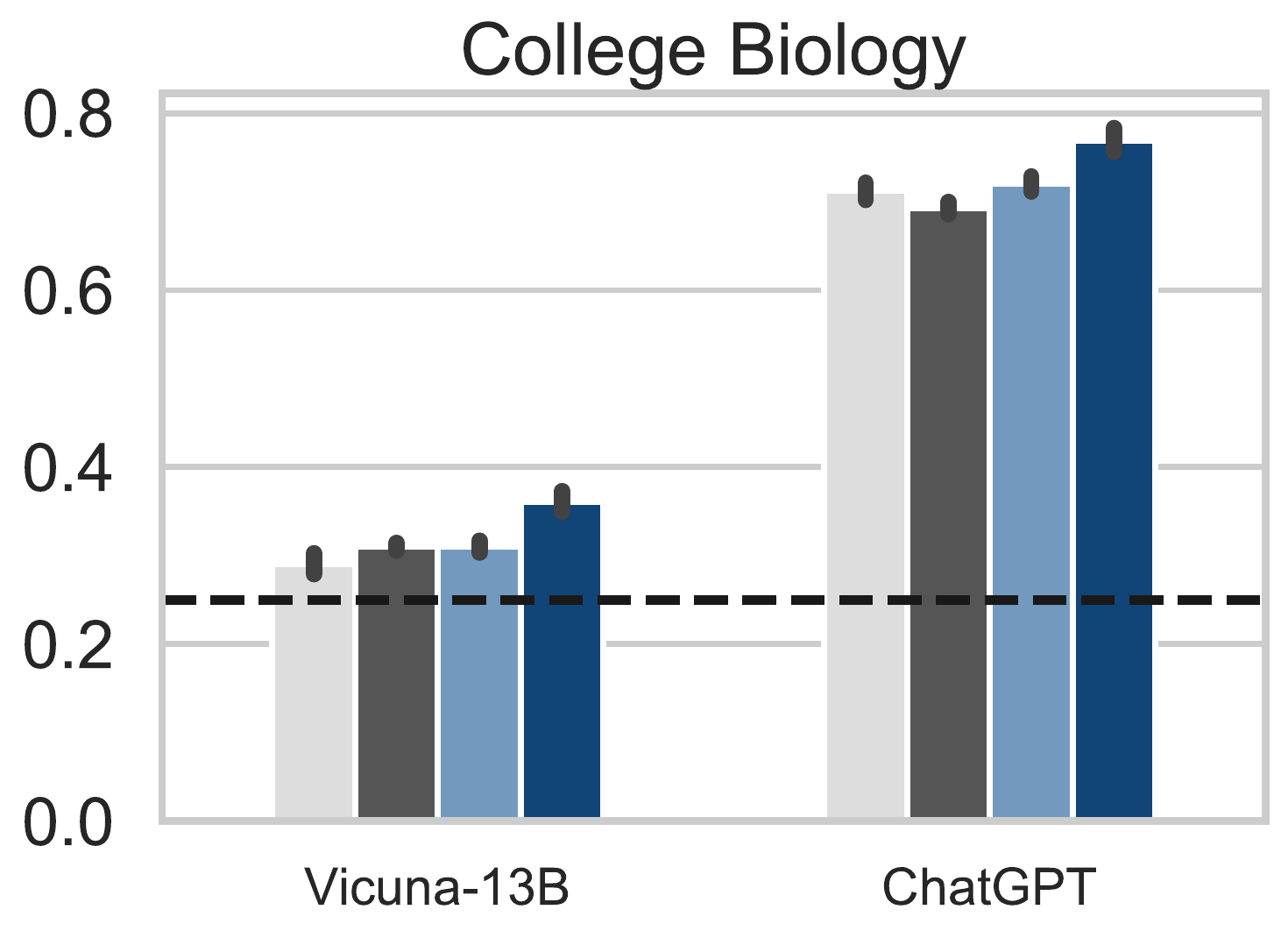}
     \end{subfigure}
     \hfill
     \begin{subfigure}[c]{0.31\textwidth}
         \centering
         \includegraphics[width=\textwidth]{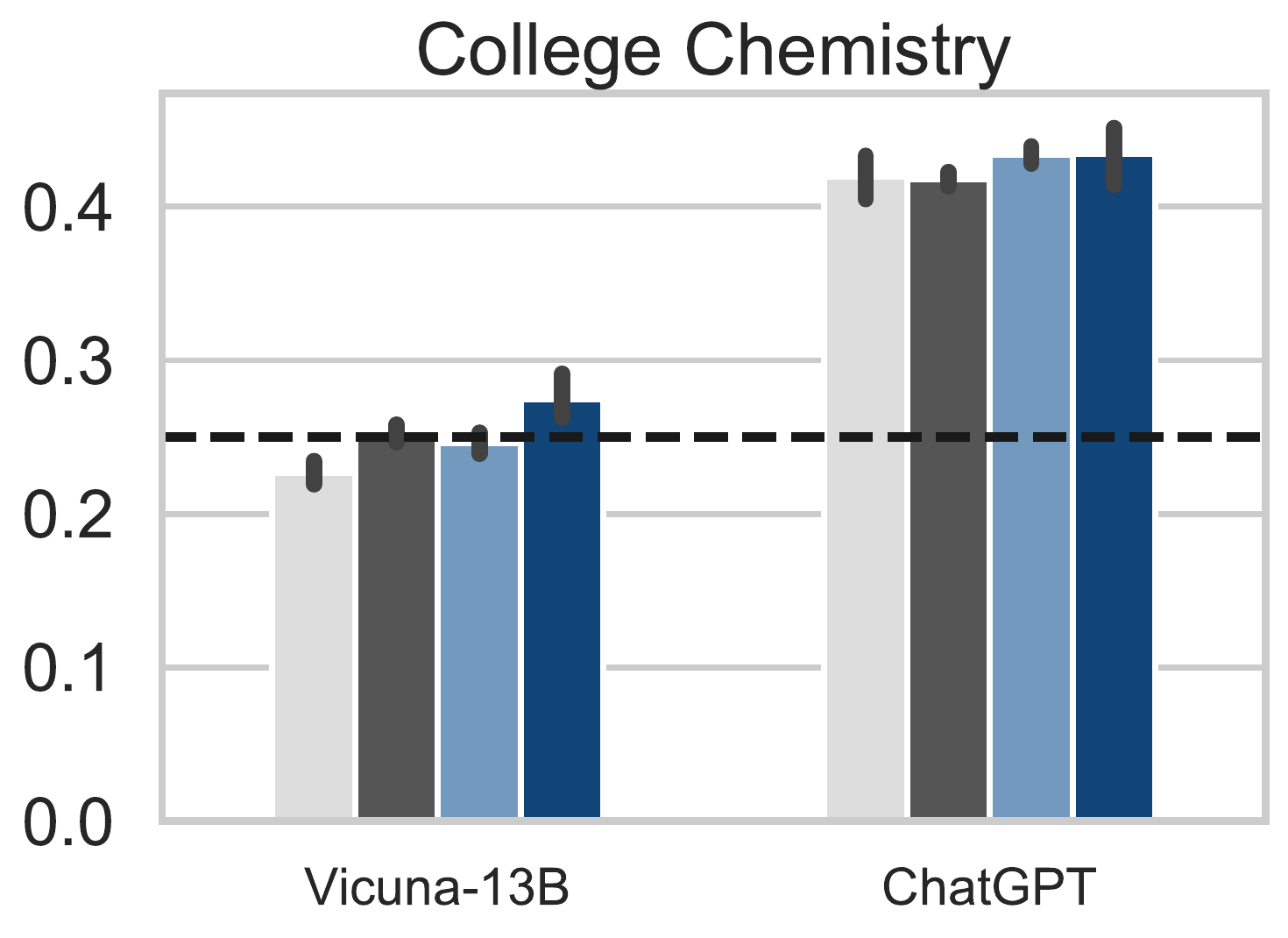}
     \end{subfigure}
     }\\
     \resizebox{\stemscale\textwidth}{!}{
     \begin{subfigure}[c]{0.31\textwidth}
         \centering
         \includegraphics[width=\textwidth]{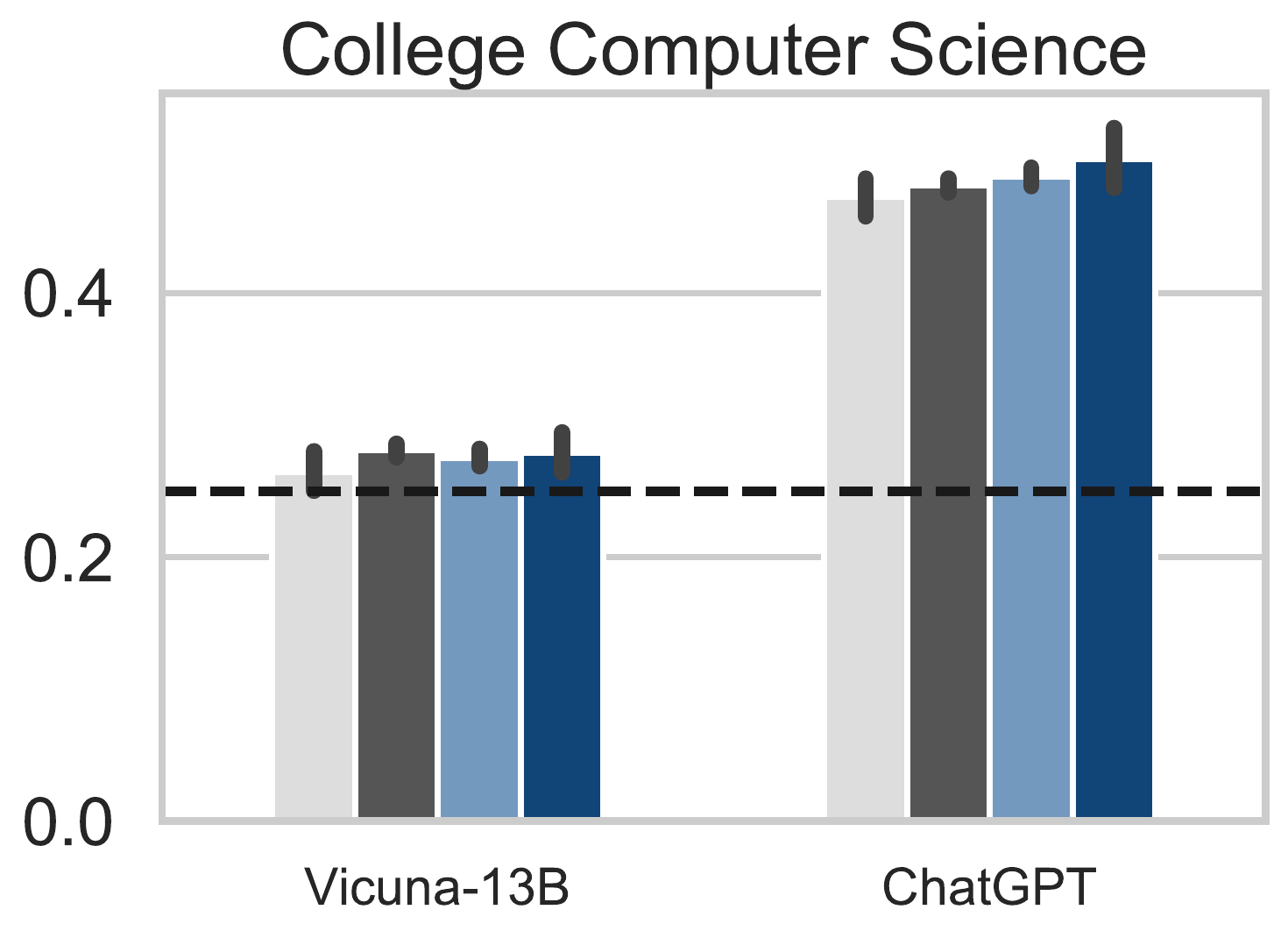}
     \end{subfigure}
      \hfill
     \begin{subfigure}[c]{0.31\textwidth}
         \centering
         \includegraphics[width=\textwidth]{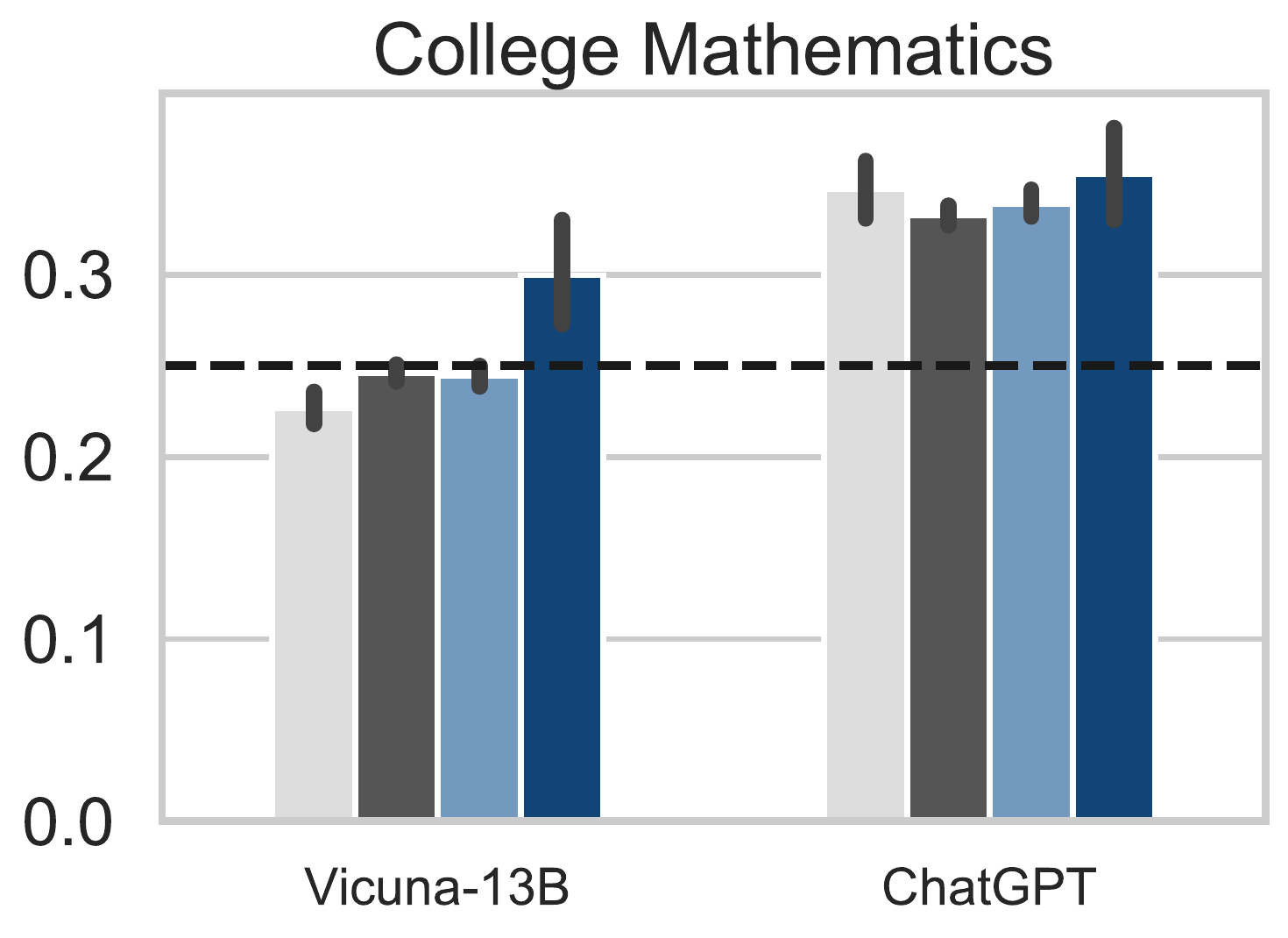}
     \end{subfigure}
     \hfill
     \begin{subfigure}[c]{0.31\textwidth}
         \centering
         \includegraphics[width=\textwidth]{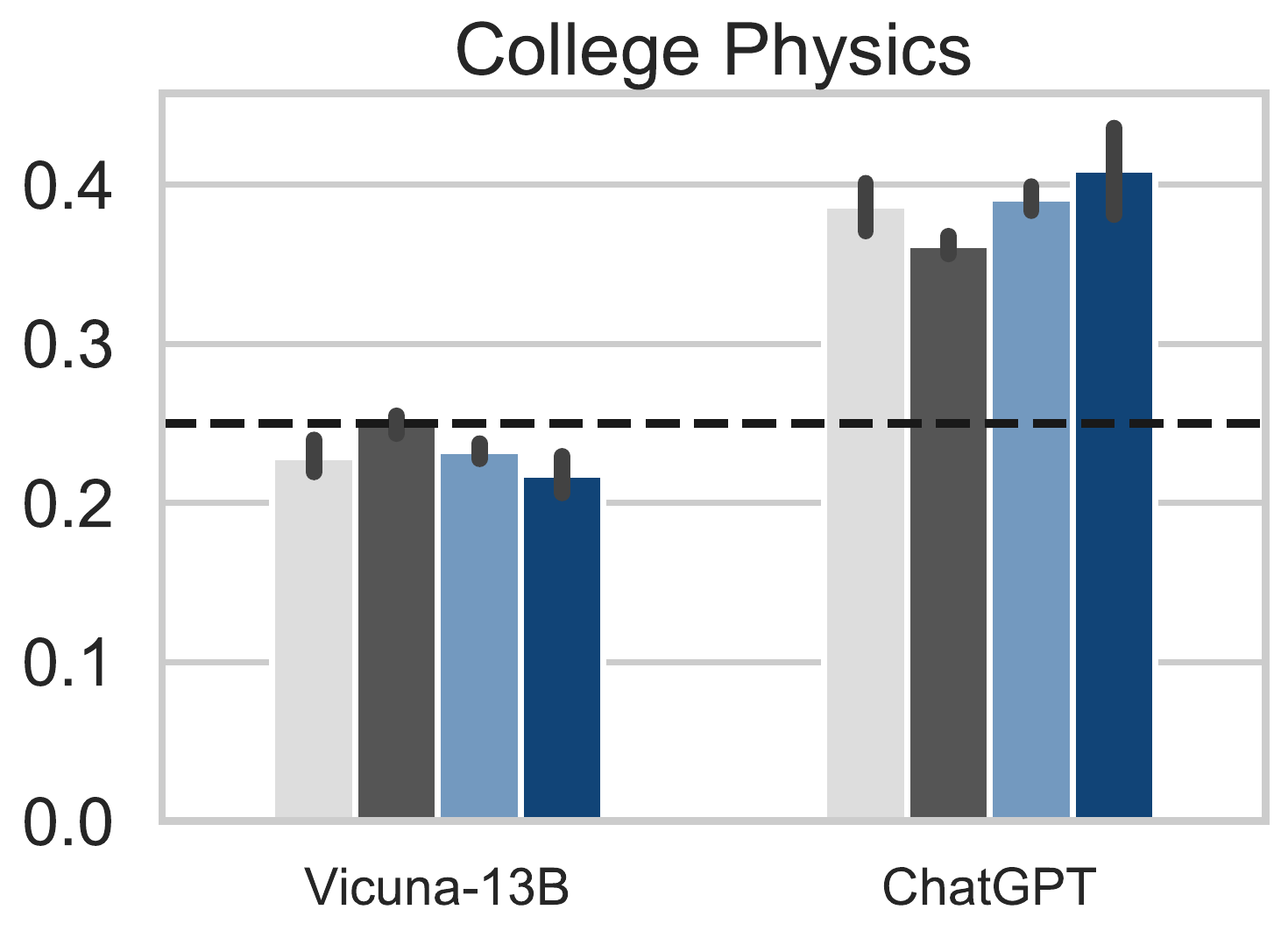}
     \end{subfigure}
     }\\
     \resizebox{\stemscale\textwidth}{!}{
      \begin{subfigure}[c]{0.31\textwidth}
         \centering
         \includegraphics[width=\textwidth]{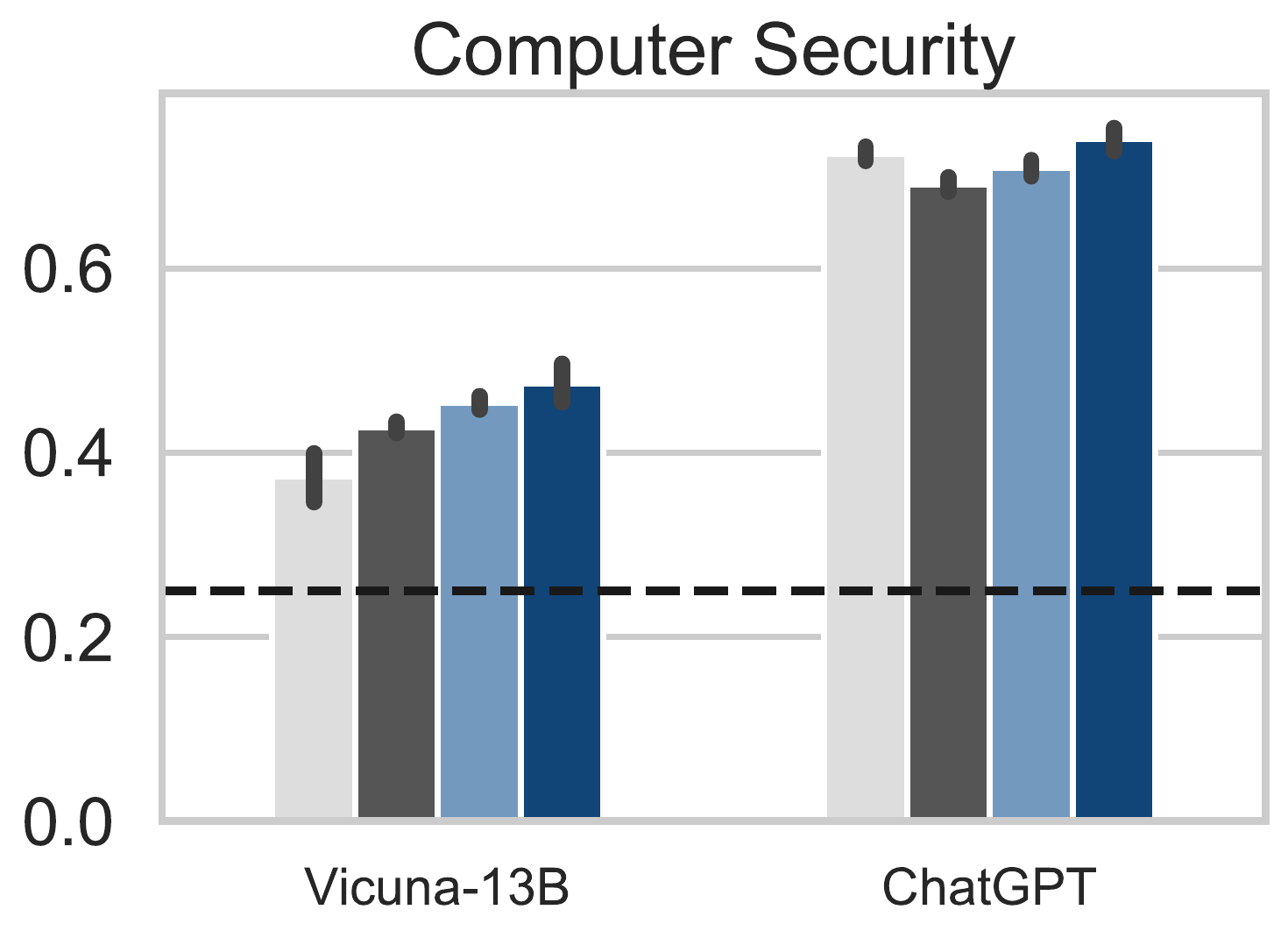}
     \end{subfigure}
      \hfill
     \begin{subfigure}[c]{0.31\textwidth}
         \centering
         \includegraphics[width=\textwidth]{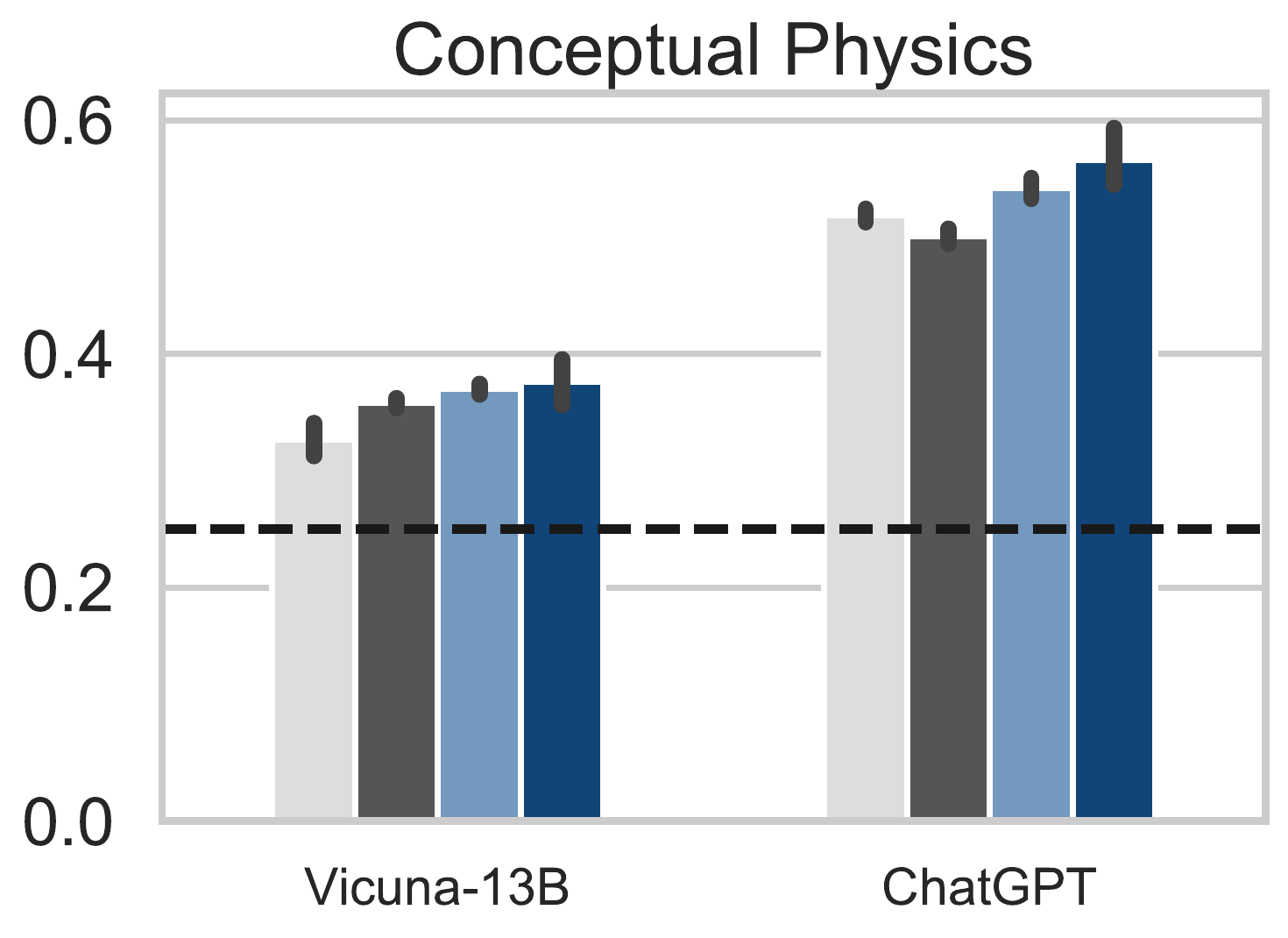}
     \end{subfigure}
     \hfill
     \begin{subfigure}[c]{0.31\textwidth}
         \centering
         \includegraphics[width=\textwidth]{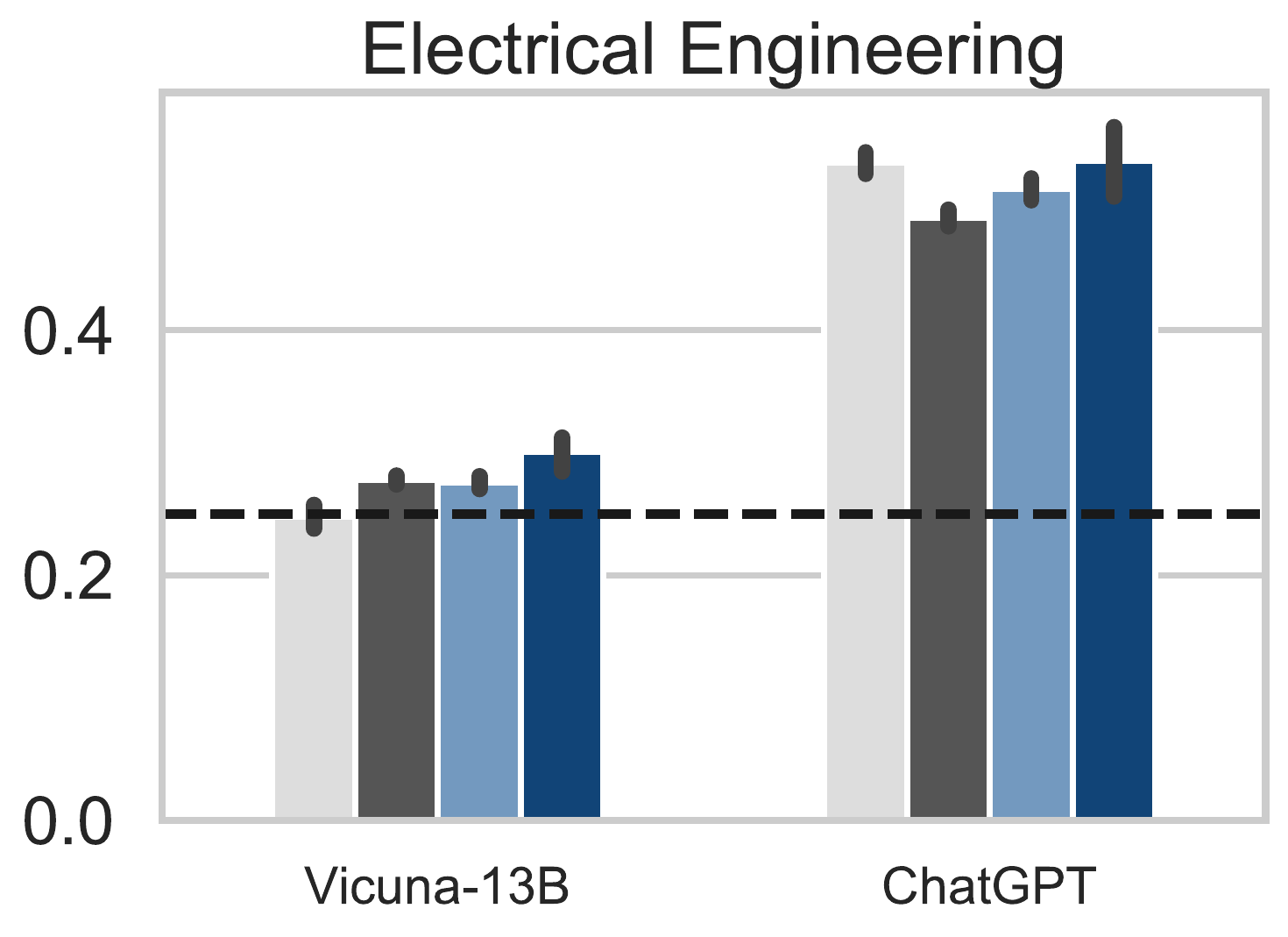}
     \end{subfigure}
     }\\
     \resizebox{\stemscale\textwidth}{!}{
     \begin{subfigure}[c]{0.31\textwidth}
         \centering
         \includegraphics[width=\textwidth]{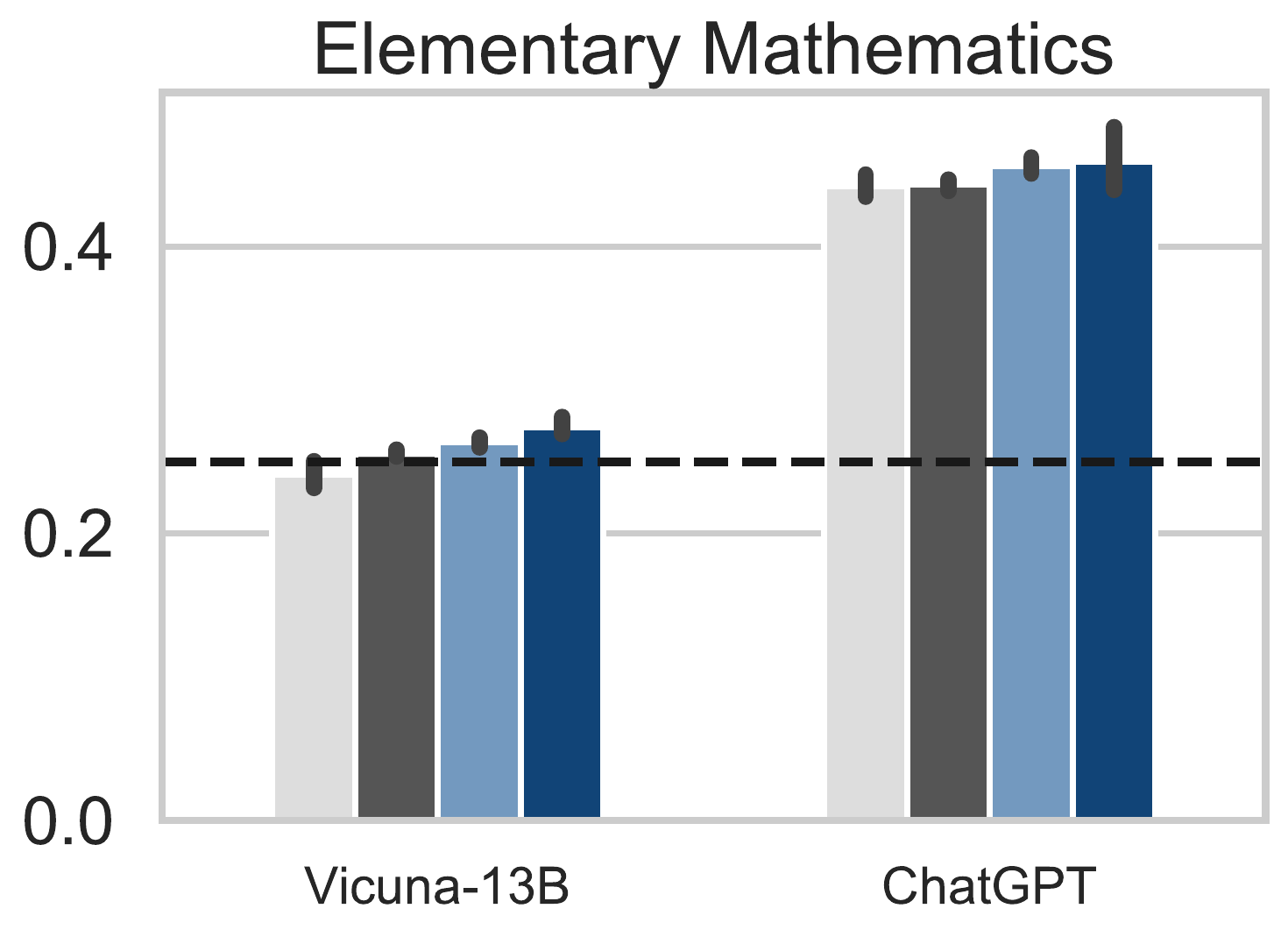}
     \end{subfigure}
     \hfill
     \begin{subfigure}[c]{0.31\textwidth}
         \centering
         \includegraphics[width=\textwidth]{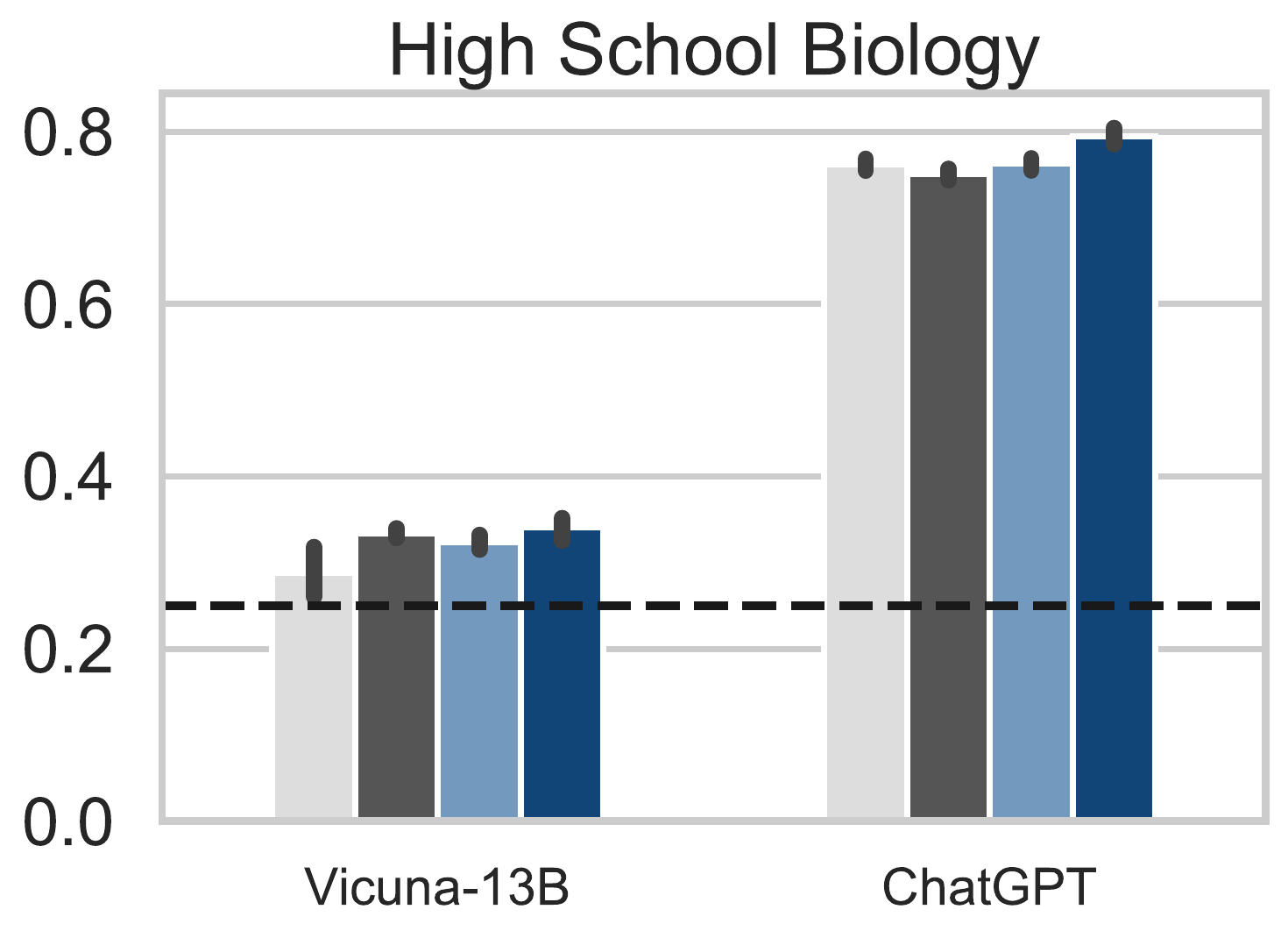}
     \end{subfigure}
      \hfill
     \begin{subfigure}[c]{0.31\textwidth}
         \centering
         \includegraphics[width=\textwidth]{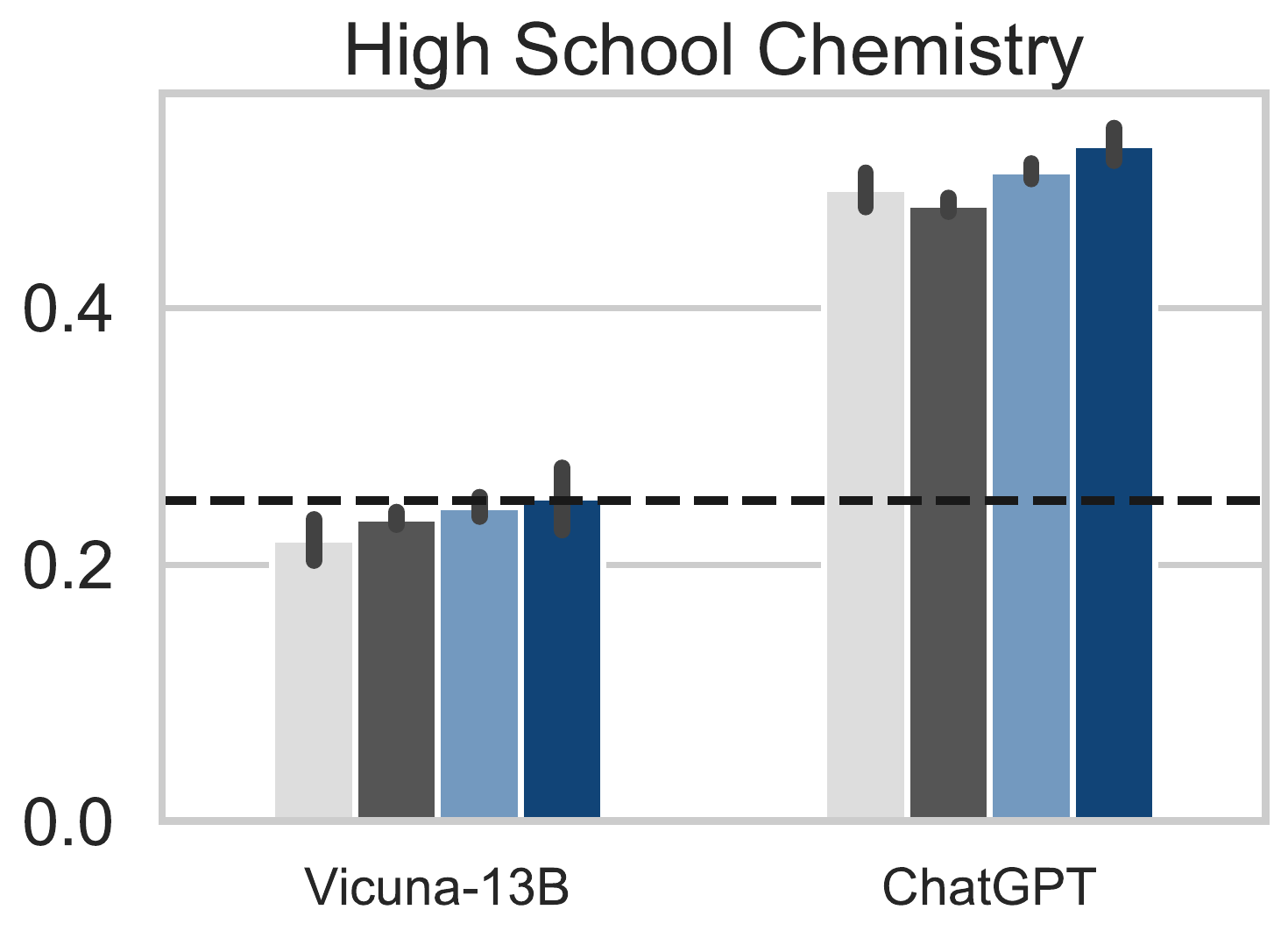}
     \end{subfigure}
     }\\
     \resizebox{\stemscale\textwidth}{!}{
     \begin{subfigure}[c]{0.31\textwidth}
         \centering
         \includegraphics[width=\textwidth]{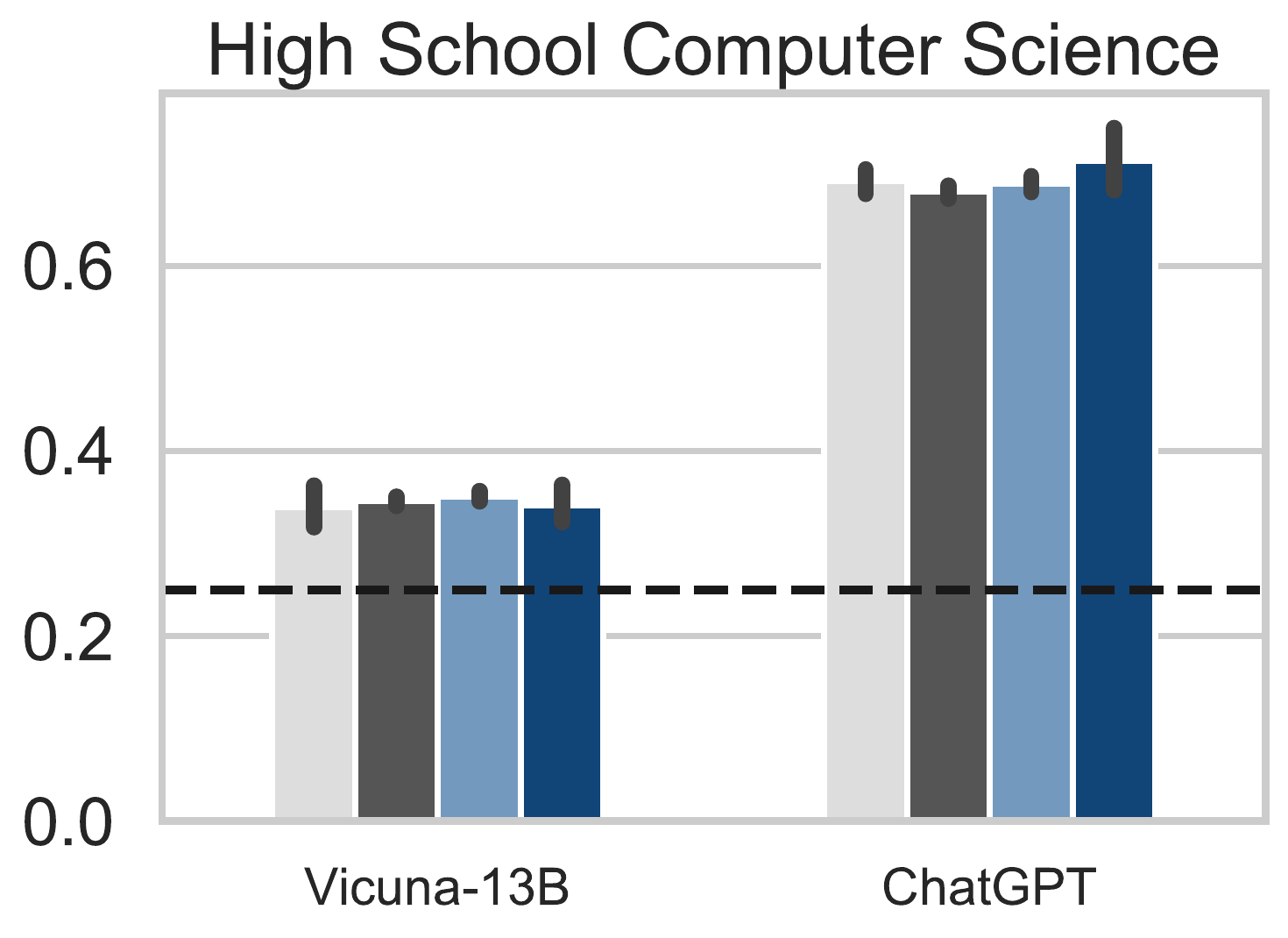}
     \end{subfigure}
      \hfill
     \begin{subfigure}[c]{0.31\textwidth}
         \centering
         \includegraphics[width=\textwidth]{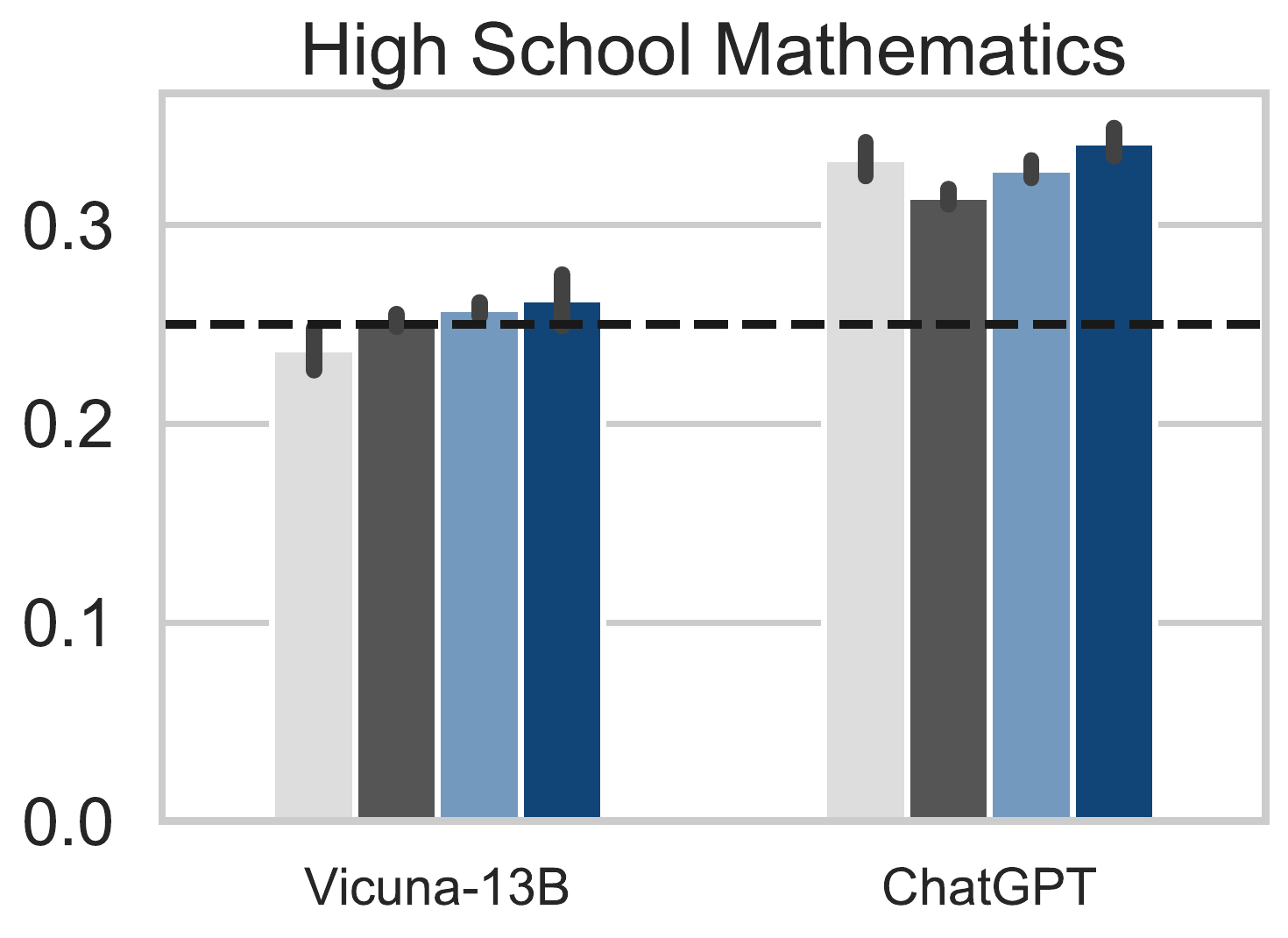}
     \end{subfigure}
     \hfill
     \begin{subfigure}[c]{0.31\textwidth}
         \centering
         \includegraphics[width=\textwidth]{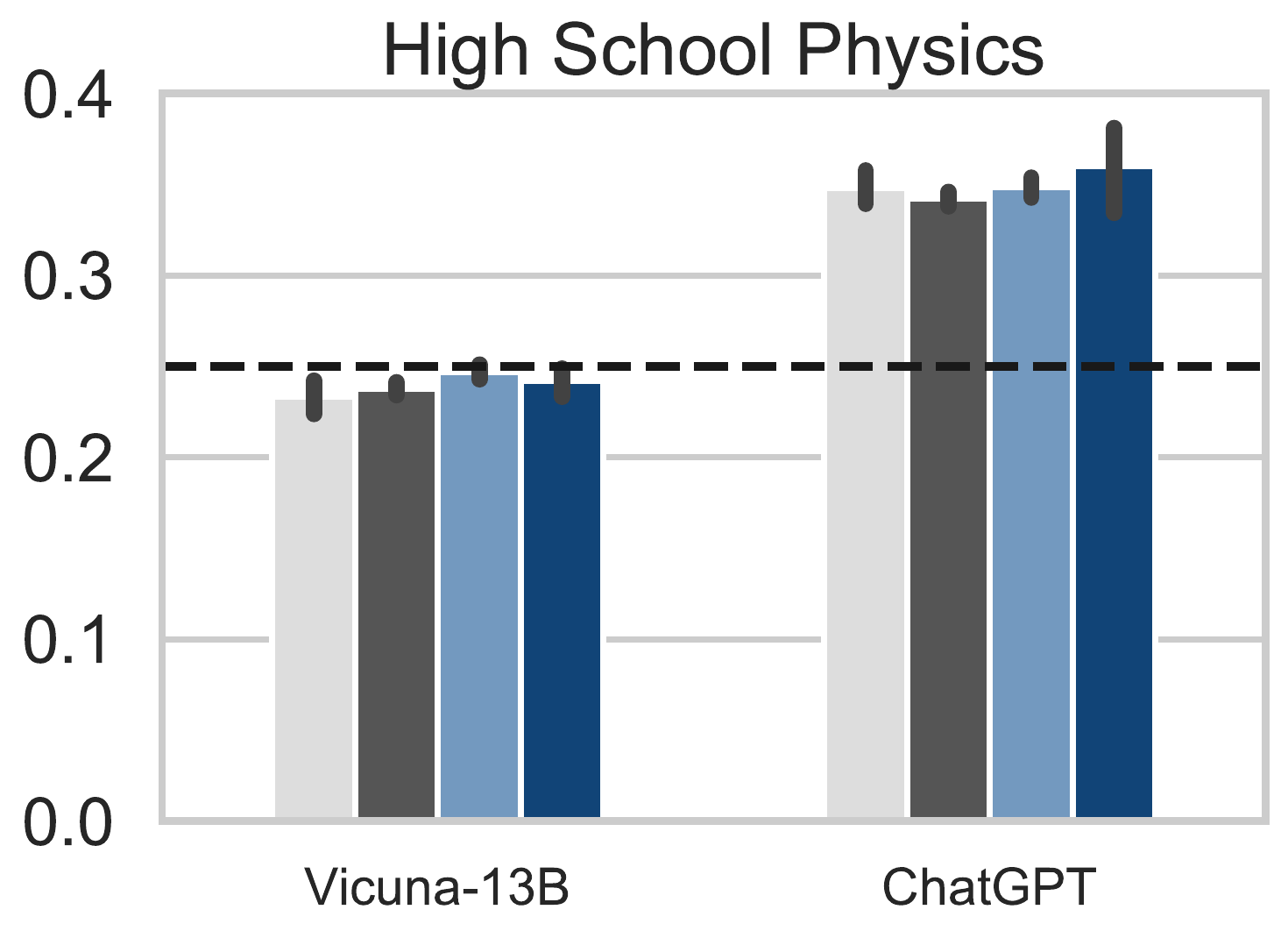}
     \end{subfigure}
     }\\
     \resizebox{\stemscale\textwidth}{!}{
     \begin{subfigure}[c]{0.31\textwidth}
         \centering
         \includegraphics[width=\textwidth]{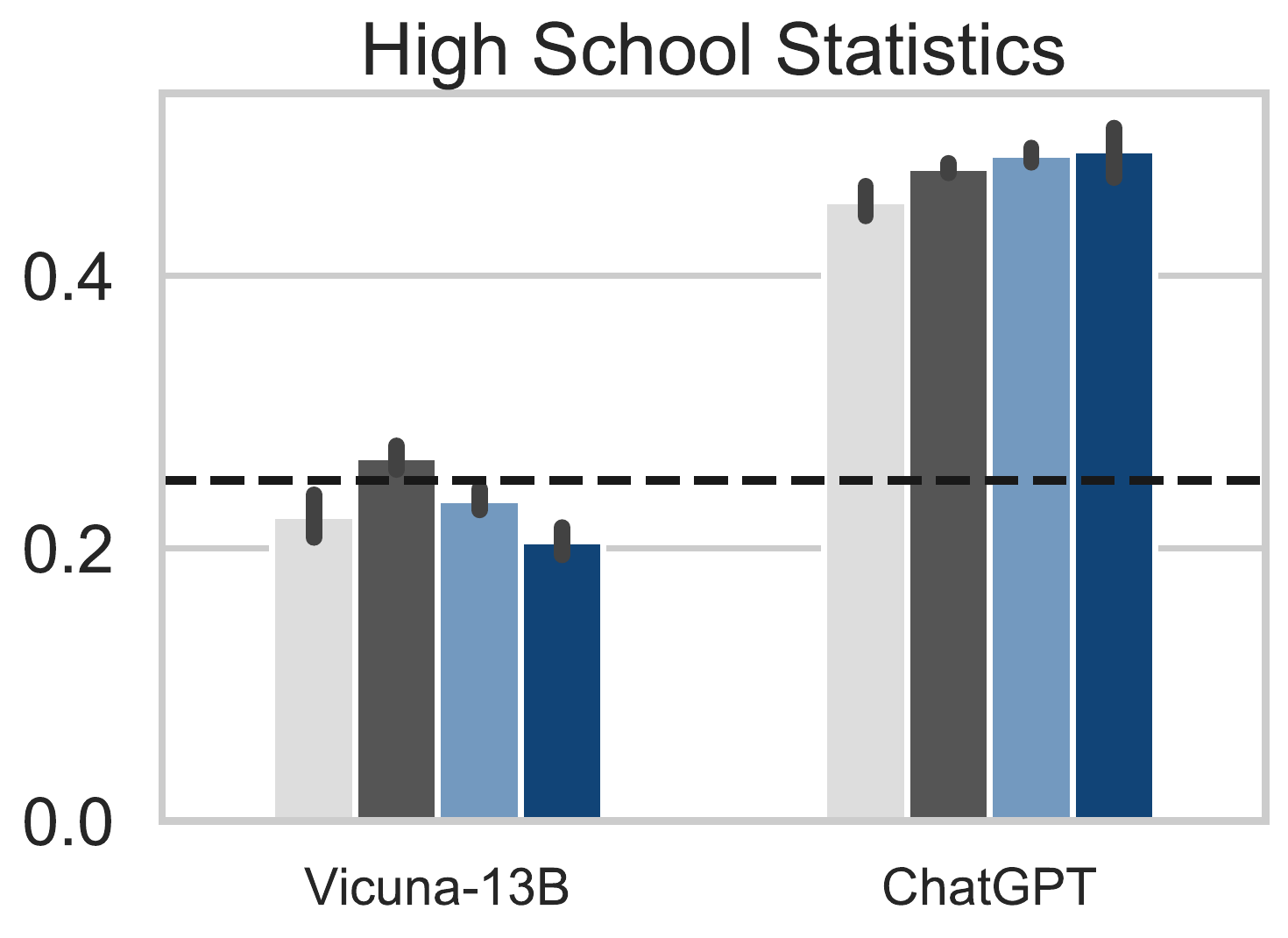}
     \end{subfigure}
     \hfill
     \begin{subfigure}[c]{0.31\textwidth}
         \centering
         \includegraphics[width=\textwidth]{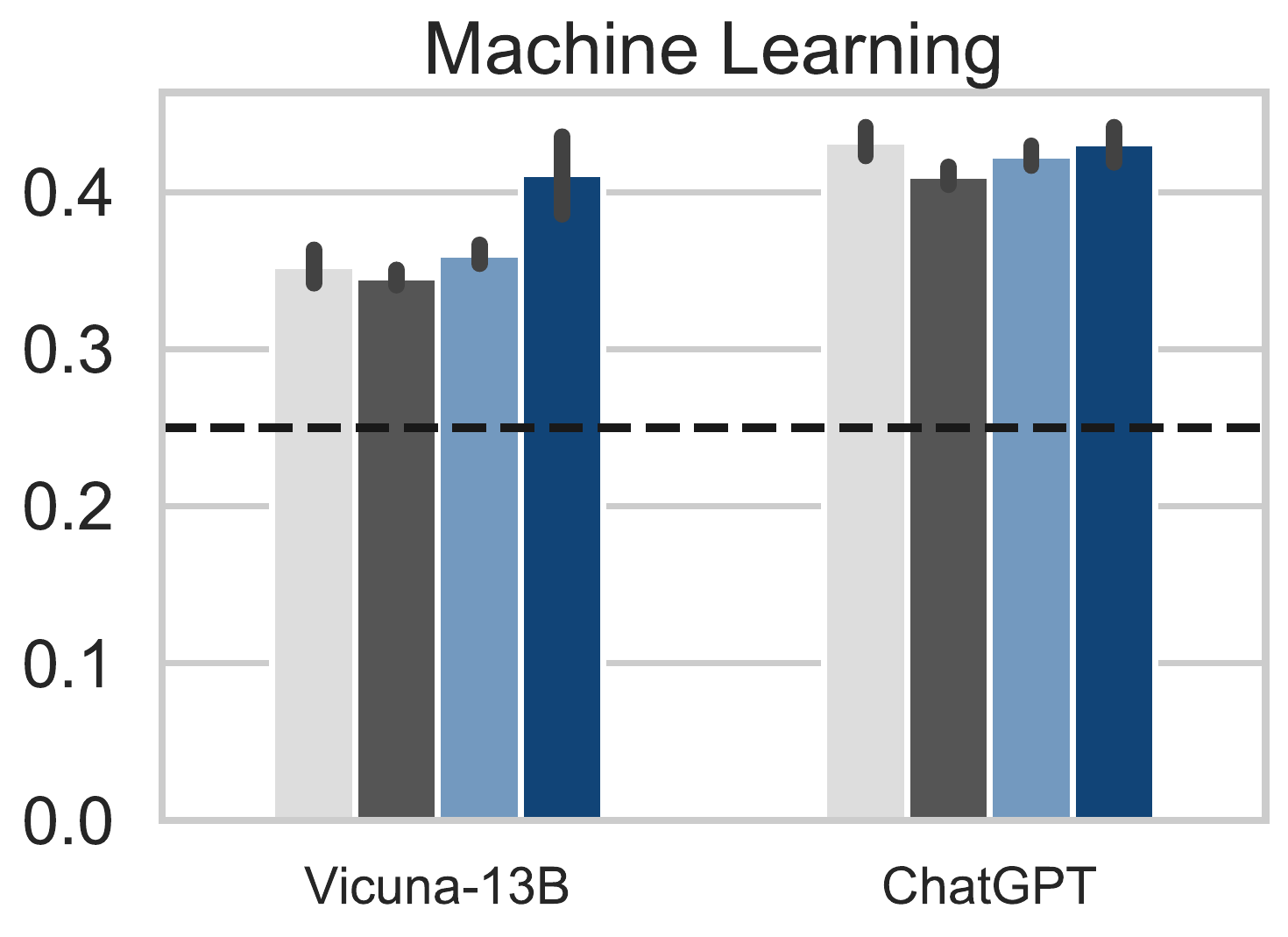}
     \end{subfigure}
     \hfill
     \begin{subfigure}[r]{0.31\textwidth}
         \centering
         \includegraphics[width=.9\textwidth]{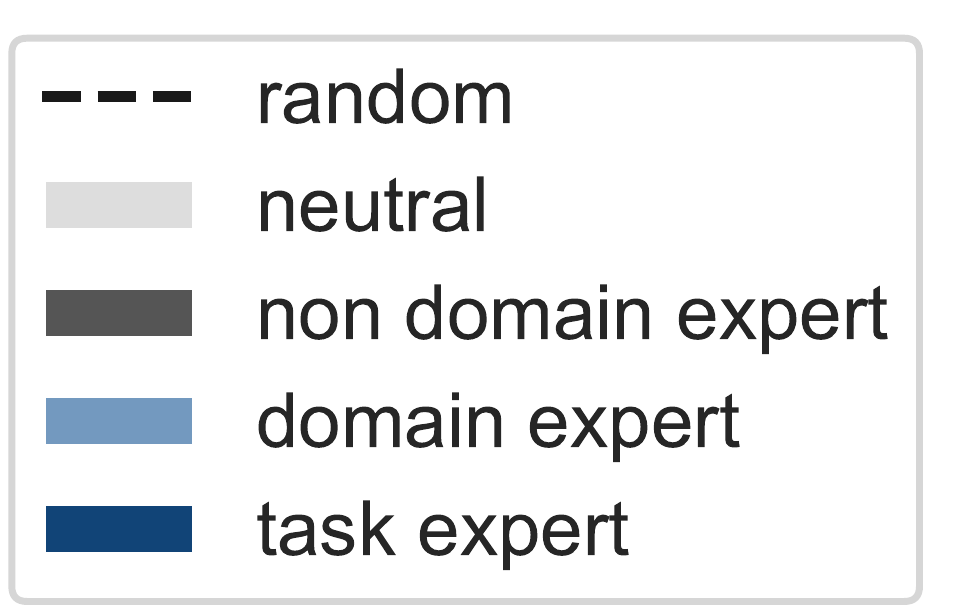}
     \end{subfigure}
     }
     \caption{Comparison between Vicuna-13B and ChatGPT for expertise-based impersonation on the STEM domain of the MMLU reasoning benchmark. We compare the task expert results with the average of all neutral personas, the average of all domain expert personas, the average of all non-domain expert personas and the random baseline (horizontal line). The first plot shows the average over all STEM tasks, while the remaining plots show the results for each STEM task individually. All 95\% confidence intervals are computed over the average task accuracy.}%
    \label{fig:mmlu_stem}
\end{figure}

\begin{figure}[h!]
     \centering
     \begin{subfigure}[c]{0.35\textwidth}
         \centering
         \includegraphics[width=\textwidth]{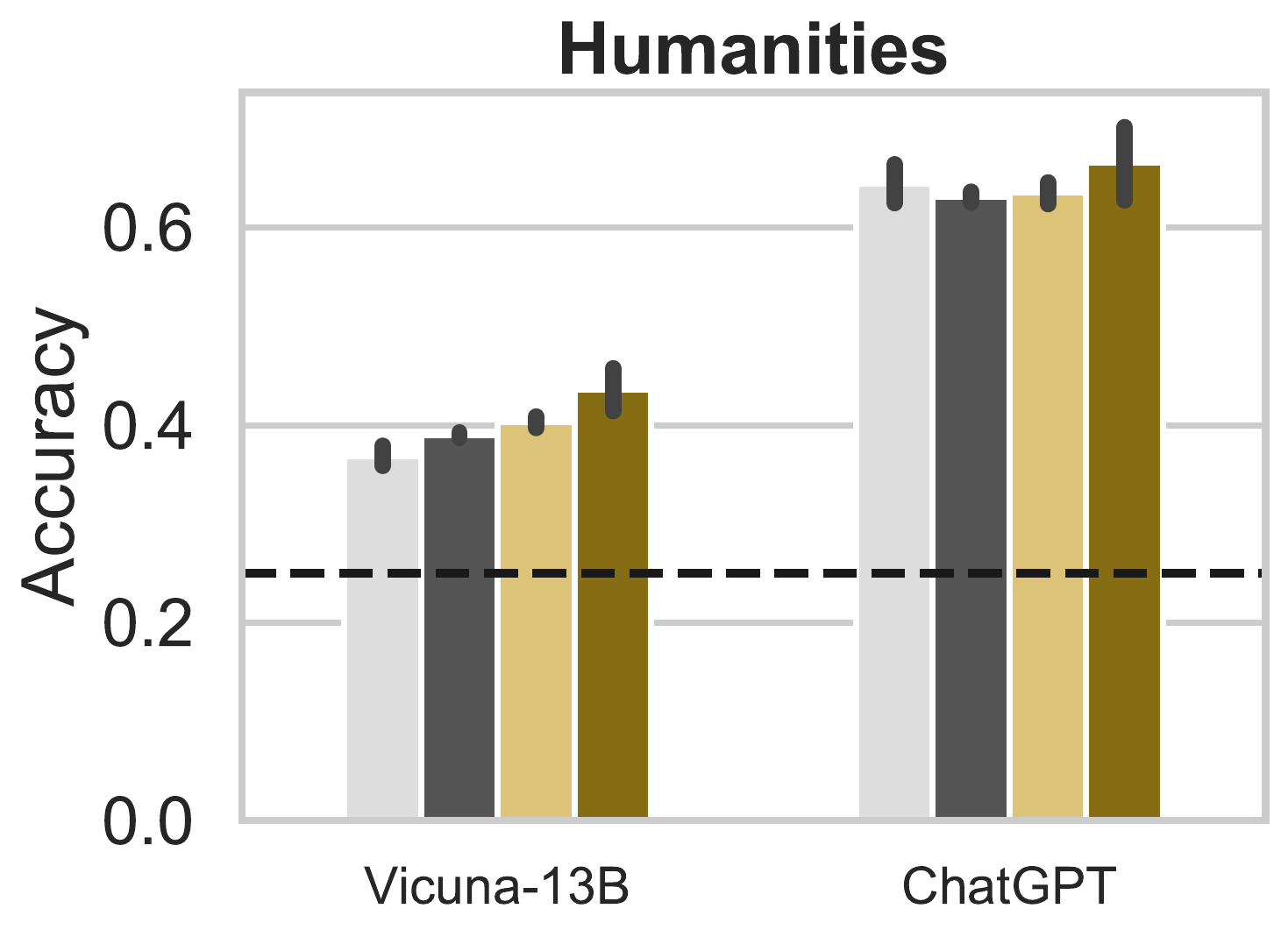}
     \end{subfigure}
     \hfill
     \begin{subfigure}[c]{0.31\textwidth}
         \centering
         \includegraphics[width=\textwidth]{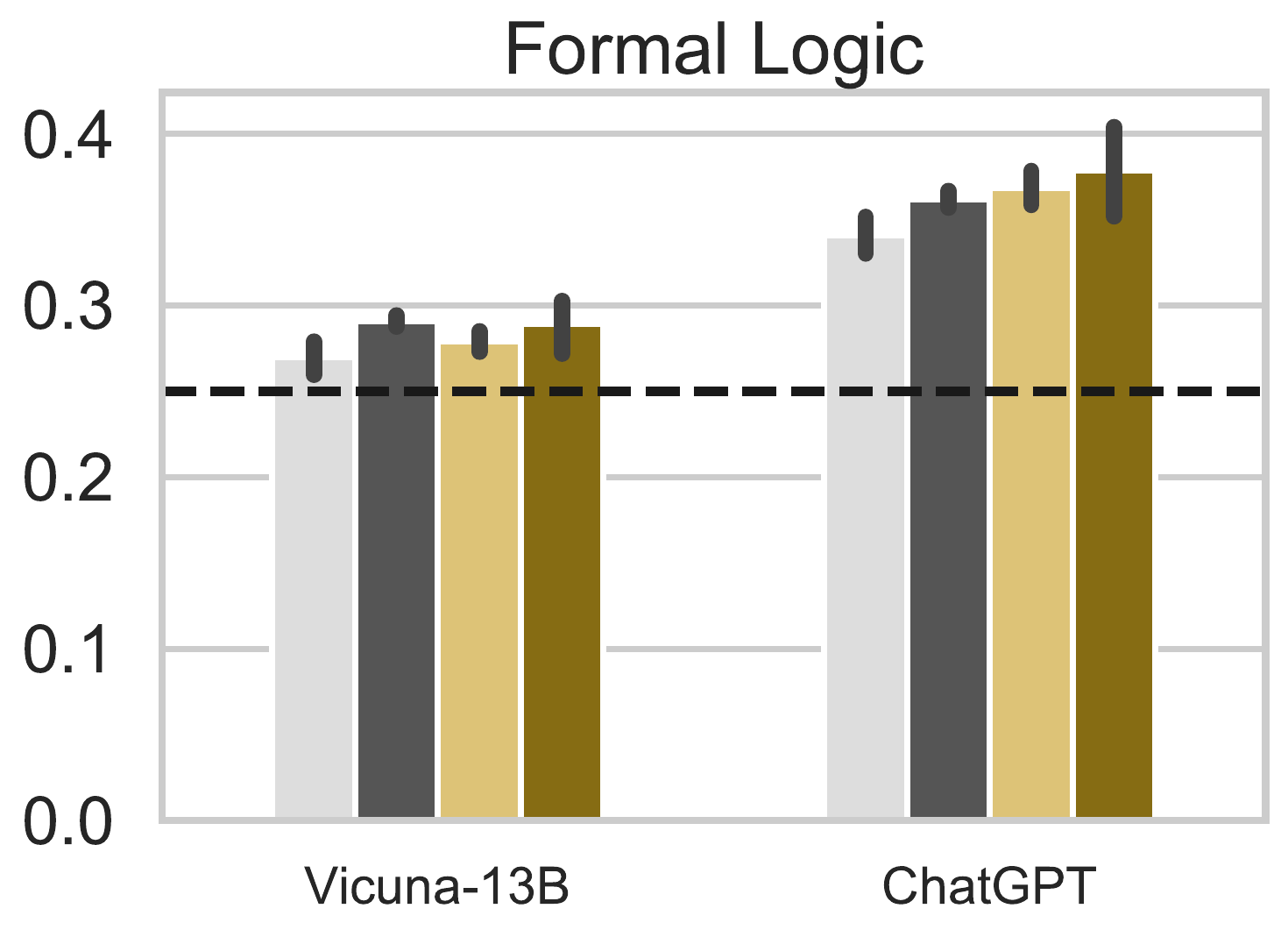}
     \end{subfigure}
     \hfill
     \begin{subfigure}[c]{0.32\textwidth}
         \centering
         \includegraphics[width=\textwidth]{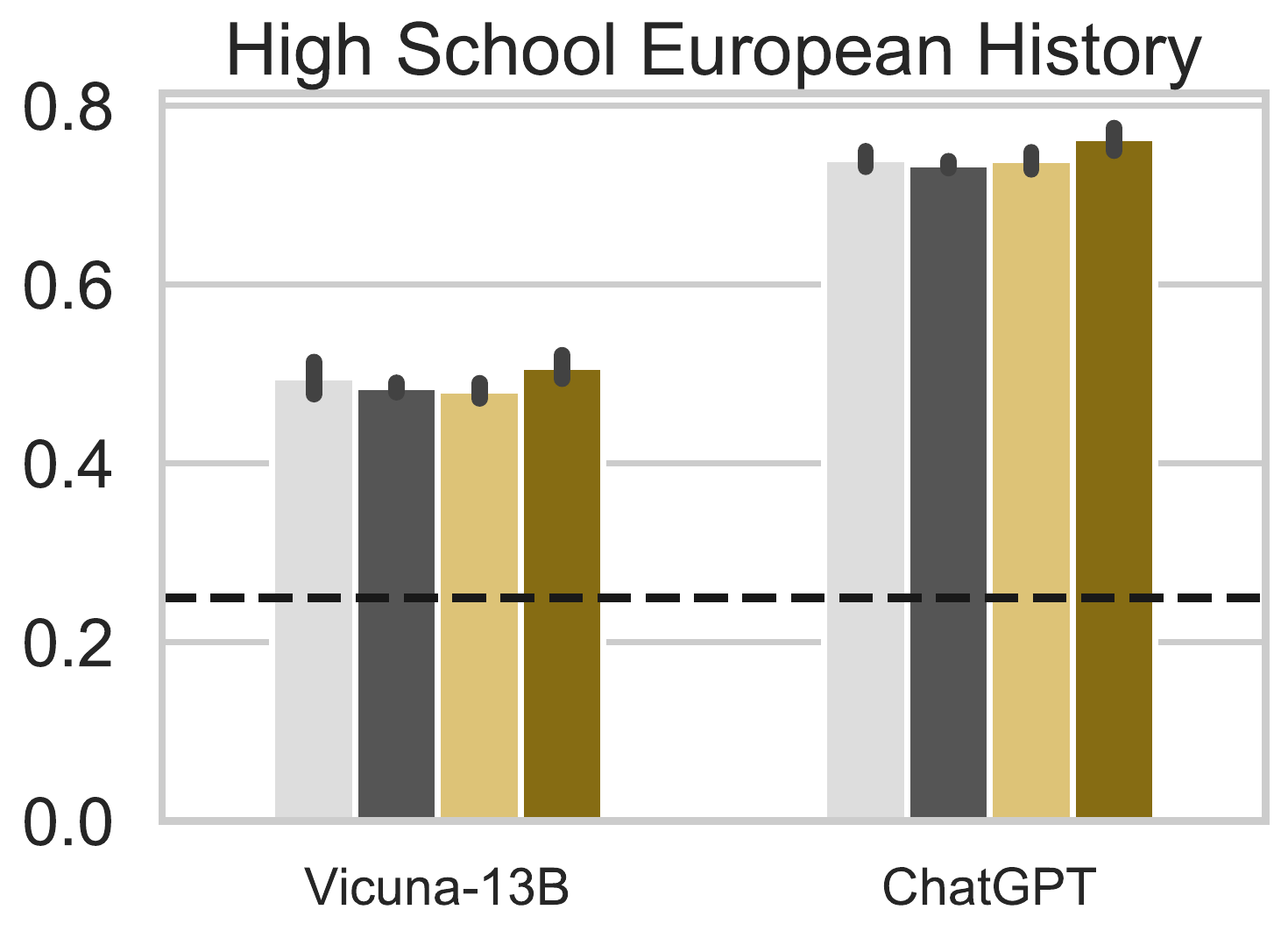}
     \end{subfigure}
     \\
     \begin{subfigure}[c]{0.31\textwidth}
         \centering
         \includegraphics[width=\textwidth]{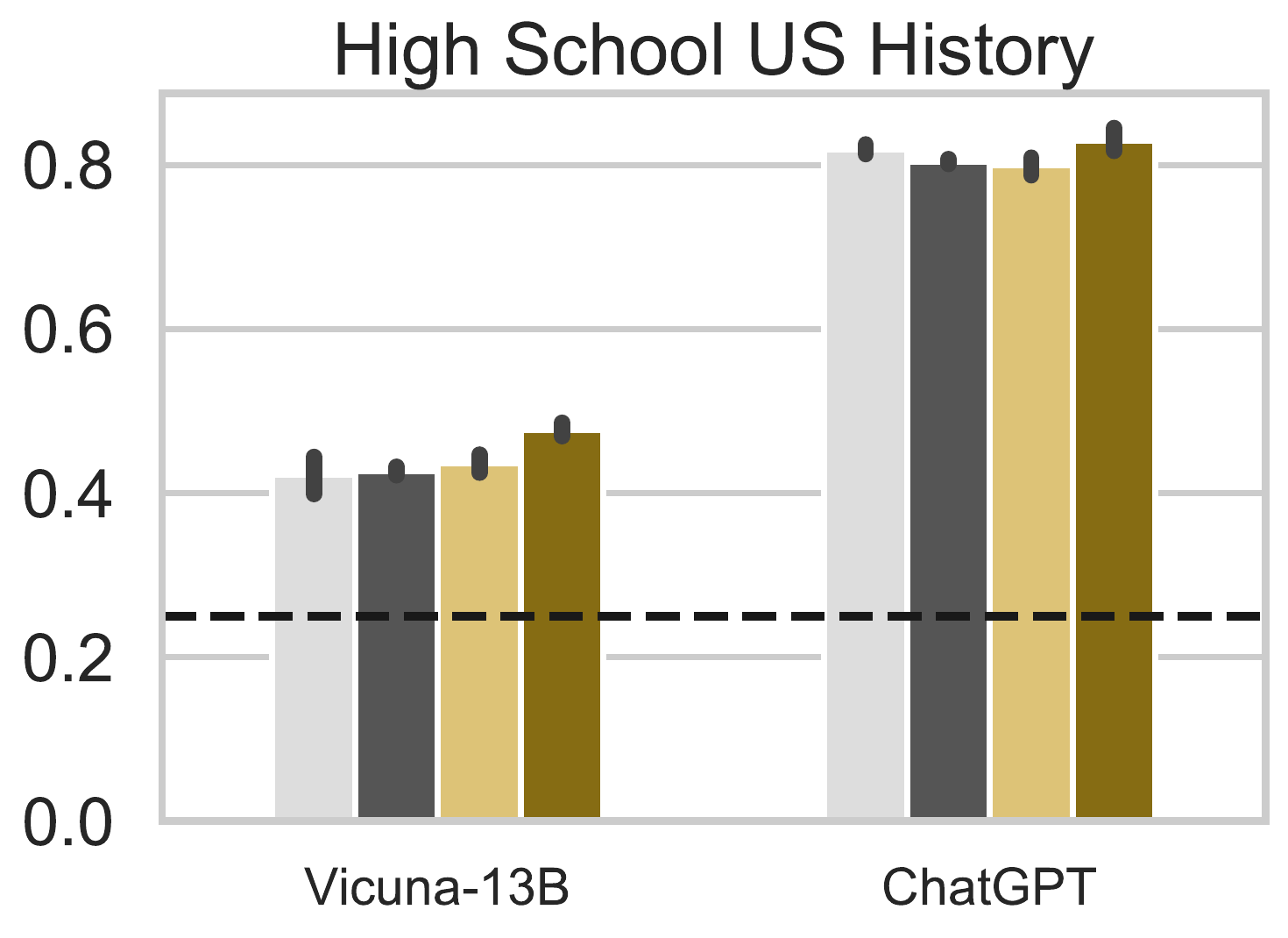}
     \end{subfigure}
     \hfill
     \begin{subfigure}[c]{0.31\textwidth}
         \centering
         \includegraphics[width=\textwidth]{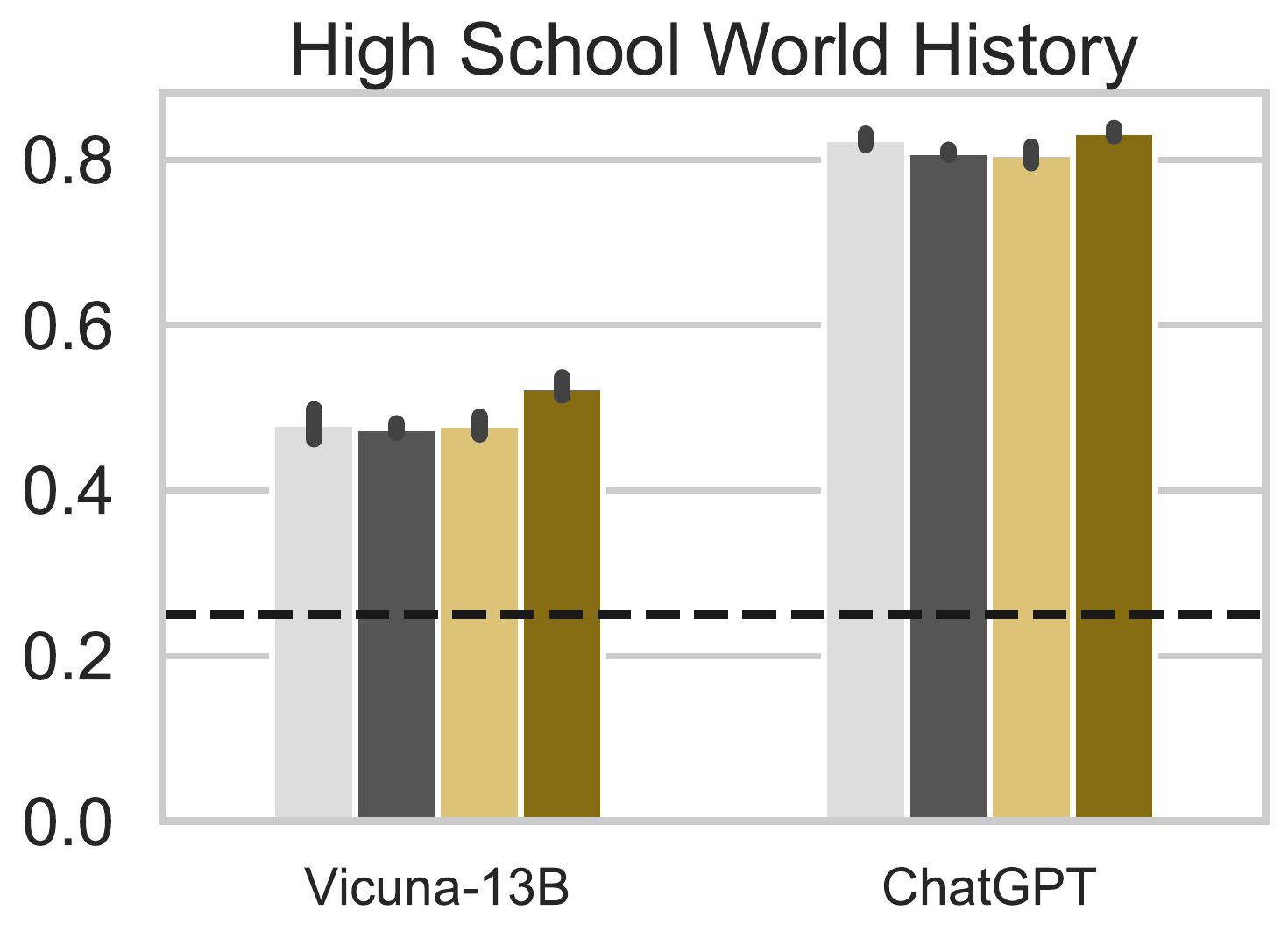}
     \end{subfigure}
     \hfill
     \begin{subfigure}[c]{0.31\textwidth}
         \centering
         \includegraphics[width=\textwidth]{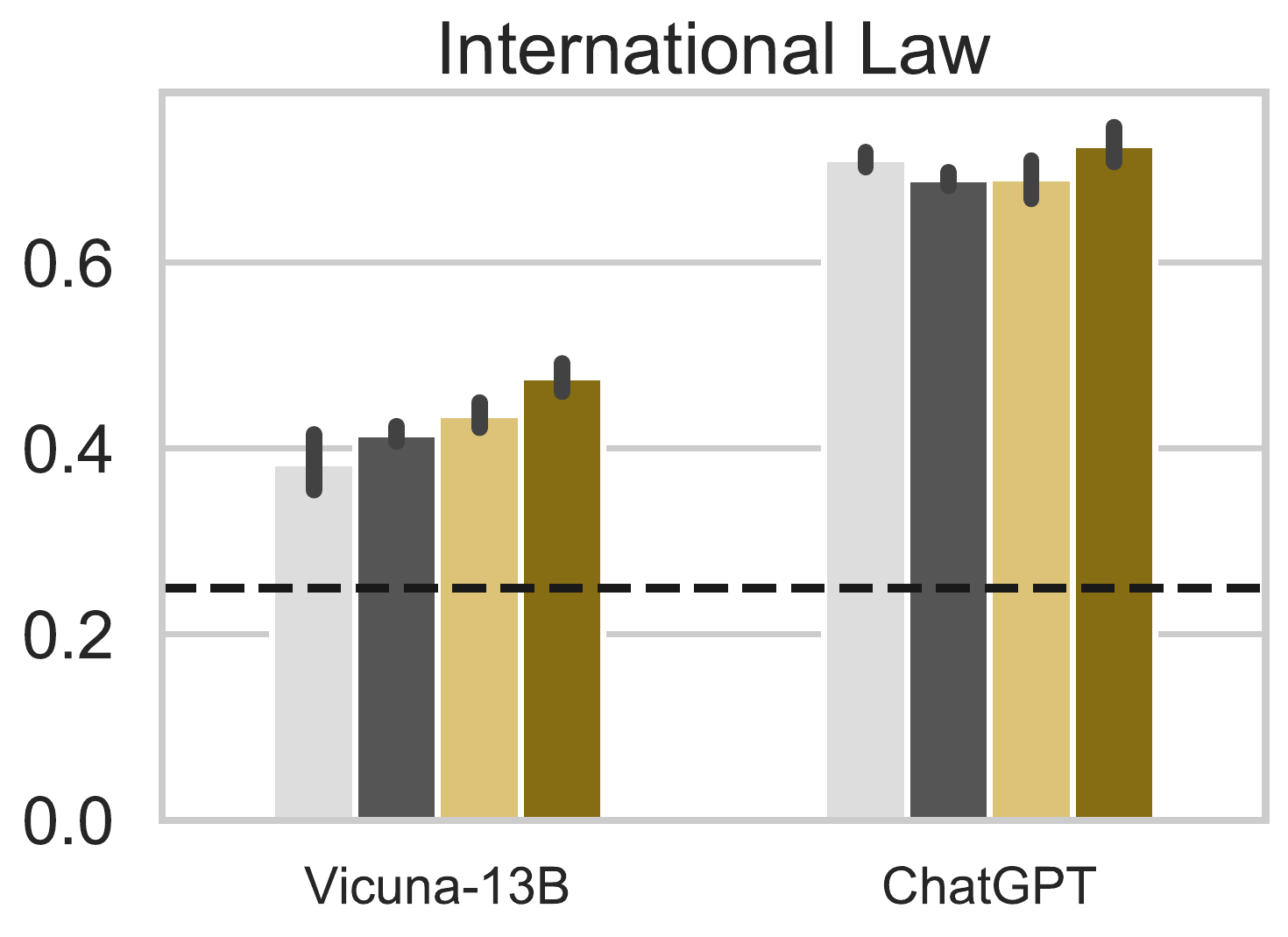}
     \end{subfigure}
     \\
     \begin{subfigure}[c]{0.31\textwidth}
         \centering
         \includegraphics[width=\textwidth]{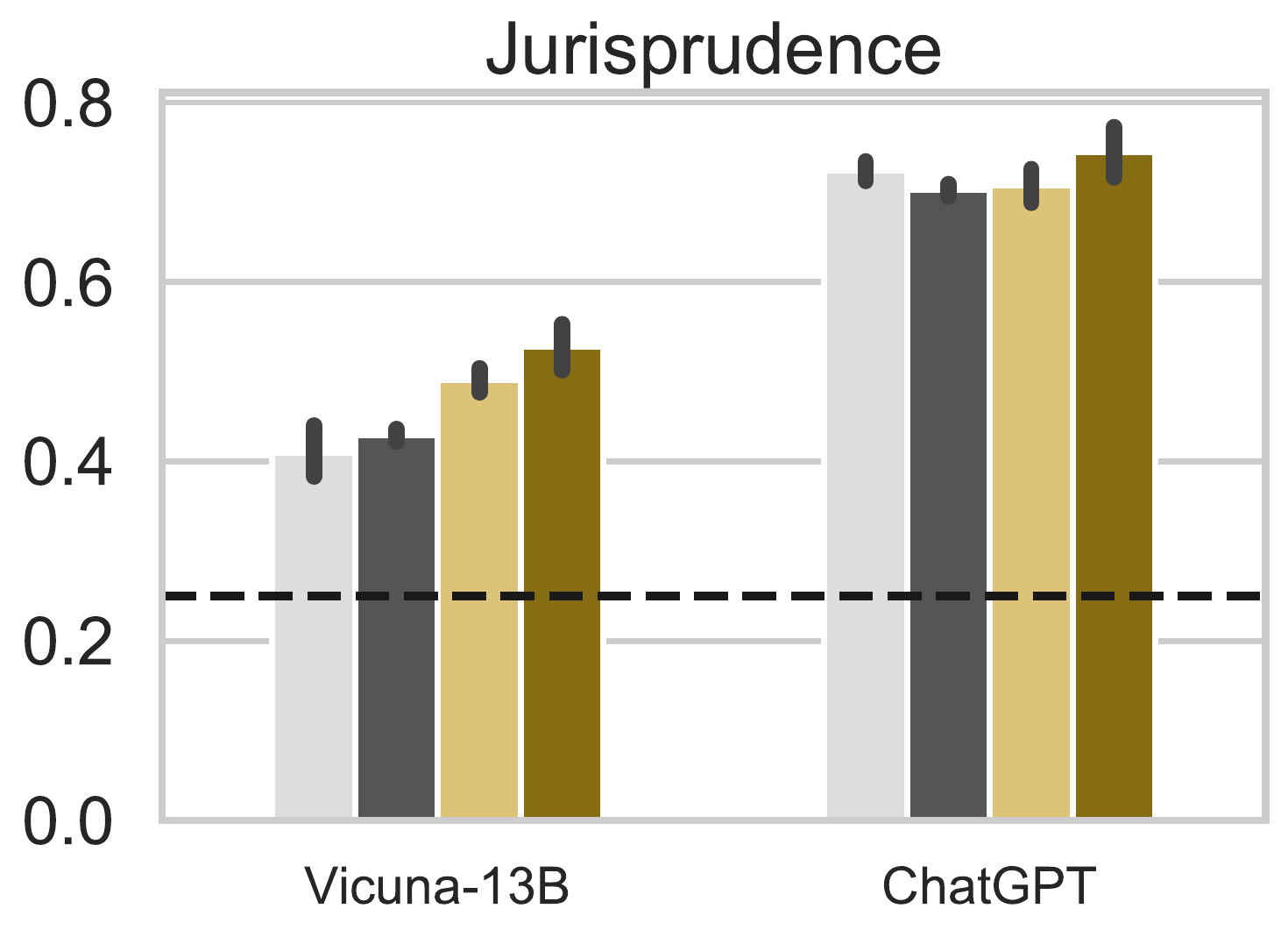}
     \end{subfigure}
      \hfill
     \begin{subfigure}[c]{0.31\textwidth}
         \centering
         \includegraphics[width=\textwidth]{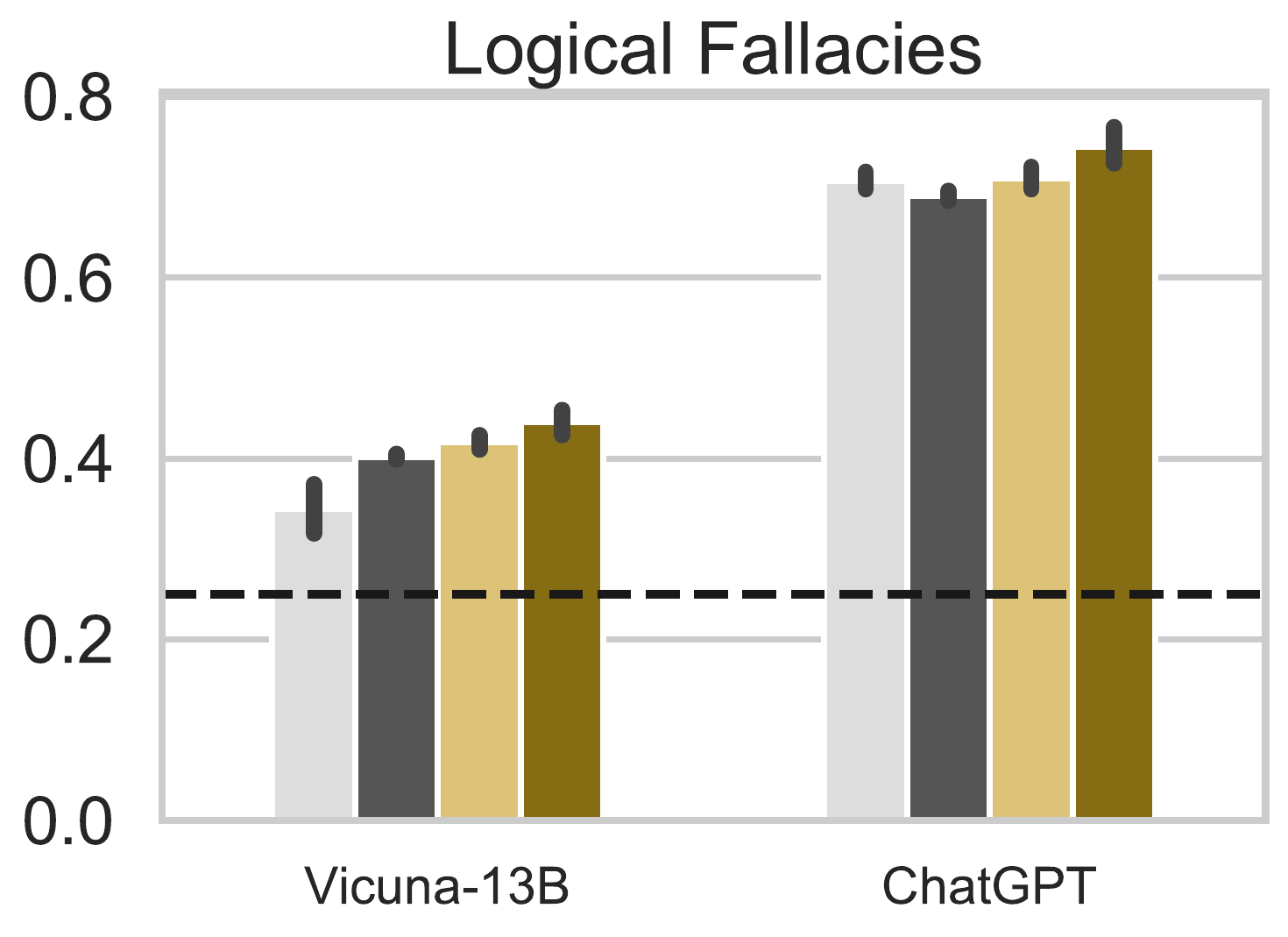}
     \end{subfigure}
     \hfill
     \begin{subfigure}[c]{0.31\textwidth}
         \centering
         \includegraphics[width=\textwidth]{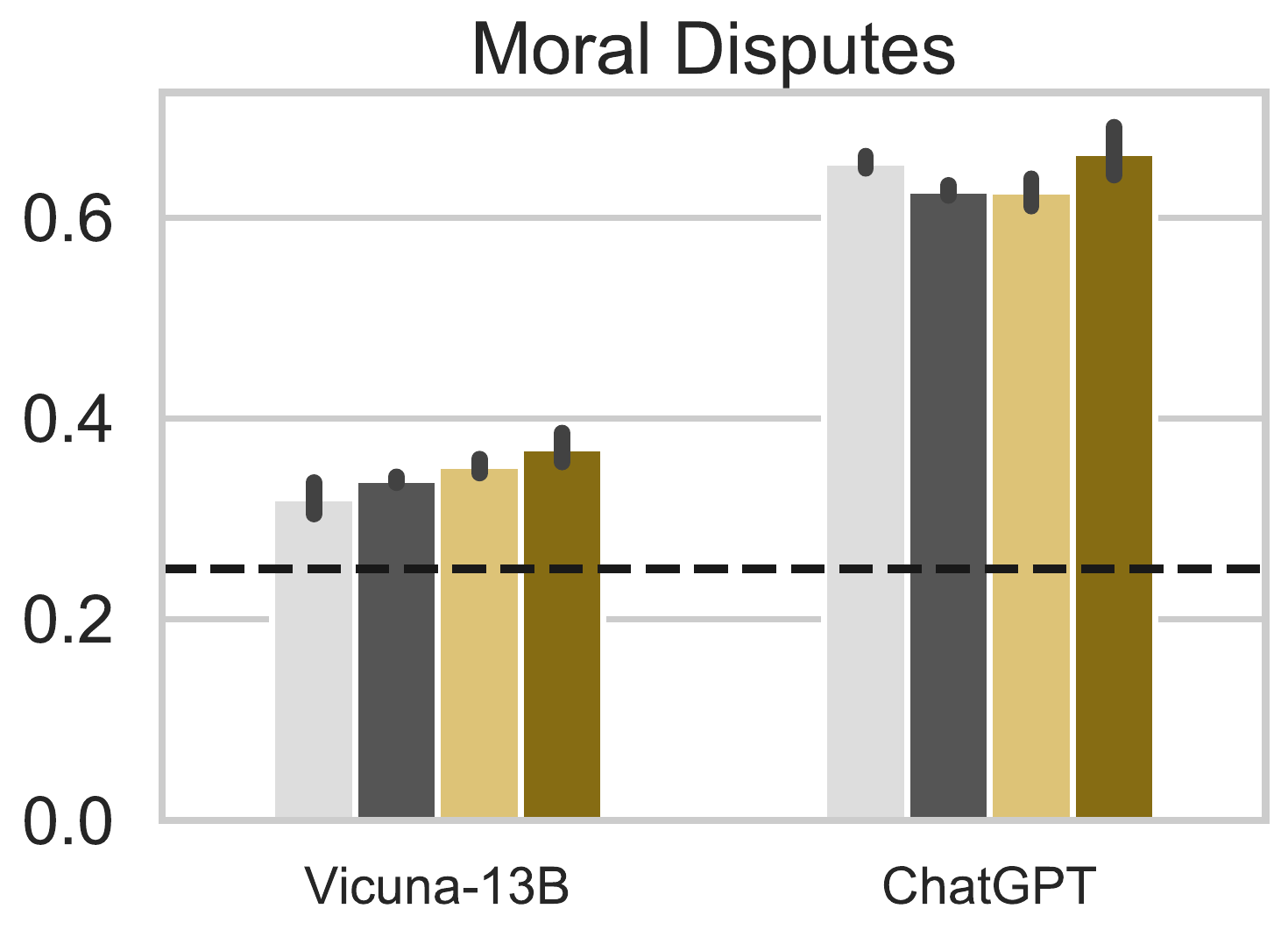}
     \end{subfigure}
     \\
      \begin{subfigure}[c]{0.31\textwidth}
         \centering
         \includegraphics[width=\textwidth]{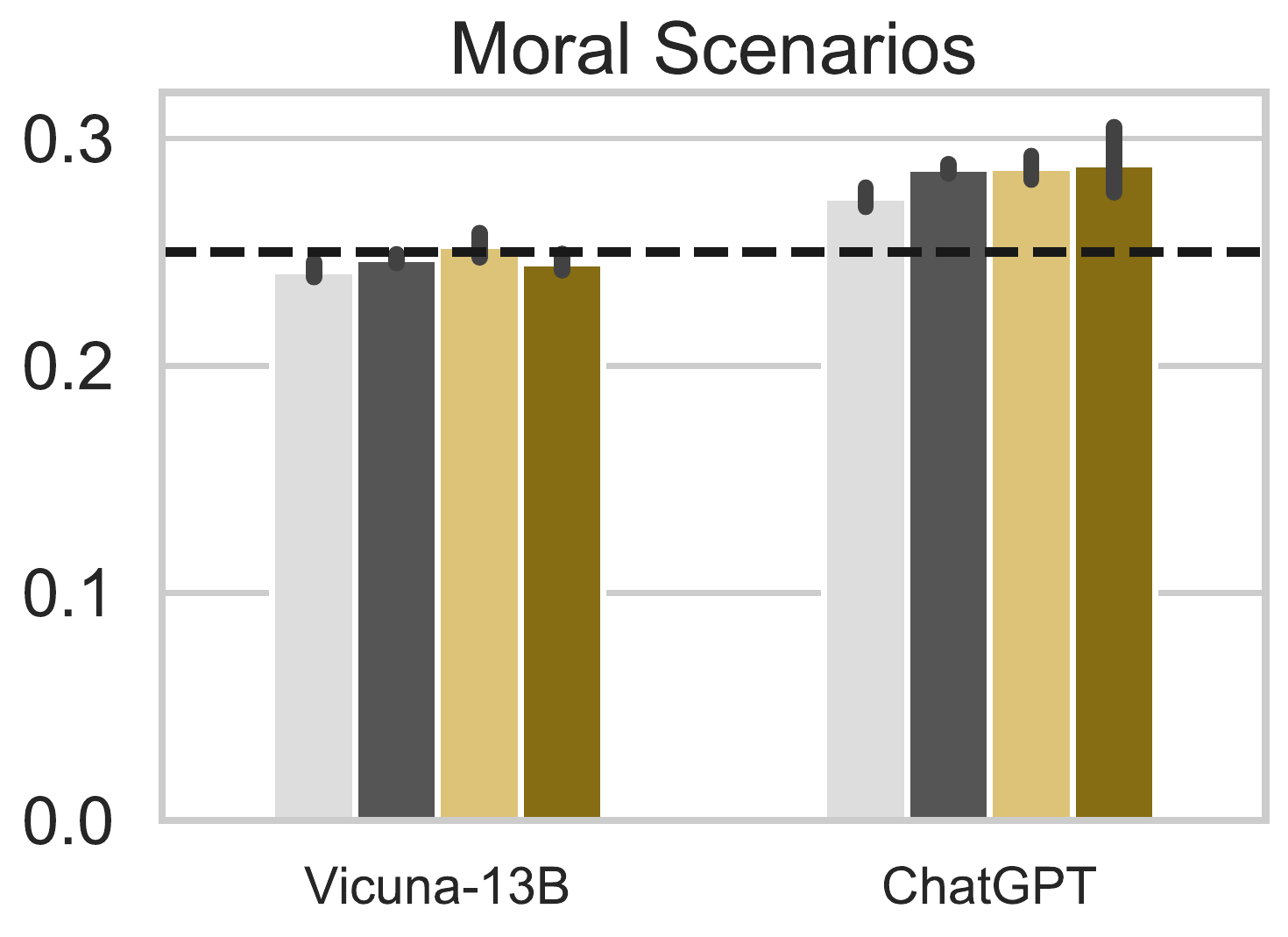}
     \end{subfigure}
      \hfill
     \begin{subfigure}[c]{0.31\textwidth}
         \centering
         \includegraphics[width=\textwidth]{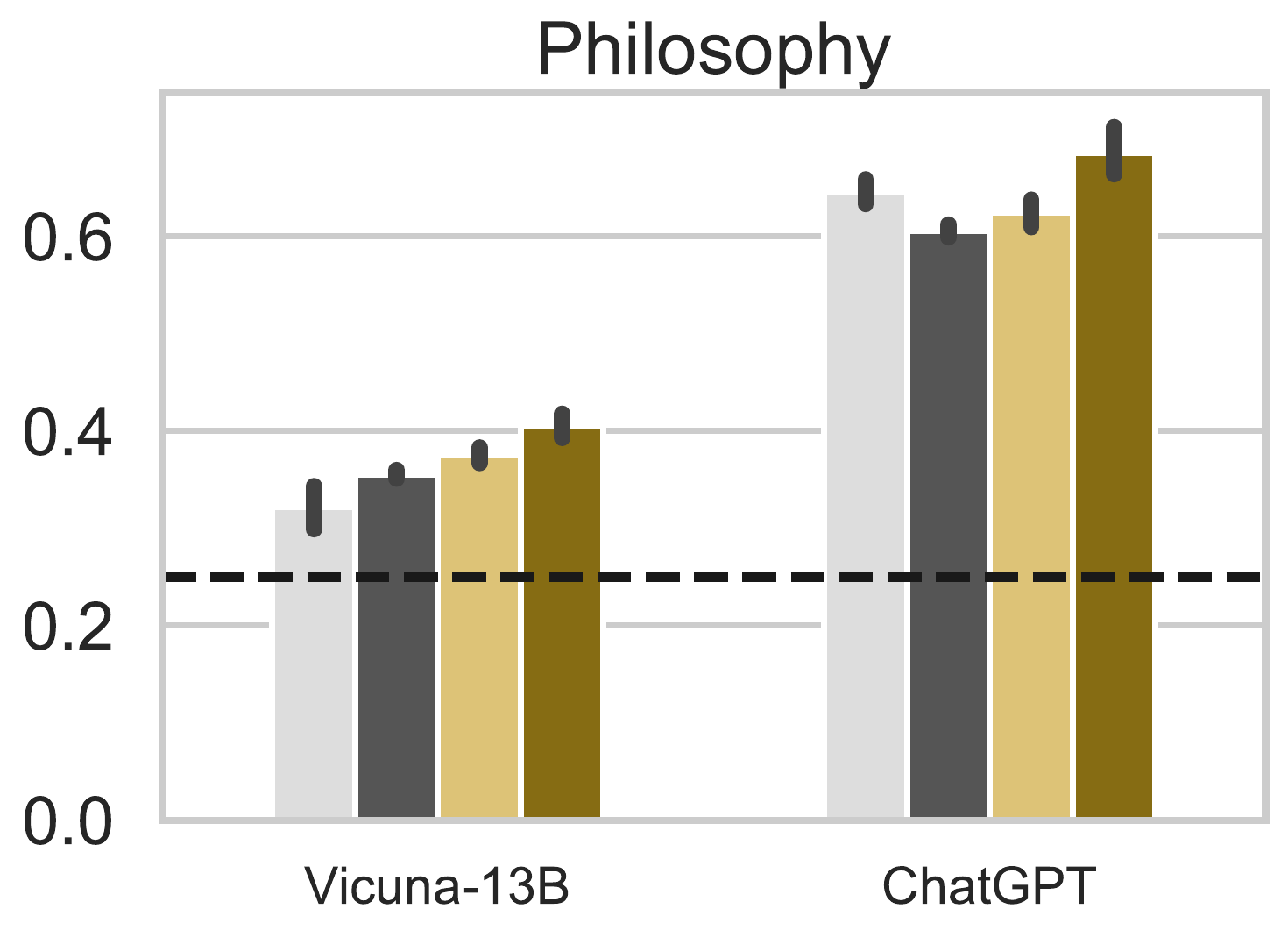}
     \end{subfigure}
     \hfill
     \begin{subfigure}[c]{0.31\textwidth}
         \centering
         \includegraphics[width=\textwidth]{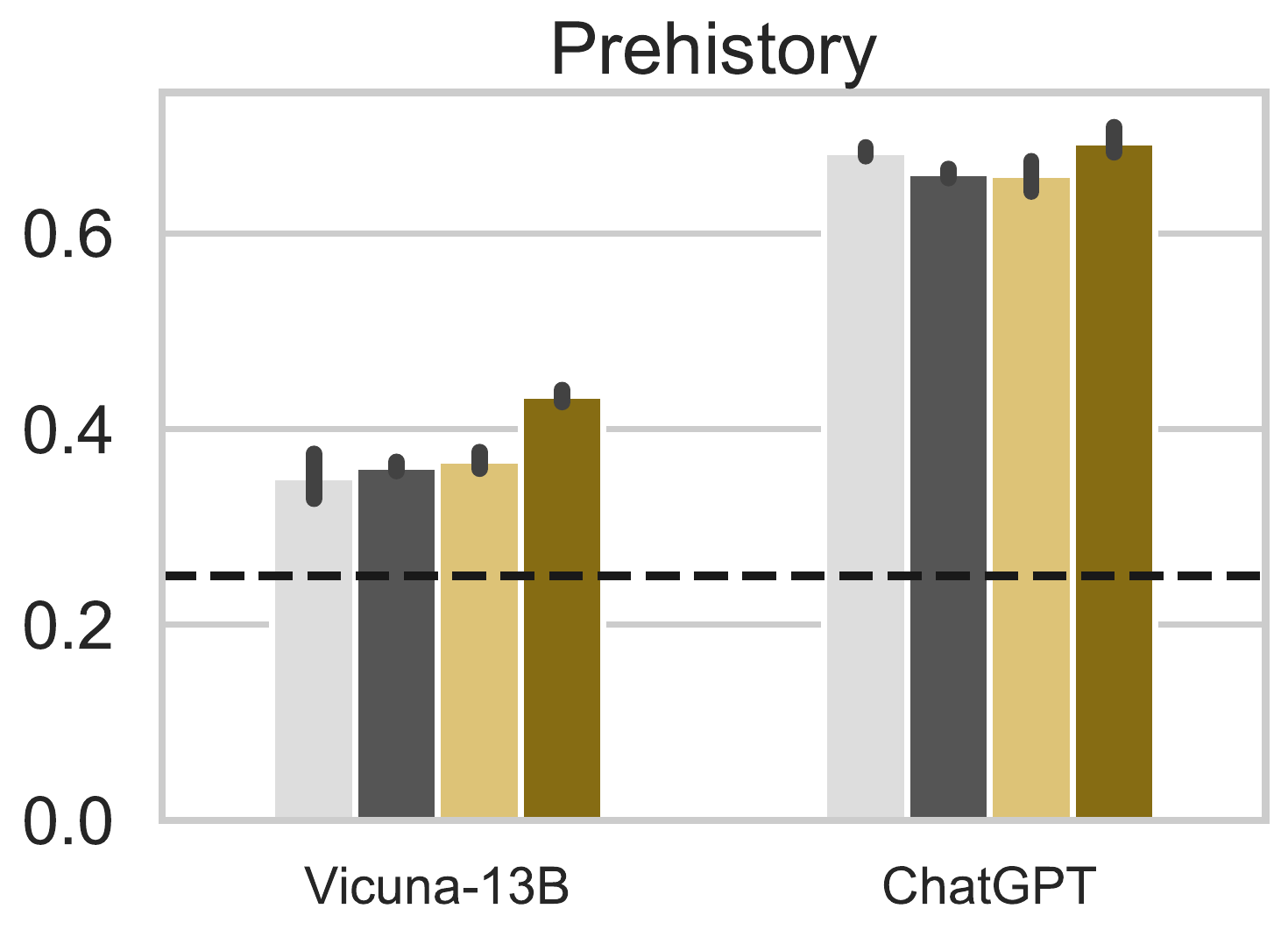}
     \end{subfigure}
     \\
     \begin{subfigure}[c]{0.31\textwidth}
         \centering
         \includegraphics[width=\textwidth]{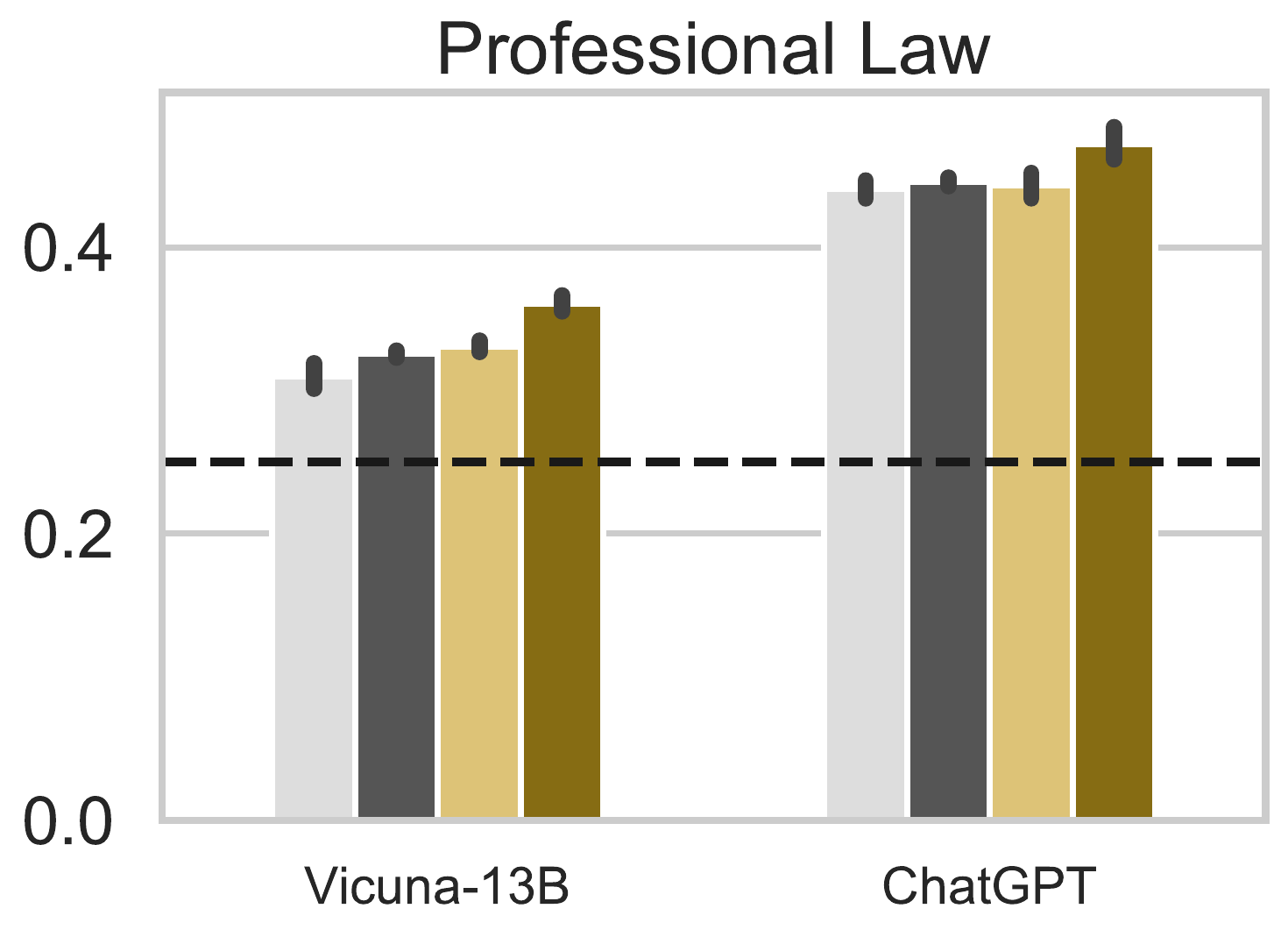}
     \end{subfigure}
     \hfill
     \begin{subfigure}[c]{0.31\textwidth}
         \centering
         \includegraphics[width=\textwidth]{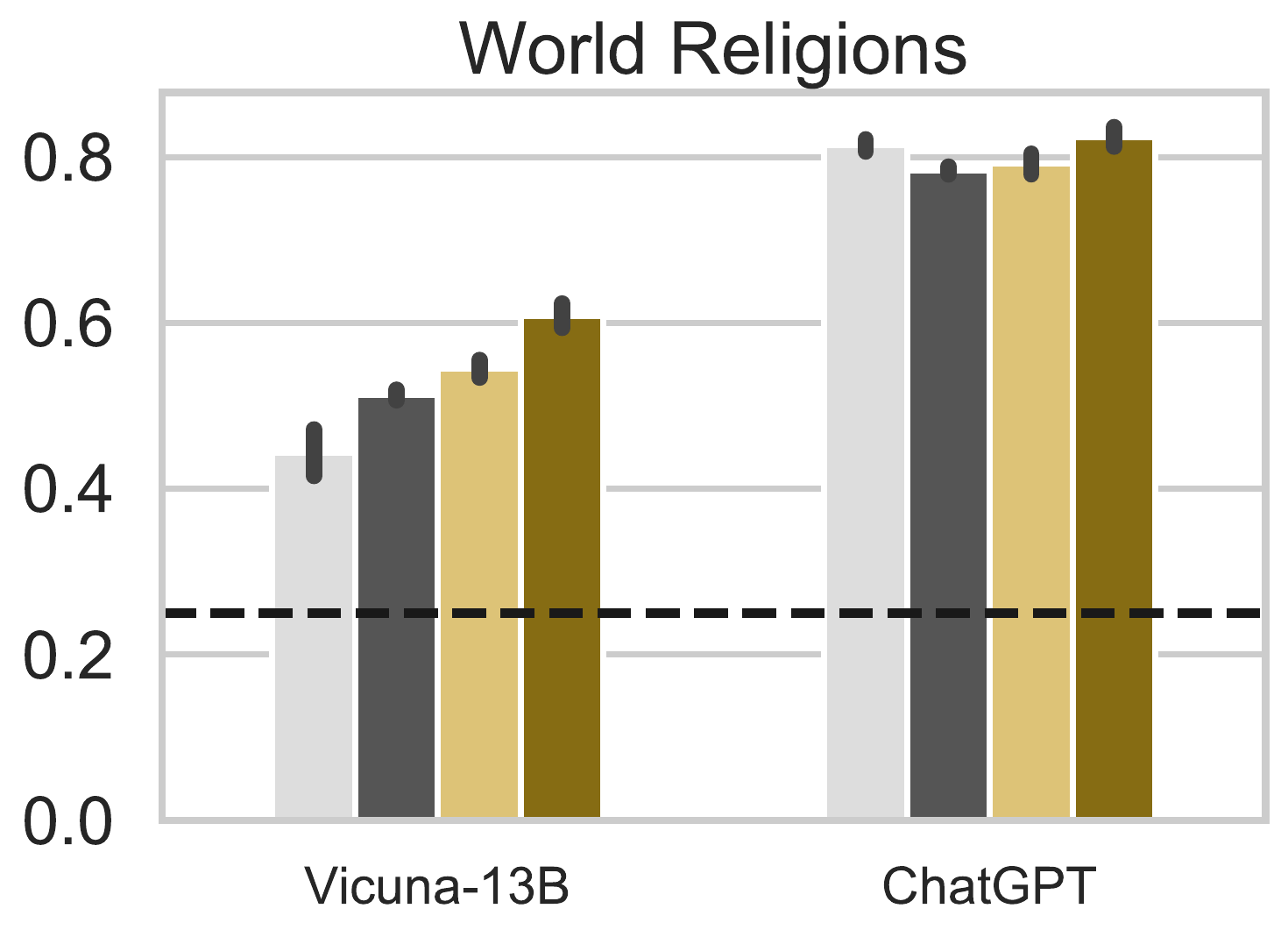}
     \end{subfigure}
     \hfill
     \begin{subfigure}[r]{0.31\textwidth}
         \centering
         \includegraphics[width=.9\textwidth]{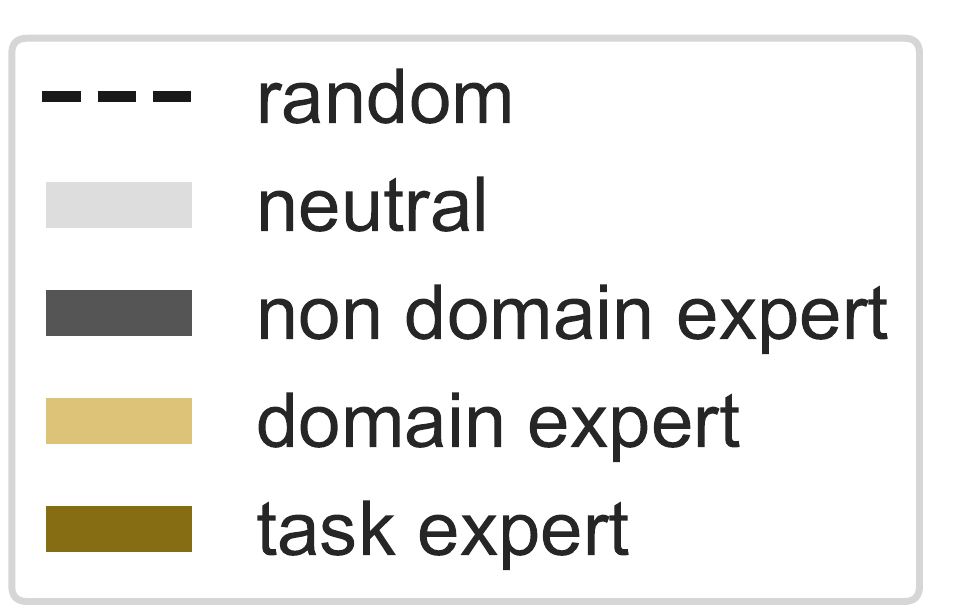}
     \end{subfigure}
     
     \caption{Comparison between Vicuna-13B and ChatGPT for expertise-based impersonation on the Humanities domain of the MMLU reasoning benchmark. We compare the task expert results with the average of all neutral personas, the average of all domain expert personas, the average of all non-domain expert personas and the random baseline (horizontal line). The first plot shows the average over all Humanities tasks, while the remaining plots show the results for each Humanities task individually. All 95\% confidence intervals are computed over the average task accuracy.}%
    \label{fig:mmlu_humanities}
\end{figure}

\begin{figure}[h!]
     \centering
     \begin{subfigure}[c]{0.35\textwidth}
         \centering
         \includegraphics[width=\textwidth]{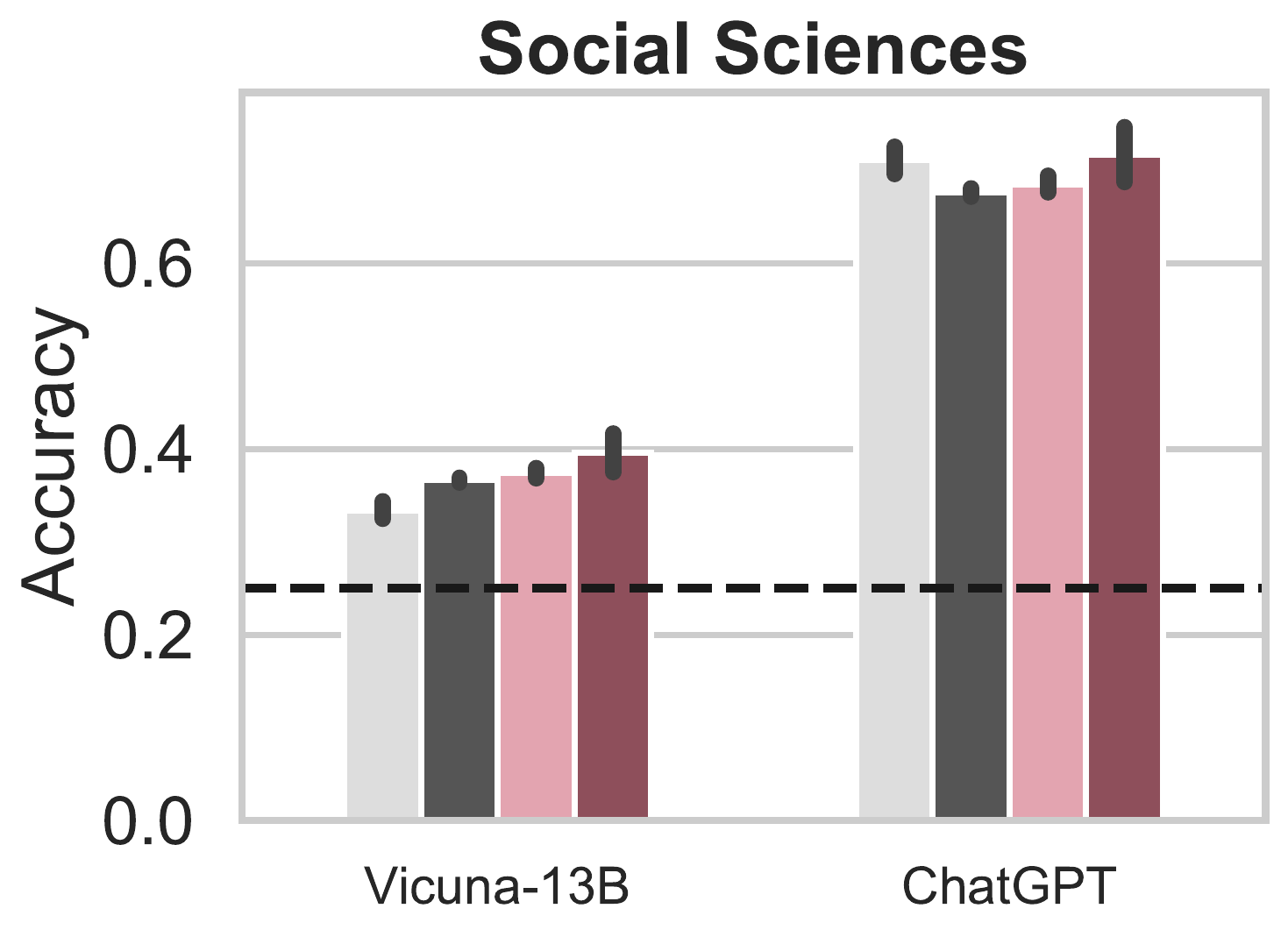}
     \end{subfigure}
     \hfill
     \begin{subfigure}[c]{0.31\textwidth}
         \centering
         \includegraphics[width=\textwidth]{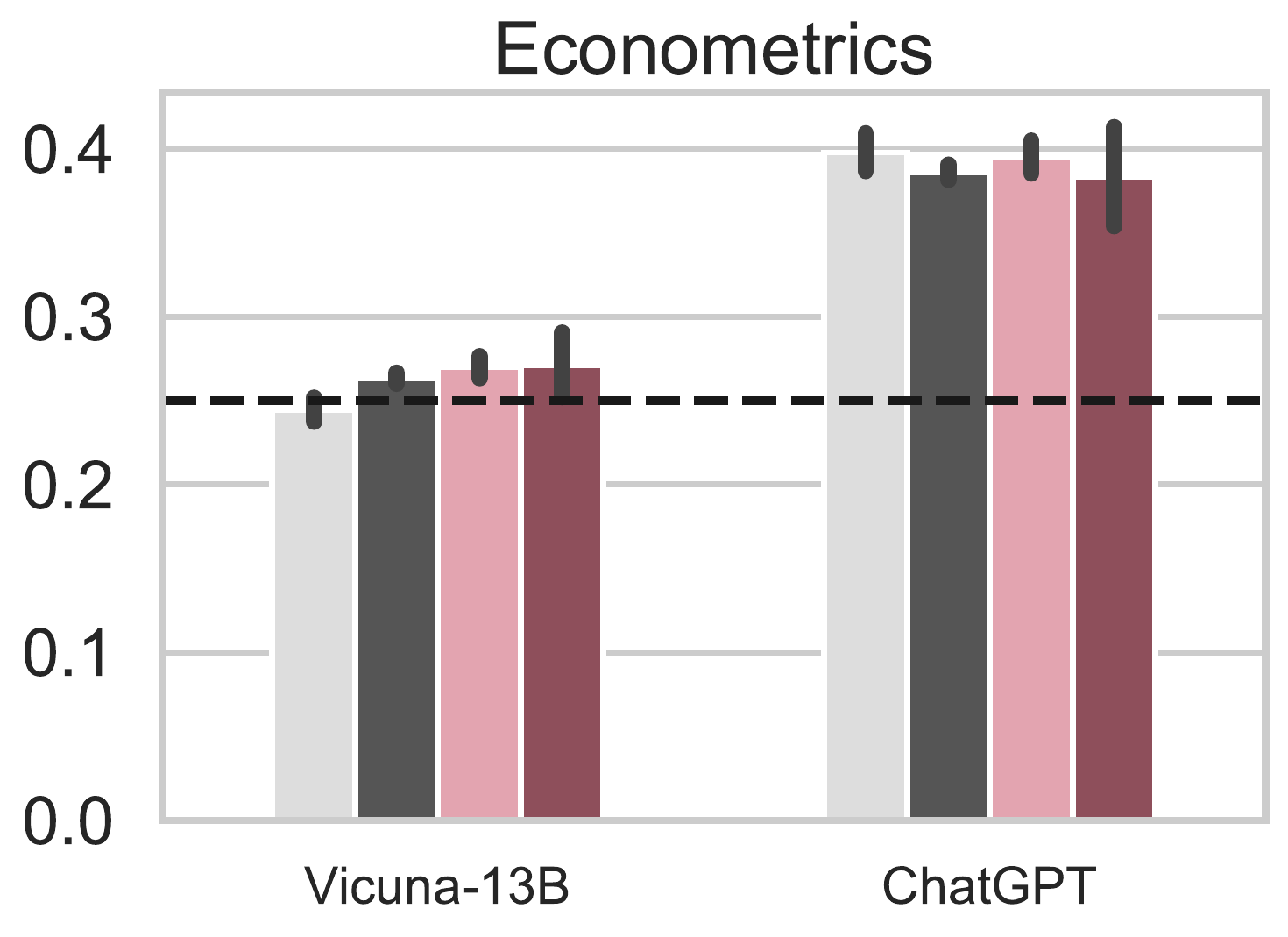}
     \end{subfigure}
     \hfill
     \begin{subfigure}[c]{0.32\textwidth}
         \centering
         \includegraphics[width=\textwidth]{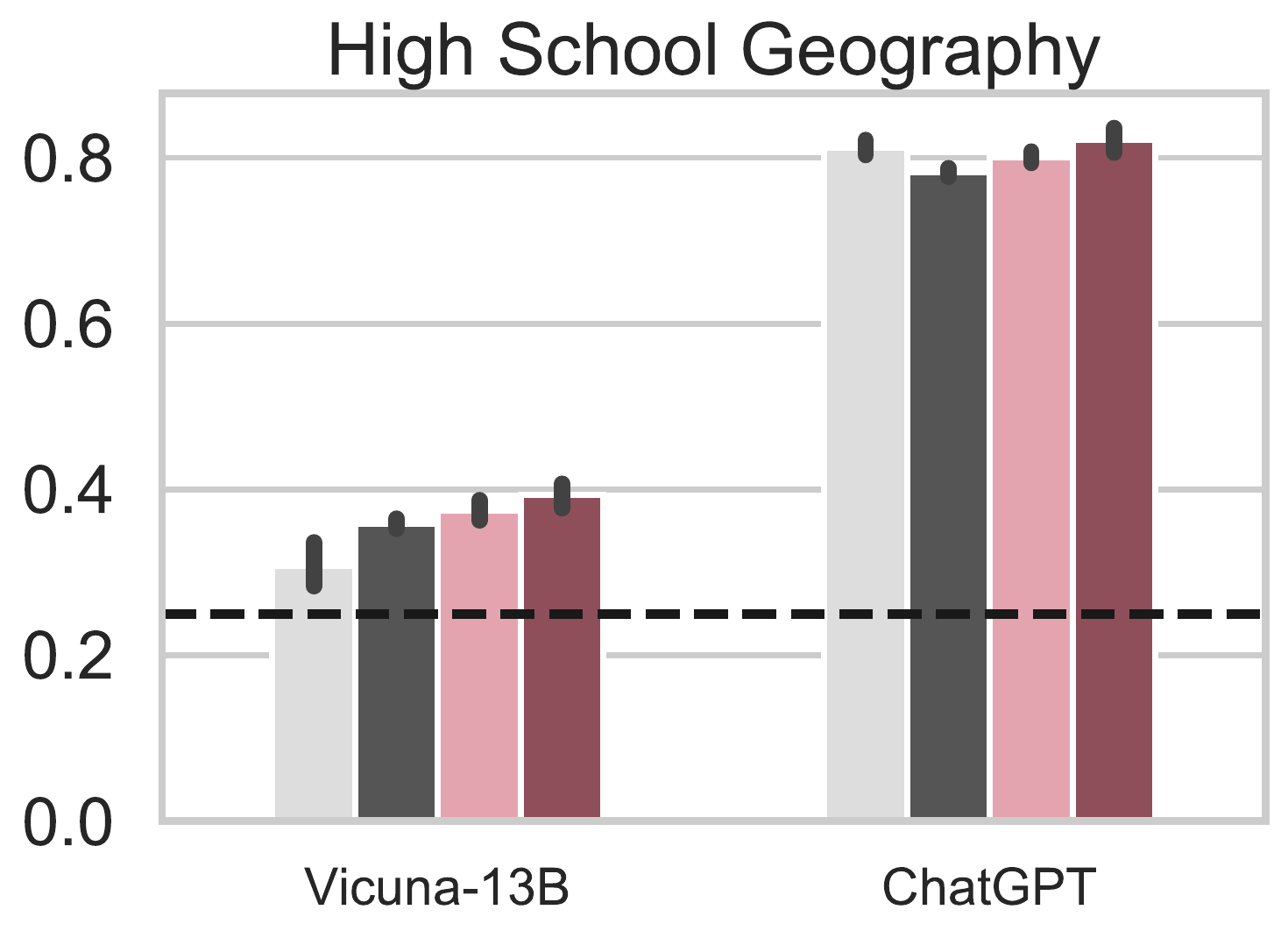}
     \end{subfigure}
     \\
     \begin{subfigure}[c]{0.31\textwidth}
         \centering
         \includegraphics[width=\textwidth]{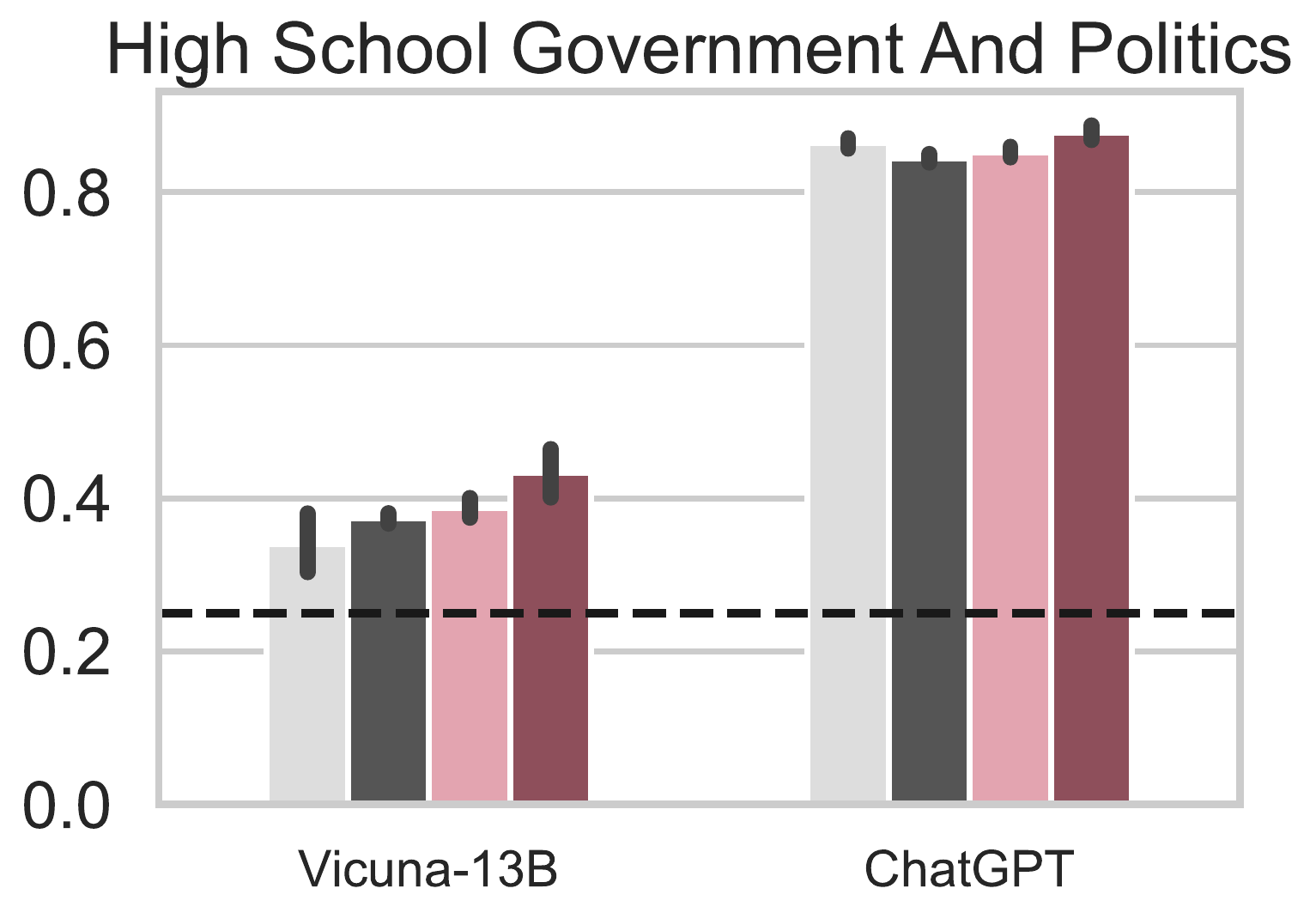}
     \end{subfigure}
     \hfill
     \begin{subfigure}[c]{0.31\textwidth}
         \centering
         \includegraphics[width=\textwidth]{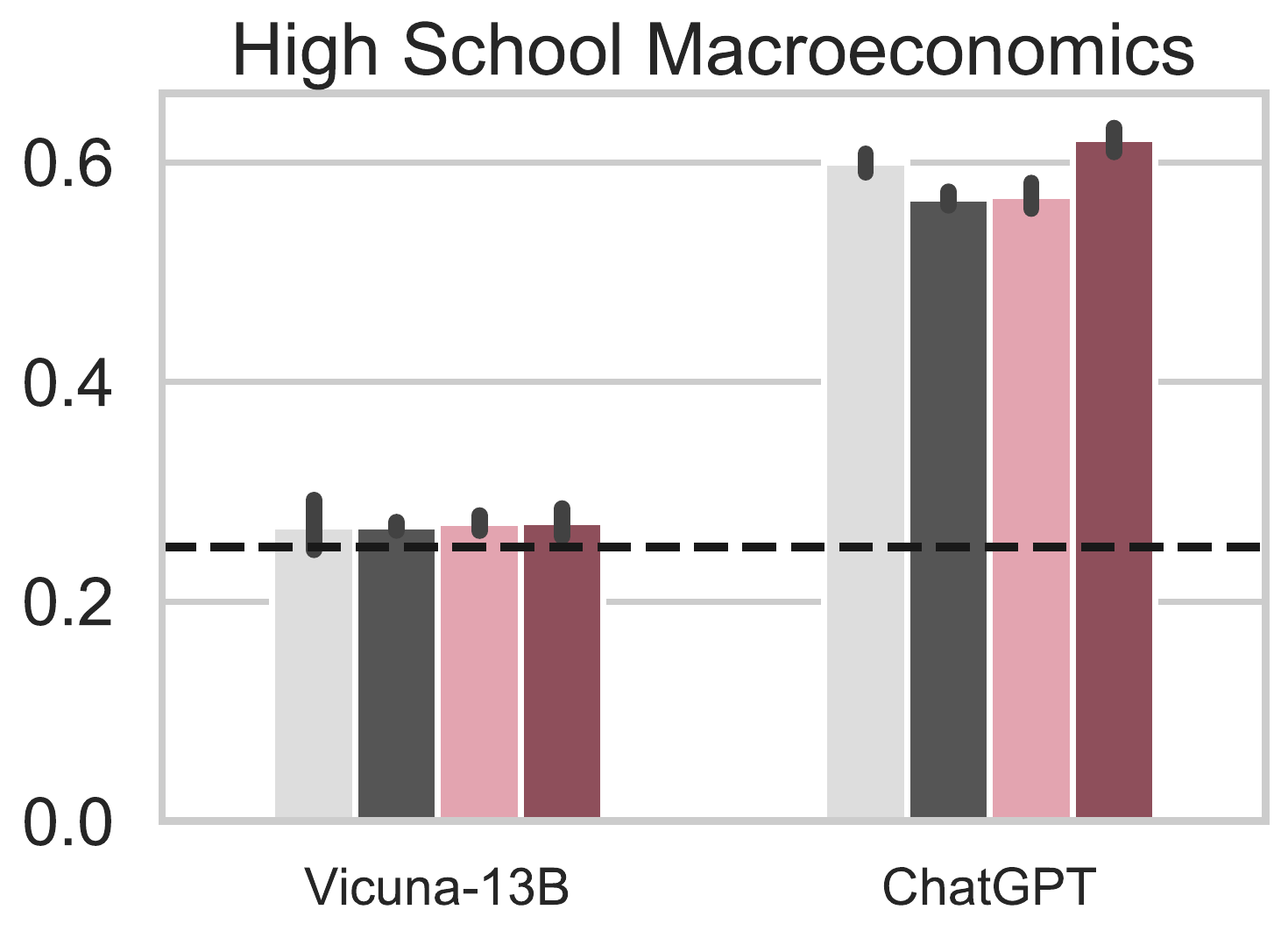}
     \end{subfigure}
     \hfill
     \begin{subfigure}[c]{0.31\textwidth}
         \centering
         \includegraphics[width=\textwidth]{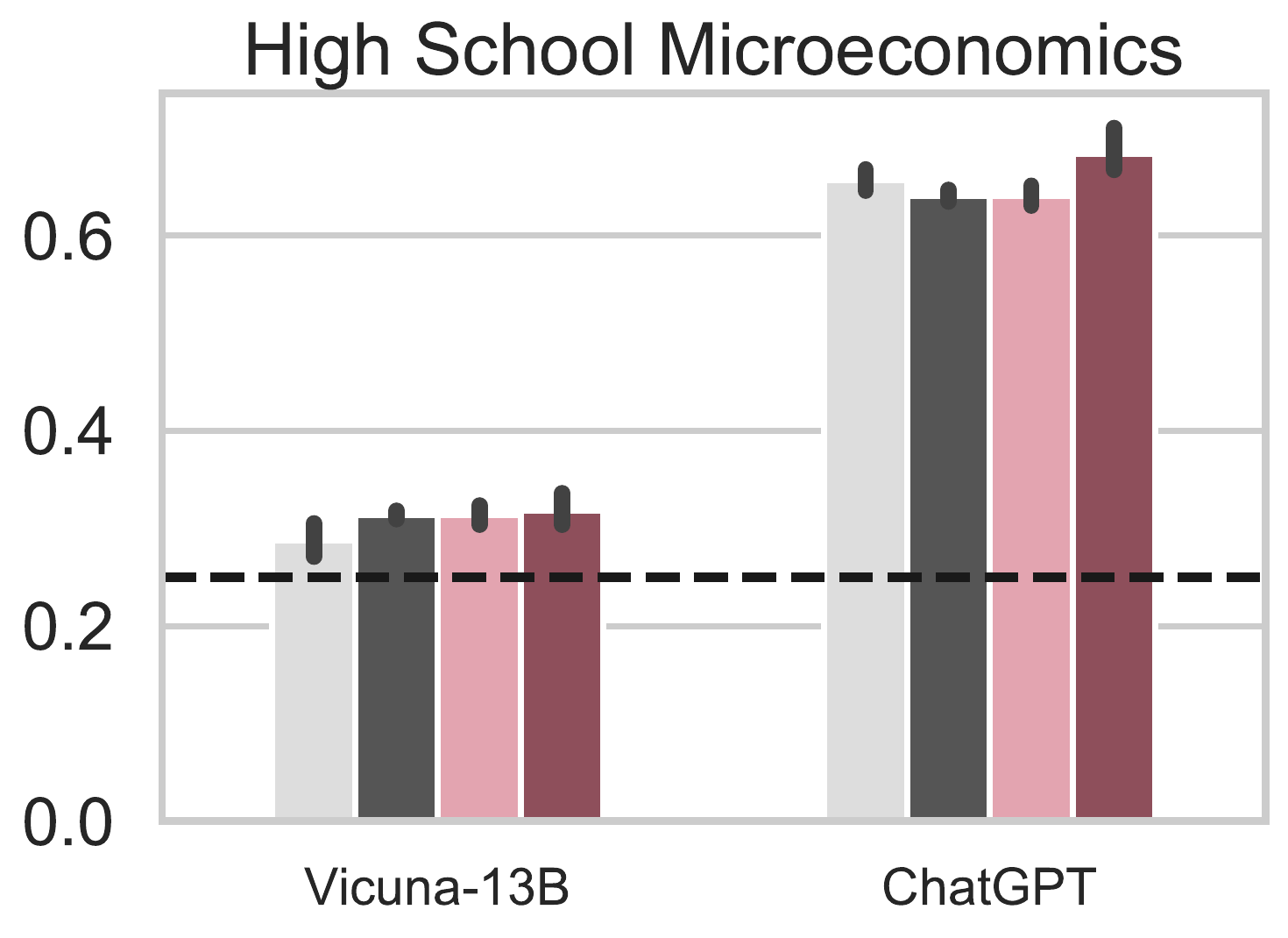}
     \end{subfigure}
     \\
     \begin{subfigure}[c]{0.31\textwidth}
         \centering
         \includegraphics[width=\textwidth]{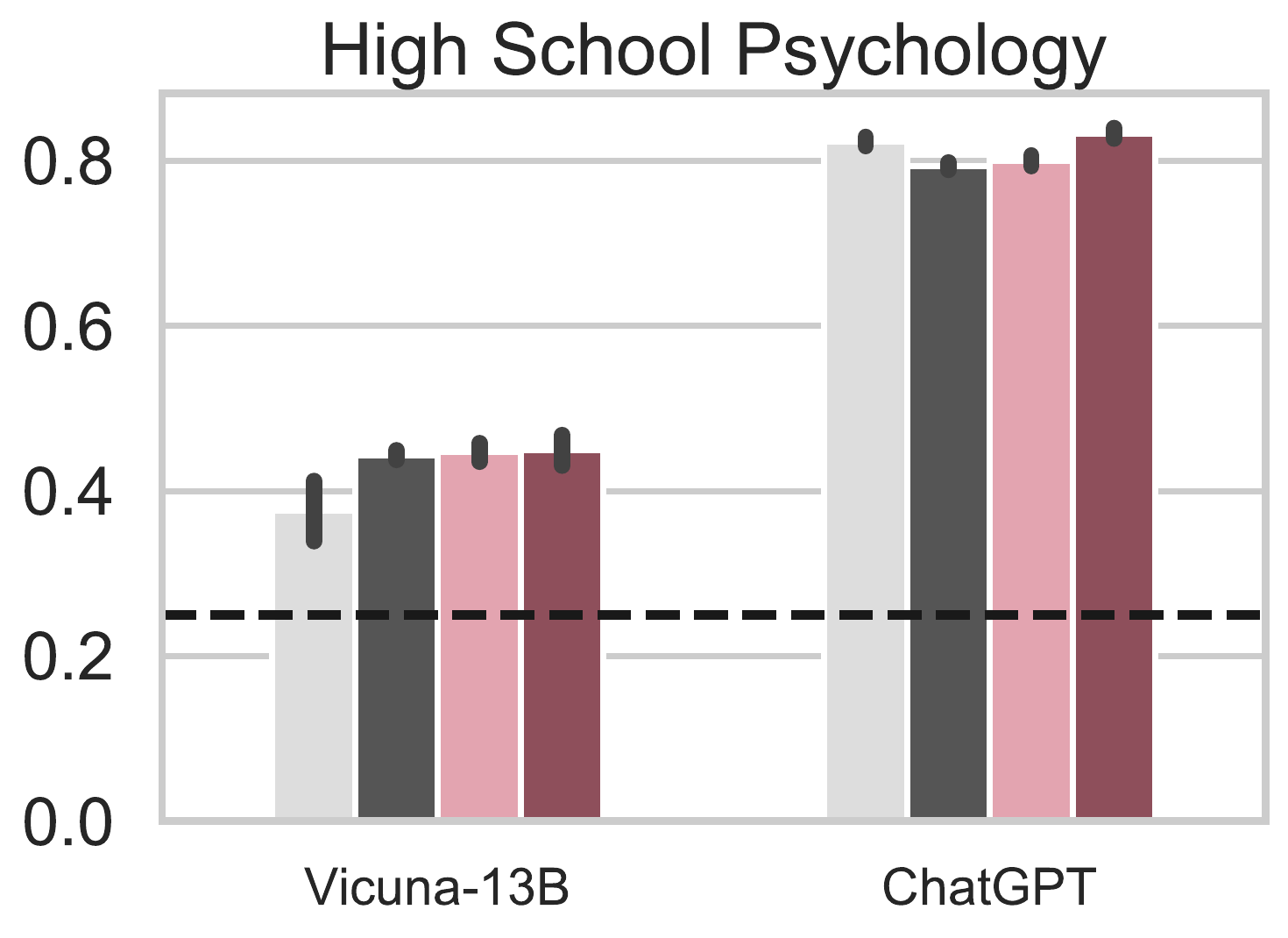}
     \end{subfigure}
      \hfill
     \begin{subfigure}[c]{0.31\textwidth}
         \centering
         \includegraphics[width=\textwidth]{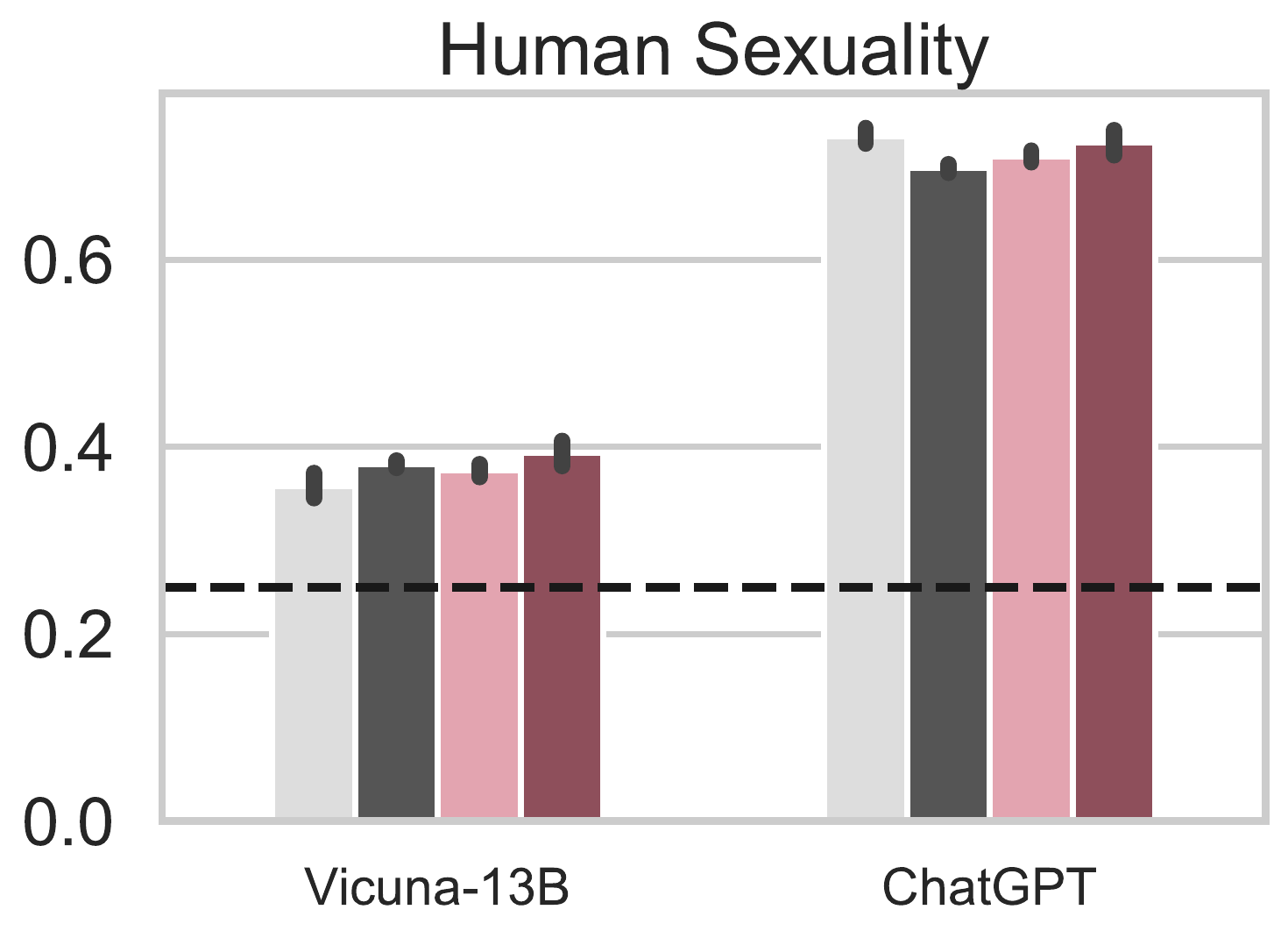}
     \end{subfigure}
     \hfill
     \begin{subfigure}[c]{0.31\textwidth}
         \centering
         \includegraphics[width=\textwidth]{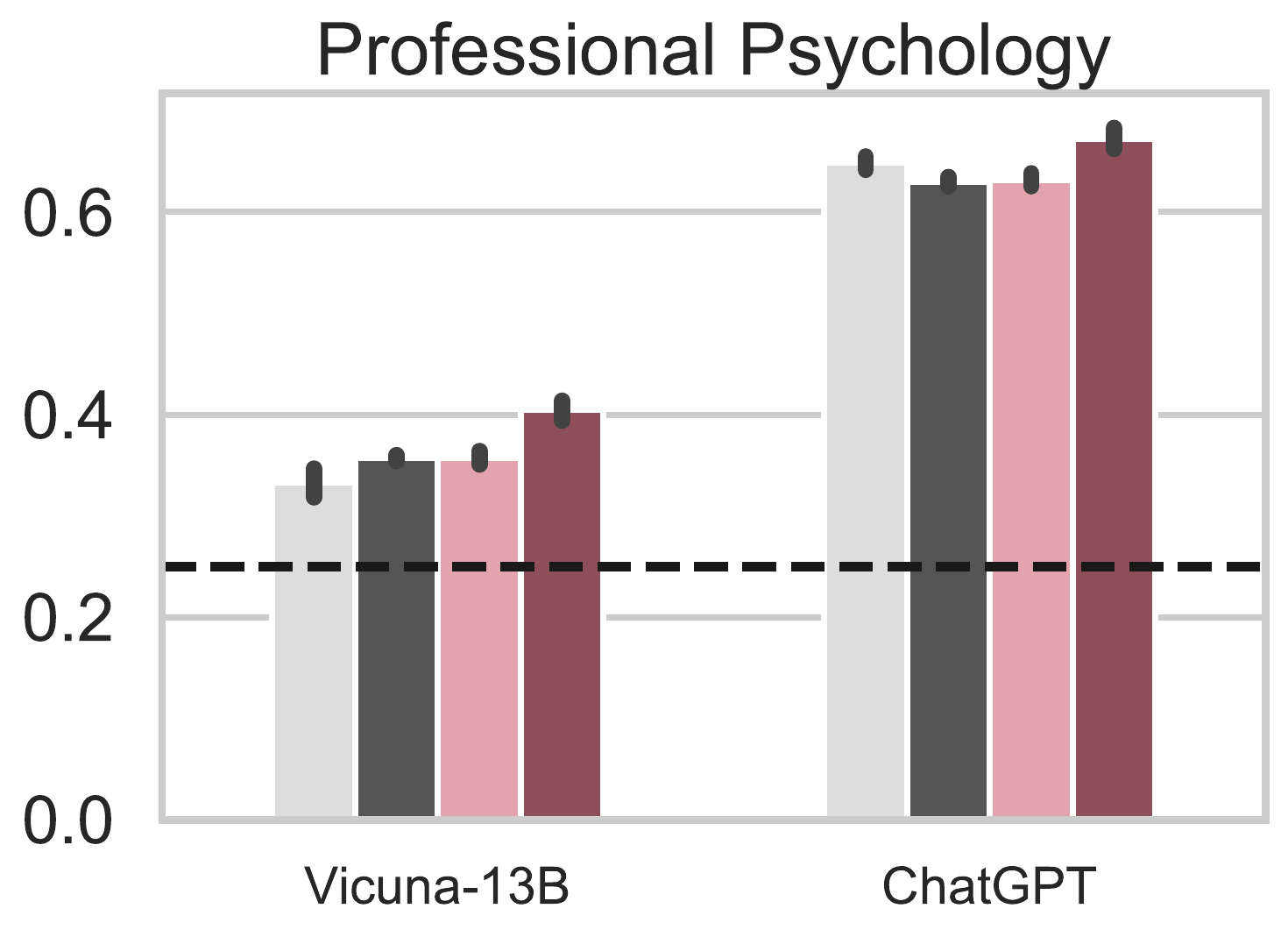}
     \end{subfigure}
     \\
      \begin{subfigure}[c]{0.31\textwidth}
         \centering
         \includegraphics[width=\textwidth]{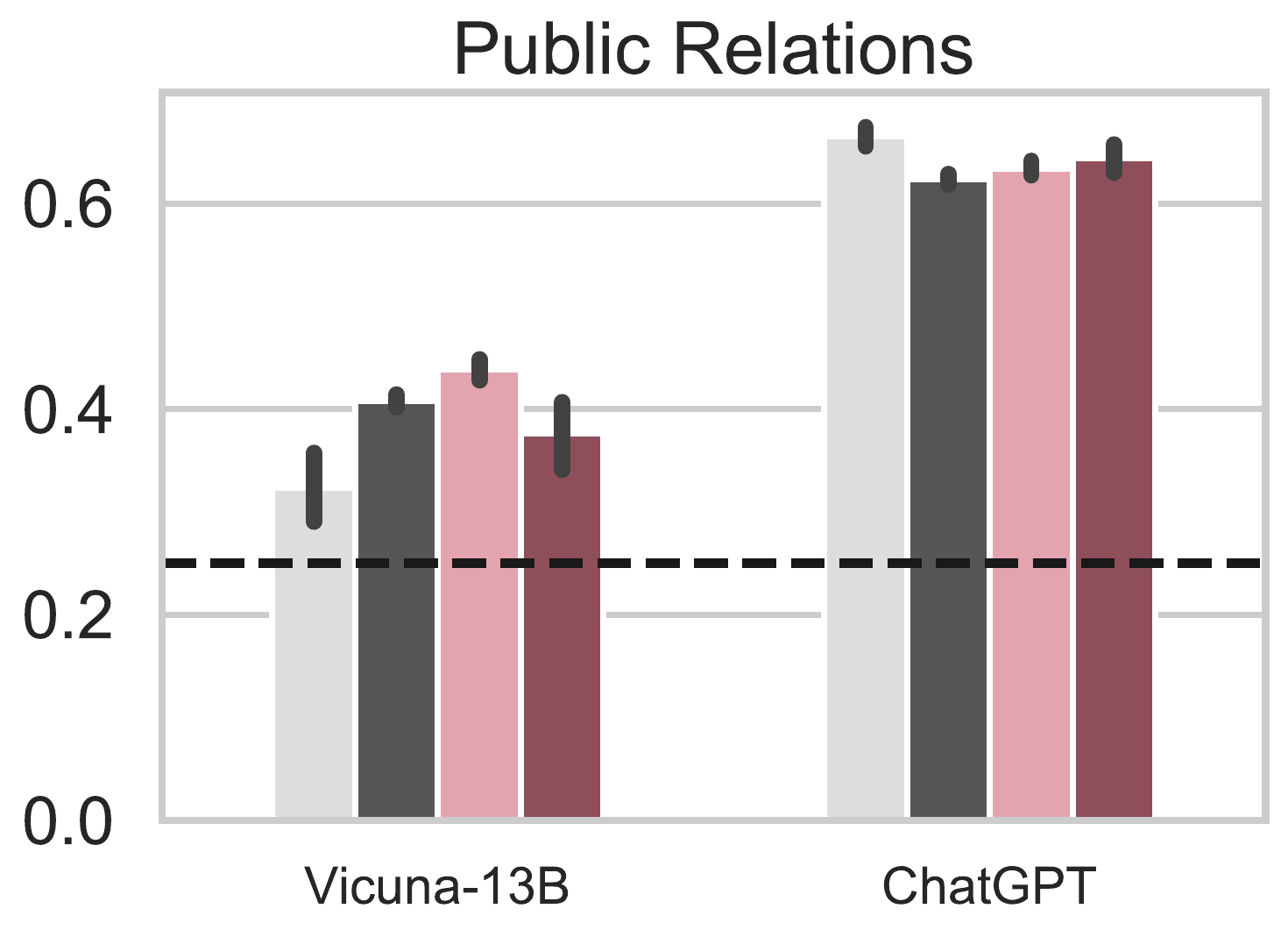}
     \end{subfigure}
      \hfill
     \begin{subfigure}[c]{0.31\textwidth}
         \centering
         \includegraphics[width=\textwidth]{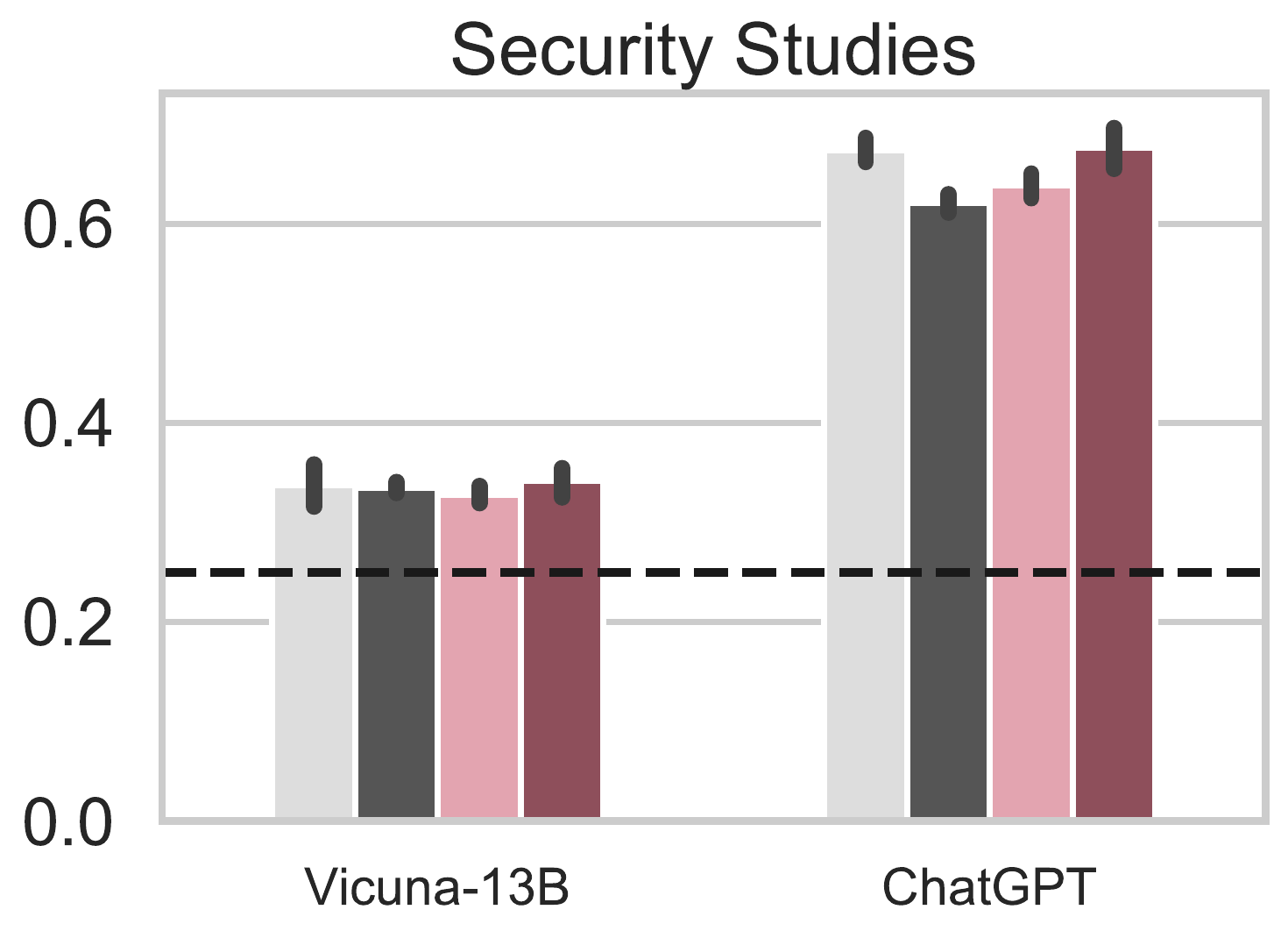}
     \end{subfigure}
     \hfill
     \begin{subfigure}[c]{0.31\textwidth}
         \centering
         \includegraphics[width=\textwidth]{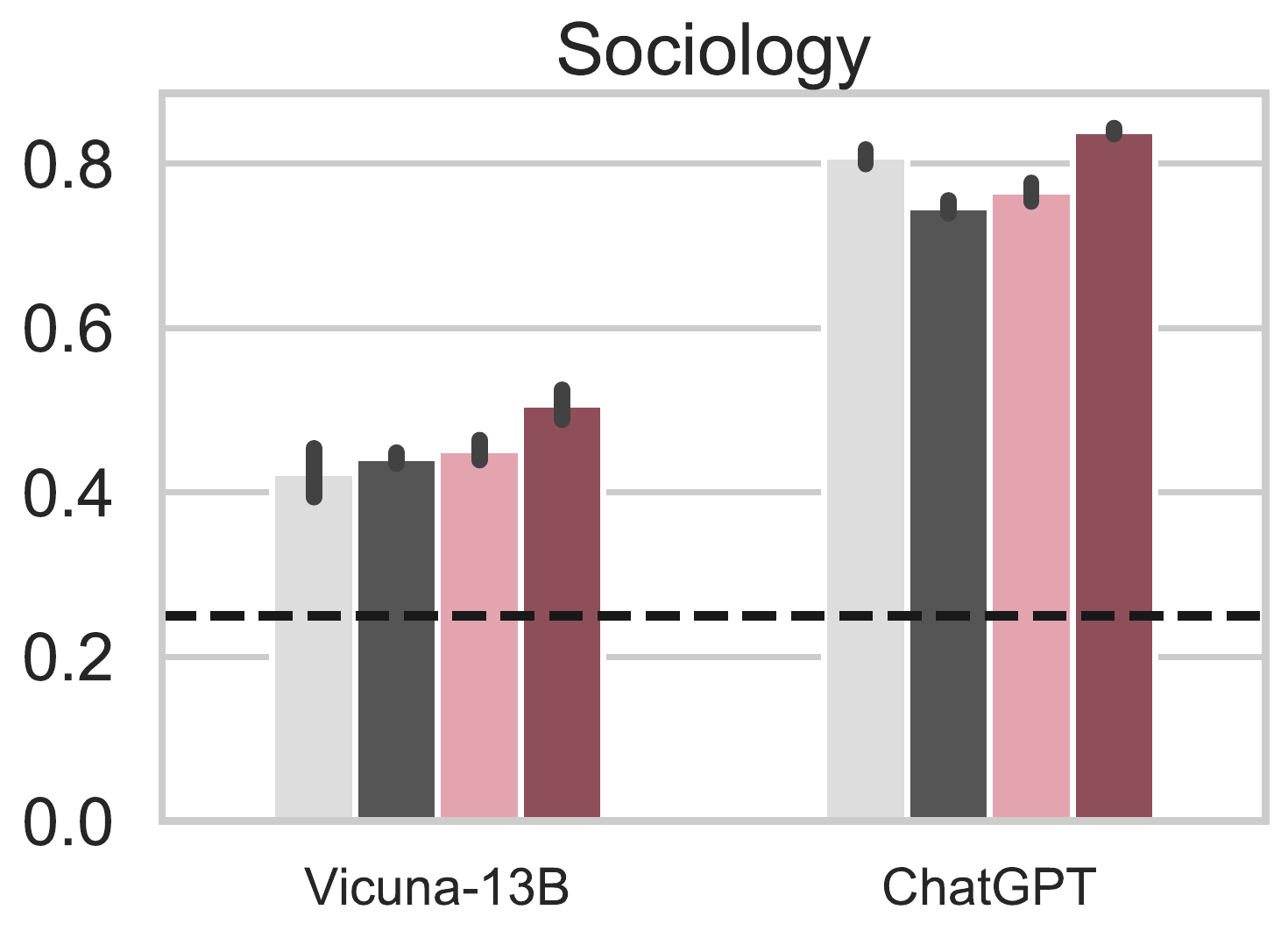}
     \end{subfigure}
     \\
     \begin{subfigure}[c]{0.31\textwidth}
         \centering
         \includegraphics[width=\textwidth]{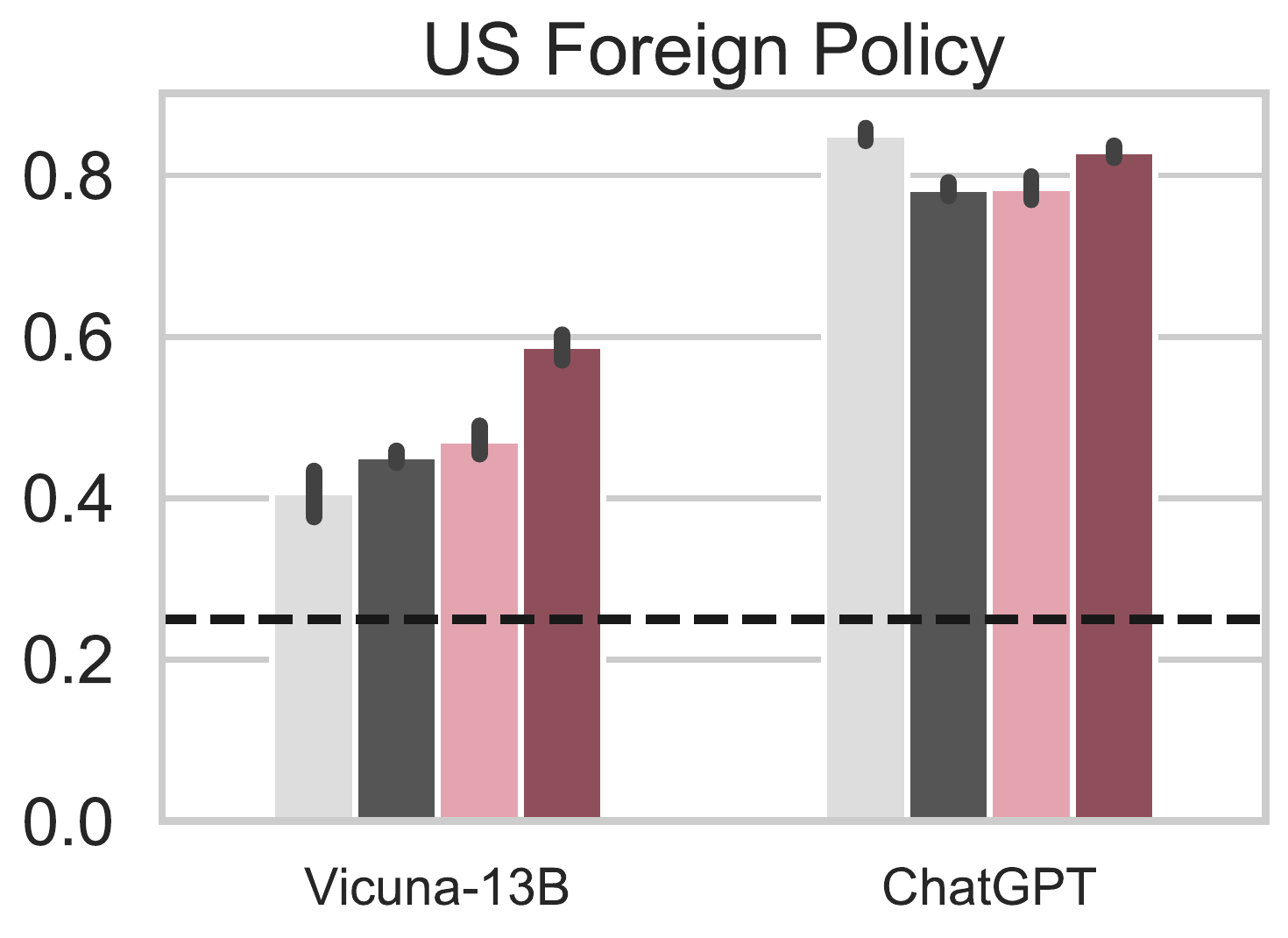}
     \end{subfigure}
     \hfill
     \begin{subfigure}[c]{0.31\textwidth}
        \centering
    \end{subfigure}
    \hfill
     \begin{subfigure}[c]{0.31\textwidth}
         \centering
         \includegraphics[width=.9\textwidth]{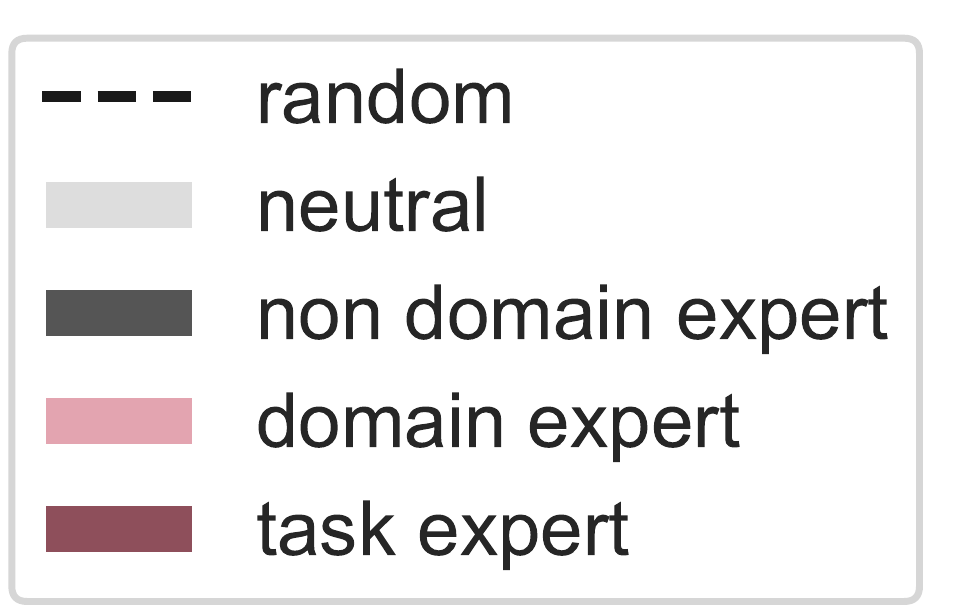}
     \end{subfigure}
     
     \caption{Comparison between Vicuna-13B and ChatGPT for expertise-based impersonation on the Social Sciences domain of the MMLU reasoning benchmark. We compare the task expert results with the average of all neutral personas, the average of all domain expert personas, the average of all non-domain expert personas and the random baseline (horizontal line). The first plot shows the average over all Social Sciences tasks, while the remaining plots show the results for each Social Sciences task individually. All 95\% confidence intervals are computed over the average task accuracy.}%
    \label{fig:mmlu_social_sciences}
\end{figure}

\begin{figure}[h!]
     \centering
     \begin{subfigure}[c]{0.35\textwidth}
         \centering
         \includegraphics[width=\textwidth]{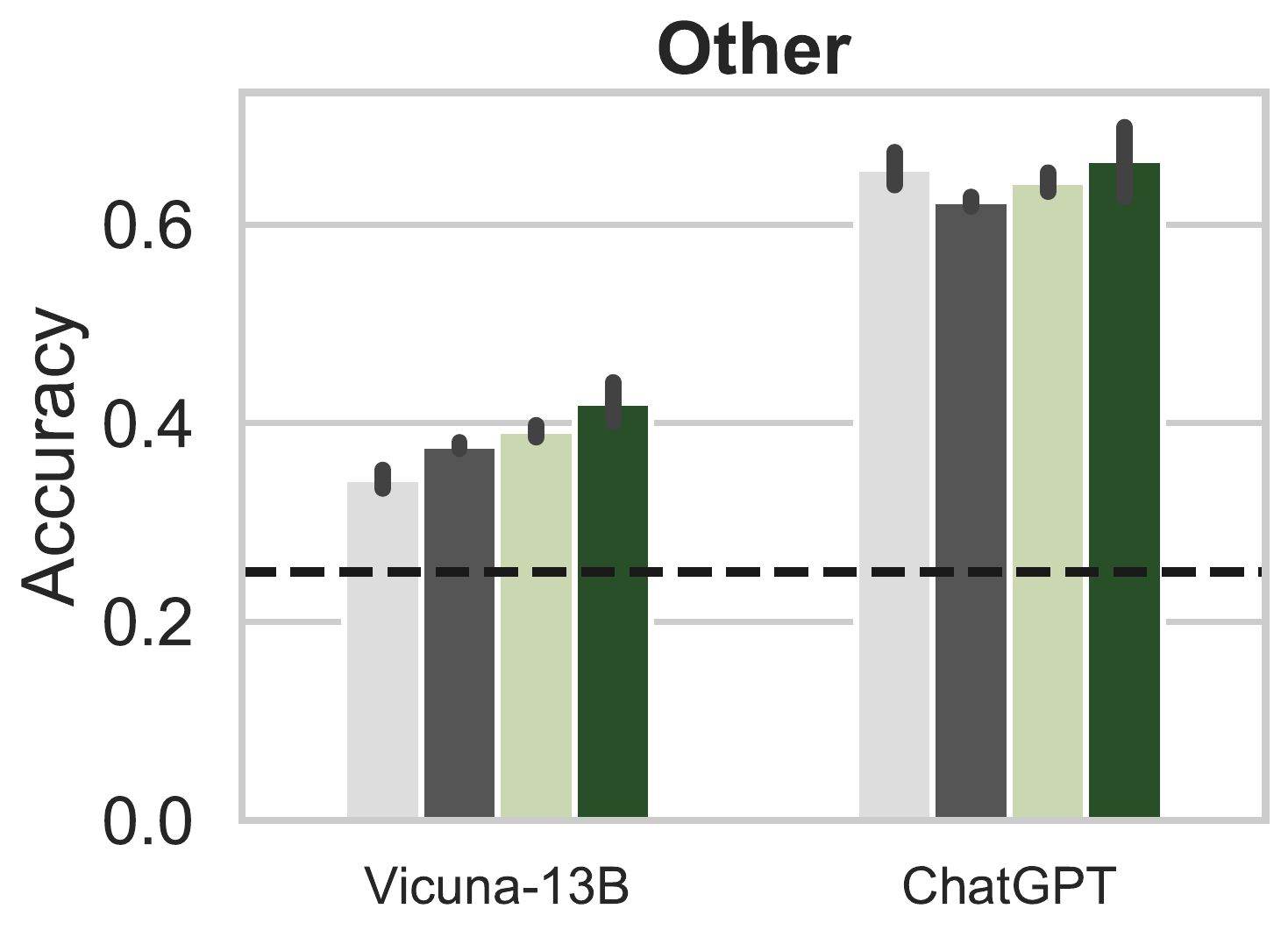}
     \end{subfigure}
     \hfill
     \begin{subfigure}[c]{0.31\textwidth}
         \centering
         \includegraphics[width=\textwidth]{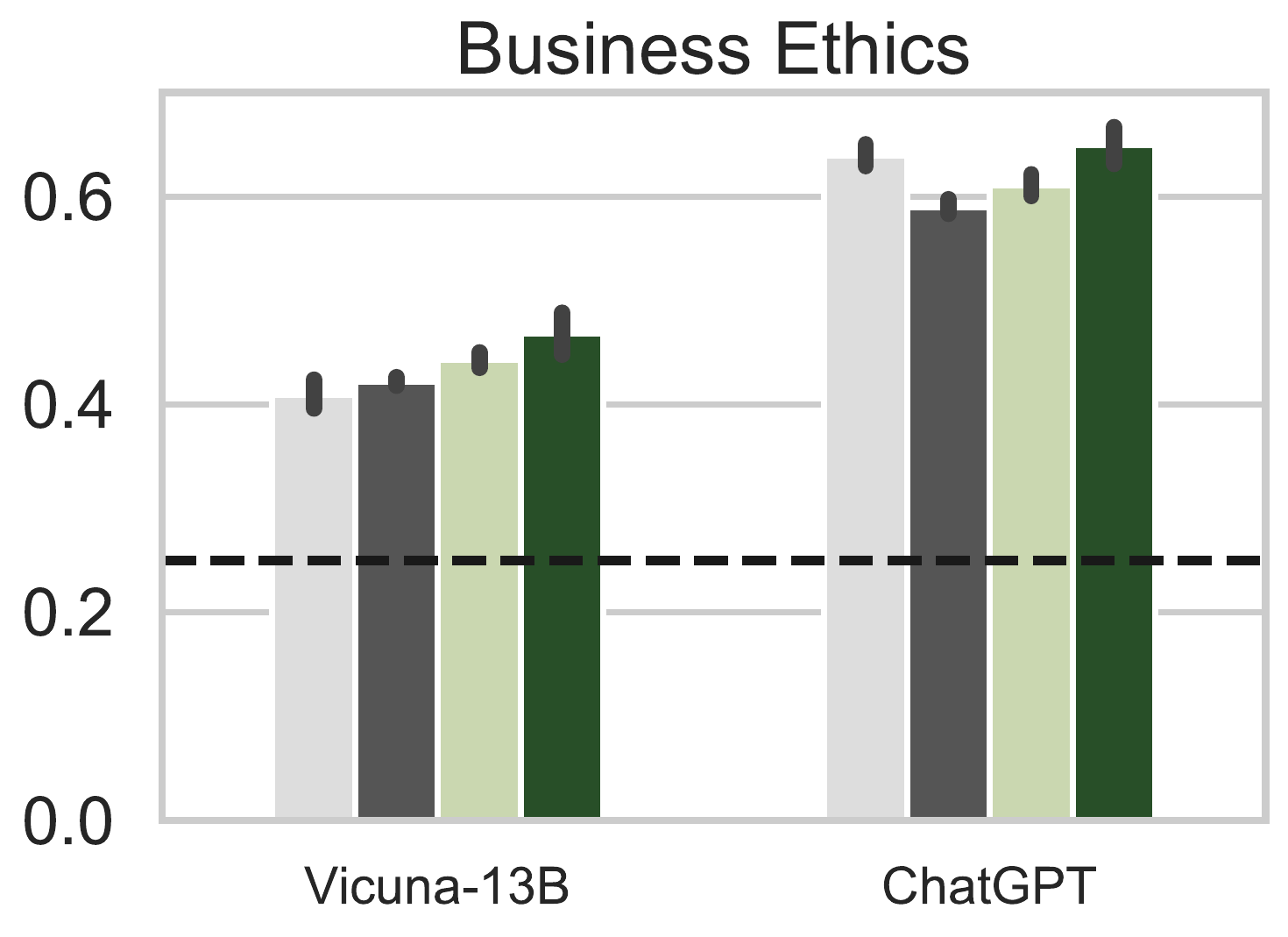}
     \end{subfigure}
     \hfill
     \begin{subfigure}[c]{0.32\textwidth}
         \centering
         \includegraphics[width=\textwidth]{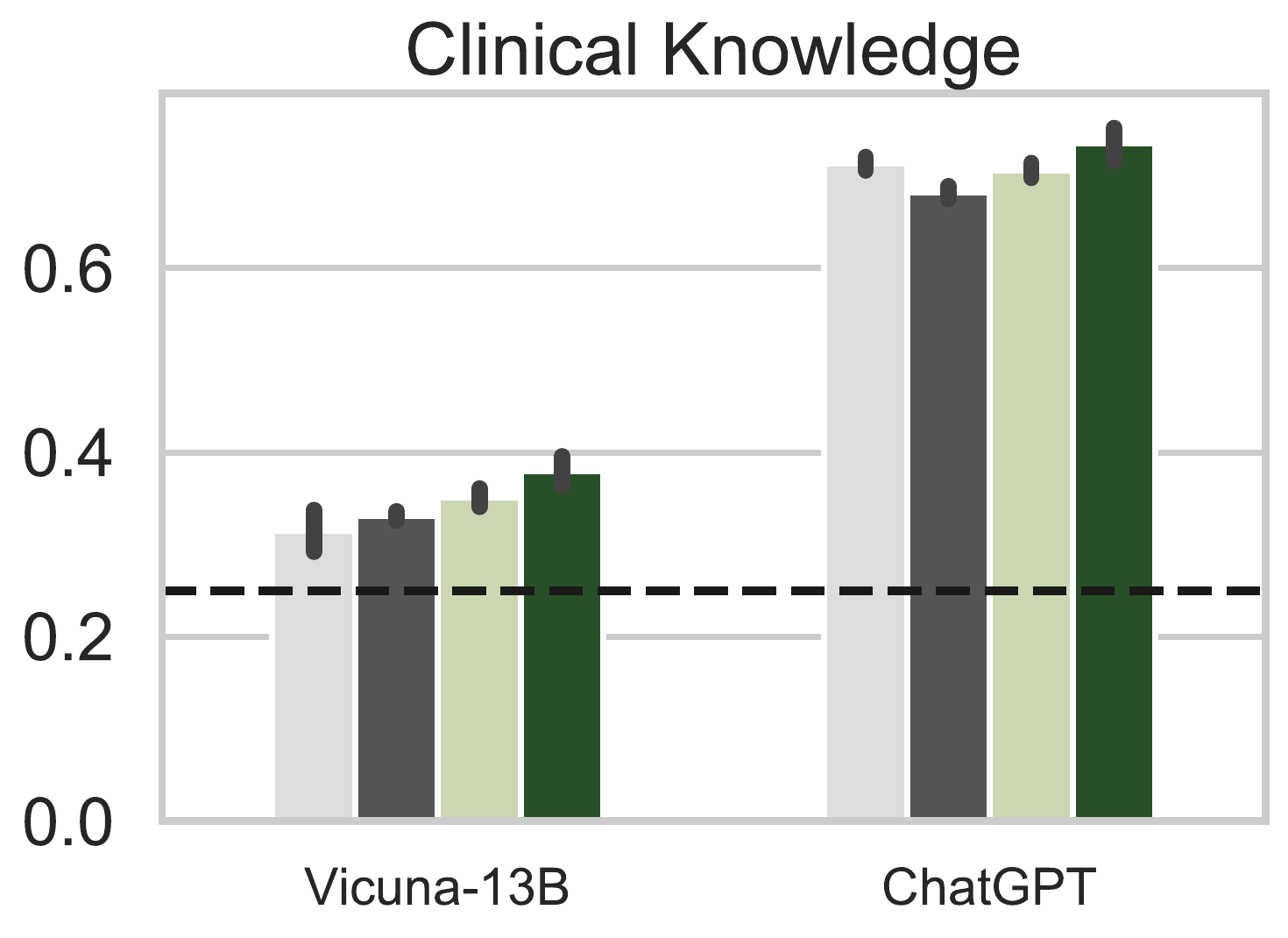}
     \end{subfigure}
     \\
     \begin{subfigure}[c]{0.31\textwidth}
         \centering
         \includegraphics[width=\textwidth]{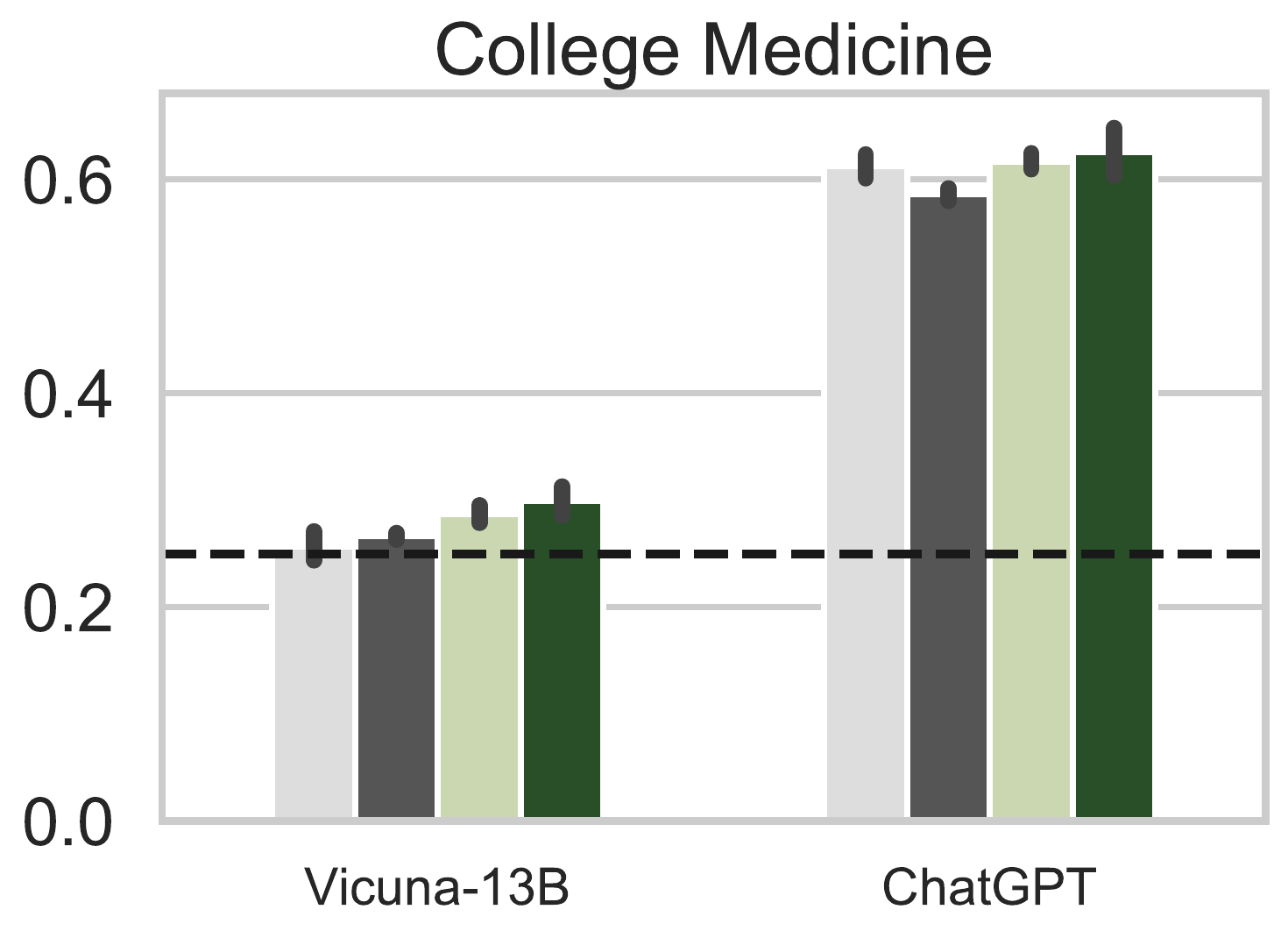}
     \end{subfigure}
     \hfill
     \begin{subfigure}[c]{0.31\textwidth}
         \centering
         \includegraphics[width=\textwidth]{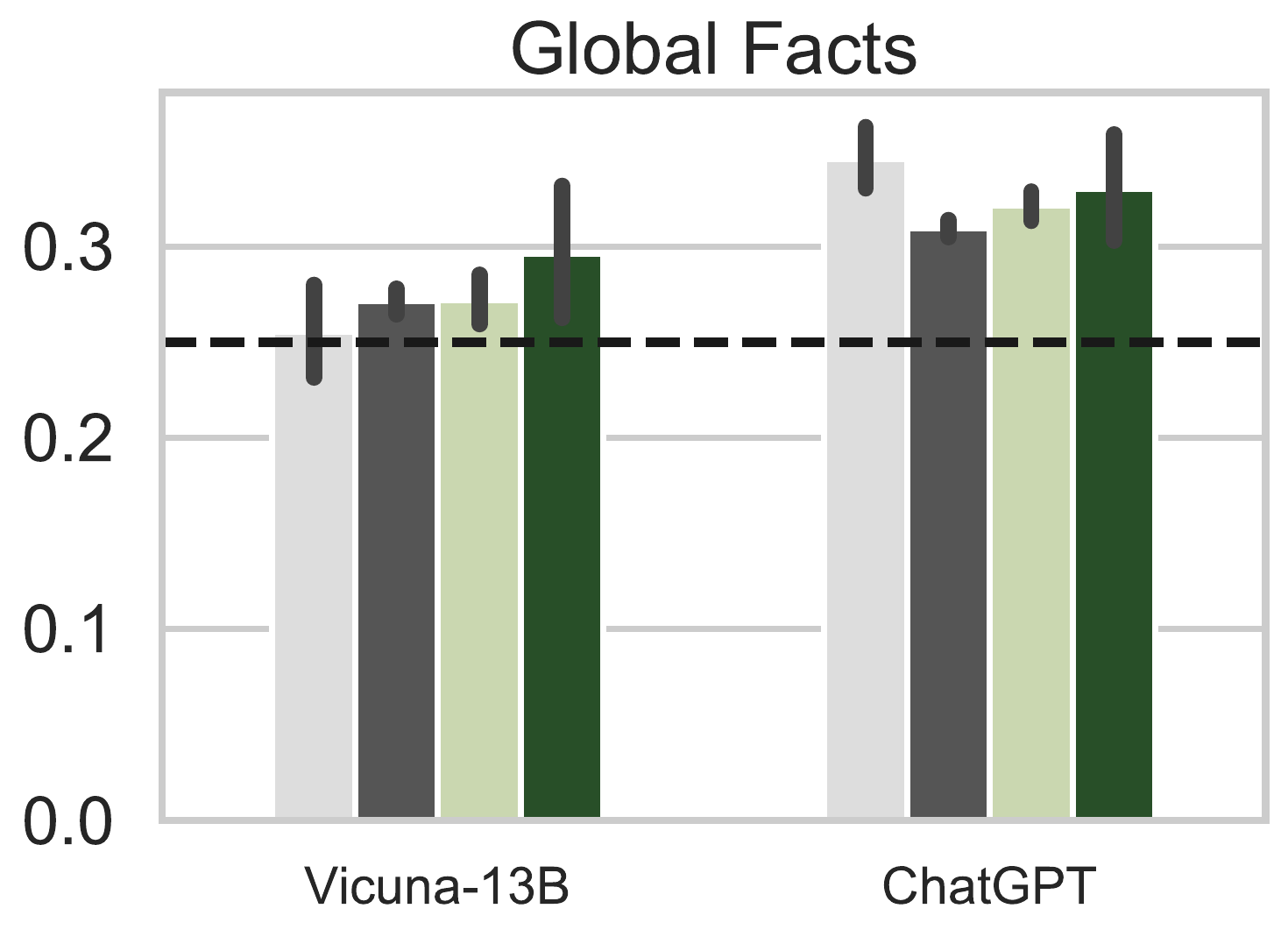}
     \end{subfigure}
     \hfill
     \begin{subfigure}[c]{0.31\textwidth}
         \centering
         \includegraphics[width=\textwidth]{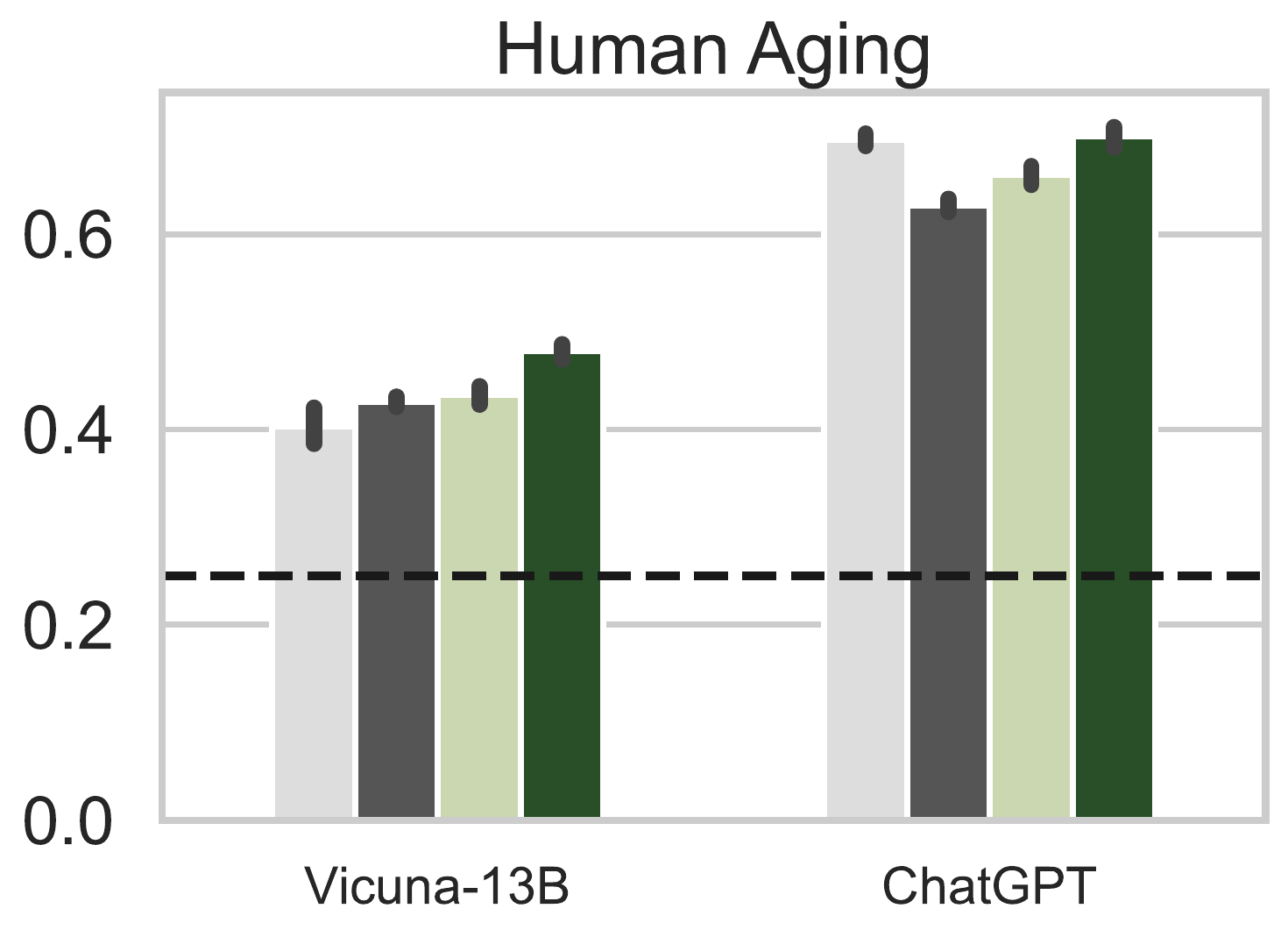}
     \end{subfigure}
     \\
     \begin{subfigure}[c]{0.31\textwidth}
         \centering
         \includegraphics[width=\textwidth]{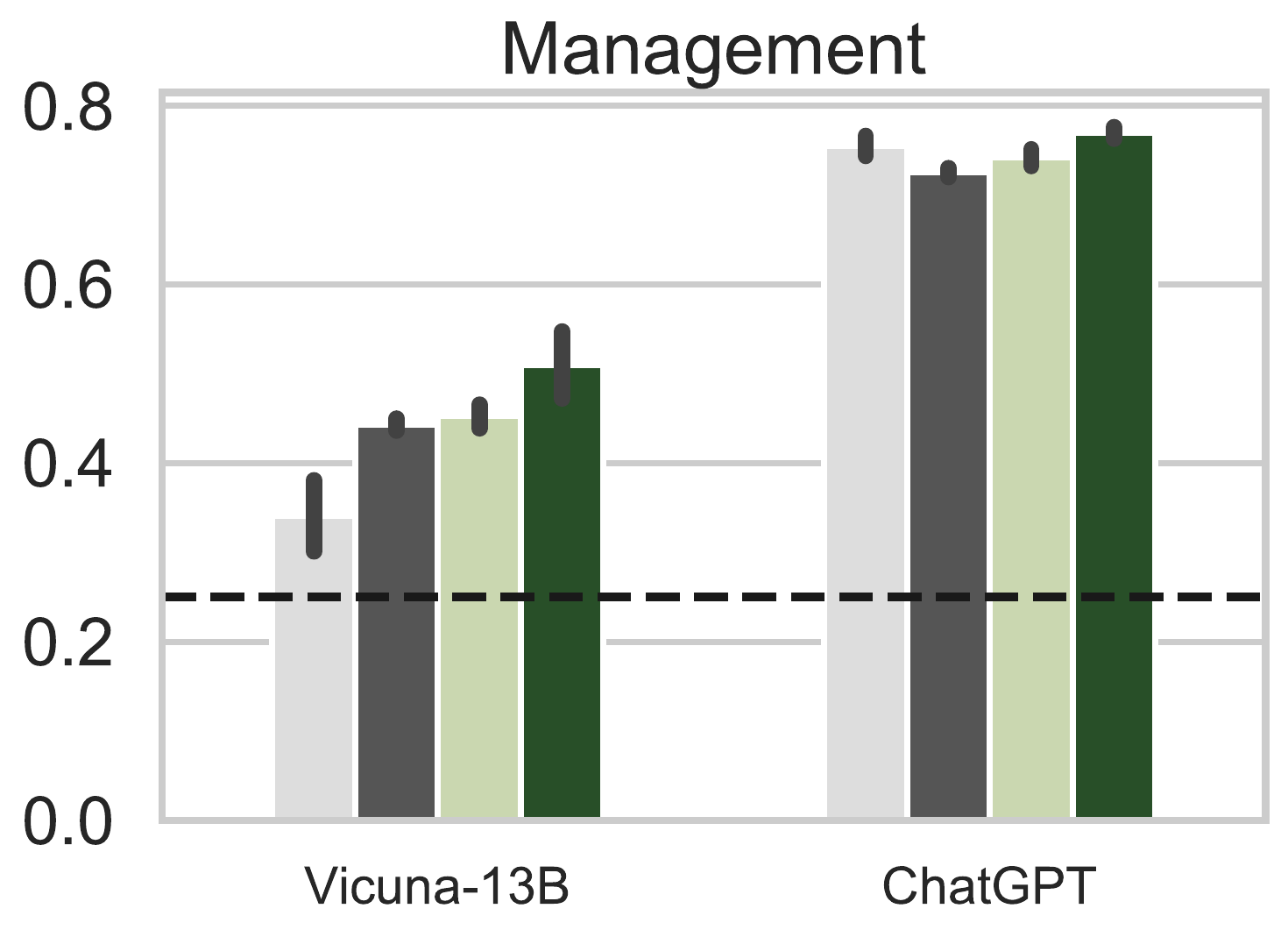}
     \end{subfigure}
      \hfill
     \begin{subfigure}[c]{0.31\textwidth}
         \centering
         \includegraphics[width=\textwidth]{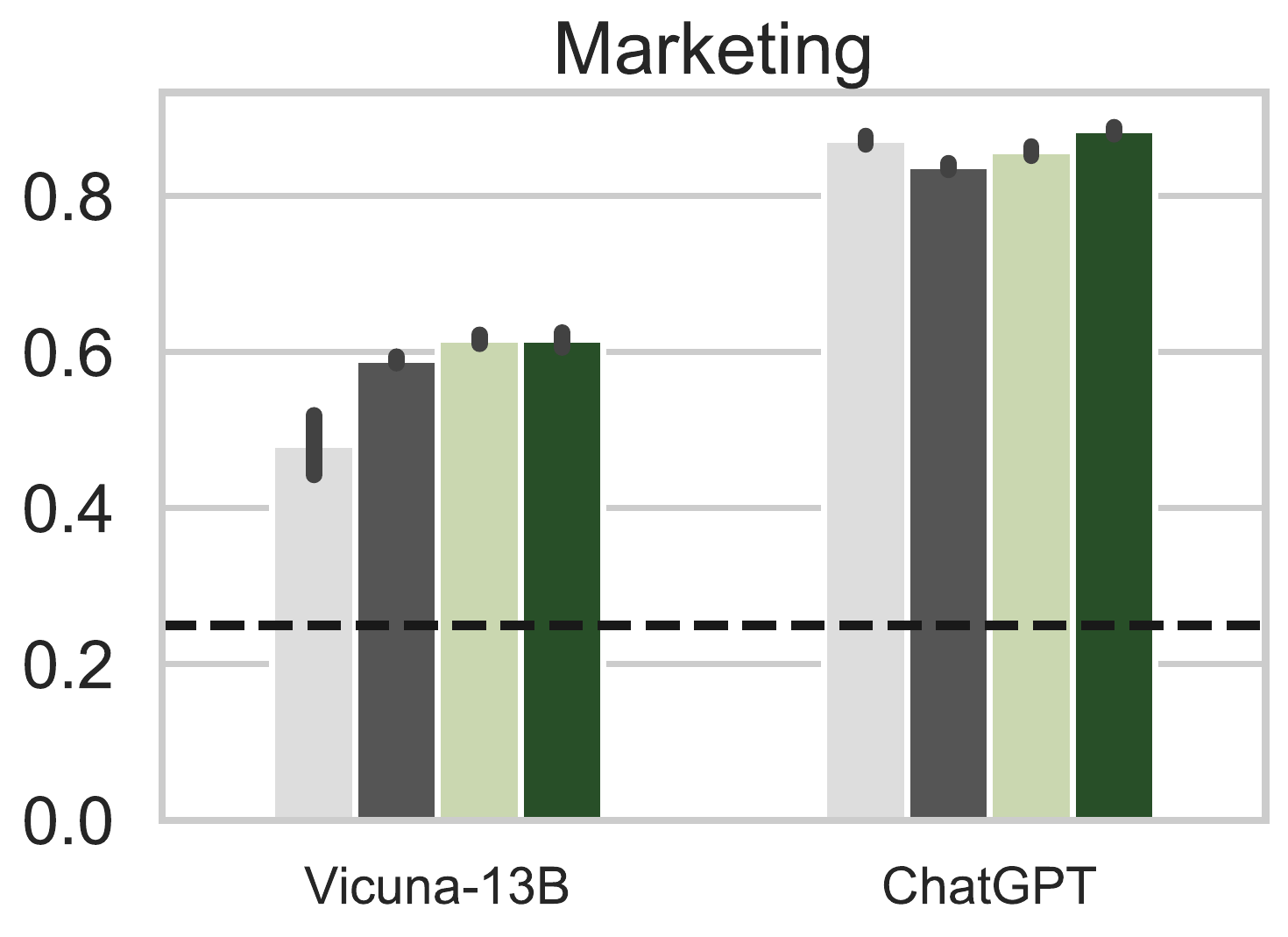}
     \end{subfigure}
     \hfill
     \begin{subfigure}[c]{0.31\textwidth}
         \centering
         \includegraphics[width=\textwidth]{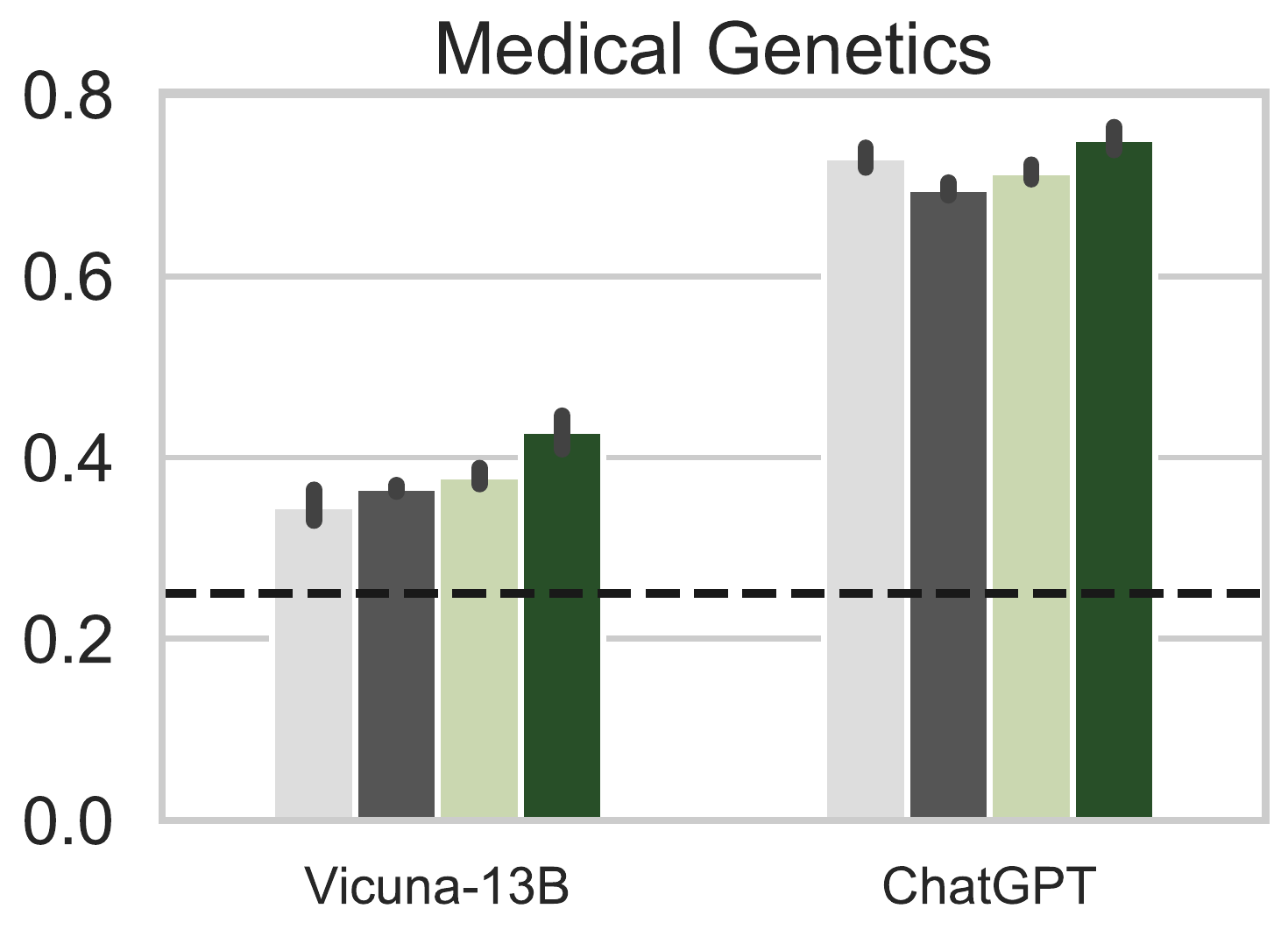}
     \end{subfigure}
     \\
      \begin{subfigure}[c]{0.31\textwidth}
         \centering
         \includegraphics[width=\textwidth]{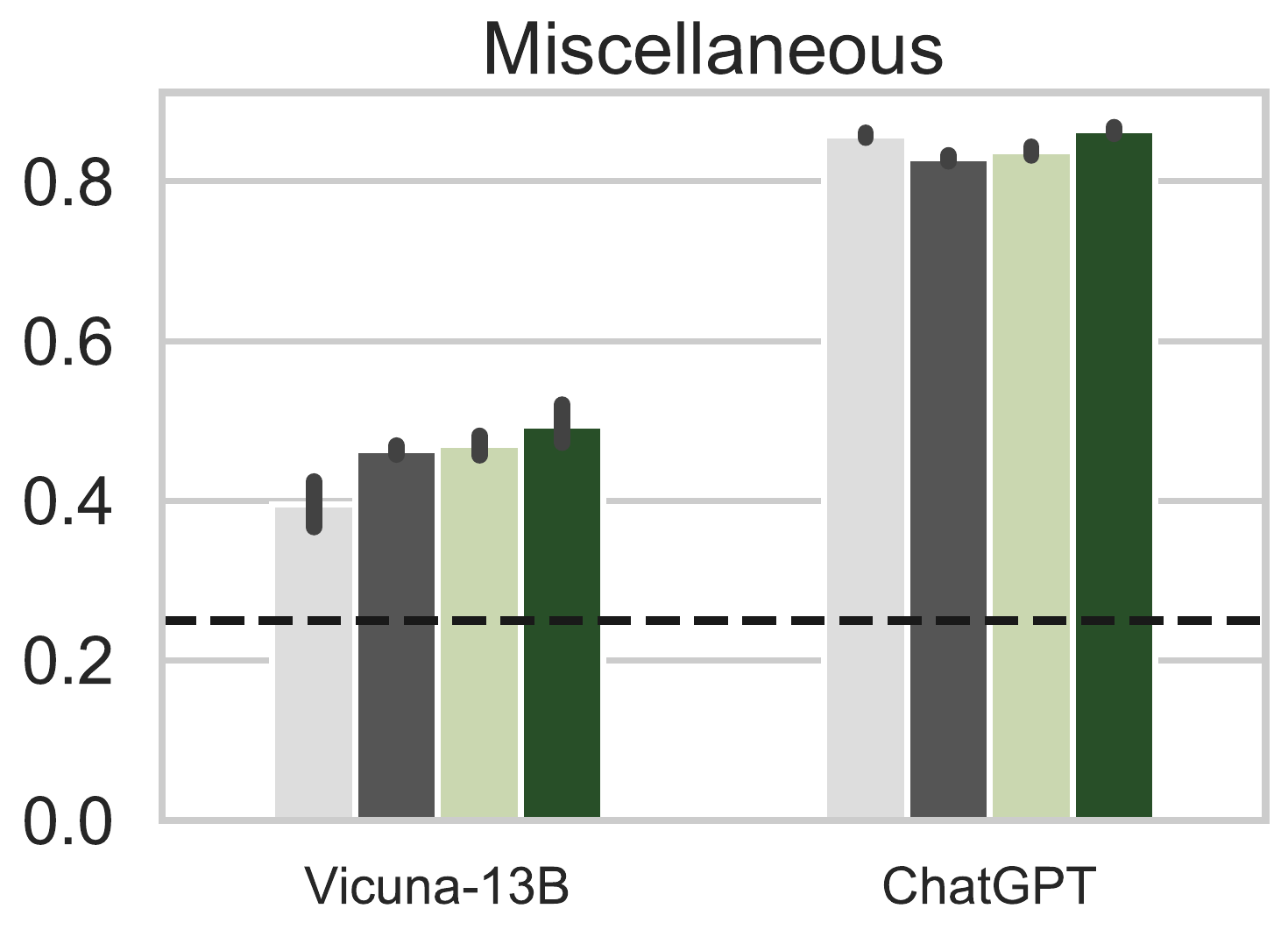}
     \end{subfigure}
      \hfill
     \begin{subfigure}[c]{0.31\textwidth}
         \centering
         \includegraphics[width=\textwidth]{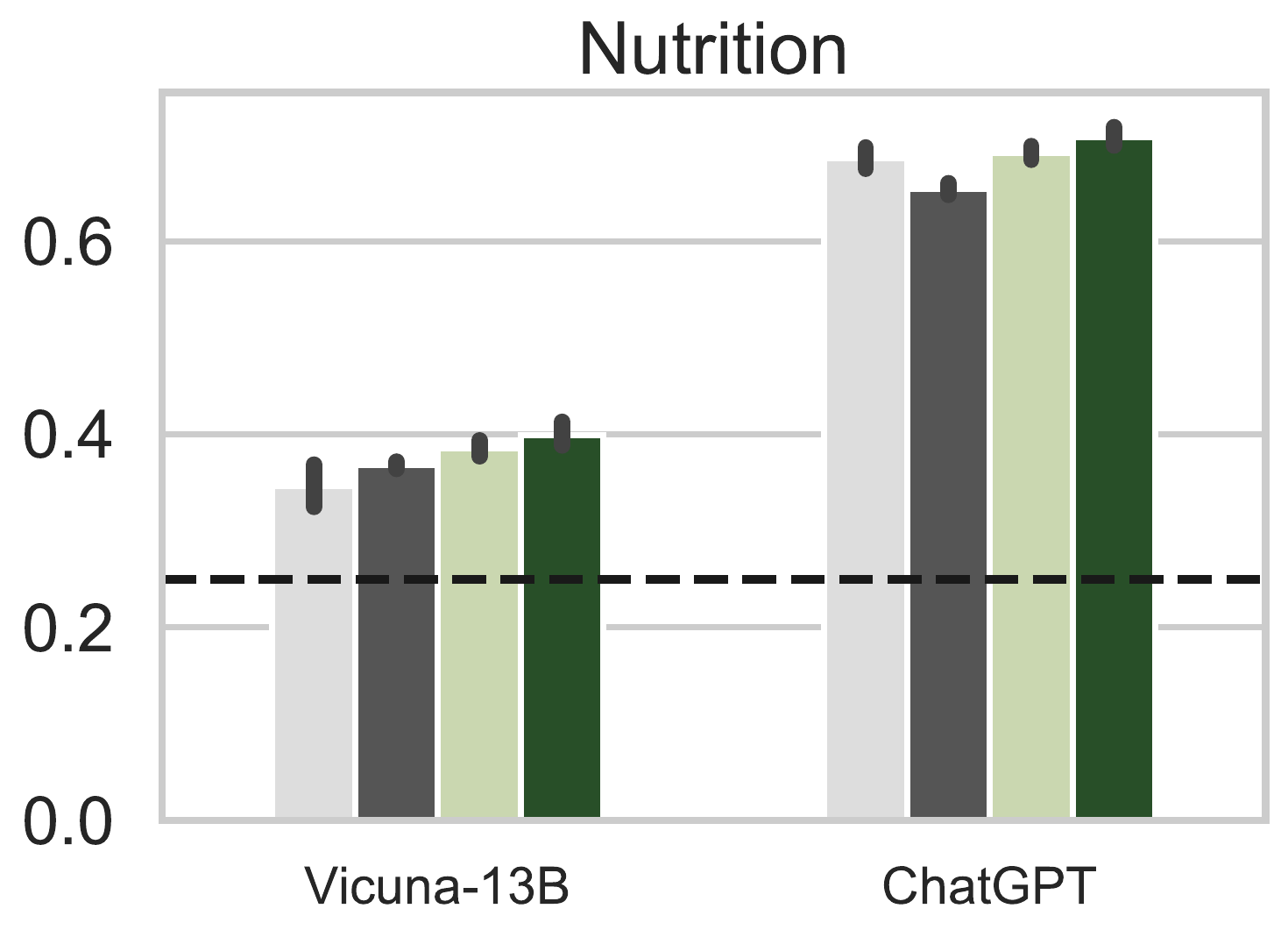}
     \end{subfigure}
     \hfill
     \begin{subfigure}[c]{0.31\textwidth}
         \centering
         \includegraphics[width=\textwidth]{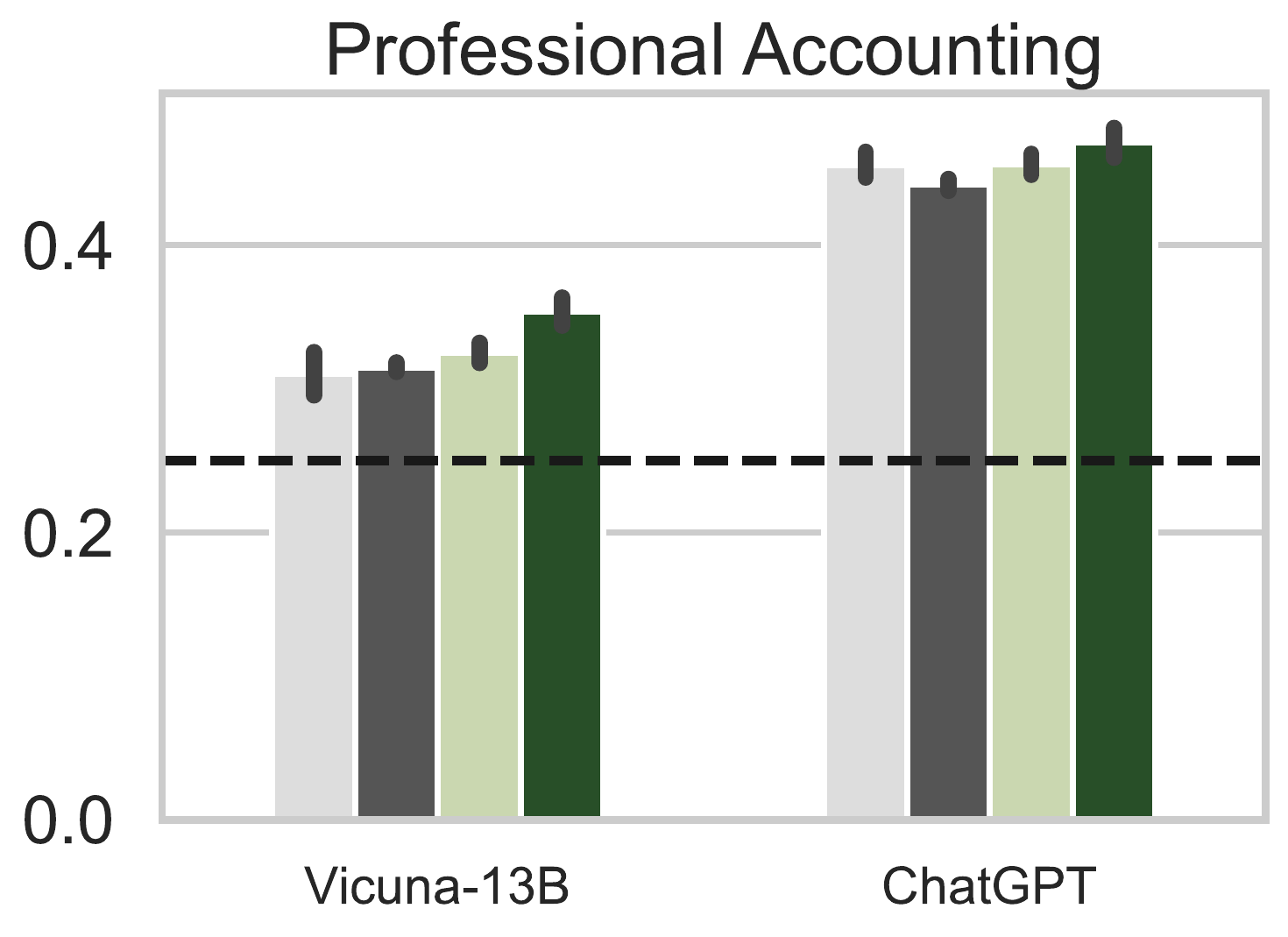}
     \end{subfigure}
     \\
     \begin{subfigure}[c]{0.31\textwidth}
         \centering
         \includegraphics[width=\textwidth]{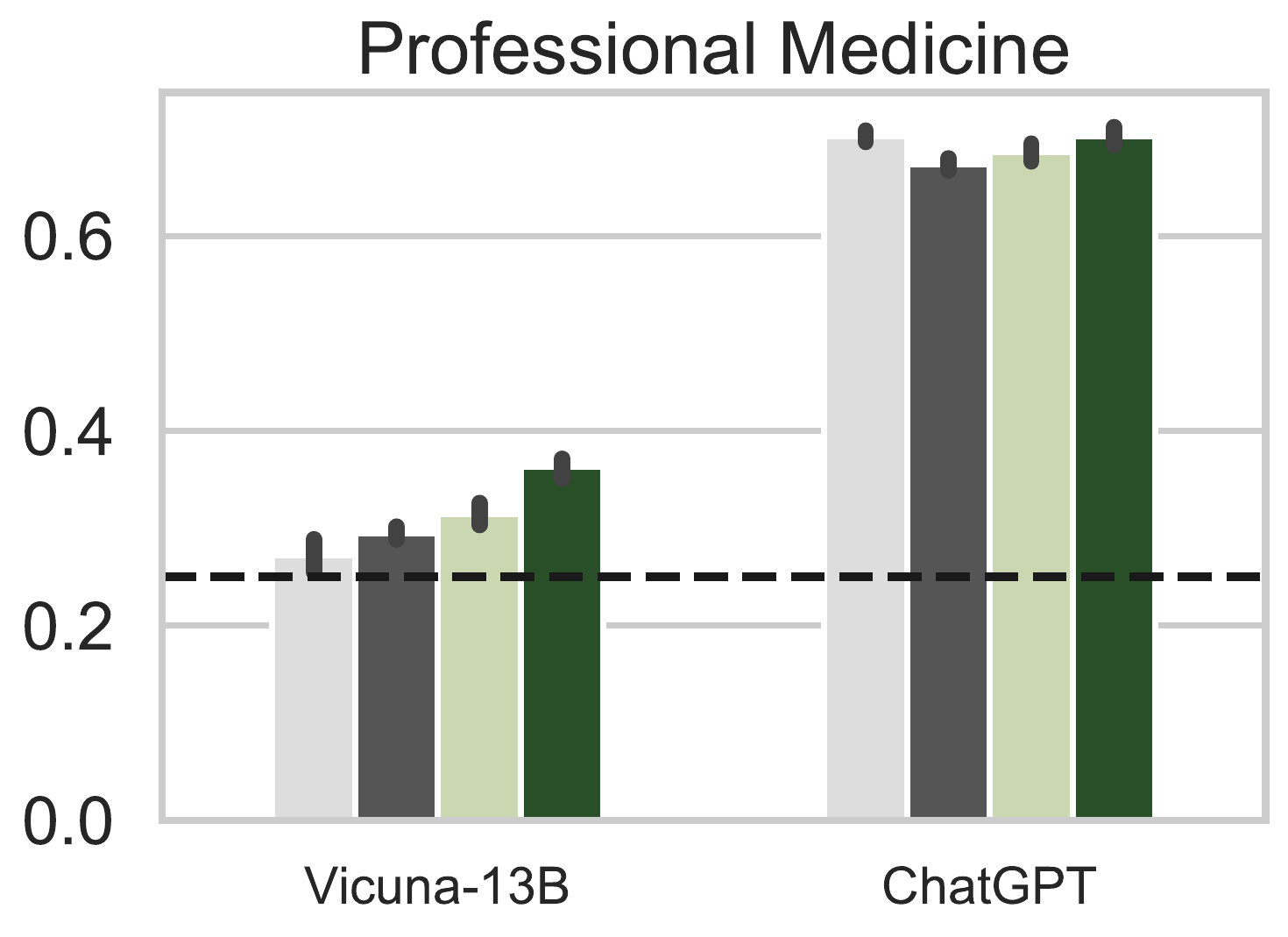}
     \end{subfigure}
     \hfill
     \begin{subfigure}[c]{0.31\textwidth}
        \centering
        \includegraphics[width=\textwidth]{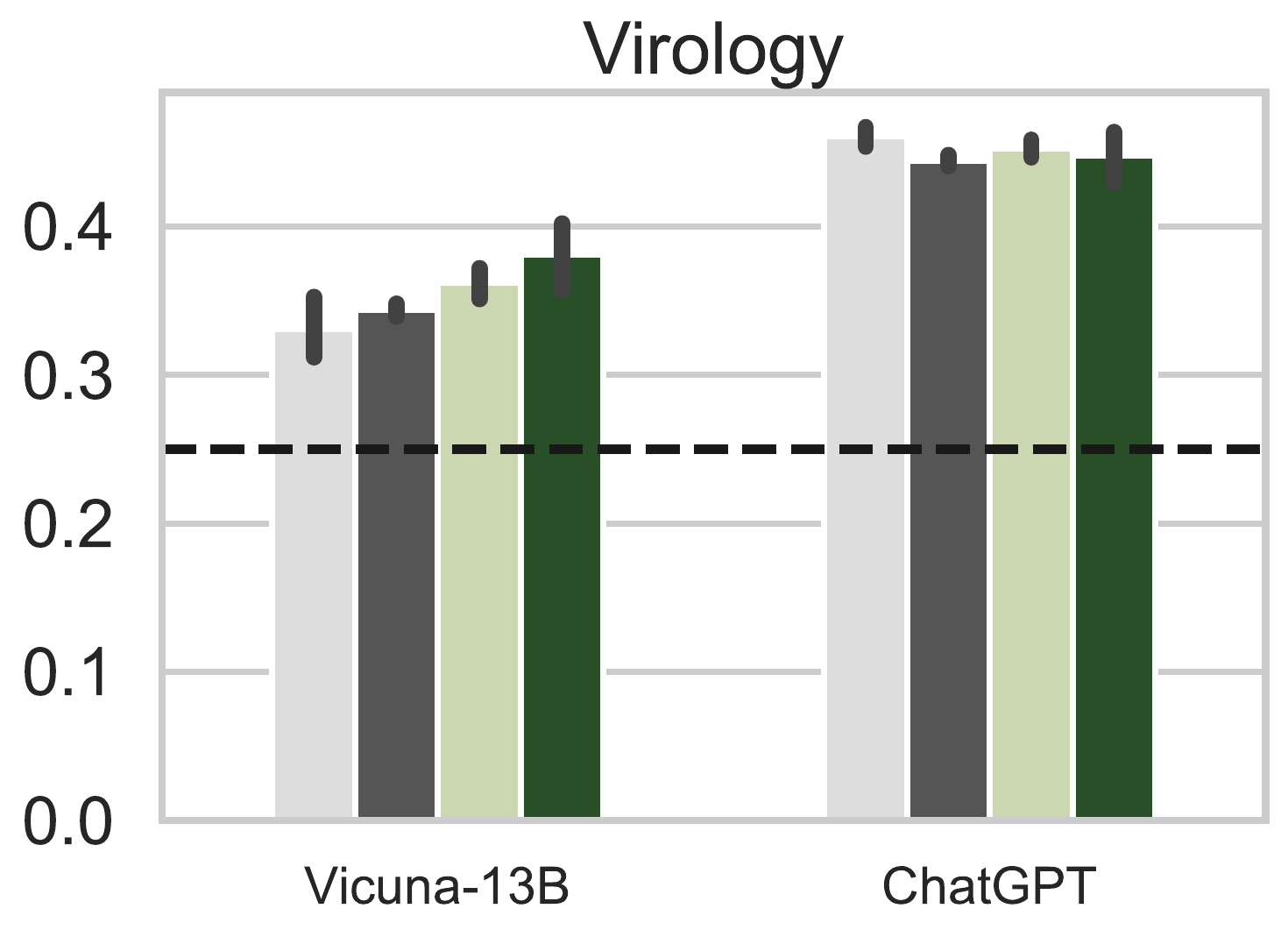}
    \end{subfigure}
    \hfill
     \begin{subfigure}[r]{0.31\textwidth}
         \centering
         \includegraphics[width=.9\textwidth]{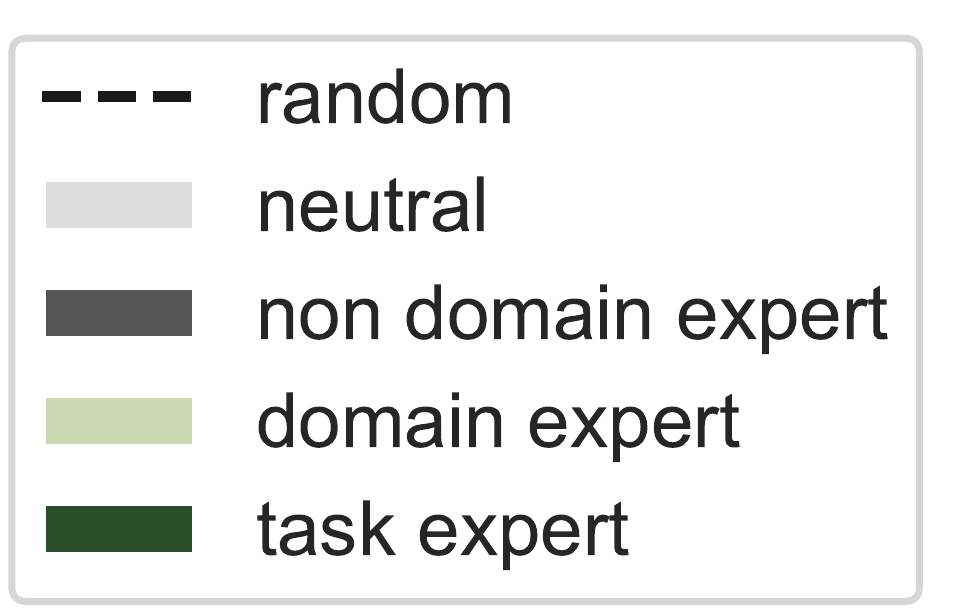}
     \end{subfigure}
     \caption{Comparison between Vicuna-13B and ChatGPT for expertise-based impersonation on the Other domain of the MMLU reasoning benchmark. We compare the task expert results with the average of all neutral personas, the average of all domain expert personas, the average of all non-domain expert personas and the random baseline (horizontal line). The first plot shows the average over all Other tasks, while the remaining plots show the results for each Other task individually. All 95\% confidence intervals are computed over the average task accuracy.}%
    \label{fig:mmlu_other}
\end{figure}

%% file: neurips_2023_arxiv_camera_ready.bbl
\begin{thebibliography}{100}
\providecommand{\natexlab}[1]{#1}
\providecommand{\url}[1]{\texttt{#1}}
\expandafter\ifx\csname urlstyle\endcsname\relax
  \providecommand{\doi}[1]{doi: #1}\else
  \providecommand{\doi}{doi: \begingroup \urlstyle{rm}\Url}\fi

\bibitem[Brown et~al.(2020)Brown, Mann, Ryder, Subbiah, Kaplan, Dhariwal, Neelakantan, Shyam, Sastry, Askell, Agarwal, Herbert{-}Voss, Krueger, Henighan, Child, Ramesh, Ziegler, Wu, Winter, Hesse, Chen, Sigler, Litwin, Gray, Chess, Clark, Berner, McCandlish, Radford, Sutskever, and Amodei]{brown2020language}
Tom~B. Brown, Benjamin Mann, Nick Ryder, Melanie Subbiah, Jared Kaplan, Prafulla Dhariwal, Arvind Neelakantan, Pranav Shyam, Girish Sastry, Amanda Askell, Sandhini Agarwal, Ariel Herbert{-}Voss, Gretchen Krueger, Tom Henighan, Rewon Child, Aditya Ramesh, Daniel~M. Ziegler, Jeffrey Wu, Clemens Winter, Christopher Hesse, Mark Chen, Eric Sigler, Mateusz Litwin, Scott Gray, Benjamin Chess, Jack Clark, Christopher Berner, Sam McCandlish, Alec Radford, Ilya Sutskever, and Dario Amodei.
\newblock Language models are few-shot learners.
\newblock \emph{NeurIPS}, 2020.

\bibitem[Webb et~al.(2022)Webb, Holyoak, and Lu]{webb2022emergent}
Taylor Webb, Keith~J Holyoak, and Hongjing Lu.
\newblock Emergent analogical reasoning in large language models.
\newblock \emph{arXiv:2212.09196}, 2022.

\bibitem[Wei et~al.(2022{\natexlab{a}})Wei, Tay, Bommasani, Raffel, Zoph, Borgeaud, Yogatama, Bosma, Zhou, Metzler, Chi, Hashimoto, Vinyals, Liang, Dean, and Fedus]{wei2022emergent}
Jason Wei, Yi~Tay, Rishi Bommasani, Colin Raffel, Barret Zoph, Sebastian Borgeaud, Dani Yogatama, Maarten Bosma, Denny Zhou, Donald Metzler, Ed~H. Chi, Tatsunori Hashimoto, Oriol Vinyals, Percy Liang, Jeff Dean, and William Fedus.
\newblock Emergent abilities of large language models.
\newblock \emph{TMLR}, 2022{\natexlab{a}}.

\bibitem[Kasneci et~al.(2023)Kasneci, Se{\ss}ler, K{\"u}chemann, Bannert, Dementieva, Fischer, Gasser, Groh, G{\"u}nnemann, H{\"u}llermeier, et~al.]{kasneci2023chatgpt}
Enkelejda Kasneci, Kathrin Se{\ss}ler, Stefan K{\"u}chemann, Maria Bannert, Daryna Dementieva, Frank Fischer, Urs Gasser, Georg Groh, Stephan G{\"u}nnemann, Eyke H{\"u}llermeier, et~al.
\newblock Chatgpt for good? on opportunities and challenges of large language models for education.
\newblock \emph{Learning and Individual Differences}, 103, 2023.

\bibitem[Bommasani et~al.(2021)Bommasani, Hudson, Adeli, Altman, Arora, von Arx, Bernstein, Bohg, Bosselut, Brunskill, et~al.]{bommasani2021opportunities}
Rishi Bommasani, Drew~A Hudson, Ehsan Adeli, Russ Altman, Simran Arora, Sydney von Arx, Michael~S Bernstein, Jeannette Bohg, Antoine Bosselut, Emma Brunskill, et~al.
\newblock On the opportunities and risks of foundation models.
\newblock \emph{arXiv:2108.07258}, 2021.

\bibitem[Tamkin et~al.(2021)Tamkin, Brundage, Clark, and Ganguli]{tamkin2021understanding}
Alex Tamkin, Miles Brundage, Jack Clark, and Deep Ganguli.
\newblock Understanding the capabilities, limitations, and societal impact of large language models.
\newblock \emph{arXiv:2102.02503}, 2021.

\bibitem[Bender et~al.(2021)Bender, Gebru, McMillan-Major, and Shmitchell]{bender2021dangers}
Emily~M Bender, Timnit Gebru, Angelina McMillan-Major, and Shmargaret Shmitchell.
\newblock On the dangers of stochastic parrots: Can language models be too big?
\newblock In \emph{ACM FAccT}, 2021.

\bibitem[Binz and Schulz(2023)]{binz2023using}
Marcel Binz and Eric Schulz.
\newblock Using cognitive psychology to understand gpt-3.
\newblock \emph{PNAS}, 120\penalty0 (6), 2023.

\bibitem[Pilault et~al.(2020)Pilault, Li, Subramanian, and Pal]{pilault2020extractive}
Jonathan Pilault, Raymond Li, Sandeep Subramanian, and Christopher Pal.
\newblock On extractive and abstractive neural document summarization with transformer language models.
\newblock In \emph{EMNLP}, 2020.

\bibitem[Wei et~al.(2022{\natexlab{b}})Wei, Wang, Schuurmans, Bosma, Ichter, Xia, Chi, Le, and Zhou]{wei2022chain}
Jason Wei, Xuezhi Wang, Dale Schuurmans, Maarten Bosma, Brian Ichter, Fei Xia, Ed~H. Chi, Quoc~V Le, and Denny Zhou.
\newblock Chain of thought prompting elicits reasoning in large language models.
\newblock In \emph{NeurIPS}, 2022{\natexlab{b}}.

\bibitem[Min et~al.(2022)Min, Lyu, Holtzman, Artetxe, Lewis, Hajishirzi, and Zettlemoyer]{min2022rethinking}
Sewon Min, Xinxi Lyu, Ari Holtzman, Mikel Artetxe, Mike Lewis, Hannaneh Hajishirzi, and Luke Zettlemoyer.
\newblock Rethinking the role of demonstrations: What makes in-context learning work?
\newblock In \emph{EMNLP}, 2022.

\bibitem[Xie et~al.(2022)Xie, Raghunathan, Liang, and Ma]{xie2022bayesianexplanation}
Sang~Michael Xie, Aditi Raghunathan, Percy Liang, and Tengyu Ma.
\newblock An explanation of in-context learning as implicit bayesian inference.
\newblock In \emph{ICLR}, 2022.

\bibitem[Deshpande et~al.(2023)Deshpande, Murahari, Rajpurohit, Kalyan, and Narasimhan]{deshpande2023toxicity}
Ameet Deshpande, Vishvak Murahari, Tanmay Rajpurohit, Ashwin Kalyan, and Karthik Narasimhan.
\newblock Toxicity in chatgpt: Analyzing persona-assigned language models.
\newblock \emph{arXiv:2304.05335}, 2023.

\bibitem[Wang et~al.(2023)Wang, Scells, Koopman, and Zuccon]{wang2023can}
Shuai Wang, Harrisen Scells, Bevan Koopman, and Guido Zuccon.
\newblock Can chatgpt write a good boolean query for systematic review literature search?
\newblock \emph{arXiv:2302.03495}, 2023.

\bibitem[Elkins and Chun(2020)]{elkins2020can}
Katherine Elkins and Jon Chun.
\newblock Can gpt-3 pass a writer’s turing test?
\newblock \emph{Journal of Cultural Analytics}, 5\penalty0 (2), 2020.

\bibitem[Binz and Schulz(2022)]{binz2022modeling}
Marcel Binz and Eric Schulz.
\newblock Modeling human exploration through resource-rational reinforcement learning.
\newblock In \emph{NeurIPS}, 2022.

\bibitem[Lampinen et~al.(2022)Lampinen, Dasgupta, Chan, Mathewson, Tessler, Creswell, McClelland, Wang, and Hill]{lampinen2022can}
Andrew Lampinen, Ishita Dasgupta, Stephanie Chan, Kory Mathewson, Mh~Tessler, Antonia Creswell, James McClelland, Jane Wang, and Felix Hill.
\newblock Can language models learn from explanations in context?
\newblock In \emph{EMNLP}. ACL, 2022.

\bibitem[Arora et~al.(2023)Arora, Narayan, Chen, Orr, Guha, Bhatia, Chami, and Re]{arora2022ask}
Simran Arora, Avanika Narayan, Mayee~F Chen, Laurel Orr, Neel Guha, Kush Bhatia, Ines Chami, and Christopher Re.
\newblock Ask me anything: A simple strategy for prompting language models.
\newblock In \emph{ICLR}, 2023.

\bibitem[Zhou et~al.(2022)Zhou, Muresanu, Han, Paster, Pitis, Chan, and Ba]{zhou2022large}
Yongchao Zhou, Andrei~Ioan Muresanu, Ziwen Han, Keiran Paster, Silviu Pitis, Harris Chan, and Jimmy Ba.
\newblock Large language models are human-level prompt engineers.
\newblock In \emph{NeurIPS Workshops}, 2022.

\bibitem[Schick and Sch{\"u}tze(2021)]{schick2020exploiting}
Timo Schick and Hinrich Sch{\"u}tze.
\newblock Exploiting cloze-questions for few-shot text classification and natural language inference.
\newblock In \emph{EACL}, 2021.

\bibitem[Sanh et~al.(2022)Sanh, Webson, Raffel, Bach, Sutawika, Alyafeai, Chaffin, Stiegler, Raja, Dey, Bari, Xu, Thakker, Sharma, Szczechla, Kim, Chhablani, Nayak, Datta, Chang, Jiang, Wang, Manica, Shen, Yong, Pandey, Bawden, Wang, Neeraj, Rozen, Sharma, Santilli, Fevry, Fries, Teehan, Scao, Biderman, Gao, Wolf, and Rush]{sanh2021multitask}
Victor Sanh, Albert Webson, Colin Raffel, Stephen Bach, Lintang Sutawika, Zaid Alyafeai, Antoine Chaffin, Arnaud Stiegler, Arun Raja, Manan Dey, M~Saiful Bari, Canwen Xu, Urmish Thakker, Shanya~Sharma Sharma, Eliza Szczechla, Taewoon Kim, Gunjan Chhablani, Nihal Nayak, Debajyoti Datta, Jonathan Chang, Mike Tian-Jian Jiang, Han Wang, Matteo Manica, Sheng Shen, Zheng~Xin Yong, Harshit Pandey, Rachel Bawden, Thomas Wang, Trishala Neeraj, Jos Rozen, Abheesht Sharma, Andrea Santilli, Thibault Fevry, Jason~Alan Fries, Ryan Teehan, Teven~Le Scao, Stella Biderman, Leo Gao, Thomas Wolf, and Alexander~M Rush.
\newblock Multitask prompted training enables zero-shot task generalization.
\newblock In \emph{ICLR}, 2022.

\bibitem[Wang et~al.(2020)Wang, Yao, Kwok, and Ni]{wang2020generalizing}
Yaqing Wang, Quanming Yao, James~T Kwok, and Lionel~M Ni.
\newblock Generalizing from a few examples: A survey on few-shot learning.
\newblock \emph{ACM computing surveys}, 53\penalty0 (3), 2020.

\bibitem[Xian et~al.(2018)Xian, Lampert, Schiele, and Akata]{xian2018zero}
Yongqin Xian, Christoph~H Lampert, Bernt Schiele, and Zeynep Akata.
\newblock Zero-shot learning—a comprehensive evaluation of the good, the bad and the ugly.
\newblock \emph{TPAMI}, 41\penalty0 (9), 2018.

\bibitem[Yuan et~al.(2023)Yuan, Yuan, Tan, Wang, and Huang]{yuan2023well}
Zheng Yuan, Hongyi Yuan, Chuanqi Tan, Wei Wang, and Songfang Huang.
\newblock How well do large language models perform in arithmetic tasks?
\newblock \emph{arXiv:2304.02015}, 2023.

\bibitem[K{\i}c{\i}man et~al.(2023)K{\i}c{\i}man, Ness, Sharma, and Tan]{kiciman2023causal}
Emre K{\i}c{\i}man, Robert Ness, Amit Sharma, and Chenhao Tan.
\newblock Causal reasoning and large language models: Opening a new frontier for causality.
\newblock \emph{arXiv:2305.00050}, 2023.

\bibitem[Reynolds and McDonell(2021)]{reynolds2021prompt}
Laria Reynolds and Kyle McDonell.
\newblock Prompt programming for large language models: Beyond the few-shot paradigm.
\newblock In \emph{CHI}, 2021.

\bibitem[Shin et~al.(2020)Shin, Razeghi, Logan~IV, Wallace, and Singh]{sshin2020autoprompt}
Taylor Shin, Yasaman Razeghi, Robert~L. Logan~IV, Eric Wallace, and Sameer Singh.
\newblock {A}uto{P}rompt: {E}liciting {K}nowledge from {L}anguage {M}odels with {A}utomatically {G}enerated {P}rompts.
\newblock In \emph{EMNLP}, 2020.

\bibitem[Hunter(2023)]{hunter2023art}
Nathan Hunter.
\newblock \emph{The art of prompt engineering with chatGPT}.
\newblock eBook, 2023.

\bibitem[Oppenlaender et~al.(2023)Oppenlaender, Linder, and Silvennoinen]{oppenlaender2023prompting}
Jonas Oppenlaender, Rhema Linder, and Johanna Silvennoinen.
\newblock Prompting ai art: An investigation into the creative skill of prompt engineering.
\newblock \emph{arXiv:2303.13534}, 2023.

\bibitem[Han et~al.(2022)Han, Kim, Yoo, Seo, Kim, Erdenee, and Chang]{han2022meet}
Seungju Han, Beomsu Kim, Jin~Yong Yoo, Seokjun Seo, Sangbum Kim, Enkhbayar Erdenee, and Buru Chang.
\newblock Meet your favorite character: Open-domain chatbot mimicking fictional characters with only a few utterances.
\newblock In \emph{NAACL-HLT}, 2022.

\bibitem[Keskar et~al.(2019)Keskar, McCann, Varshney, Xiong, and Socher]{keskar2019ctrl}
Nitish~Shirish Keskar, Bryan McCann, Lav~R Varshney, Caiming Xiong, and Richard Socher.
\newblock Ctrl: A conditional transformer language model for controllable generation.
\newblock \emph{arXiv:1909.05858}, 2019.

\bibitem[Shanahan et~al.(2023)Shanahan, McDonell, and Reynolds]{Shanahan2023RolePlayWL}
Murray Shanahan, Kyle McDonell, and Laria Reynolds.
\newblock Role-play with large language models.
\newblock \emph{ArXiv:2305.16367}, 2023.

\bibitem[Yang et~al.(2018)Yang, Hu, Dyer, Xing, and Berg-Kirkpatrick]{yang2018unsupervised}
Zichao Yang, Zhiting Hu, Chris Dyer, Eric~P Xing, and Taylor Berg-Kirkpatrick.
\newblock Unsupervised text style transfer using language models as discriminators.
\newblock \emph{NeurIPS}, 2018.

\bibitem[Crowson et~al.(2022)Crowson, Biderman, Kornis, Stander, Hallahan, Castricato, and Raff]{Crowson2022VQGANCLIPOD}
Katherine Crowson, Stella~Rose Biderman, Daniel Kornis, Dashiell Stander, Eric Hallahan, Louis Castricato, and Edward Raff.
\newblock Vqgan-clip: Open domain image generation and editing with natural language guidance.
\newblock In \emph{ECCV}, 2022.

\bibitem[Nichol et~al.(2021)Nichol, Dhariwal, Ramesh, Shyam, Mishkin, McGrew, Sutskever, and Chen]{Nichol2021GLIDETP}
Alex Nichol, Prafulla Dhariwal, Aditya Ramesh, Pranav Shyam, Pamela Mishkin, Bob McGrew, Ilya Sutskever, and Mark Chen.
\newblock Glide: Towards photorealistic image generation and editing with text-guided diffusion models.
\newblock In \emph{ICML}, 2021.

\bibitem[Saharia et~al.(2022)Saharia, Chan, Saxena, Li, Whang, Denton, Ghasemipour, Gontijo~Lopes, Karagol~Ayan, Salimans, et~al.]{saharia2022photorealistic}
Chitwan Saharia, William Chan, Saurabh Saxena, Lala Li, Jay Whang, Emily~L Denton, Kamyar Ghasemipour, Raphael Gontijo~Lopes, Burcu Karagol~Ayan, Tim Salimans, et~al.
\newblock Photorealistic text-to-image diffusion models with deep language understanding.
\newblock \emph{NeurIPS}, 2022.

\bibitem[Dehouche and Dehouche(2023)]{Dehouche2023WhatsIA}
Nassim Dehouche and Kullathida Dehouche.
\newblock What’s in a text-to-image prompt? the potential of stable diffusion in visual arts education.
\newblock \emph{Heliyon}, 9, 2023.

\bibitem[Brack et~al.(2022)Brack, Schramowski, Friedrich, Hintersdorf, and Kersting]{Brack2022TheSA}
Manuel Brack, Patrick Schramowski, Felix Friedrich, Dominik Hintersdorf, and Kristian Kersting.
\newblock The stable artist: Steering semantics in diffusion latent space.
\newblock \emph{arXiv:2212.06013}, 2022.

\bibitem[Witteveen and Andrews(2022)]{Witteveen2022InvestigatingPE}
Sam Witteveen and Martin Andrews.
\newblock Investigating prompt engineering in diffusion models.
\newblock \emph{arXiv:2211.15462}, 2022.

\bibitem[Lin et~al.(2022)Lin, Hilton, and Evans]{lin2021truthfulqa}
Stephanie Lin, Jacob Hilton, and Owain Evans.
\newblock {T}ruthful{QA}: Measuring how models mimic human falsehoods.
\newblock In \emph{ACL}, 2022.

\bibitem[Wolf et~al.(2023)Wolf, Wies, Levine, and Shashua]{wolf2023fundamental}
Yotam Wolf, Noam Wies, Yoav Levine, and Amnon Shashua.
\newblock Fundamental limitations of alignment in large language models.
\newblock \emph{arXiv:2304.11082}, 2023.

\bibitem[Aher et~al.(2022{\natexlab{a}})Aher, Arriaga, and Kalai]{aher2022using}
Gati Aher, Rosa~I Arriaga, and Adam~Tauman Kalai.
\newblock Using large language models to simulate multiple humans.
\newblock \emph{arXiv:2208.10264}, 2022{\natexlab{a}}.

\bibitem[Pellert et~al.(2023)Pellert, Lechner, Wagner, Rammstedt, and Strohmaier]{pellert2023ai}
Max Pellert, Clemens~M Lechner, Claudia Wagner, Beatrice Rammstedt, and Markus Strohmaier.
\newblock Ai psychometrics: Using psychometric inventories to obtain psychological profiles of large language models.
\newblock 2023.

\bibitem[Park et~al.(2023{\natexlab{a}})Park, Schoenegger, and Zhu]{park2023correct}
Peter~S. Park, Philipp Schoenegger, and Chongyang Zhu.
\newblock "correct answers" from the psychology of artificial intelligence.
\newblock \emph{arXiv:2302.07267}, 2023{\natexlab{a}}.

\bibitem[Karra et~al.(2022)Karra, Nguyen, and Tulabandhula]{karra2022ai}
Saketh~Reddy Karra, Son Nguyen, and Theja Tulabandhula.
\newblock Ai personification: Estimating the personality of language models.
\newblock \emph{arXiv:2204.12000}, 2022.

\bibitem[Coda-Forno et~al.(2023)Coda-Forno, Witte, Jagadish, Binz, Akata, and Schulz]{coda2023inducing}
Julian Coda-Forno, Kristin Witte, Akshay~K Jagadish, Marcel Binz, Zeynep Akata, and Eric Schulz.
\newblock Inducing anxiety in large language models increases exploration and bias.
\newblock \emph{arXiv:2304.11111}, 2023.

\bibitem[Dominguez-Olmedo et~al.(2023)Dominguez-Olmedo, Hardt, and Mendler-Dunner]{DominguezOlmedo2023QuestioningTS}
Ricardo Dominguez-Olmedo, Moritz Hardt, and Celestine Mendler-Dunner.
\newblock Questioning the survey responses of large language models.
\newblock \emph{arXiv:2306.07951}, 2023.

\bibitem[Argyle et~al.(2023)Argyle, Busby, Fulda, Gubler, Rytting, and Wingate]{argyle2022out}
Lisa~P. Argyle, Ethan~C. Busby, Nancy Fulda, Joshua~R. Gubler, Christopher Rytting, and David Wingate.
\newblock Out of one, many: Using language models to simulate human samples.
\newblock \emph{Political Analysis}, 2023.

\bibitem[Jiang et~al.(2023)Jiang, Zhang, Cao, Kabbara, and Roy]{jiang2023personallm}
Hang Jiang, Xiajie Zhang, Xubo Cao, Jad Kabbara, and Deb Roy.
\newblock Personallm: Investigating the ability of gpt-3.5 to express personality traits and gender differences.
\newblock \emph{arXiv:2305.02547}, 2023.

\bibitem[Aher et~al.(2022{\natexlab{b}})Aher, Arriaga, and Kalai]{Aher2022UsingLL}
Gati Aher, RosaI. Arriaga, and Adam~Tauman Kalai.
\newblock Using large language models to simulate multiple humans and replicate human subject studies.
\newblock In \emph{ICML}, 2022{\natexlab{b}}.

\bibitem[Caliskan et~al.(2017)Caliskan, Bryson, and Narayanan]{caliskan2017semantics}
Aylin Caliskan, Joanna~J Bryson, and Arvind Narayanan.
\newblock Semantics derived automatically from language corpora contain human-like biases.
\newblock \emph{Science}, 356\penalty0 (6334), 2017.

\bibitem[Abid et~al.(2021)Abid, Farooqi, and Zou]{abid2021persistent}
Abubakar Abid, Maheen Farooqi, and James Zou.
\newblock Persistent anti-muslim bias in large language models.
\newblock In \emph{AAAI/ACM AEIS}, 2021.

\bibitem[Kang et~al.(2023)Kang, Li, Stoica, Guestrin, Zaharia, and Hashimoto]{kang2023exploiting}
Daniel Kang, Xuechen Li, Ion Stoica, Carlos Guestrin, Matei Zaharia, and Tatsunori Hashimoto.
\newblock Exploiting programmatic behavior of llms: Dual-use through standard security attacks.
\newblock \emph{arXiv:2302.05733}, 2023.

\bibitem[Jia et~al.(2021)Jia, Yang, Xia, Chen, Parekh, Pham, Le, Sung, Li, and Duerig]{jia2021align}
Chao Jia, Yinfei Yang, Ye~Xia, Yi{-}Ting Chen, Zarana Parekh, Hieu Pham, Quoc~V. Le, Yun{-}Hsuan Sung, Zhen Li, and Tom Duerig.
\newblock Scaling up visual and vision-language representation learning with noisy text supervision.
\newblock In \emph{{ICML}}, 2021.

\bibitem[Singh et~al.(2022)Singh, Hu, Goswami, Couairon, Galuba, Rohrbach, and Kiela]{singh2022flava}
Amanpreet Singh, Ronghang Hu, Vedanuj Goswami, Guillaume Couairon, Wojciech Galuba, Marcus Rohrbach, and Douwe Kiela.
\newblock {FLAVA:} {A} foundational language and vision alignment model.
\newblock In \emph{CVPR}, 2022.

\bibitem[Radford et~al.(2021)Radford, Kim, Hallacy, Ramesh, Goh, Agarwal, Sastry, Askell, Mishkin, Clark, Krueger, and Sutskever]{radford2021clip}
Alec Radford, Jong~Wook Kim, Chris Hallacy, Aditya Ramesh, Gabriel Goh, Sandhini Agarwal, Girish Sastry, Amanda Askell, Pamela Mishkin, Jack Clark, Gretchen Krueger, and Ilya Sutskever.
\newblock Learning transferable visual models from natural language supervision.
\newblock In \emph{{ICML}}, 2021.

\bibitem[Menon and Vondrick(2023)]{menon2023visual}
Sachit Menon and Carl Vondrick.
\newblock Visual classification via description from large language models.
\newblock In \emph{ICLR}, 2023.

\bibitem[Yang et~al.(2022{\natexlab{a}})Yang, Panagopoulou, Zhou, Jin, Callison-Burch, and Yatskar]{Yang2022LanguageIA}
Yue Yang, Artemis Panagopoulou, Shenghao Zhou, Daniel Jin, Chris Callison-Burch, and Mark Yatskar.
\newblock Language in a bottle: Language model guided concept bottlenecks for interpretable image classification.
\newblock \emph{arXiv:2211.11158}, 2022{\natexlab{a}}.

\bibitem[Yang et~al.(2022{\natexlab{b}})Yang, Gan, Wang, Hu, Lu, Liu, and Wang]{yang2022vqa}
Zhengyuan Yang, Zhe Gan, Jianfeng Wang, Xiaowei Hu, Yumao Lu, Zicheng Liu, and Lijuan Wang.
\newblock An empirical study of {GPT-3} for few-shot knowledge-based {VQA}.
\newblock In \emph{{AAAI}}, 2022{\natexlab{b}}.

\bibitem[Chiang et~al.(2023)Chiang, Li, Lin, Sheng, Wu, Zhang, Zheng, Zhuang, Zhuang, Gonzalez, et~al.]{chiang2023vicuna}
Wei-Lin Chiang, Zhuohan Li, Zi~Lin, Ying Sheng, Zhanghao Wu, Hao Zhang, Lianmin Zheng, Siyuan Zhuang, Yonghao Zhuang, Joseph~E Gonzalez, et~al.
\newblock Vicuna: An open-source chatbot impressing gpt-4 with 90\%* chatgpt quality, 2023.

\bibitem[Touvron et~al.(2023{\natexlab{a}})Touvron, Lavril, Izacard, Martinet, Lachaux, Lacroix, Rozi{\`e}re, Goyal, Hambro, Azhar, et~al.]{touvron2023llama}
Hugo Touvron, Thibaut Lavril, Gautier Izacard, Xavier Martinet, Marie-Anne Lachaux, Timoth{\'e}e Lacroix, Baptiste Rozi{\`e}re, Naman Goyal, Eric Hambro, Faisal Azhar, et~al.
\newblock Llama: Open and efficient foundation language models.
\newblock \emph{arXiv:2302.13971}, 2023{\natexlab{a}}.

\bibitem[Eccleston(2023)]{shareGPT2023}
Dom Eccleston.
\newblock {ShareGPT: Share your wildest conversations with one click.}
\newblock \url{https://sharegpt.com/}, 2023.
\newblock [Online; accessed 15-May-2023].

\bibitem[Zheng et~al.(2023)Zheng, Chiang, Sheng, Zhuang, Wu, Zhuang, Lin, Li, Li, Xing, Zhang, Gonzalez, and Stoica]{Zheng2023JudgingLW}
Lianmin Zheng, Wei-Lin Chiang, Ying Sheng, Siyuan Zhuang, Zhanghao Wu, Yonghao Zhuang, Zi~Lin, Zhuohan Li, Dacheng Li, Eric~P. Xing, Haotong Zhang, Joseph Gonzalez, and Ion Stoica.
\newblock Judging llm-as-a-judge with mt-bench and chatbot arena.
\newblock \emph{arXiv:2306.05685}, 2023.

\bibitem[Ouyang et~al.(2022)Ouyang, Wu, Jiang, Almeida, Wainwright, Mishkin, Zhang, Agarwal, Slama, Ray, et~al.]{Ouyang2022TrainingLM}
Long Ouyang, Jeffrey Wu, Xu~Jiang, Diogo Almeida, Carroll Wainwright, Pamela Mishkin, Chong Zhang, Sandhini Agarwal, Katarina Slama, Alex Ray, et~al.
\newblock Training language models to follow instructions with human feedback.
\newblock \emph{NeurIPS}, 2022.

\bibitem[Gershman(2018)]{gershman2018deconstructing}
Samuel~J Gershman.
\newblock Deconstructing the human algorithms for exploration.
\newblock \emph{Cognition}, 173, 2018.

\bibitem[Schulz and Gershman(2019)]{schulz2019algorithmic}
Eric Schulz and Samuel~J Gershman.
\newblock The algorithmic architecture of exploration in the human brain.
\newblock \emph{Current opinion in neurobiology}, 55, 2019.

\bibitem[Hendrycks et~al.(2021)Hendrycks, Burns, Basart, Zou, Mazeika, Song, and Steinhardt]{hendrycks2021measuring}
Dan Hendrycks, Collin Burns, Steven Basart, Andy Zou, Mantas Mazeika, Dawn Song, and Jacob Steinhardt.
\newblock Measuring massive multitask language understanding.
\newblock In \emph{ICLR}, 2021.

\bibitem[Cherti et~al.(2022)Cherti, Beaumont, Wightman, Wortsman, Ilharco, Gordon, Schuhmann, Schmidt, and Jitsev]{Cherti2022ReproducibleSL}
Mehdi Cherti, Romain Beaumont, Ross Wightman, Mitchell Wortsman, Gabriel Ilharco, Cade Gordon, Christoph Schuhmann, Ludwig Schmidt, and Jenia Jitsev.
\newblock Reproducible scaling laws for contrastive language-image learning.
\newblock \emph{arXiv:2212.07143}, 2022.

\bibitem[Arora et~al.(2022)Arora, Narayan, Chen, Orr, Guha, Bhatia, Chami, Sala, and R'e]{Arora2022AskMA}
Simran Arora, Avanika Narayan, Mayee~F. Chen, Laurel~J. Orr, Neel Guha, Kush~S Bhatia, Ines Chami, Frederic Sala, and Christopher R'e.
\newblock Ask me anything: A simple strategy for prompting language models.
\newblock \emph{arXiv:2210.02441}, 2022.

\bibitem[Nussenbaum and Hartley(2019)]{nussenbaum2019reinforcement}
Kate Nussenbaum and Catherine~A Hartley.
\newblock Reinforcement learning across development: What insights can we draw from a decade of research?
\newblock \emph{Developmental cognitive neuroscience}, 40, 2019.

\bibitem[Liquin and Gopnik(2020)]{Liquin2020ChildrenAM}
Emily~G. Liquin and Alison Gopnik.
\newblock Children are more exploratory and learn more than adults in an approach-avoid task.
\newblock \emph{Cognition}, 218, 2020.

\bibitem[Schulz et~al.(2019)Schulz, Wu, Ruggeri, and Meder]{schulz2019searching}
Eric Schulz, Charley~M Wu, Azzurra Ruggeri, and Bj{\"o}rn Meder.
\newblock Searching for rewards like a child means less generalization and more directed exploration.
\newblock \emph{Psychological science}, 30\penalty0 (11), 2019.

\bibitem[Giron et~al.(2022)Giron, Ciranka, Schulz, van~den Bos, Azzurra, Ruggeri, Meder, and Wu]{Giron2022DevelopmentalCI}
Anna~P. Giron, Simon Ciranka, Eric Schulz, Wouter van~den Bos, Azzurra, Ruggeri, Bj{\"o}rn Meder, and Charley~M. Wu.
\newblock Developmental changes in learning resemble stochastic optimization.
\newblock 2022.

\bibitem[Blanco and Sloutsky(2019)]{Blanco2019SystematicEA}
Nathaniel~J. Blanco and Vladimir~M. Sloutsky.
\newblock Systematic exploration and uncertainty dominate young children's choices.
\newblock \emph{Developmental science}, 2019.

\bibitem[Chowdhery et~al.(2022)Chowdhery, Narang, Devlin, Bosma, Mishra, Roberts, Barham, Chung, Sutton, Gehrmann, et~al.]{chowdhery2022palm}
Aakanksha Chowdhery, Sharan Narang, Jacob Devlin, Maarten Bosma, Gaurav Mishra, Adam Roberts, Paul Barham, Hyung~Won Chung, Charles Sutton, Sebastian Gehrmann, et~al.
\newblock Palm: Scaling language modeling with pathways.
\newblock \emph{arXiv:2204.02311}, 2022.

\bibitem[Hoffmann et~al.(2022)Hoffmann, Borgeaud, Mensch, Buchatskaya, Cai, Rutherford, Casas, Hendricks, Welbl, Clark, et~al.]{hoffmann2022chinchilla}
Jordan Hoffmann, Sebastian Borgeaud, Arthur Mensch, Elena Buchatskaya, Trevor Cai, Eliza Rutherford, Diego de~Las Casas, Lisa~Anne Hendricks, Johannes Welbl, Aidan Clark, et~al.
\newblock Training compute-optimal large language models.
\newblock \emph{arXiv:2203.15556}, 2022.

\bibitem[Liang et~al.(2023)Liang, Bommasani, Lee, Tsipras, Soylu, Yasunaga, Zhang, Narayanan, Wu, Kumar, Newman, Yuan, Yan, Zhang, Cosgrove, Manning, R'e, Acosta-Navas, Hudson, Zelikman, Durmus, Ladhak, Rong, Ren, Yao, Wang, Santhanam, Orr, Zheng, Yuksekgonul, Suzgun, Kim, Guha, Chatterji, Khattab, Henderson, Huang, Chi, Xie, Santurkar, Ganguli, Hashimoto, Icard, Zhang, Chaudhary, Wang, Li, Mai, Zhang, and Koreeda]{Liang2023HolisticEO}
Percy Liang, Rishi Bommasani, Tony Lee, Dimitris Tsipras, Dilara Soylu, Michihiro Yasunaga, Yian Zhang, Deepak Narayanan, Yuhuai Wu, Ananya Kumar, Benjamin Newman, Binhang Yuan, Bobby Yan, Ce~Zhang, Christian Cosgrove, Christopher~D. Manning, Christopher R'e, Diana Acosta-Navas, Drew~A. Hudson, E.~Zelikman, Esin Durmus, Faisal Ladhak, Frieda Rong, Hongyu Ren, Huaxiu Yao, Jue Wang, Keshav Santhanam, Laurel~J. Orr, Lucia Zheng, Mert Yuksekgonul, Mirac Suzgun, Nathan~S. Kim, Neel Guha, Niladri~S. Chatterji, Omar Khattab, Peter Henderson, Qian Huang, Ryan Chi, Sang~Michael Xie, Shibani Santurkar, Surya Ganguli, Tatsunori Hashimoto, Thomas~F. Icard, Tianyi Zhang, Vishrav Chaudhary, William Wang, Xuechen Li, Yifan Mai, Yuhui Zhang, and Yuta Koreeda.
\newblock Holistic evaluation of language models.
\newblock \emph{Annals of the New York Academy of Sciences}, 1525, 2023.

\bibitem[Wah et~al.(2011)Wah, Branson, Welinder, Perona, and Belongie]{Wah2011TheCB}
Catherine Wah, Steve Branson, Peter Welinder, Pietro Perona, and Serge~J. Belongie.
\newblock The caltech-ucsd birds-200-2011 dataset.
\newblock 2011.

\bibitem[Krause et~al.(2013)Krause, Stark, Deng, and Fei-Fei]{Krause20133DOR}
Jonathan Krause, Michael Stark, Jia Deng, and Li~Fei-Fei.
\newblock 3d object representations for fine-grained categorization.
\newblock \emph{ICCV Workshops}, 2013.

\bibitem[Maji et~al.(2013)Maji, Kannala, Rahtu, Blaschko, and Vedaldi]{maji13fine-grained}
S.~Maji, J.~Kannala, E.~Rahtu, M.~Blaschko, and A.~Vedaldi.
\newblock Fine-grained visual classification of aircraft.
\newblock Technical report, 2013.

\bibitem[Nilsback and Zisserman(2008)]{Nilsback08flower}
Maria-Elena Nilsback and Andrew Zisserman.
\newblock Automated flower classification over a large number of classes.
\newblock In \emph{Indian Conference on Computer Vision, Graphics and Image Processing}, 2008.

\bibitem[Dosovitskiy et~al.(2021)Dosovitskiy, Beyer, Kolesnikov, Weissenborn, Zhai, Unterthiner, Dehghani, Minderer, Heigold, Gelly, Uszkoreit, and Houlsby]{Dosovitskiy2020AnII}
Alexey Dosovitskiy, Lucas Beyer, Alexander Kolesnikov, Dirk Weissenborn, Xiaohua Zhai, Thomas Unterthiner, Mostafa Dehghani, Matthias Minderer, Georg Heigold, Sylvain Gelly, Jakob Uszkoreit, and Neil Houlsby.
\newblock An image is worth 16x16 words: Transformers for image recognition at scale.
\newblock In \emph{ICLR}, 2021.

\bibitem[Schuhmann et~al.(2022)Schuhmann, Beaumont, Vencu, Gordon, Wightman, Cherti, Coombes, Katta, Mullis, Wortsman, Schramowski, Kundurthy, Crowson, Schmidt, Kaczmarczyk, and Jitsev]{Schuhmann2022LAION5BAO}
Christoph Schuhmann, Romain Beaumont, Richard Vencu, Cade Gordon, Ross Wightman, Mehdi Cherti, Theo Coombes, Aarush Katta, Clayton Mullis, Mitchell Wortsman, Patrick Schramowski, Srivatsa Kundurthy, Katherine Crowson, Ludwig Schmidt, Robert Kaczmarczyk, and Jenia Jitsev.
\newblock Laion-5b: An open large-scale dataset for training next generation image-text models.
\newblock \emph{arXiv:2210.08402}, 2022.

\bibitem[Oates and Grayson(2004)]{oates2004cognitive}
John~Ed Oates and Andrew~Ed Grayson.
\newblock \emph{Cognitive and language development in children.}
\newblock Open University Press, 2004.

\bibitem[Durrant and Brenchley(2019)]{durant2023sophistication}
Philip Durrant and Mark Brenchley.
\newblock Development of vocabulary sophistication across genres in english children’s writing.
\newblock \emph{Springer Reading Writing}, 32, 2019.

\bibitem[Burnell et~al.(2023)Burnell, Schellaert, Burden, Ullman, Martinez-Plumed, Tenenbaum, Rutar, Cheke, Sohl-Dickstein, Mitchell, et~al.]{burnell2023rethink}
Ryan Burnell, Wout Schellaert, John Burden, Tomer~D Ullman, Fernando Martinez-Plumed, Joshua~B Tenenbaum, Danaja Rutar, Lucy~G Cheke, Jascha Sohl-Dickstein, Melanie Mitchell, et~al.
\newblock Rethink reporting of evaluation results in ai.
\newblock \emph{Science}, 380\penalty0 (6641), 2023.

\bibitem[Kaplan et~al.(2020)Kaplan, McCandlish, Henighan, Brown, Chess, Child, Gray, Radford, Wu, and Amodei]{kaplan2020scaling}
Jared Kaplan, Sam McCandlish, Tom Henighan, Tom~B Brown, Benjamin Chess, Rewon Child, Scott Gray, Alec Radford, Jeffrey Wu, and Dario Amodei.
\newblock Scaling laws for neural language models.
\newblock \emph{arXiv:2001.08361}, 2020.

\bibitem[Ziegler et~al.(2019)Ziegler, Stiennon, Wu, Brown, Radford, Amodei, Christiano, and Irving]{ziegler2019fine}
Daniel~M Ziegler, Nisan Stiennon, Jeffrey Wu, Tom~B Brown, Alec Radford, Dario Amodei, Paul Christiano, and Geoffrey Irving.
\newblock Fine-tuning language models from human preferences.
\newblock \emph{arXiv:1909.08593}, 2019.

\bibitem[Park et~al.(2023{\natexlab{b}})Park, O'Brien, Cai, Morris, Liang, and Bernstein]{park2023generative}
Joon~Sung Park, Joseph~C O'Brien, Carrie~J Cai, Meredith~Ringel Morris, Percy Liang, and Michael~S Bernstein.
\newblock Generative agents: Interactive simulacra of human behavior.
\newblock \emph{arXiv:2304.03442}, 2023{\natexlab{b}}.

\bibitem[Wang et~al.(2022)Wang, Li, Li, He, Huang, Zhao, Zhang, Xu, Liu, Wang, et~al.]{wang2022internvideo}
Yi~Wang, Kunchang Li, Yizhuo Li, Yinan He, Bingkun Huang, Zhiyu Zhao, Hongjie Zhang, Jilan Xu, Yi~Liu, Zun Wang, et~al.
\newblock Internvideo: General video foundation models via generative and discriminative learning.
\newblock \emph{arXiv:2212.03191}, 2022.

\bibitem[Touvron et~al.(2023{\natexlab{b}})Touvron, Martin, Stone, Albert, Almahairi, Babaei, Bashlykov, Batra, Bhargava, Bhosale, Bikel, Blecher, Ferrer, Chen, Cucurull, Esiobu, Fernandes, Fu, Fu, Fuller, Gao, Goswami, Goyal, Hartshorn, Hosseini, Hou, Inan, Kardas, Kerkez, Khabsa, Kloumann, Korenev, Koura, Lachaux, Lavril, Lee, Liskovich, Lu, Mao, Martinet, Mihaylov, Mishra, Molybog, Nie, Poulton, Reizenstein, Rungta, Saladi, Schelten, Silva, Smith, Subramanian, Tan, Tang, Taylor, Williams, Kuan, Xu, Yan, Zarov, Zhang, Fan, Kambadur, Narang, Rodriguez, Stojnic, Edunov, and Scialom]{Touvron2023Llama2O}
Hugo Touvron, Louis Martin, Kevin~R. Stone, Peter Albert, Amjad Almahairi, Yasmine Babaei, Nikolay Bashlykov, Soumya Batra, Prajjwal Bhargava, Shruti Bhosale, Daniel~M. Bikel, Lukas Blecher, Cristian~Cant{\'o}n Ferrer, Moya Chen, Guillem Cucurull, David Esiobu, Jude Fernandes, Jeremy Fu, Wenyin Fu, Brian Fuller, Cynthia Gao, Vedanuj Goswami, Naman Goyal, Anthony~S. Hartshorn, Saghar Hosseini, Rui Hou, Hakan Inan, Marcin Kardas, Viktor Kerkez, Madian Khabsa, Isabel~M. Kloumann, A.~V. Korenev, Punit~Singh Koura, Marie-Anne Lachaux, Thibaut Lavril, Jenya Lee, Diana Liskovich, Yinghai Lu, Yuning Mao, Xavier Martinet, Todor Mihaylov, Pushkar Mishra, Igor Molybog, Yixin Nie, Andrew Poulton, Jeremy Reizenstein, Rashi Rungta, Kalyan Saladi, Alan Schelten, Ruan Silva, Eric~Michael Smith, R.~Subramanian, Xia Tan, Binh Tang, Ross Taylor, Adina Williams, Jian~Xiang Kuan, Puxin Xu, Zhengxu Yan, Iliyan Zarov, Yuchen Zhang, Angela Fan, Melanie Kambadur, Sharan Narang, Aurelien Rodriguez, Robert Stojnic, Sergey Edunov, and Thomas Scialom.
\newblock Llama 2: Open foundation and fine-tuned chat models.
\newblock \emph{arXiv:2307.09288}, 2023{\natexlab{b}}.

\bibitem[Honnibal and Montani(2017)]{spacy2017Natural}
Matthew Honnibal and Ines Montani.
\newblock {spaCy 2}: Natural language understanding with {B}loom embeddings, convolutional neural networks and incremental parsing.
\newblock 2017.
\newblock To appear.

\bibitem[Kincaid et~al.(1975)Kincaid, Fishburne, Rogers, and Chissom]{Kincaid1975DerivationON}
J.~Peter Kincaid, Robert~P. Fishburne, Richard~L. Rogers, and Brad~S. Chissom.
\newblock Derivation of new readability formulas (automated readability index, fog count and flesch reading ease formula) for navy enlisted personnel.
\newblock 1975.

\bibitem[Mclaughlin(1969)]{Mclaughlin1969SMOGG}
G.~Harry Mclaughlin.
\newblock Smog grading - a new readability formula.
\newblock \emph{The Journal of Reading}, 1969.

\bibitem[Coleman and Liau(1975)]{Coleman1975ACR}
Meri Coleman and Ta~Lin Liau.
\newblock A computer readability formula designed for machine scoring.
\newblock \emph{Journal of Applied Psychology}, 60, 1975.

\bibitem[Dale and Chall(1948)]{Dale1948AFF}
Edgar Dale and Jeanne~Sternlicht Chall.
\newblock A formula for predicting readability.
\newblock 1948.

\bibitem[Chall and Dale(1995)]{Chall1995ReadabilityR}
Jeanne~Sternlicht Chall and Edgar Dale.
\newblock Readability revisited : the new dale-chall readability formula.
\newblock 1995.

\bibitem[Klare(1974)]{Klare1974Assessing}
George~R. Klare.
\newblock Assessing readability.
\newblock \emph{Reading Research Quarterly}, 1974.

\bibitem[Gunning(1968)]{Gunning1968TheTO}
Robbie Gunning.
\newblock The technique of clear writing.
\newblock 1968.

\bibitem[Anil et~al.(2023)Anil, Dai, Firat, Johnson, Lepikhin, Passos, Shakeri, Taropa, Bailey, Chen, Chu, Clark, Shafey, Huang, Meier-Hellstern, Mishra, Moreira, Omernick, Robinson, Ruder, Tay, Xiao, Xu, Zhang, Abrego, Ahn, Austin, Barham, Botha, Bradbury, Brahma, Brooks, Catasta, Cheng, Cherry, Choquette-Choo, Chowdhery, Cr{\'e}py, Dave, Dehghani, Dev, Devlin, D'iaz, Du, Dyer, Feinberg, Feng, Fienber, Freitag, Garc{\'i}a, Gehrmann, Gonz{\'a}lez, Gur-Ari, Hand, Hashemi, Hou, Howland, Hu, Hui, Hurwitz, Isard, Ittycheriah, Jagielski, Jia, Kenealy, Krikun, Kudugunta, Lan, Lee, Lee, Li, Li, Li, Li, Li, Lim, Lin, Liu, Liu, Maggioni, Mahendru, Maynez, Misra, Moussalem, Nado, Nham, Ni, Nystrom, Parrish, Pellat, Polacek, Polozov, Pope, Qiao, Reif, Richter, Riley, Ros, Roy, Saeta, Samuel, Shelby, Slone, Smilkov, So, Sohn, Tokumine, Valter, Vasudevan, Vodrahalli, Wang, Wang, Wang, Wang, Wieting, Wu, Xu, Xu, Xue, Yin, Yu, Zhang, Zheng, Zheng, Zhou, Zhou, Petrov, and Wu]{Anil2023PaLM2T}
Rohan Anil, Andrew~M. Dai, Orhan Firat, Melvin Johnson, Dmitry Lepikhin, Alexandre~Tachard Passos, Siamak Shakeri, Emanuel Taropa, Paige Bailey, Z.~Chen, Eric Chu, J.~Clark, Laurent~El Shafey, Yanping Huang, Kathleen~S. Meier-Hellstern, Gaurav Mishra, Erica Moreira, Mark Omernick, Kevin Robinson, Sebastian Ruder, Yi~Tay, Kefan Xiao, Yuanzhong Xu, Yujing Zhang, Gustavo~Hernandez Abrego, Junwhan Ahn, Jacob Austin, Paul Barham, Jan~A. Botha, James Bradbury, Siddhartha Brahma, Kevin~Michael Brooks, Michele Catasta, Yongzhou Cheng, Colin Cherry, Christopher~A. Choquette-Choo, Aakanksha Chowdhery, C~Cr{\'e}py, Shachi Dave, Mostafa Dehghani, Sunipa Dev, Jacob Devlin, M.~C. D'iaz, Nan Du, Ethan Dyer, Vladimir Feinberg, Fan Feng, Vlad Fienber, Markus Freitag, Xavier Garc{\'i}a, Sebastian Gehrmann, Lucas Gonz{\'a}lez, Guy Gur-Ari, Steven Hand, Hadi Hashemi, Le~Hou, Joshua Howland, An~Ren Hu, Jeffrey Hui, Jeremy Hurwitz, Michael Isard, Abe Ittycheriah, Matthew Jagielski, Wen~Hao Jia, Kathleen Kenealy, Maxim Krikun, Sneha Kudugunta, Chang Lan, Katherine Lee, Benjamin Lee, Eric Li, Mu-Li Li, Wei Li, Yaguang Li, Jun~Yu Li, Hyeontaek Lim, Han Lin, Zhong-Zhong Liu, Frederick Liu, Marcello Maggioni, Aroma Mahendru, Joshua Maynez, Vedant Misra, Maysam Moussalem, Zachary Nado, John Nham, Eric Ni, Andrew Nystrom, Alicia Parrish, Marie Pellat, Martin Polacek, Alex Polozov, Reiner Pope, Siyuan Qiao, Emily Reif, Bryan Richter, Parker Riley, Alexandra Ros, Aurko Roy, Brennan Saeta, Rajkumar Samuel, Renee~Marie Shelby, Ambrose Slone, Daniel Smilkov, David~R. So, Daniela Sohn, Simon Tokumine, Dasha Valter, Vijay Vasudevan, Kiran Vodrahalli, Xuezhi Wang, Pidong Wang, Zirui Wang, Tao Wang, John Wieting, Yuhuai Wu, Ke~Xu, Yunhan Xu, Lin~Wu Xue, Pengcheng Yin, Jiahui Yu, Qiaoling Zhang, Steven Zheng, Ce~Zheng, Wei Zhou, Denny Zhou, Slav Petrov, and Yonghui Wu.
\newblock Palm 2 technical report.
\newblock \emph{arXiv:2305.10403}, 2023.

\end{thebibliography}
